\documentclass[12pt,PhD,twoside]{muthesis} 

\usepackage{verbatim}
\usepackage{graphicx}
\usepackage{url} 
\usepackage{listings}
\usepackage{pslatex} 

\usepackage{natbib}
\usepackage{times}
\usepackage{latexsym}
\usepackage{graphicx}
\usepackage{booktabs}
\usepackage{multirow}
\usepackage{amsmath}
\usepackage{amsfonts}
\usepackage{subcaption}
\usepackage{tikz}
\usepackage{float}
\usepackage{tcolorbox}
\usepackage{ulem}
\usepackage{float}
\usepackage{soul}
\usepackage{algorithm}
\usepackage[hidelinks]{hyperref}
\usepackage[linguistics]{forest}
\usepackage{algpseudocode}
\usetikzlibrary{arrows.meta}
\usepackage{pgfplots}
\usetikzlibrary{bayesnet}
\usepackage{wrapfig}
\usepackage{bbm}
\usepackage{arydshln}
\usepackage{multicol}
\usepackage{pifont}
\usepackage{enumitem}
\newcommand{\questionA}{\textbf{RQ1:} \textit{``How can we localise linguistic semantic features with geometrical properties in the latent space?''}}
\newcommand{\questionB}{\textbf{RQ2:} \textit{``How can semantic features be effectively separated within the latent space?''}}
\newcommand{\questionC}{\textbf{RQ3:} \textit{``How can we represent syntax in latent space?''}}
\newcommand{\questionD}{\textbf{RQ4:} \textit{``How can we learn a latent discrete space for natural language?''}}
\newcommand{\questionE}{\textbf{RQ5:} \textit{``How can we systematically define explanatory reasoning patterns?''}}
\newcommand{\questionF}{\textbf{RQ6:} \textit{``How can reasoning patterns be learned and disentangled within the latent space?''}}

\begin{document}

\title{Formal Semantic Control over Language Models}
\author{Yingji Zhang}
\stuid{10875677}
\principaladviser{Andre Freitas}

\beforeabstract
\prefacesection{Basic Concepts, Terms, and Abbreviations}
\begin{enumerate}[label={}]

\item \textbf{Concept:} an abstract mental representation that groups together objects, events, or ideas sharing common properties, enabling humans to categorise and reason about the world.

\item \textbf{Natural Language Explanation:} a human-readable justification or clarification that describes why or how a concept holds or relates to others.

\item \textbf{Natural Language Definition:} a human-readable description of a concept, typically structured as X is a Y that Z, where the concept (X) is classified under a general category (Y) and further specified by distinguishing properties (Z).

\item \textbf{Natural Language Inference (NLI):} the logical connection of (and the progression through) statements to reach a conclusion.

\item \textbf{Deductive Inference:} a type of reasoning where you start from general principles, rules, or known facts and apply them to reach a specific, logically certain conclusion.

\item \textbf{Predicate (PRED):} an expression (often a verb) that denotes an action, event, or state, in a sentence. 

\item \textbf{Argument (ARG):} an argument is an entity that fills a role required by the predicate in that action, event, or state, in a sentence.

\item \textbf{Predicate-Argument Structure:} a linguistic structure represents the relationship between an action (the predicate) and its participants (the arguments) in a sentence.

\item \textbf{Syntactic Tree:} a diagram that shows the grammatical structure of a sentence according to the rules of a formal grammar. It represents how words group together into phrases and how those phrases combine to form a complete sentence.

\item \textbf{Semantic Role Labelling (SRL):} identifying the predicate–argument structure of a sentence by determining ``who did what to whom, when, where, and how.'' where each word is assigned a semantic role.

\item \textbf{Abstract Meaning Representation (AMR):} a directed, rooted graph that represents the predicate-argument structure of a sentence, capturing its meaning as concepts (nodes) and the relationships between them (edges).

\item \textbf{Formal Inference:} reasoning where validity is determined solely by logical structure, independent of the specific meanings of the terms involved.

\item \textbf{Abstraction:} in formal inference, abstraction involves isolating the essential properties of objects while disregarding incidental details, thereby enabling general inference across classes of entities~\cite{1532611}.

\item \textbf{Compositionality:} the meaning of a whole expression is determined by the meanings of its parts and the rules (structures) used to combine them.

\item \textbf{Localisation:} adjusting the meaning of sentences locally in their context during the process of building up sentence meaning.

\item \textbf{Formal Semantics:} the study of natural language meaning using structure, logic, and mathematics, where expressions are mapped to precise interpretations in a model according to the principle of compositionality.

\item \textbf{Distributional (or Latent) Semantics:} the meaning of words or sentences, or phrases is modelled as vectors in a high-dimensional space, learned from their distribution in large text corpora using neural networks.

\item \textbf{Latent Space:} a hidden vector space learned by a neural network where complex data is represented by lower-dimensional embeddings that capture its essential, meaningful features.

\item \textbf{Latent Space Geometry:} the structure and relationships of features in the latent space a neural network uses to encode information.

\item \textbf{Subspace:} a subset of a vector space that is closed under vector addition and scalar multiplication.

\item \textbf{Convex Cone:} a ``fan-shaped'' region of space that starts at the origin and expands outward, which is closed under nonnegative linear combinations of its elements (vectors) in a vector space. 

\item \textbf{Disentanglement:} the process of learning representations where different, independent factors of variation in the data are separated into distinct, interpretable components in latent space.

\item \textbf{Information Bottleneck:} a principle that learns compact representations of data by preserving only the information most relevant to a target.

\item \textbf{Sentence Bottleneck:} a neural architecture constraint where an entire sentence is compressed into a single fixed-size vector.

\item \textbf{Variational AutoEncoder (VAE):} an encoder-bottleneck-decoder architecture, where the bottleneck is a latent space (single sentence embedding in this thesis), with each dimension following a known distribution, such as the Gaussian distribution.

\item \textbf{Normalising Flow (NF):} a lightweight model for building complex probability distributions by applying a sequence of invertible transformations to simple ones (e.g., Gaussian).

\item \textbf{Vector Quantised Variational Autoencoder (VQ-VAE):} an encoder-bottleneck-decoder architecture, where the bottleneck is a latent space (a sequence of token embeddings in this thesis), learned by a codebook.

\end{enumerate}

\prefacesection{Abstract}
\abstracttitle
%
{\singlespacing




This thesis advances \textit{semantic representation learning} to render language representations or models more semantically and geometrically interpretable, and to enable localised, quasi-symbolic, compositional control through deliberate shaping of their latent space geometry. We pursue this goal within a VAE framework, exploring two complementary research directions: \textbf{(i)} Sentence-level learning and control: disentangling and manipulating specific semantic features in the latent space to guide sentence generation, with explanatory text serving as the testbed; and
\textbf{(ii)} Reasoning-level learning and control: isolating and steering inference behaviours in the latent space to control NLI. In this direction, we focus on \textit{Explanatory NLI} tasks, in which two premises (explanations) are provided to infer a conclusion. The overarching objective is to move toward language models whose internal semantic representations can be systematically interpreted, precisely structured, and reliably directed. We introduce a set of novel theoretical frameworks and practical methodologies, together with corresponding experiments, to demonstrate that our approaches enhance both the interpretability and controllability of latent spaces for natural language across the thesis.
}

\prefacesection{Declaration}

\begin{enumerate}
    \item No portion of the work referred to in this thesis has been submitted in support of an application for another degree or qualification of this or any other university or other institute of learning.
    \item The material presented in this thesis represents the candidate’s own work except where stated otherwise.
\end{enumerate}
\begin{figure*}[ht!]
    \includegraphics[width=0.3\linewidth]{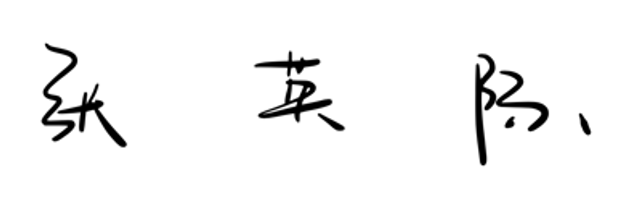}
\end{figure*}

\prefacesection{Acknowledgements}

Looking back over the past four years, it has been a spiritual journey that began with the confusion of choosing a research topic, passed through the anxiety of repeated paper rejections, and finally settled into a calm state as everything gradually came into place. I am deeply grateful to everyone who has accompanied me along this journey.

First and foremost, I would like to sincerely thank my two advisors, Andre Freitas and Ian Pratt-Hartmann, for their patient guidance and encouragement in both research and life. Andre’s catchphrase “Super Good” often gave me the courage to keep moving forward in moments of self-doubt when experiments failed and papers were rejected. Ian’s rigor in scholarship has gradually shaped me into a more critical thinker. I am also grateful to my senior colleagues and friends, especially Danilo, Marco, and Mohan, whose help and discussions made each piece of work more solid and refined. Finally, I would like to express my deepest gratitude to my parents; this is a kind of gratitude that, when expressed, almost feels like a debt. Your unconditional love and silent support have been the greatest source of strength and motivation throughout my entire journey.

\afterpreface

\chapter{Introduction} \label{cha:intro}
\section{Motivation}
\begin{figure*}[t]
    \includegraphics[width=\linewidth]{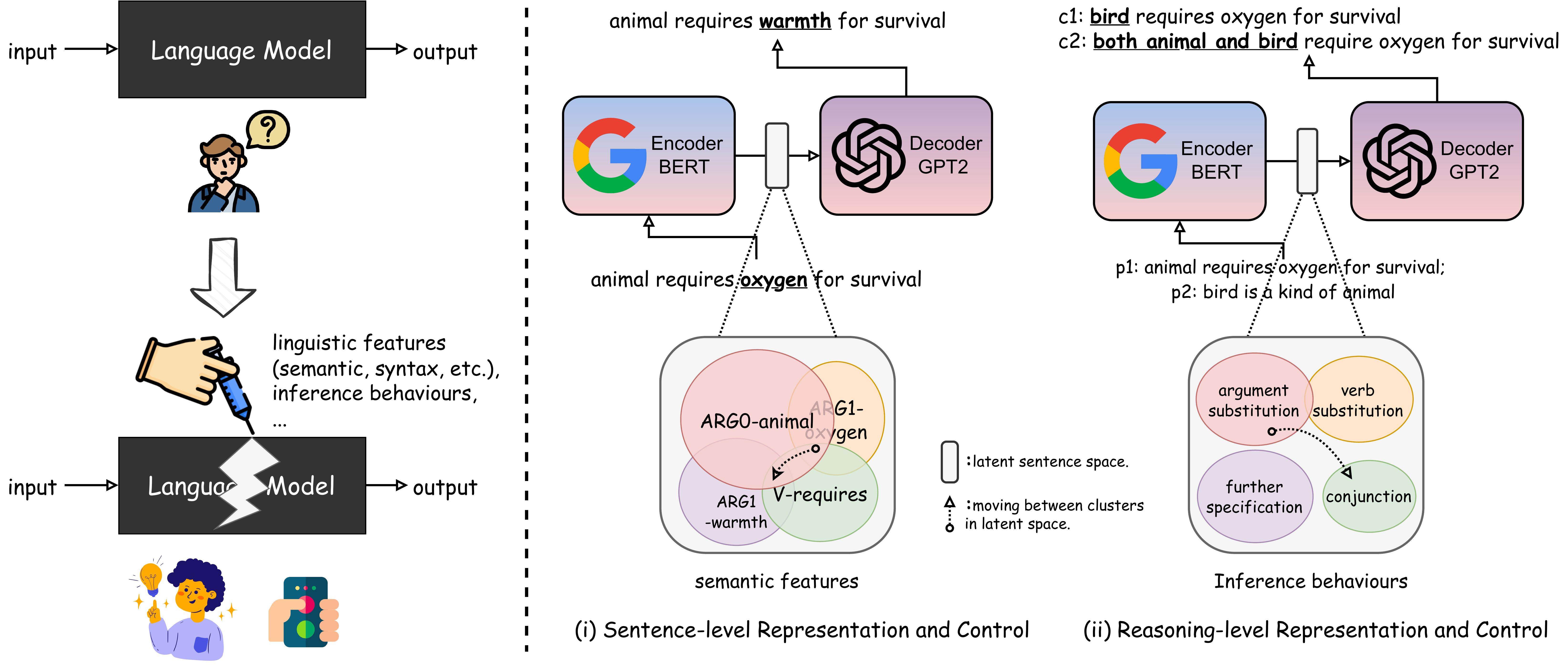}
    \caption{This thesis focuses on the \textit{``semantic and geometrical''} interpretability and \textit{``fine-grained, localised, quasi-symbolic, compositional''} generative control of Transformer Language Models (TLMs) by shaping the latent space geometry within the Variational AutoEncoder (VAE)  framework.}
    \label{fig:motivation}
\end{figure*}
Over the past decade, language models (LMs) have evolved from statistical n-gram systems, Recurrent Neural Networks~\cite{sherstinsky2020fundamentals}, to massive Transformer-based architectures~\cite{vaswani2017attention}, such as BERT \cite{devlin2019bert} and GPT \cite{radford2019language}, capable of understanding and generating human-like text across a vast range of tasks~\cite{yang2025qwen3}. These models now underpin applications in natural language understanding, inference, and generation. Their ability to learn from vast corpora and capture subtle statistical regularities in language has made them the foundation of modern Natural Language Processing (NLP). 

However, this remarkable progress has come at a cost: we understand little about the internal semantic mechanisms these models use. They operate as “black boxes”, where the representations they learn are difficult to interpret and control. Specifically, \textbf{\textit{Interpretability:}} Neural networks discover useful features automatically, and their latent spaces can capture the underlying geometry of features in data~\cite{Bengio2013RepresentationLA,alshammariunifying}. For Transformer LMs (TLMs), however, this latent space is high-dimensional, complex, with generative features for natural language poorly defined and geometrically entangled~\cite{scherlis2022polysemanticity}, thereby lacking \textit{semantic and geometrical} interpretation. \textbf{\textit{Controllability:}} Without a disentangled latent space, \textit{fine-grained, quasi-symbolic, localised} generative control of semantic content is difficult, limiting the ability to guide model outputs.

These challenges have direct implications for the reliability and trustworthiness of NLP systems in tasks such as Natural Language Learning (NLL) and Natural Language Inference (NLI). Consider, for instance, the reasoning process with the following premises (explanations): \textit{p1: \underline{animal} requires oxygen for survival} and \textit{p2: bird is a kind of \underline{animal}}. A human reasoner can infer \textit{c: bird requires oxygen for survival} by localising the argument \textit{animal} in \textit{p1} and substituting it with \textit{bird}. Alternatively, one might derive \textit{c: both animal and bird require oxygen for survival} by conjunctively combining the two arguments. In human (formal) reasoning, delivering this localised, quasi-symbolic, compositional inference control by identifying and substituting specific semantic arguments is straightforward because “subject”, “predicate”, and “object” are explicitly represented in linguistic cognition. This supports interpretable reasoning processes. 
In TLMs, however, such manipulation is far harder. Semantic features, such as \textit{subject-animal}, \textit{object-oxygen}, and \textit{predicate-require}, are not explicitly encoded as independent variables. Instead, they exist as distributed patterns in a high-dimensional latent space, with multiple features entangled across many parameters~\cite{scherlis2022polysemanticity}. Changing a single semantic factor (e.g., replacing “animal” with “bird”) can inadvertently alter unrelated aspects of meaning, making fine-grained reasoning control challenging.

A promising path toward interpretability and control lies in understanding and shaping the geometry of the latent space, where latent geometry refers to the structure and relationships of features in the hidden representations a network uses to encode information. An interpretable latent geometry exhibits meaningful clustering and separation aligned with underlying generative features, enabling localised control via targeted manipulations.

The Variational Autoencoder (VAE)~\cite{kingma2013auto} offers a compelling framework for this. Its encoder–bottleneck–decoder architecture compresses inputs into a low-dimensional latent sentence space, encouraging the capture of salient generative factors. When combined with an autoregressive TLM as the decoder, these latent variables can be explicitly manipulated to guide generation. Crucially, we can intentionally shape this geometry for greater interpretability and controllability.

Theoretically speaking, latent generative factors can be potentially aligned with semantic features grounded in formal semantic theory, such as Montague Semantics \cite{dowty2012introduction} and Semantic Role Labelling~\cite{palmer2010semantic}, which offer canonical, fine-grained, and systematic representations of meaning. By mapping latent variables to such formal semantic features, we can enhance the interpretability and safety of TLMs from a semantic perspective. Moreover, those generative factors can also be aligned with the inference behaviours, such as \textit{substitution} and \textit{conjunction} in the previous example. By separating the inference behaviours, we can deliver a controllable NLI process. In practice, targeted latent space optimisation via architectural modifications or specialised objectives can geometrically separate or disentangle these factors, enabling localised generative control. In the mechanistic interpretability context, localisation refers to identifying the specific features or components most responsible for a given behaviour~\cite{mueller2025mib}. Viewed geometrically, this means pinpointing and manipulating the relevant areas in latent space.

Overall, this thesis advances \textit{semantic representation learning} to render language representations or models more semantically and geometrically interpretable, and to enable localised, quasi-symbolic, compositional control through deliberate shaping of their latent space geometry. We pursue this goal within a VAE framework, exploring two complementary research directions: \textbf{(i)} Sentence-level learning and control: disentangling and manipulating specific semantic features in the latent space to guide sentence generation, with explanatory text serving as the testbed; and
\textbf{(ii)} Reasoning-level learning and control: isolating and steering inference behaviours in the latent space to control NLI. In this direction, we focus on \textit{Explanatory NLI} tasks, in which two premises (explanations) are provided to infer a conclusion. 
Our objective is to move toward LMs whose internal semantic representations can be systematically interpreted, precisely shaped, and reliably directed.

\subsection{Scientific Explanations}

This thesis focuses on scientific natural language explanations, drawing on WorldTree \cite{jansen2018worldtree} and EntailmentBank~\cite{dalvi2021explaining} datasets, as a rich framework for advancing research in \textit{semantic representation learning}. The motivation behind this focus on the explanation is two-fold:

\begin{enumerate}
    \item \textbf{Cognitive Perspective.} Scientific explanations, such as \textit{an animal is a kind of living thing}, reflect taxonomic structures grounded in conceptual categorisation, a fundamental cognitive process by which humans organise knowledge hierarchically. This involves mechanisms such as reasoning, analogy, and abstraction. Human cognition relies on structured schemas and ontologies that support pattern recognition, inference-making, and generalisation. By modelling these explanatory forms, AI systems can achieve more interpretable reasoning aligned with human cognitive expectations.

    \item \textbf{Linguistic Perspective.} Scientific explanations offer a semantically rich yet controlled experimental setting. They exhibit a well-scoped range of linguistic structures, combining compositional complexity with syntactic regularity. This makes them ideal for probing how language models encode and disentangle formal semantic relations, such as causal dependencies, taxonomic hierarchies, and conditional reasoning patterns.
    
\end{enumerate}

Given these properties, scientific explanations serve as an effective testbed for evaluating the quality of latent semantic representations, particularly in the context of compositionality, interpretability, and generalisation.


\section{Research Questions and Objectives}

\begin{figure*}[t]
    \includegraphics[width=\linewidth]{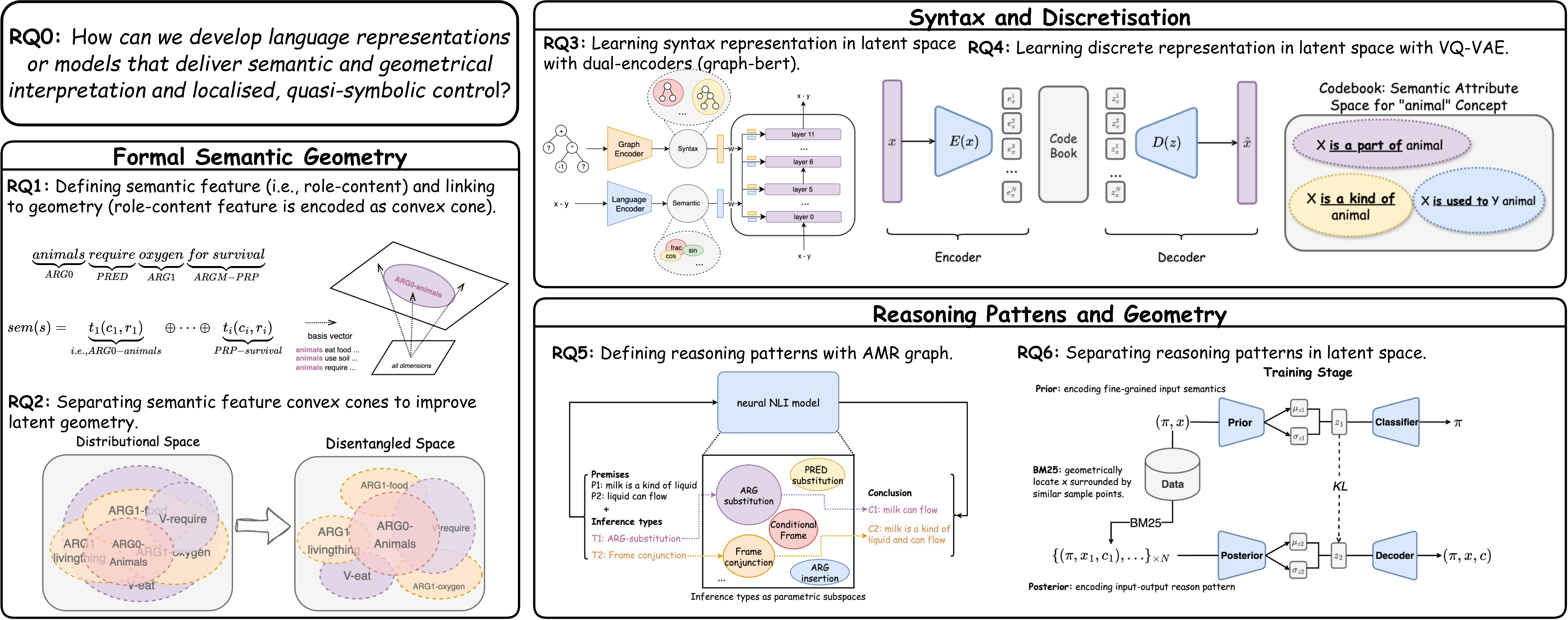}
    \caption{Overview of research questions and their solutions.}
    \label{fig:overview_intro}
\end{figure*}

Given the general objective of the thesis, the main high-level research question can be formulated as follows:

\begin{itemize}[leftmargin=1.5em]
    \item \textbf{RQ0:} \textit{``How can we develop language representations or models that deliver semantic and geometrical interpretation and localised, quasi-symbolic control?''}
\end{itemize}

We further break down the overall objective of the thesis into more specific research questions, from Sentence-level learning and control (\textbf{RQ1-4}) to Reasoning-level learning and control (\textbf{RQ5-6}), which will be explored in details in each chapter. The following sections discuss the specific research questions along with a general outline of the methodology adopted to answer them.

\subsection{Formal Semantic Geometry and Disentanglement}

The first part of the thesis aims at proposing a general theoretical framework to unveil the geometrical semantic structure of natural language expressions, especially but not limited to scientific explanations, within the latent space of the VAE architecture.
\begin{itemize}[leftmargin=1.5em]
    \item \questionA
\end{itemize}
This research question represents a fundamental step toward understanding how semantic features shape the geometry of sentence representations. To address \textbf{RQ1}, we begin by identifying general semantic features rooted in formal semantics theory, especially \textit{Semantic Role Labelling} \cite{palmer2010semantic} and \textit{Argument Structure Theory} \cite{jackendoff1992semantic}. We then formalise the connection between these linguistic semantic features and mathematical geometric concepts, such as directions, convex cones, and subspaces, in the latent space. This theoretical grounding allows us to bridge formal and latent semantics through a unified representation framework that is both interpretable and controllable.

In Chapter~\ref{cha:geo}, we demonstrated that semantic features, \textit{semantic role-word content}, can be formalised as \textit{convex cones} within the latent space (see Figure \ref{fig:overview_intro}). These features, grounded in formal semantics, compose to form the overall meaning of a sentence. Geometrically, each semantic feature corresponds to a cone-shaped region in the latent space, and the composition of these cones contributes to the positioning of the sentence representation. However, due to the nature of distributional semantics and contextual embedding, these convex cones of semantic features often \textit{overlap}, leading to entanglement between features. This overlap reflects the ambiguity and context dependence in natural language, and presents challenges for precise feature separation and interpretability within the latent geometry. To address this challenge, we pose the second research question:
\begin{itemize}[leftmargin=1.5em]
    \item \questionB
\end{itemize}
To answer \textbf{RQ2}, we demonstrated an effective approach to separate those semantic features, leading to better geometrical interpretability and controllability in Chapter~\ref{cha:dis}.

\subsection{Syntax Representation}

Syntactic structure plays a foundational role in shaping sentence meaning. In formal semantics theory, syntactic tree information serves as an essential first step in determining the semantic roles and relational structure of a sentence. Motivated by this observation, this thesis investigates how to explicitly inject syntactic representations into the latent space of a VAE, thereby enabling a more structured and interpretable model of semantics. This leads us to the third research question:
\begin{itemize}[leftmargin=1.5em]
    \item \questionC
\end{itemize}

To address \textbf{RQ3}, Chapter~\ref{cha:syntax} focuses on a bi-encoder VAE architecture in which syntactic and semantic information are encoded separately. Specifically, we use a graph encoder to capture the hierarchical syntactic structure (e.g., constituency trees) and a pre-trained TLM encoder (BERT) to represent sentence-level semantic content. These two encoders feed into a shared latent space designed to disentangle and integrate both structural and semantic information.

Experimental results demonstrate that explicit injection of syntactic structure into the latent space helps alleviate the information bottleneck commonly observed in VAE architectures. This approach enables the model to better capture nuanced sentence semantics, promotes more interpretable and controllable latent representations.

\subsection{Semantic Discretisation}

In addition to the VAE architecture with its continuous latent sentence space, natural language representation often exhibits an inherently discrete structure, reflected in symbolic elements such as words, syntactic categories, and semantic roles. To better align with this discretised nature of language, this thesis also investigates learning in a \textit{discrete latent space} using the Vector Quantised Variational Autoencoder (VQ-VAE) architecture. By mapping continuous encoder outputs to a finite set of learnable codebook entries, the VQ-VAE enables more symbolic and interpretable representations.

\begin{itemize}[leftmargin=1.5em]
    \item \questionD
\end{itemize}

To address \textbf{RQ4}, in Chapter~\ref{cha:discrete}, we present a baseline model, T5-VQVAE, which integrates the T5 encoder-decoder architecture with a vector-quantised latent space. This model supports the goal of \textit{symbolic semantic control}. Through this framework, we explore how discrete representations can facilitate fine-grained semantic manipulation and controlled generation.

\subsection{Explanatory Inference Pattern and Geometry}

Before that, our primary motivation is to investigate \textit{sentence-level learning and control} over the latent space. In Chapter~\ref{cha:reason}, we aim to propose both a theoretical and practical framework for \textit{reasoning-level learning and control} within the latent space. This investigation consists of two key stages: (1) Systematically annotate the reasoning behaviours involved in explanatory reasoning tasks, capturing patterns such as \textit{substitution} and \textit{conjunction}. (2) Encode and disentangle these annotated reasoning behaviours within the latent space, enabling the model to leverage this structured representation to guide and control the decoder during generation. For the first stage, we pose the research question:

\begin{itemize}[leftmargin=1.5em]
    \item \questionE
\end{itemize}

To address this question, we aim to develop a systematic annotation methodology for capturing explanatory reasoning behaviours, grounded in a formal semantic framework, specifically, \textit{Abstract Meaning Representation (AMR)}~\cite{jackendoff1992semantic}. We focus in particular on \textit{explanatory inference}, in which two premises are used to derive a conclusion. This approach allows us to represent the underlying logical and semantic structure of explanations in a linguistically consistent and interpretable form.

By defining explicit reasoning behaviour labels through this annotation framework, we can subsequently inject these labels into the latent space in the second stage of our investigation, which leads us to the final research question:

\begin{itemize}[leftmargin=1.5em]
    \item \questionF
\end{itemize}

Answering this question enables the model to learn clustered and disentangled latent representations that correspond to distinct types of reasoning behaviours. Such structured latent geometry supports more interpretable model behaviour over reasoning patterns during language generation, which is demonstrated in Chapter~\ref{cha:reasondis}.

\begin{figure*}[t]
    \includegraphics[width=\linewidth]{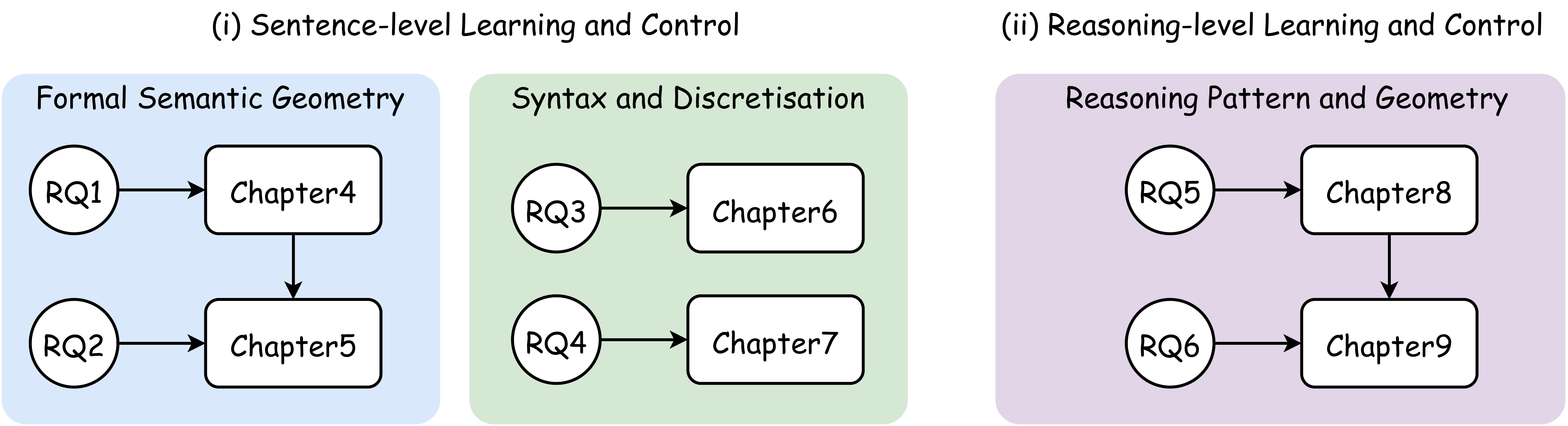}
    \caption{Overview of the thesis structure, including dependencies between research questions and chapters. Chapters 2 and 3 relate directly to all chapters by investigating the background and literature review.}
    \label{fig:rq}
\end{figure*}

\section{Contribution}

The research described in this thesis can be summarised by the following key contributions:

\begin{enumerate}
    \item \textbf{C1:} A comprehensive literature review on \textit{latent semantic geometry}, covering three major autoencoder architectures: Variational AutoEncoder (VAE), Vector Quantised Variational AutoEncoder (VQ-VAE), and Sparse AutoEncoder (SAE). This review establishes a unified perspective on how different architectures induce and structure semantic representations.
    
    \item \textbf{C2:} A novel theoretical and practical framework for \textit{formal semantic geometry and disentanglement} within the Language VAE architecture, aiming to bridge the gap between distributional and symbolic semantics through structured latent representations.
    
    \item \textbf{C3:} A novel framework for \textit{reasoning pattern annotation, learning, and disentanglement} within the Language VAE architecture, designed to bridge the gap between rule-based reasoning and memorised pattern recognition in generative models.
    
    \item \textbf{C4:} A new Language VAE architecture that disentangles heterogeneous \textit{syntax-semantic spaces}, leveraging dual encoders to incorporate structured syntactic information alongside semantic content.
    
    \item \textbf{C5:} The development of \textit{Transformer-based Language VQ-VAE baselines}, T5-VQVAE model, to support symbolic semantic control through discrete latent representations.
\end{enumerate}

\section{Thesis Outline}









The thesis is organised as follows:

\textbf{Chapter~\ref{cha:back}} present the technical background for the remainder of this thesis to further clarify the motivation of this thesis, focusing on Formal Semantic Theory, Distributional Semantics, TLMs, and VAEs.

\textbf{Chapter~\ref{cha:survey}} presents the literature review. It surveys foundational work in latent semantic representation from the perspective of latent geometry, highlighting the research gap, which this thesis focuses on.

\textbf{Chapter~\ref{cha:geo}} introduces a theoretical and practical framework for \textit{formal semantic geometry}. It defines formal semantic features and establishes a principled mapping between these linguistic features and mathematical geometric concepts within the latent space.

\textbf{Chapter~\ref{cha:dis}} addresses the separation of formal semantic features to enhance the geometric structure of the latent space, aiming to reduce feature entanglement and improve interpretability and controllability.

\textbf{Chapter~\ref{cha:syntax}} explores the integration of syntactic structure into the latent space through a \textit{dual-encoder VAE architecture}, combining graph-based syntax encoders with pretrained TLM encoders.

\textbf{Chapter~\ref{cha:discrete}} investigates learning discrete representations of natural language via the \textit{Vector Quantised VAE (VQ-VAE)} architecture, enabling symbolic semantic control through quantised latent codes.

\textbf{Chapter~\ref{cha:reason}} proposes a systematic framework for \textit{explanatory reasoning pattern annotation} using Abstract Meaning Representation (AMR), identifying key reasoning types such as taxonomic, causal, and definitional relations.

\textbf{Chapter~\ref{cha:reasondis}} builds on the annotated reasoning patterns to learn and disentangle these behaviours within the latent space, enabling fine-grained control over reasoning during generation.

\textbf{Chapter~\ref{cha:con}} concludes the thesis by revisiting the core research questions and summarising the key findings. It also discusses limitations and outlines open challenges and future directions for semantic representation learning.

\section{Publications}
The chapters presented in this thesis are based on the following publications:
\begin{itemize}
    \item \textbf{Chapters 2 \& 3} Zhang, Y., Carvalho, D. S., \& Freitas, A. (2025). \textit{Bridging Compositional and Distributional Semantics: A Survey on Latent Semantic Geometry via AutoEncoder.} (Under Review). The thesis author designed the methodology, leading the writing of the manuscript. Danilo Carvalho provided fundamental support for experimental and methodological design. Andre Freitas provided support and supervision throughout the publication process.
    \item \textbf{Chapter 4} Zhang, Y., Carvalho, D. S., \& Freitas, A. (2025). \textit{Formal Semantic Geometry over Transformer-based Variational AutoEncoder.} In the 29th Conference on Computational Natural Language Learning: CoNLL 2025 (\textit{Best Paper nomination}) \cite{zhang2022}. The thesis author designed the methodology, leading the writing of the manuscript. Danilo Carvalho provided fundamental support for experimental and methodological design. Andre Freitas provided support and supervision throughout the publication process.

    \item \textbf{Chapter 5} Zhang, Y., Carvalho, D. S., \& Freitas, A. (2024). \textit{Learning disentangled semantic spaces of explanations via invertible neural networks.} In Association for Computational Linguistics: ACL 2024 \cite{zhang2023learning}. The thesis author designed the methodology and experiment, leading the writing of the manuscript. Danilo Carvalho and Andre Freitas provided support and supervision throughout the publication process.
    
    \item \textbf{Chapter 6} Zhang, Y., Valentino, M., Carvalho, D. S., Pratt-Hartmann, I., \& Freitas, A. (2024). \textit{Graph-Induced Syntactic-Semantic Spaces in Transformer-Based Variational AutoEncoders.} In Findings of the Association for Computational Linguistics: NAACL 2024 \cite{zhang2023graph}. The thesis author designed the methodology and experiment, leading the writing of the manuscript. Marco Valentino provided fundamental support for experimental and methodological design. Danilo Carvalho and Andre Freitas provided support and supervision throughout the publication process.
    
    \item \textbf{Chapter 7} Zhang, Y., Carvalho, D. S., Pratt-Hartmann, I., \& Freitas, A. 2024. \textit{Improving Semantic Control in Discrete Latent Spaces with Transformer Quantised Variational Autoencoders.} In Findings of the Association for Computational Linguistics: EACL 2024 \cite{zhang-etal-2024-improving}. The thesis author designed the methodology and experiment, leading the writing of the manuscript. Danilo Carvalho, Ian Pratt-Hartmann, and Andre Freitas provided support and supervision throughout the publication process.

    \item \textbf{Chapter 8} Zhang, Y., Carvalho, D. S., \& Freitas, A. (2023). \textit{Guiding Explanation-based NLI through Symbolic Inference Types.} (Under Review) \cite{zhang2023type}. The thesis author designed the methodology and experiment, leading the writing of the manuscript. Danilo Carvalho provided fundamental support for annotation design. Andre Freitas provided support and supervision throughout the publication process.

    \item \textbf{Chapter 9} Zhang, Y., Valentino, M., Carvalho, D. S., \& Freitas, A. (2026). \textit{Learning to Disentangle Latent Reasoning Rules with Language VAEs: A Systematic Study.} The Fortieth AAAI Conference on Artificial Intelligence: AAAI 2026 \cite{zhang2025learningdisentanglelatentreasoning}. The thesis author designed the methodology and experiment, leading the writing of the manuscript. Marco Valentino, Danilo Carvalho, and Andre Freitas provided support and supervision throughout the publication process.
\end{itemize}

There are several other papers in which I participated that contribute to and support the investigation presented in this study.
\begin{itemize}
\item Carvalho, D. S., Mercatali, G., Zhang, Y., \& Freitas, A. 2023. \textit{Learning Disentangled Representations for Natural Language Definitions.} In Findings of the Association for Computational Linguistics: EACL 2023 \cite{carvalho2023learning}. The thesis author implements the experiment of the Transformer VAE baseline.

\item Carvalho, D. S., Zhang, Y., Unsworth, H., \& Freitas, A. (2025). \textit{LangVAE and LangSpace: Building and Probing for Language Model VAEs.} In Proceedings of the 2025 Conference on Empirical Methods in Natural Language Processing: EMNLP 2025 (System Demonstration) \cite{carvalho2025langvae}. The thesis author implements the probing tools of the language VAEs.
\end{itemize}

\chapter{Background} \label{cha:back}

In this Chapter, we present the technical background for the remainder of this thesis, focusing on the Formal Semantic Theory, Distributional Semantics, Transformer Language Models (TLMs), Variational AutoEncoders (VAEs), and Evaluation Approaches.

\section{Formal Semantic Theory}

This section provides essential background in formal semantic theory, including formal and material inference, semantic role labelling, and automatic semantic structure annotation, to further illustrate the motivation and objectives of this thesis.

\subsection{NLI: Formal \& Material Inference} 

An NLI system should be capable of two fundamental reasoning capabilities, including \textbf{formal inference} and \textbf{material inference}.

\paragraph{Formal Inference.} Formal inference refers to reasoning where validity is determined solely by logical structure, independent of the specific meanings of the terms involved. For example, consider the classic syllogism in Aristotle’s system:
\textit{p1: All A are B}. \textit{p2: all B are C}. \textit{c: all A are C}. In this case, the truth of the conclusion follows purely from the abstract form, regardless of what \textit{A}, \textit{B}, or \textit{C} actually refer to.
In Aristotle’s system, there are 24 abstracted reasoning patterns to guarantee validity. This form of abstraction reflects core aspects of human cognition~\cite{wang2003discovering,wang2003layered,wang2002cognitive}, where human abstract knowledge about the world may be described by a set of objects and their relations. For seeking generality and universal truth, either the objects or the relations can only be abstractly described and rigorously inferred by abstract models rather than by real-world details.

In natural language, the syntactic structure is important to abstraction, because it organises meaning in a way that can be detached from specific content and applied universally. In the example above, the reasoning pattern can be schematically represented as follows:
\[
\frac{
  S[\; NP[All \; A] \; VP[are \; B] \;] \quad 
  S[\; NP[All \; B] \; VP[are \; C] \;]
}{
  S[\; NP[All \; A] \; VP[are \; C] \;]
}
\]
Based on the syntactic representation, the reasoning form can then be formally and logically represented in predicate logic, e.g., first-order logic (FOL) \cite{fitting2012first} as:
\[
\frac{\forall x\,(A(x)\!\rightarrow\!B(x)) \quad \forall x\,(B(x)\!\rightarrow\!C(x))}
{\forall x\,(A(x)\!\rightarrow\!C(x))}
\]
In set-theoretic terms, such as Montague Semantics \cite{dowty2012introduction}, the same structure is captured as $A \subseteq B$, $B \subseteq C$, hence $A \subseteq C$.

By defining universal reasoning patterns over syntactic representations, formal inference ensures rigor, generality, and logical soundness through abstract models of objects and relations, rather than relying on specific real-world details. This approach enables the localisation of sentence meaning, thereby allowing reasoning to be generalised, controlled, and interpreted across categories and domains.

\begin{figure*}[ht!]
    \includegraphics[width=0.9\linewidth]{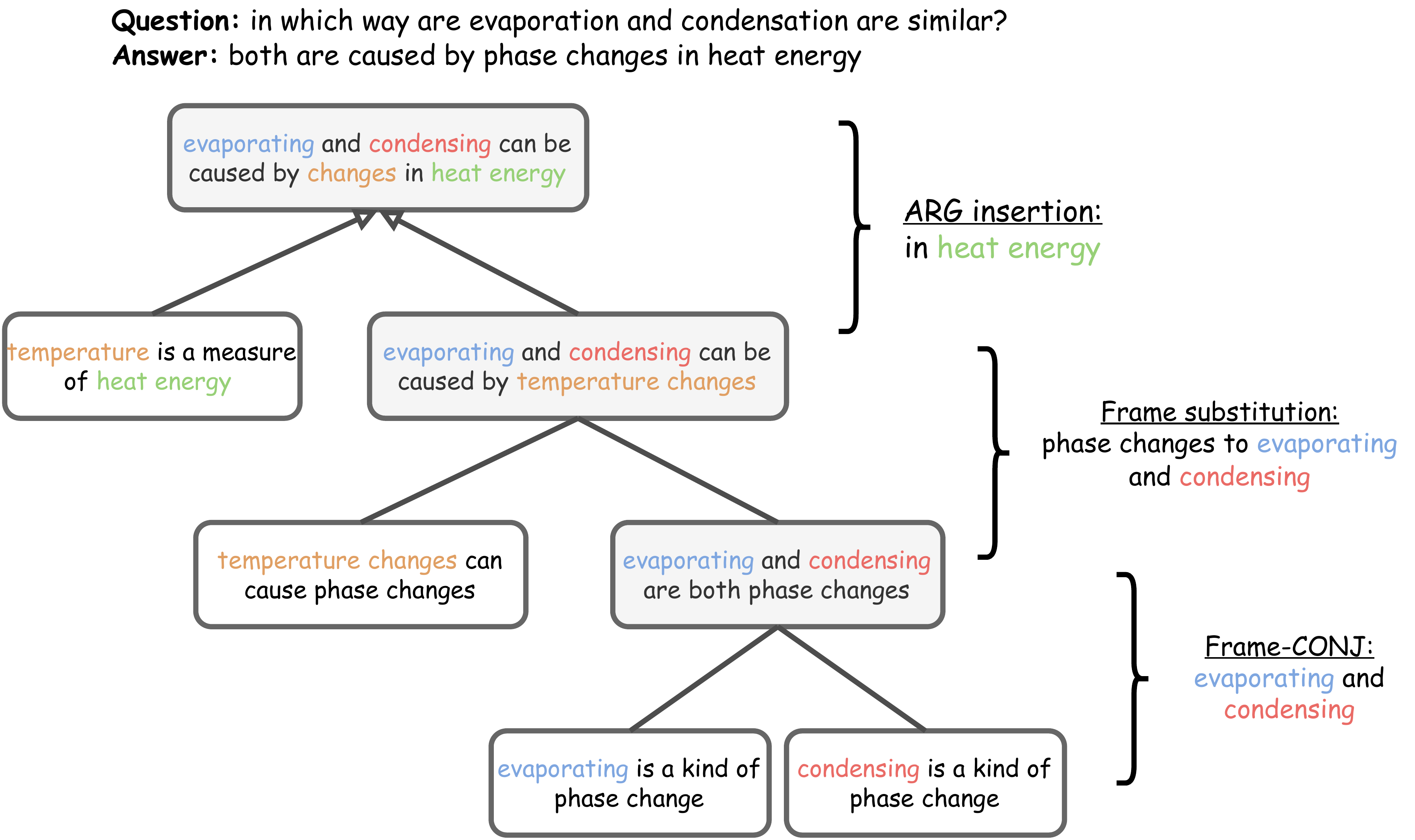}
    \caption{An example of an explanation-based entailment tree from EntailmentBank~\cite{dalvi2021explaining}. In this thesis, the focus is restricted to single-step explanatory inference, where two premises are combined to infer a conclusion.}
    \label{fig:sur_tree}
\end{figure*}

\paragraph{Material Inference.} Material inference \cite{ae1d0a0a-8dea-323c-89d0-72c600931595,brandom1994making} (or content-based inference, which commonly appeared in current TLMs), on the other hand, depends on the semantic content of the terms and is not strictly guaranteed by logical form alone. These inferences are context-sensitive, defeasible, and capable of capturing richer, real-world reasoning. For example, the last reasoning step in Figure~\ref{fig:sur_tree}:

\textit{p1: \underline{temperature} is a measure of \underline{heat energy}.}

\textit{p2: \underline{evaporating and condensing} can be caused by \underline{temperature changes}.} 

\textit{c: \underline{evaporating and condensing} can be caused by \underline{changes in heat energy}.}

This inference holds due to background physical knowledge about the relationship between temperature and heat energy. However, the same logical reasoning pattern can yield invalid conclusions if the content is changed: \textit{p1: \underline{Microscopes} are a measure of \underline{viruses}.} \textit{p2: \underline{A disease} can be caused by \underline{changes in viruses}.} \textit{c: \underline{A disease} can be caused by \underline{changes in microscopes}.} Despite the similar reasoning pattern, the conclusion in this case is clearly invalid, illustrating how material inferences are grounded in content and domain knowledge.

This distinction highlights a trade-off central to NLI systems. Bringing these two together is crucial not only for improving reasoning accuracy but also for interpretability, since explicit inference patterns can be inspected, and for controllability, since different reasoning forms can be selectively invoked.

\subsection{SRL: Interface between Formal-Material Inferences}

Semantic Role Labelling (SRL) plays a pivotal role as an interface between formal and material reasoning. Its development can be traced back to \citet{Fillmore1976} theory of case grammar and his subsequent work on frame semantics \cite{Fillmore1982}, which introduced the idea that predicates evoke conceptual frames with characteristic participant roles. These insights motivated large-scale lexical resources, such as PropBank \cite{Palmer2005}, which provided annotated corpora of predicate–argument structures. Early computational models, most notably \citet{Gildea2002}, demonstrated automatic role assignment using syntactic features, paving the way for machine learning approaches that matured through the CoNLL-2004/2005 shared tasks \cite{Carreras2004,Carreras2005}. With the advent of deep learning, models shifted from feature-rich classifiers to end-to-end neural encoders \cite{Zhou2015,He2017} and, more recently, contextual pretraining with ELMo \cite{Peters2018} and BERT \cite{Shi2019} has established SRL as a central component of modern shallow semantic parsing.

SRL provides a structured, domain-independent semantic representation by recovering predicate–argument structure (“who did what to whom, when, where, why”). This representation is more abstract than surface-level words and syntax but less rigid than pure logical forms. Specifically, it can bind both lexical semantics and semantic structure by identifying arguments of a predicate and labelling them with semantic roles such as Agent, Patient, Instrument, or Location, thereby capturing the underlying meaning of a sentence. For example:
$$\underbrace{animals}_{ARG0}~\underbrace{require}_{PRED}~\underbrace{oxygen}_{ARG1}~\underbrace{for~survival}_{ARGM-PRP}$$
In this example, ``animals" is labelled as ARG0 (Agent), indicating the doer of the action; ``require" is marked as the PRED (predicate); ``oxygen" is ARG1 (Patient), representing the entity being required; and the prepositional phrase ``for survival" is labelled ARGM-PRP, denoting a purpose adjunct.
\begin{table}[ht!]
\centering
\resizebox{7cm}{!}{
\begin{forest}
for tree={
    if n children=0{
      inner sep=1pt,
      align=center,
      base=top,
      s sep=0mm,
      l sep=100mm,
      minimum size=1em,
      tier=terminal
    }{},
  }
  [S [NP [NN [ \textit{\underline{ARG0 (Agent)}} [\textit{animals} $\rightarrow$ \textit{birds}] ] ]] [VP [VBZ [ \underline{\textit{PRED}} [\textit{require}]] ] [NP [ NP [\underline{\textit{ARG1 (Patient)}} [\textit{oxygen}] ] ] 
  [PP [ARGM-PRP [\underline{\textit{PRP}} [\textit{for}] ] 
  [\underline{\textit{PRP}} [\textit{survival}] ]
  ]]]]]
\end{forest}
}
\caption{Constituency parse tree enriched with SRL roles, showing how syntactic structure can be combined with semantic annotations. By learning the structural representation, we can provide semantic interpretation and quasi-symbolic control.} \label{tab:syntax_srl}
\end{table}

From a surface-level perspective, SRL abstracts away from syntactic and lexical variation to generate consistent, role-based representations of meaning. This abstraction enables generalisation across different linguistic constructions that express the same underlying relations, as shown in Table \ref{tab:syntax_srl}. Toward formal inference, SRL structures can be systematically translated into logical forms, supporting deductive reasoning grounded in predicate logic. These logical forms preserve structural regularities necessary for formal validity. Toward material inference, SRL provides precisely the semantic hooks, such as predicate types and argument roles, that enable integration with commonsense or domain-specific knowledge bases. This facilitates defeasible reasoning about real-world scenarios, where meaning and background knowledge are crucial. For example:

\textit{p1: \underline{temperature}(ARG0) is a measure of \underline{heat energy}(ARG1).}

\textit{p2: \underline{ARG0} can be caused by \underline{temperature changes}(ARG1).} 

\textit{c: \underline{ARG0} can be caused by \underline{changes in heat energy}(ARG1).}

According to SRL, in \textit{p1}, we know: 
$\text{temperature} \equiv f(\text{heat\_energy})$
where $f$ denotes a measure function. Therefore,
$
\text{temperature\_changes} \;\mapsto\; \text{heat\_energy\_changes}.
$
This shows how SRL identifies the predicate \textit{cause} and aligns its arguments, making the \textit{argument substitution} reasoning form explicit. Thus, SRL enables the material inference by grounding the mapping in semantic roles, while still maintaining a structure that can be treated formally. In this case, we can approximately define reasoning patterns grounded on semantic structural transformation to assist the control and interpretation of NLI systems.

\subsection{Automatic Semantic Structure Annotation}
\begin{figure}[ht!]
    \centering
    \includegraphics[scale=0.62]{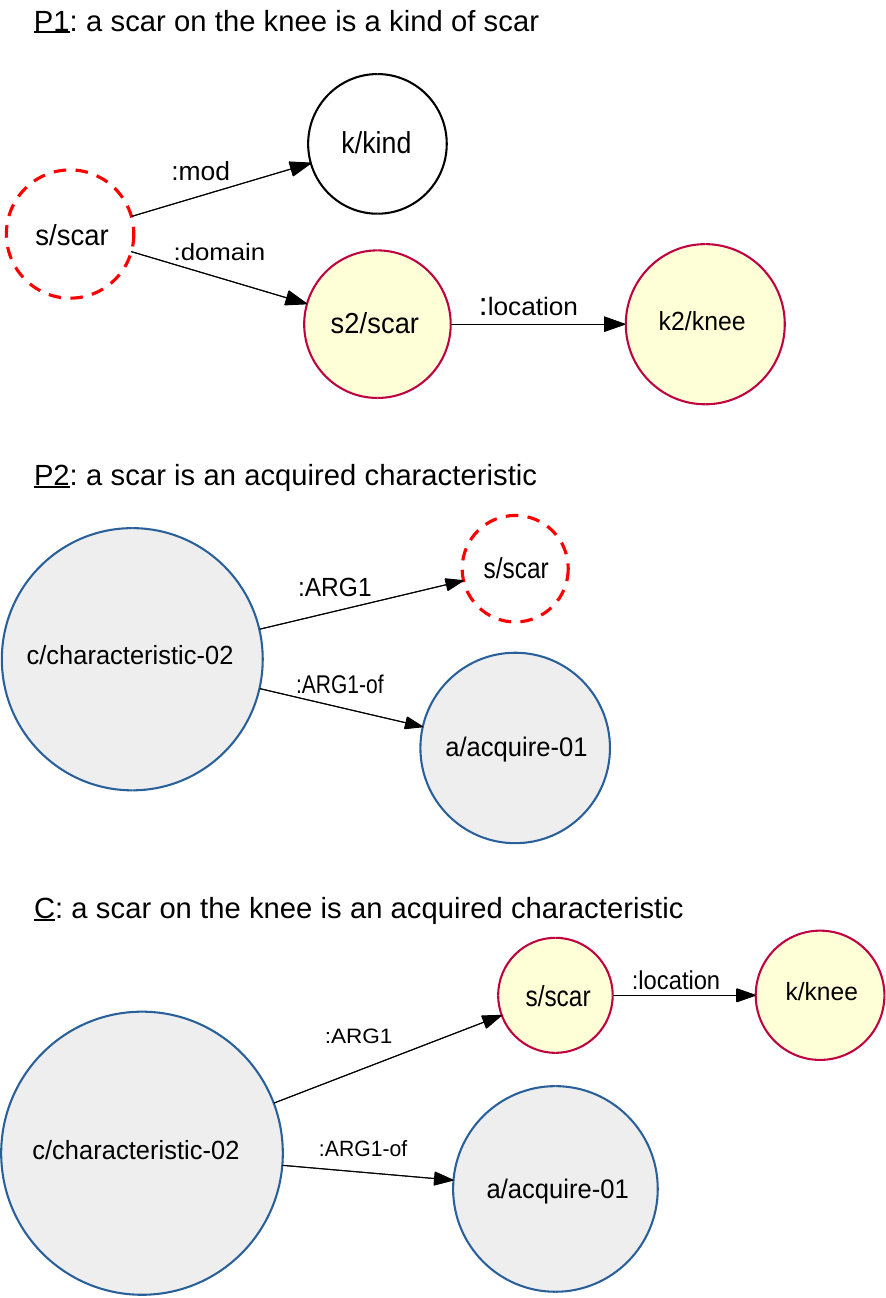}
    \caption{AMR argument substitution: the explanatory inference behaviour can be defined as subgraph substitution \cite{zhang2023type}.}
    \label{fig:survey_amr_argsub}
\end{figure}
To learn the language representations or models with the properties of linguistic features, such as syntax, SRL, and inference patterns, semantic annotations should be leveraged as supervisory signals to guide and structure the geometry of the latent space. This section introduces three types of semantic annotation within predicate-argument structure (\textbf{Constituency Syntax Representation}, \textbf{Semantic Role Labelling}, and \textbf{Abstract Meaning Representation}) to facilitate such learning.

\paragraph{Syntax Representation.} Syntax representation refers to the structured depiction of the grammatical relationships between elements within a sentence. These representations are essential for understanding how words combine to form meaningful phrases and sentences, enabling both linguistic analysis and computational language modelling. Among the various methods of syntactic representation, constituency parsing plays a pivotal role. A Constituency Parser analyses the syntactic structure of a sentence by identifying its constituent parts, such as noun phrases (NP), verb phrases (VP), and prepositional phrases (PP), and arranging them into a hierarchical tree structure. This tree reflects the nested, phrase-based organisation of natural language, based on phrase structure grammars such as those defined by the Penn Treebank \cite{marcus-etal-1993-building}.

\paragraph{Semantic Role Labelling.} Building on syntactic analysis, SRL extends the understanding of sentence structure by assigning roles to constituents based on their semantic relationships with the main predicates, typically verbs. In this work, an off-the-shelf annotation tool, i.e., \textit{AllenNLP library}, is employed for efficient semantic annotation, encompassing both syntactic structure and SRL~\cite{gardner-etal-2018-allennlp}.

\paragraph{Abstract Meaning Representation.} Abstract Meaning Representation (AMR) \cite{banarescu2013abstract} can be viewed as a graph-based formalism that builds on the predicate–argument structure and SRL. Like predicate–argument structures, AMR represents events and states through predicates and explicitly encodes their arguments (e.g., agent, patient, theme) as graph nodes connected via labelled edges. These edges correspond closely to semantic roles in SRL, such as \textit{:ARG0} (typically the agent) or \textit{:ARG1} (typically the patient/theme), providing a systematic mapping between natural language expressions and underlying meaning. However, AMR goes beyond traditional SRL by unifying disparate surface forms into canonical graph representations, abstracting away from tense, aspect, and syntactic variation, and integrating additional semantic information such as coreference, negation, and modality. An example of using AMR graph to support reasoning patterns annotation is provided in Figure \ref{fig:survey_amr_argsub}. In this thesis, we use an off-the-shelf parser \cite{damonte-17} for automatic AMR annotation.

\subsection{Summary}
In summary, this section has introduced key concepts from formal semantics, highlighting their strengths in supporting NLI systems. These theoretical foundations illustrate the potential for structured, meaning-preserving representations in computational models. Such representations can potentially improve the semantic interpretability and localised generative control. In the next section, we will introduce the distributional semantics learned in current TLMs and highlight their limitations when compared with formal semantics.


\section{Distributional Semantics}

\paragraph{Overview.} 
Distributional semantics is a computational approach to modelling meaning that rests on the empirical observation that words occurring in similar linguistic contexts tend to have related meanings over vector space \cite{firth2020papers}. Rather than defining meaning in terms of logical form, as demonstrated in Formal Semantic Theory, distributional semantics represents lexical items as vectors in high-dimensional spaces derived from their patterns of co-occurrence in large corpora. Early implementations relied on co-occurrence matrices and latent semantic analysis, but the field advanced significantly with neural word embedding models such as Word2Vec \cite{mikolov2013efficientestimationwordrepresentations}, which efficiently learn dense, low-dimensional vector representations that capture fine-grained semantic relationships. More recently, research has extended these ideas beyond words to sentence embeddings, which encode entire phrases or sentences as vectors, enabling semantic similarity and inference at a higher level of granularity. Models such as Sentence-BERT \cite{reimers2019sentencebertsentenceembeddingsusing} exemplify this development, showing how distributional principles can scale from lexical semantics to discourse-level meaning. Together, these methods have established distributional semantics as a cornerstone of modern computational linguistics and NLP.

\paragraph{Limitations.} 
Despite its success, distributional sentence semantics faces several important limitations when compared with formal semantics. While formal semantics provides explicit symbolic control through syntactic structures, distributional sentence representations generally lack mechanisms to systematically combine meanings in a rule-governed way, making it difficult to deliver semantic interpretation and fine-grained, localised semantic control. Furthermore, distributional sentence embeddings are often opaque: their high-dimensional vectors resist straightforward interpretability, offering little insight into which aspects of meaning are being represented and how they correspond to encode semantic structures.
\begin{figure}[t]
    \centering
    \includegraphics[width=0.9\linewidth]{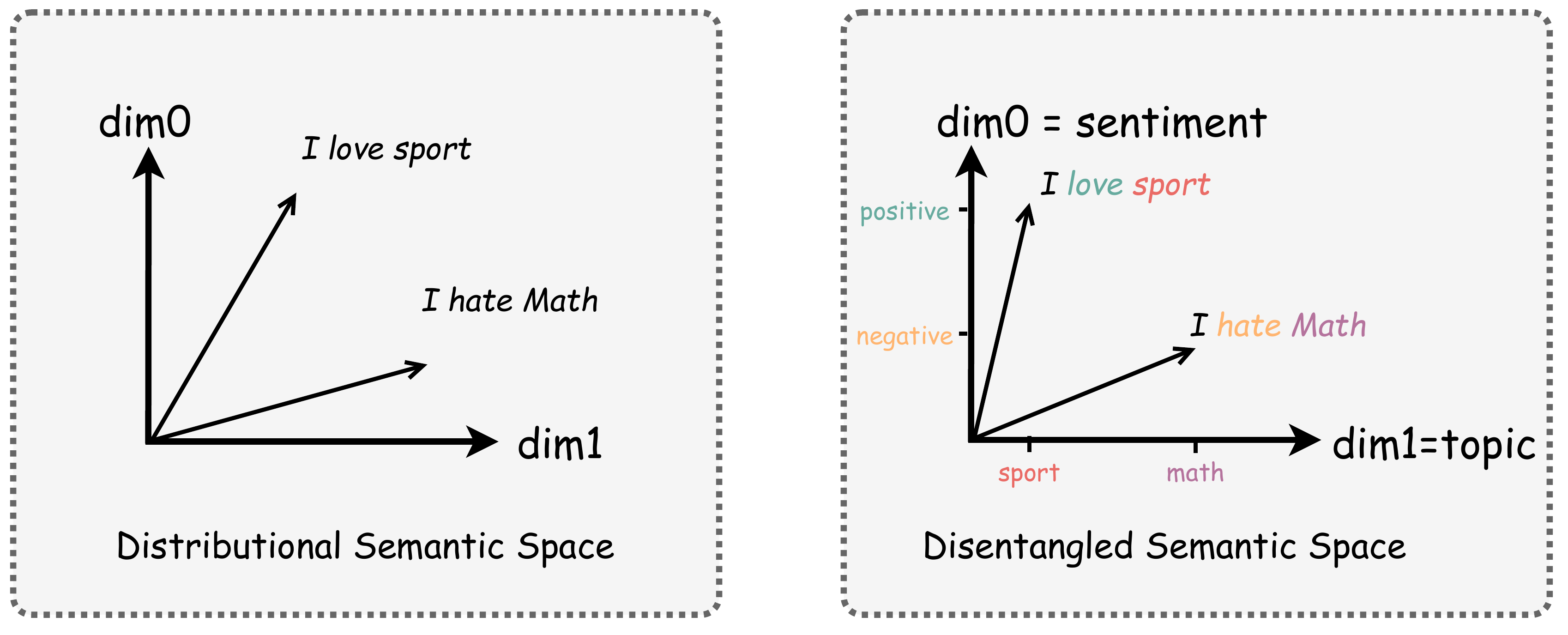}
    \caption{An example of separating semantic features (e.g., sentiment and topic) within the two-dimensional semantic space.}
    \label{fig:survey_vector}
\end{figure}


\paragraph{Semantic Space via AutoEncoder.} 
The AutoEncoder (AE) architecture, consisting of an encoder–bottleneck–decoder structure, learns to represent input sentences within a low-dimensional latent space. While the encoder and decoder capture fine-grained sentence information, the bottleneck layer (i.e., latent space) abstracts this into a more structural representation \cite{ldrdd2025}. Building on this framework, the Variational AutoEncoder (VAE) introduces an additional objective that promotes greater geometric separability of the latent space, most notably through disentanglement \cite{bengio2013deep, higgins2016beta,kim2018disentangling,locatello2019challengingcommonassumptionsunsupervised}.

In the NLP domain, latent space can be explicitly designed to capture distinct linguistic generative features~\cite{mercatali-freitas-2021-disentangling-generative,felhi2022towards}, or separated features such as sentiment and topic (see Figure~\ref{fig:survey_vector}) can be learned within distributional semantic spaces~\cite{liu-etal-2023-composable,gu-etal-2023-controllable,gu-etal-2022-distributional,hu2021causal,vasilakes-etal-2022-learning}. Such disentanglement or separation not only enables localised generative control but also enhances the interpretability of latent space geometry. These properties motivate the adoption of the VAE framework in this work as a means of achieving controllability over linguistic features in TLMs.


The following sections will present practical architectures designed to learn natural language representations through the TLM-VAE framework that can potentially embody formal properties.

\section{Transformer Language Models (TLMs)}

\paragraph{Overview.} Transformer Language Models (TLMs) pretrained via reconstruction can be broadly classified into three architectural categories, including encoder-only, decoder-only, and encoder–decoder, each optimised for distinct objectives. 

\begin{figure*}[ht!]
    \includegraphics[width=\linewidth]{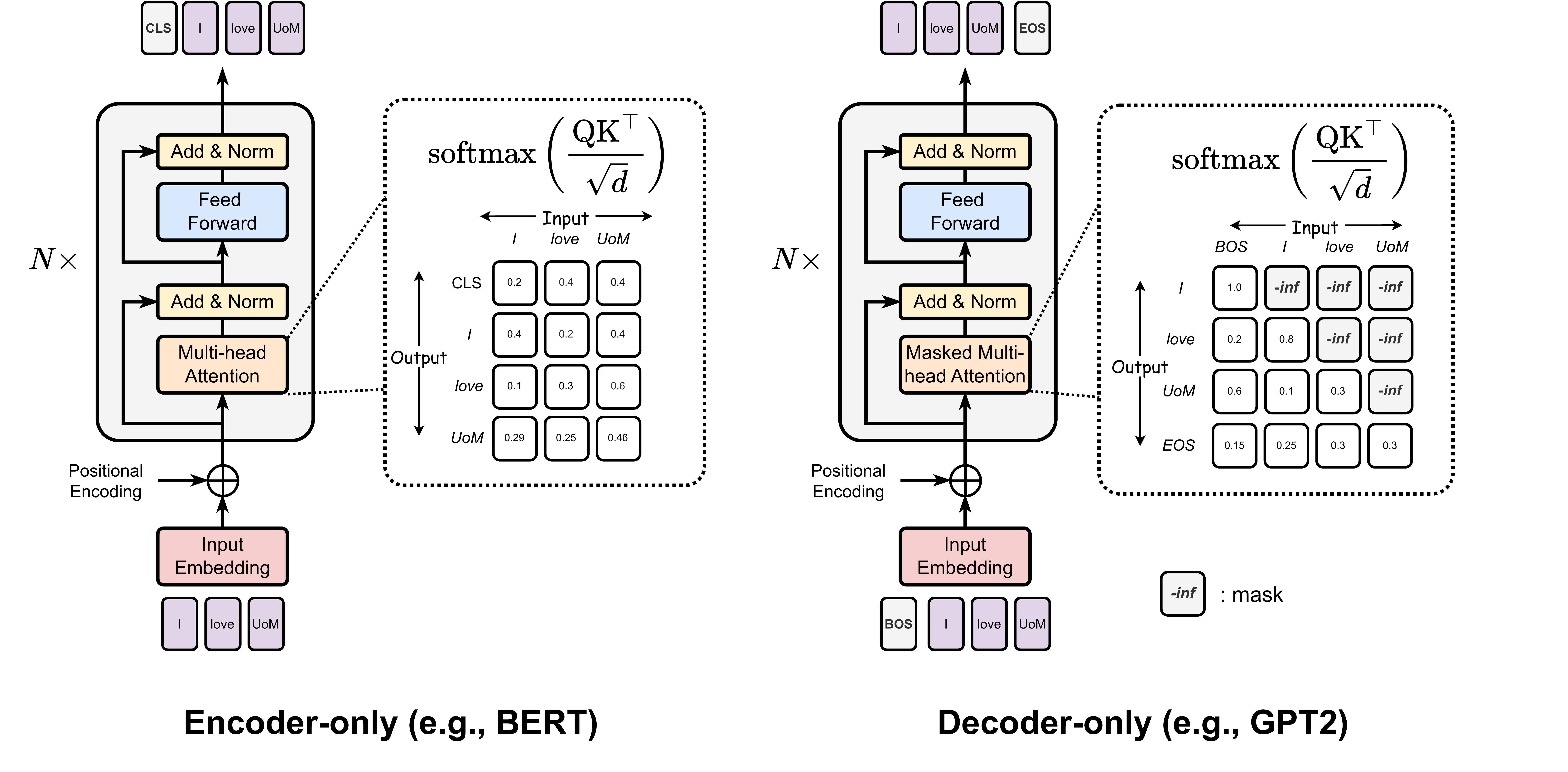}
    \caption{Encoder-only models use bidirectional attention, while decoder-only models rely on unidirectional attention to maintain autoregressive behaviour.}
    \label{fig:sur_arc}
\end{figure*}

\textbf{(i)~Encoder-Only:} Models such as BERT~\cite{devlin2018bert} specialise in \textit{text understanding and representation}. Leveraging bidirectional attention, they capture contextual information from both left and right, making them well-suited for natural language learning and representation tasks. E.g., the first output special token [CLS] is usually considered as the text representation, which contains sufficient semantic information.
\textbf{(ii) Decoder-Only:} Models such as GPT2~\cite{radford2019language} are designed for \textit{text generation}, using unidirectional (causally masked) attention to predict the next token left-to-right. During the pre-training stage, the input text is prefixed with BOS and ended by EOS tokens, marking the beginning and end of text generation. 
\textbf{(iii) Encoder–Decoder:} Architectures such as T5~\cite{raffel2020exploring} and BART~\cite{https://doi.org/10.48550/arxiv.1910.13461} integrate \textit{text understanding} in the encoder with \textit{text generation} in the decoder. This design supports conditional generation tasks, such as translation, summarisation, and text-to-text transformation, and enables flexible use of intermediate token representations.

\paragraph{Architecture.} The transformer layer is a core component in those TLMs. Although different TLMs vary in their architectural configurations, they typically consist of two main sublayers: multi-head self-attention and a position-wise feed-forward network, each followed by residual connections and layer normalisation, as shown in Figure \ref{fig:sur_arc}. Formally, given an input sequence represented as a matrix $X \in \mathbb{R}^{n \times d}$, where $n$ is the sequence length and $d$ is the hidden dimension, the queries, keys, and values are computed as $$Q = XW^Q, \quad K = XW^K, \quad V = XW^V$$, with learnable projection matrices 
$
W^Q, W^K, W^V \in \mathbb{R}^{d \times d_k}
$
The scaled dot-product attention is defined as
\[
\text{Attention}(Q,K,V) = \text{softmax}\!\left(\frac{QK^\top}{\sqrt{d_k}}\right)V
\]
In multi-head attention (MHA), $h$ such attentions are computed in parallel and concatenated:
$
\text{MHA}(X) = \text{Concat}(\text{head}_1, \dots, \text{head}_h) W^O
$
where each $\text{head}_i = \text{Attention}(Q_i,K_i,V_i)$.
A residual connection and layer normalisation are applied:
\[
X' = \text{LayerNorm}(X + \text{MHA}(X))
\]
The Feed-Forward Network (FFN) is applied independently to each position:
\[
\text{FFN}(X') = \sigma(X'W_1 + b_1) W_2 + b_2
\]
where $W_1 \in \mathbb{R}^{d \times d_{ff}}, \, W_2 \in \mathbb{R}^{d_{ff} \times d}$, and $\sigma$ is typically ReLU or GELU. Another residual connection and normalisation yield the output of the Transformer layer:
\[
Y = \text{LayerNorm}(X' + \text{FFN}(X'))
\]

\section{Variational AutoEncoder}

The Variational Autoencoder (VAE) framework facilitates the integration of encoder-only and decoder-only architectures by introducing a low-dimensional, regularised latent sentence space, where syntax, semantic content, and inference properties can be potentially packaged as a vector, thereby enhancing the interpretability and controllability of the latent representations.

\subsection{Variational AutoEncoder (VAE)}

\paragraph{Overview.} The VAE \cite{kingma2013auto} is a generative model that learns a continuous and probabilistic latent space through an encoder-decoder architecture. The encoder maps input data to a distribution over latent variables, typically modelled as a multivariate Gaussian, while the decoder reconstructs the original input from samples drawn from this latent distribution. By jointly optimising a reconstruction loss and a regularisation term (the Kullback-Leibler (KL) divergence between the approximate posterior and a prior distribution), the VAE encourages the latent space to be smooth and structured, making it well-suited for capturing semantic variations in a principled and interpretable way. The VAE is trained by minimising the evidence lower bound (ELBO), defined as:
\begin{equation}
\mathcal{L}_{\text{VAE}}(x; \theta, \phi) 
= \mathrm{KL}\!\left( q_\phi(z \mid x) \,\|\, p(z) \right) - \mathbb{E}_{q_\phi(z \mid x)} \big[ \log p_\theta(x \mid z) \big]
\end{equation}
Where $q_\phi(z \mid x)$ (i.e., posterior distribution) represents the distribution learned from the Encoder; $p(z)$ is the prior distribution (e.g., Gaussian); $p_\theta(x \mid z)$ means the distribution from the Decoder. Intuitively, the left term forces the latent space close to the prior distribution (ideally, each dimensions follow an independent prior distribution). The right term ensures the reconstruction quality.


\paragraph{Language VAEs.} With the advancement of TLMs, recent Language VAEs commonly adopt Transformer-based pretrained models as both encoder and decoder components. Notable examples include Optimus \cite{li2020optimus}, Della \cite{hu-etal-2022-fuse}, AdaVAE \cite{tu2022adavae}, LlamaVAE \cite{zhang2023llamavae}, and a general Language VAE framework designed to support flexible variable combinations between encoder and decoder \cite{carvalho2025langvae}.
\begin{figure*}[ht!]
    \includegraphics[width=\linewidth]{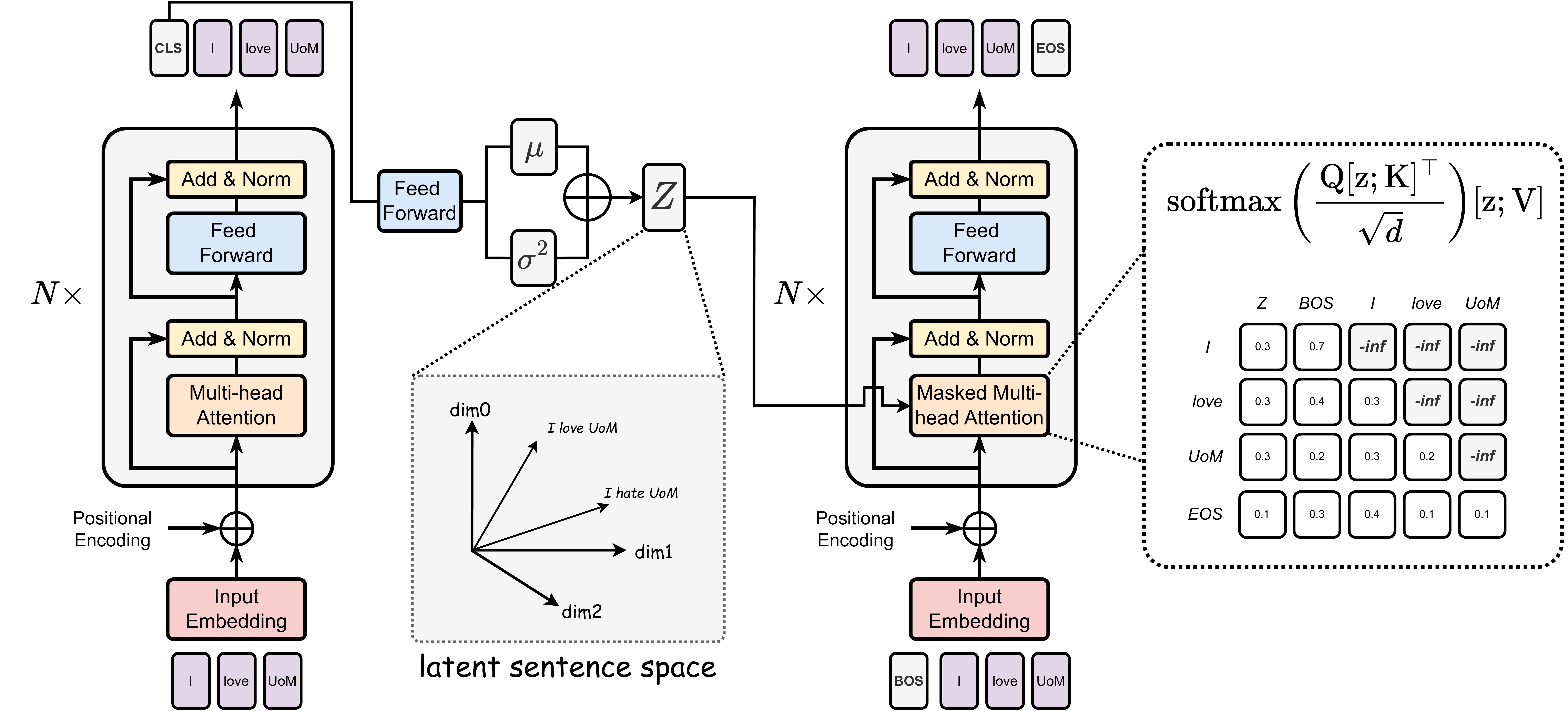}
    \caption{Optimus: BERT and GPT-2 are integrated within a VAE framework using a memory injection mechanism. During inference, sentence generation can be initiated with only the latent sentence space $z$ and a BOS token, as $z$ encapsulates all essential information required for generation.}
    \label{fig:sur_optimus}
\end{figure*}

Although more complex latent space manifolds, such as those employed in hyperbolic VAEs \cite{cho2023hyperbolic} and hyperspherical VAEs \cite{davidson2018hyperspherical}, have been proposed to encode different geometrical properties, the majority of Transformer-based language VAE models continue to adopt a Gaussian distribution as the prior over the latent space, due to its mathematical tractability and ease of optimisation. The Gaussian prior also facilitates effective sampling and interpolation within the latent space, which is essential for tasks like conditional generation and semantic manipulation.

One representative VAE-based model is Optimus~\cite{li2020optimus}, which employs BERT~\cite{devlin2019bert} as the encoder and GPT2~\cite{radford2019language} as the decoder. The model is designed to learn rich latent representations of sentences by integrating pretrained language models into the VAE framework.

Specifically, BERT encodes the input sentence \( x \) and produces a fixed-length representation using the special \texttt{[CLS]} token. This representation is used to parameterise a Gaussian latent distribution \( \mathcal{N}(\mu, \Sigma) \), where both \( \mu \) and \( \Sigma \) are learned through training. A latent vector \( z \sim \mathcal{N}(\mu, \Sigma) \) is then sampled and passed through a multi-layer perceptron \( W \), which projects \( z \) into a high-dimensional embedding \( h \in \mathbb{R}^{L \times H} \). Here, \( L \) and \( H \) denote the number of hidden layers and the dimension of hidden token representation, respectively. In GPT2, for example, \( L \) is 12, \( H \) is 768. \( H \) can be further split into $12 \times 64$, where 12 is the number of heads, 64 is the dimensions of token representation in the attention network. Therefore, for each attention head at each hidden layer, there is a corresponding latent representation $z \in \mathbb{R}^{1 \times 64} $. Please note that those configurations are different in distinct TLMs.

Each $z \in \mathbb{R}^{1 \times 64} $ is then incorporated as additional key and value in the self-attention of its corresponding head at the hidden layer in the decoder, a configuration referred to as the \textit{memory} setup. This design allows the latent variable to influence every layer of the GPT2 decoder, enabling fine-grained control over generation. The attention operation at each layer, each head can be summarised as:
\begin{equation}
\texttt{kv\_mem}: \text{softmax}\left(\frac{\text{Q}[\text{z}; \text{K}]^\top}{\sqrt{d}}\right)[\text{z}; \text{V}]
\end{equation}
where \( Q \), \( K \), and \( V \) represent the query, key, and value matrices with shape $\mathbb{R}^{seq \times 64} $ ($seq$ is the variable length of input sentence) commonly used in the Transformer architecture \cite{vaswani2017attention}. $[\text{z}; \text{V}]$ or $[\text{z}; \text{K}]$ with shape $\mathbb{R}^{(1+seq) \times 64}$.

\subsection{Normalising Flow AutoEncoder (NF-AE)}
\begin{figure*}[ht!]
\centering
    \includegraphics[width=\linewidth]{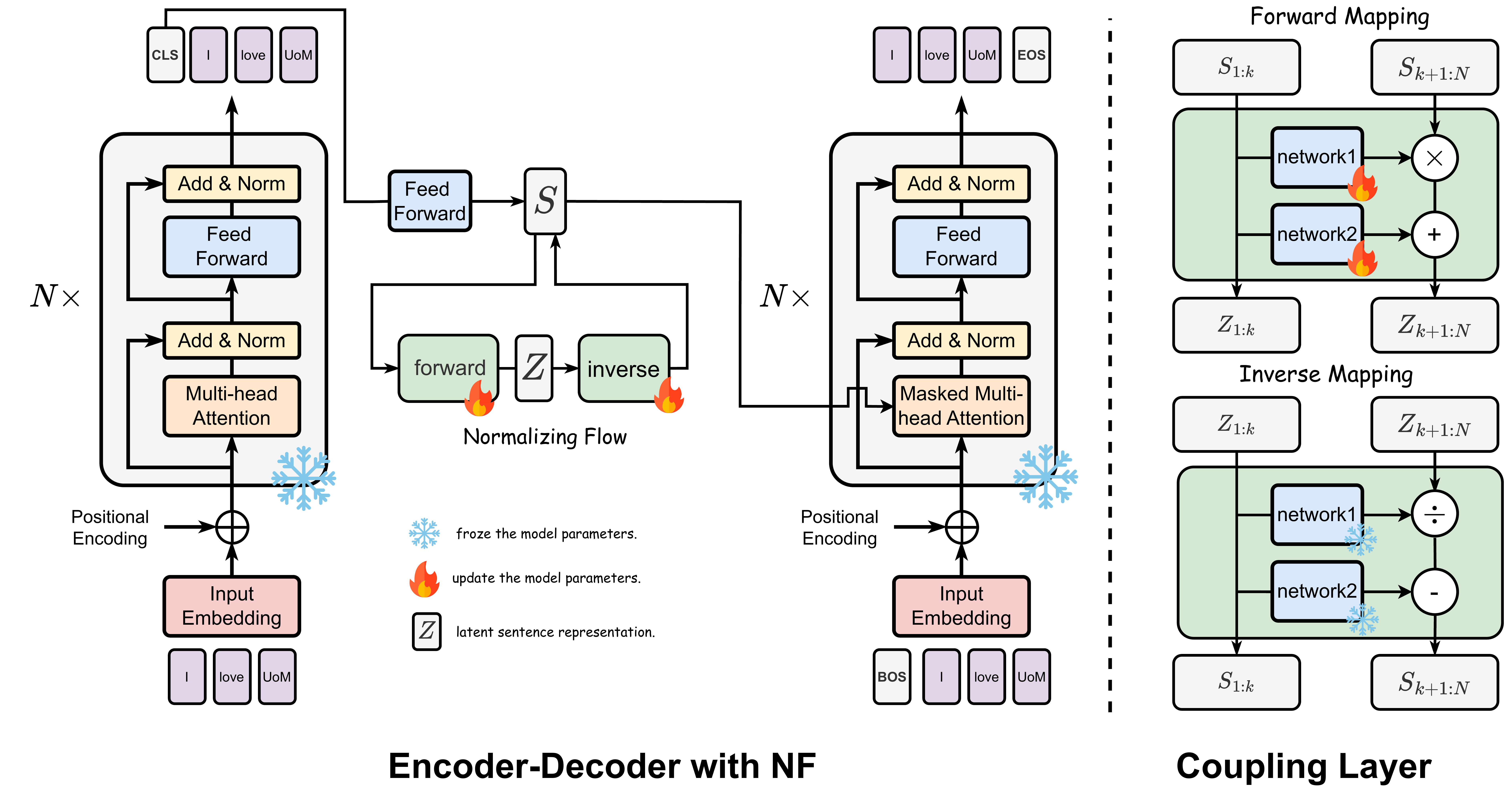}
    \caption{NF-AE: Plugging Normalising Flow (NF) into AE architecture. The lightweight NF can be plugged into the frozen pretrained AE model to enhance latent space interpretability and control. The core component of the NF is the coupling layer, which enables invertible transformations.}
    \label{fig:sur_flow}
\end{figure*}
A Normalising Flow (NF) \cite{dinh2014nice,kingma2018glow} is a lightweight generative model that transforms a simple probability distribution (e.g., a standard Gaussian) into a more complex one through a sequence of invertible and differentiable mappings. As shown in Figure \ref{fig:sur_flow}, within the Optimus-based AutoEncoder architecture, the latent sentence representation $s$ learned from [cls] token can be passed through the NF to learn a latent sentence space $z$ (typically modelled as a Gaussian distribution). Owing to the invertibility of the NF, $z$ can then be mapped back to $s$. This bidirectional mapping enables controlled manipulation of the latent representation by operating directly on the NF model.

The objective function of NF is to maximise the likelihood of the observed data under the transformed distribution. Given an invertible mapping $f$ that transforms data $s$ to a latent variable $z = f(s)$ drawn from a simple base distribution $p_Z(z)$, the probability density of $s$ is obtained using the change-of-variables formula:
\[
\log p_X(s) = \log p_Z(f(s)) + \log \left| \det \frac{\partial f(s)}{\partial s} \right|
\]
By minimising the negative log-likelihood, the flow learns transformations that warp the base distribution into one that matches the data. A common design for $f$ is the coupling layer, which splits the input into two parts: one remains unchanged, while the other is transformed using scale and translation functions conditioned on the first part. This structure ensures invertibility and keeps the Jacobian determinant easy to compute, making coupling layers both efficient and expressive.

\subsection{Vector Quantised Variational AutoEncoder (VQ-VAE)}
\begin{figure*}[ht!]
\centering
    \includegraphics[width=0.7\linewidth]{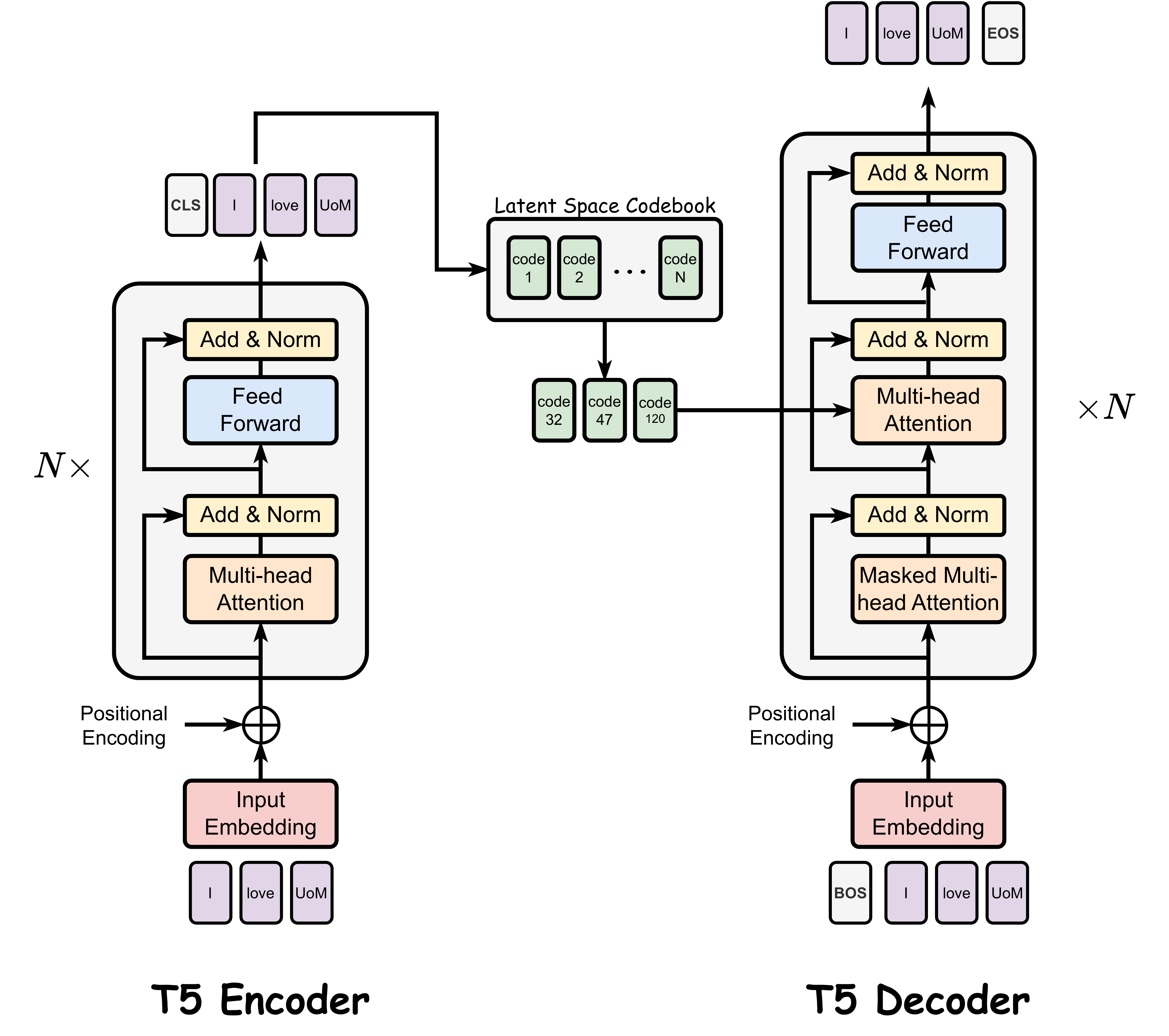}
    \caption{VQ-VAE: Integrating T5 encoder-decoder into VQ-VAE architecture. Specifically, the token embeddings produced by the encoder are replaced with their corresponding approximated codebook embeddings, which are then passed as inputs to the decoder.}
    \label{fig:sur_vqvae}
\end{figure*}
The Vector Quantised Variational Autoencoder (VQ-VAE) \cite{van2017neural} is a variant of the traditional VAE that replaces the continuous latent space with a discrete latent codebook. Rather than sampling from a Gaussian prior, the encoder outputs are mapped to the nearest vector in a learnable set of embedding vectors, known as the codebook. This discrete codebook serves as a fundamental representation space that can be efficiently transferred across architectures. As a result, VQ-VAE has been widely adopted in domains such as speech (e.g., Fish-Speech \cite{fish-speech-v1.4}) and multimodal learning (e.g., Janus-Pro \cite{chen2025janus}).

Architecturally, the model comprises three main components: an encoder that transforms input data $x$ into continuous latent token vectors, a quantisation module that replaces each token vector with its nearest neighbour in the codebook, and a decoder that reconstructs the input from these quantised representations. Training involves minimising a combination of reconstruction loss, a vector quantisation loss to align encoder outputs with codebook entries, and a commitment loss that encourages encoder outputs to stay close to selected codebook vectors. Figure \ref{fig:sur_vqvae} visualises the integration of the T5 encoder-decoder \cite{Raffel2020t5} within the VQ-VAE framework~\cite{zhang-etal-2024-improving}.

\subsection{Summary}
In summary, this section introduces three architectures, including VAEs, NF-AEs, and VQ-VAEs, for integrating latent spaces with TLMs (Encoder-Only, Decoder-Only, and Encoder-Decoder architectures). 
The following section presents evaluation methods for both the latent space representations and generative capabilities.

\section{Evaluation Approaches}
In this section, we introduce two primary categories for evaluating the performance of VAE models: latent space evaluation (visualisation, traversal, interpolation, and arithmetic) and generative evaluation (BLEU, BLEURT, and LLM evaluator).

\subsection{Latent Space Evaluation}
\begin{figure*}[ht!]
    \includegraphics[width=\linewidth]{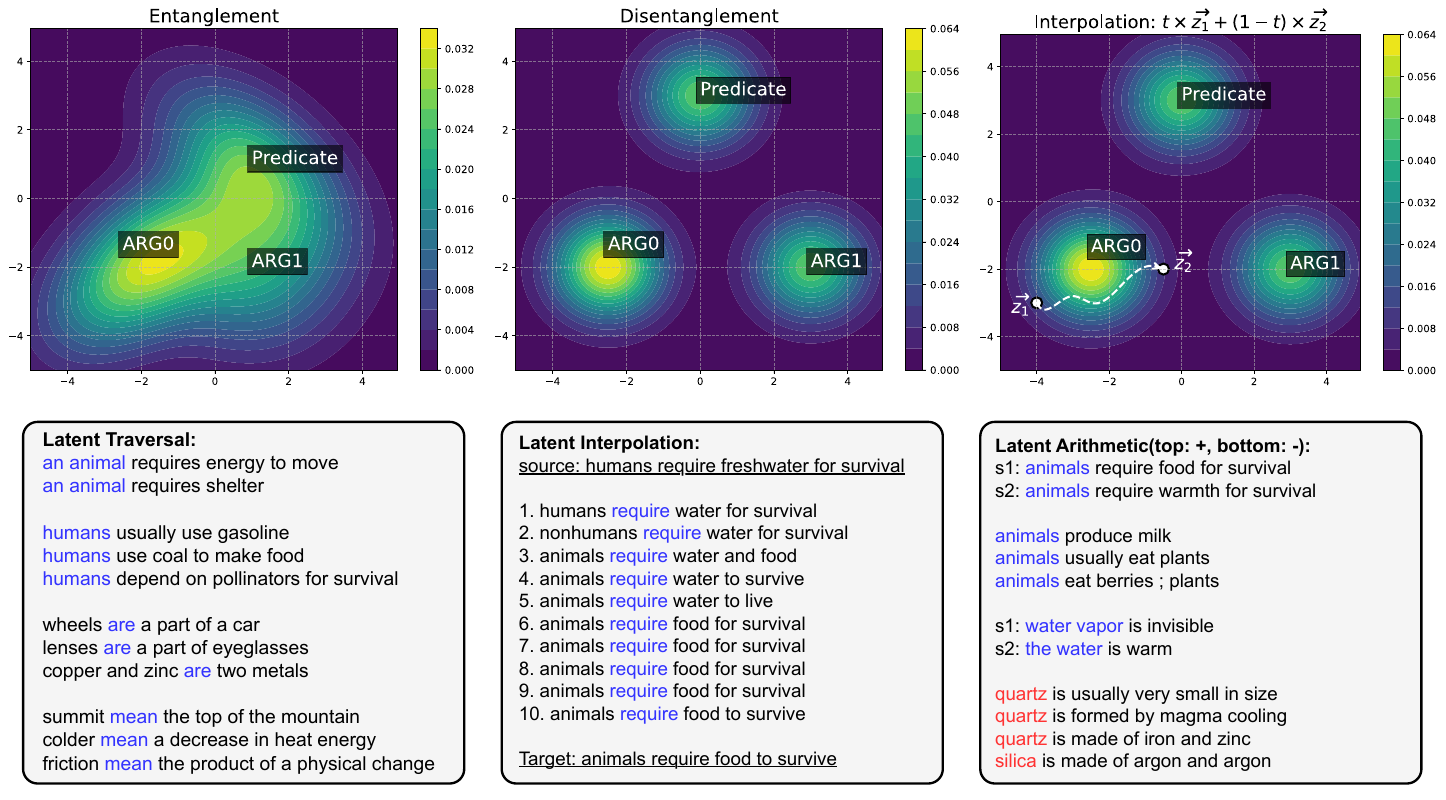}
    \caption{An example of latent space evaluation. By manipulating the latent space geometry (top), we can deliver semantic control to language generation (bottom).}
    \label{fig:latent_eval}
\end{figure*}

\paragraph{Latent Visualisation.} 
The structure of the latent space can be examined by visualising feature clusters and their separability. Dimensionality reduction techniques such as \textit{Principal Component Analysis (PCA)}, \textit{t-SNE}~\cite{van2008visualizing}, and \textit{UMAP}~\cite{mcinnes2018umap} are commonly employed to project high-dimensional latent representations into two or three dimensions. These visualisations help reveal semantic clustering and separation across different latent factors. Those visualisations are widely utilised during the experiments in this thesis.

\paragraph{Latent Traversal.} In VAE and VQ-VAE, latent traversal is a standard procedure for evaluating the geometry of latent spaces~\citep{higgins2016beta, kim2018disentangling}. It involves performing a random walk, such as \textit{Ornstein-Uhlenbeck}, across the latent space. If the latent space exhibits well-structured geometrical properties, the traversed examples are expected to display consistent and uniform patterns. As illustrated in Figure~\ref{fig:latent_eval} (left), decoding latent vectors sampled around the initial input sentence yields outputs with similar semantic features, suggesting that the latent space possesses well-structured geometric properties. For example, given an initial vector, we can randomly sample each latent dimension to traverse the latent space.
$$\text{z}_{neighbour} = \text{z}[i] \sim N(0, 1)_{\forall i \in \{0,..,size(\text{z})\}}$$

\paragraph{Latent Interpolation.} The interpolation mechanism operates by selecting two input sentences, \( x_1 \) and \( x_2 \), and encoding them into their corresponding latent representations \( z_1 \) and \( z_2 \). A linear interpolation is then performed along the latent space trajectory defined by \( z_t = z_1 \cdot (1 - t) + z_2 \cdot t \), where \( t \) varies from 0 to 1 in increments of 0.1. Consequently, nine intermediate sentences are generated along the interpolation path. As demonstrated in Figure~\ref{fig:latent_eval} (middle), the intermediate sentences generated during interpolation consistently preserve the semantic feature \textit{predicate-require}, indicating that the latent space exhibits well-structured geometric properties.

\textit{Interpolation smoothness metrics (IS)} \cite{zhang-etal-2024-improving} provide a quantitative way to calculate the smoothness of interpolation, which calculates the ratio between the ideal semantic distance (i.e., the aligned semantic distance between source and target sentences) and the actual semantic distance (i.e., the sum of semantic distance between adjacent sentences during interpolation). A higher ratio indicates that the actual path aligns better with the ideal path, suggesting better semantic-geometric properties. The metric is defined as:
$$\text{IS} = \mathbb{E}_{(s_0, ..., s_T) \sim P} \frac{\delta(\text{align}(s_0, s_T))}{\sum^T_{t=0} \delta(\text{align}(s_t, s_{t+0.1}))}$$
where $s_0, ..., s_T$ is the sequence of sentences during interpolation, $\delta$ and $\text{align}$ are sentence similarity and alignment functions, respectively, which are performed via Word Mover’s Distance \citep{zhao-etal-2019-moverscore}.

\paragraph{Latent Arithmetic.} 
Latent arithmetic, such as addition and subtraction, has been widely regarded as a lightweight approach to controlling TLM~\cite{shen2020educating,mercatali-freitas-2021-disentangling-generative,zhang-etal-2024-improving,zhang2024formalsemanticgeometrytransformerbased}, indicating the geometrical separation and linearity in the latent or parametric space.

In VAE, for example, given two sentences that share the concept animal, \textit{animals require food for survival} and \textit{animals require warmth for survival}, their vector addition in the latent space may yield a new sentence such as \textit{animals produce milk}, which also preserves the concept animal. This suggests the existence of a ``convex cone'' corresponding to the semantic concept animal within the latent space~\cite{zhang-etal-2024-improving,zhang2024formalsemanticgeometrytransformerbased}. 
The effectiveness of latent arithmetic can be quantitatively assessed by measuring the proportion of generated sentences that exhibit the intended target features.

\subsection{Text Generation Evaluation}

In addition to evaluating the latent geometry, several metrics can be employed to assess the generation quality, including BLEU, BLEURT, and LLM evaluators. Below, we provide a brief overview of the evaluation metrics used in this study.

BLEU \cite{papineni2002bleu} is a widely used automatic metric that measures n-gram overlap between generated and reference texts. It emphasises precision and includes a brevity penalty, making it effective for tasks like machine translation. However, BLEU often struggles to capture semantic similarity, especially when valid paraphrases are present. BLEURT \cite{https://doi.org/10.48550/arxiv.2004.04696} addresses these limitations by using transformer-based embeddings fine-tuned on human ratings. It better reflects fluency and meaning preservation, making it more suitable for evaluating reconstruction tasks where surface forms may vary. Moreover, previous studies have revealed that the Large Language Models (LLM) evaluation can be consistent with the results obtained by expert human evaluation \citep{chiang-lee-2023-large, liu-etal-2023-g, wang-etal-2023-chatgpt, Huang_2023}. Thus, we also conduct a quantitative analysis to measure whether the generated conclusion contradicts the premises through LLM evaluators, such as ChatGPT4o, in the NLI tasks.

\section{Conclusion} \label{sec:concl}

In this chapter, we present the technical background that underpins the remainder of this thesis, with a focus on Formal Semantic Theory, Distributional Semantics, Transformer Language Models (TLMs), and Variational Autoencoders (VAEs). We further elaborate on the motivation for integrating semantic structure into modern language representations and models, emphasising its potential to enhance interpretability and reasoning capabilities. The following chapter will review existing approaches to learning latent space geometry and will identify the key research gaps that this thesis aims to address.

\chapter{Literature Review} \label{cha:survey}

To develop a general understanding of \textbf{RQ0:} \textit{``How can we develop language representations or models that deliver semantic and geometrical interpretation and localised, quasi-symbolic control?''}, this chapter provides a systematic survey of research in \textit{semantic representation learning}, with a particular focus on the latent geometry induced by three different autoencoder architectures. We review three major classes of models: the Variational AutoEncoder (VAE), the Vector Quantised Variational AutoEncoder (VQ-VAE), and the Sparse AutoEncoder (SAE). For each architecture, we discuss how semantic information is encoded in the latent space, with special attention to the resulting geometric properties (i.e., feature clustering and separation).

\section{Semantic Representation Learning} \label{sec:related}

\begin{figure}[ht]
    \centering
    \includegraphics[scale=0.65]{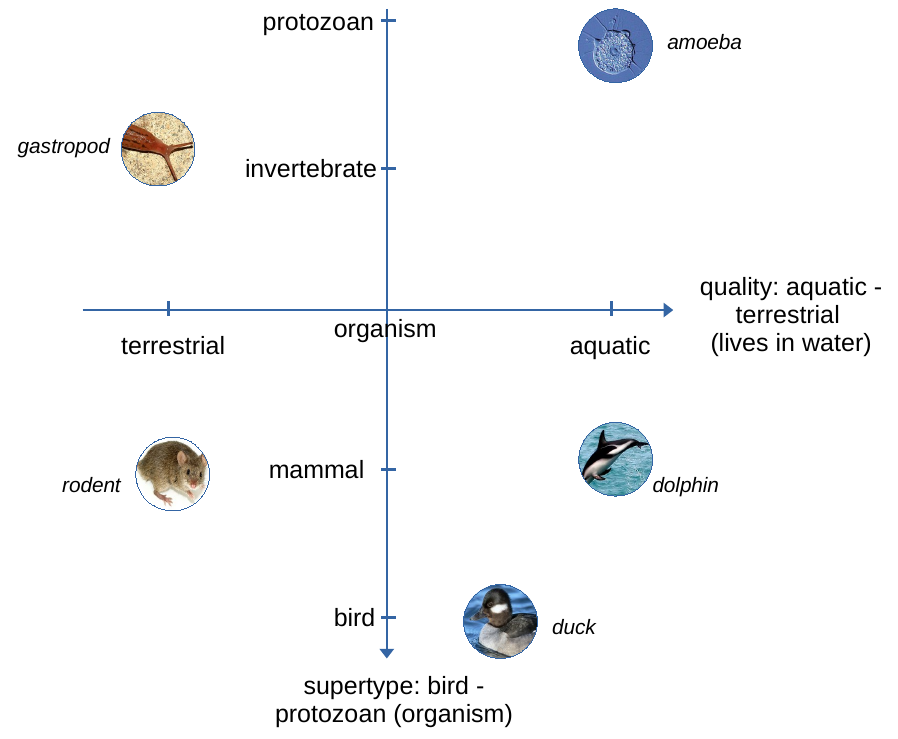}
    \caption{\citet{carvalho2023learning} proposed the conceptualisation of the latent geometry of concept \textit{organism} in the latent space, where different categories are located in different areas, thus improving latent space interpretation.}
    \label{fig:latent_concept}
\end{figure}

\paragraph{Task Definition.} 
This thesis focuses on the \textit{Semantic representation learning} task, which seeks to discover, structure, and shape latent representations of language such that they capture interpretable and separated semantic features. The aim is not just to encode statistical regularities for prediction or generation, but to align internal model representations with human-interpretable semantic structures (e.g., subject, predicate, object roles; substitution and conjunction in inference). We characterise this task along two complementary dimensions:

\paragraph{The Scope of Interpretability.}
One promising direction to interpret the representation space geometry is \textit{Linear Representation Hypothesis}, which suggests that high-level concepts, such as ``mammal'' and ``reptile'', can be encoded within causally separated subspaces (i.e., directions) of the high-dimensional latent space. An illustrative example is shown in Figure~\ref{fig:latent_concept}. This hypothesis has been supported across various domains, including word embeddings \citep{mikolov2013distributed, arora-etal-2016-latent}, sentence embeddings \citep{li2020sentence}, and TLMs \citep{meng2022locating, merullo-etal-2024-language, chanin-etal-2024-identifying, park2024the, park2024the1}.

Despite these advances, a deeper understanding of how such high-level concepts are encoded remains limited from the perspective of sentence semantics. In this survey, we emphasise that a well-structured representation space should exhibit interpretable geometric properties, such as meaningful directions, distances, and clustering, that correspond to underlying ``semantic'' features. For instance, semantically similar expressions should be mapped to nearby vectors, while distinct semantic roles or categories should occupy separable regions or align along distinct axes in the latent space. Such geometric interpretability offers valuable insights into how language models internally organise and encode linguistic meaning.

\paragraph{The Scope of Controllability.}
Building on this geometric view, \textit{controllability} concerns the ability to manipulate interpretable structures in latent space to achieve \textit{fine-grained}, \textit{quasi-symbolic}, and \textit{localised} control of model behaviour. This entails adjusting specific semantic attributes (e.g., tense, sentiment, or argument structure) by modifying their corresponding latent variables, thereby enabling targeted control over language generation. For instance, shifting sentiment from \textit{I love this movie} to \textit{I hate this movie}, or altering argument roles from \textit{animal is a kind of living thing} to \textit{tiger is a kind of living thing}.

The survey is organised as follows: (a) Section 3.1 frames the scope of the survey, stating a definition of semantic representation learning; (b) Section 3.2 reviews the main data resources for tasks requiring compositional semantics; (c) Section 3.3 provides a detailed classification of the main architectural patterns and approaches proposed for semantic representation learning.

\section{Linguistics and Data Resources}
In this section, we primarily introduce two types of natural language expressions that are widely employed in tasks involving \textit{formal semantics}. Both types of data exhibit semantically rich yet structurally controlled properties, making them suitable for evaluating the compositional capabilities of representation spaces. In addition, we briefly discuss other resources commonly used to study \textit{informal semantics}\footnote{Formal vs Informal Semantics: the difference between formal and informal semantics depends on the level of precision and mathematical rigour used to describe meaning. In this survey, we distinguish the informal and formal semantics by considering whether the structural representation, such as syntax, semantic roles, reasoning patterns, etc., is explicitly encoded in the latent space.}. A summarisation for data resources is illustrated in Table \ref{tab:bench}.
\begin{table*}[ht!]
\centering
\resizebox{\textwidth}{!}{
\small
\centering
\renewcommand\arraystretch{1.2}
\begin{tabular}{lll}
\toprule
\textbf{Expression} & \textbf{Compositional Semantic Tasks} & \textbf{Datasets} \\ \hline \hline
\multirow{3}{*}{Definitions} & Definition Modelling(vec2def) & \citet{noraset2016definition}, \citet{yang2019incorporating}, \citet{mickus-etal-2022-semeval} \\
& Definition Modelling(word2def) & \citet{noraset2016definition}, \textit{WordNet}, \textit{Wiktionary}, \textit{Wikipedia} \\
& Reverse Dictionary(def2vec) & \textit{Semeval-2022 Task 1: CODWOE} \cite{mickus-etal-2022-semeval} \\ \hline
\multirow{4}{*}{Explanations} & QA with explanation graph & \textit{WorldTree} \cite{jansen2018worldtree} \\
& QA with entailment tree & \textit{EntailmentBank}~\cite{dalvi2021explaining} \\
& Syllogistic NLI & \textit{Rule-based Explanatory NLI corpus}~\cite{zhang2024controllablenaturallanguageinference} \\
& commonsense QA & \textit{Zebra-KB}~\cite{molfese-etal-2024-zebra} \\ 
& ... & \textit{explanations corpora survey}~\cite{thayaparan2020survey} \\ \hline
\multirow{6}{*}{Others}
& \multirow{3}{*}{Sentiment Style-transfer} & \textit{Yelp reviews}~\cite{shen2017style}, \textit{Yahoo answers}~\cite{yang2017improved} \\
& & \textit{IMDb movie review}~\cite{diao2014jointly,maas-etal-2011-learning,li-etal-2018-delete} \\
& & \textit{Stanford Sentiment Treebank}~\cite{socher-etal-2013-recursive} \\
& Formality Style-transfer & \textit{GYAFC}~\cite{rao-tetreault-2018-dear} \\
& Codeswitching Style-transfer & \textit{LinCE}~\cite{aguilar-etal-2020-lince}\\
& Topic Style-transfer & \textit{AGNews dataset}~\cite{zhang2015character}, \textit{DBpedia}~\cite{zhang2015character} \\
& Toxic Style-transfer & \textit{Jigsaw Toxic Comment} \\
& Disentanglement & \textit{dSentences} \\
& ... & \textit{style-transfer corpora survey}~\cite{jin-etal-2022-deep} \\
\toprule
\end{tabular}
}
\caption{Summarisation of data sources and tasks. Please note that our motivation is to investigate the utilisation of those expressions in fundamental representation learning, rather than the downstream tasks we mentioned here.} \label{tab:bench}
\end{table*}

\paragraph{Natural Language Definitions.} Natural language definitions are linguistic expressions that explain the meaning of terms using everyday language, allowing individuals to comprehend, categorise, and communicate abstract concepts. They serve as bridges between raw lexical forms and the conceptual knowledge they denote, facilitating understanding across contexts and domains. These definitions are often characterised by a high level of abstraction and structured phrasing, which mirrors the hierarchical organisation of concepts in human cognition. This structured nature plays a key role in supporting reasoning, knowledge transfer, and the construction of mental models, making definitions an essential component of both human learning and machine understanding. Table \ref{tab:DSR} is an illustrative example of a definition annotated with definition semantic roles (DSR) \cite{silva2018recognizing}.

\begin{table}[ht!]
\begin{tcolorbox}[fontupper=\small, fontlower=\small, middle=0.3cm, top=1pt, bottom=1pt]
Definition: \textcolor{blue}{Homologous recombination repair} is a \textcolor{red}{DNA repair} \underline{process} \textcolor{green}{that includes the invasion of an undamaged DNA molecule by a damaged molecule of identical or very similar sequence.}

DSR: \textcolor{blue}{DEFINIENDUM} \textcolor{red}{DIFFERENTIA QUALITY} \underline{SUPERTYPE} \textcolor{green}{DIFFERENTIA-EVENT}
\end{tcolorbox}
\caption{An example of a definition and its DSR annotation. The definition of DSR is provided in Table \ref{tab:dsr_define}.}
\label{tab:DSR}
\end{table}

\begin{table}[h!]
\centering
\small
\begin{tabular}{lp{10cm}} \toprule
\textbf{Role} & \textbf{Description} \\
\hline
Supertype & the immediate or ancestral entity’s superclass \\
\hline
Differentia quality & a quality that distinguishes the entity from the others under the same supertype \\
\hline
Differentia event & an event (action, state or process) in which the entity participates and that is mandatory to distinguish it from the others under the same supertype \\
\hline
Event location & the location of a differentia event \\
\hline
Event time & the time in which a differentia event happens \\
\hline
Origin location & the entity’s location of origin \\
\hline
Quality modifier & degree, frequency or manner modifiers that constrain a differentia quality \\
\hline
Purpose & the main goal of the entity’s existence or occurrence \\
\hline
Associated fact & a fact whose occurrence is/was linked to the entity’s existence or occurrence \\
\hline
Accessory determiner & a determiner expression that doesn’t constrain the supertype-differentia scope \\
\hline
Accessory quality & a quality that is not essential to characterize the entity \\
\hline
\textit{[Role]} particle & a particle, such as a phrasal verb complement, non-contiguous to the other role components \\
\bottomrule
\end{tabular}
\caption{Semantic roles for dictionary definitions. For more detailed illustrations and explanations of DSR, readers are referred to the original work by \citet{silva2018recognizing}.}\label{tab:dsr_define}
\end{table}

Natural language definitions are widely applied in the task, such as definition modelling \cite{noraset2016definition} where the goal is to generate a natural language definition given the word to be defined (definiendum) or its word embedding, and reversed dictionary (definition-to-vector) \cite{mickus-etal-2022-semeval}. These tasks are closely connected to compositional semantics, as they require the model to parse, interpret, and reconstruct meaning based on the combination of words and structures in the input or output. By grounding lexical meaning in natural language, these approaches contribute to more interpretable and semantically aware representation learning.

\paragraph{Natural Language Explanations.} Natural language explanations, such as atomic sentence: \textit{animal is a kind of living thing} and condition sentence: \textit{if something is a living thing, it needs water to survival}, are designed for formal/compositional/symbolic semantic inference task in natural language form, which provides a semantically complex and yet controlled experimental setting, containing a both well-scoped and diverse set of target concepts and sentence structures. A number of scientific or commonsense reasoning benchmarks leverage natural language explanations as core resources, including WorldTree \cite{jansen2018worldtree}, EntailmentBank \cite{dalvi2021explaining}, and Zebra \cite{molfese-etal-2024-zebra}. 

In this thesis, we use explanations from both EntailmentBank and WorldTree. Table \ref{tab:visua_details} and \ref{tab:srl_silva} summarise the topic information and statistical information of the SRL in the Natural Language Explanations dataset, respectively.

Both definitions and explanations, reflecting aspects of human knowledge organisation, cognition, and inference, have clear semantic structure and regularity, which makes them central to the study of formal semantics.
\begin{table}[ht!]
\begin{tcolorbox}[fontupper=\small, fontlower=\small, middle=0.3cm, top=1pt, bottom=1pt]
Single Attribute Style-Transfer: I \textcolor{blue}{love} this movie $\rightarrow$ I \textcolor{red}{hate} this movie \\
Multiple Attributes Style-Transfer: I \textcolor{blue}{love} \textcolor{green}{Physics} $\rightarrow$ I \textcolor{red}{hate} \textcolor{green}{Math}
\end{tcolorbox}
\caption{An example of the Style-Transfer task.}
\label{tab:others}
\end{table}
\paragraph{Other Resources.} In addition to the two main types of expressions we focus on in formal semantics, the NLP domain includes a wide range of corpora that explore compositional semantics across various downstream tasks, such as style transfer and paraphrasing. Table \ref{tab:others} illustrates an example of semantics that can be composed in the style-transfer task.

These datasets often capture nuanced variations in meaning associated with task-related attributes such as sentiment, and serve as valuable resources for investigating how language models handle semantic composition in more flexible and context-dependent settings. Typical corpora include \textit{Yelp reviews}~\cite{shen2017style}, \textit{Yahoo answers}~\cite{yang2017improved}, \textit{dSentences}~\footnote{\url{https://github.com/mcharrak/discreteVAE}}, \textit{IMDb movie review}~\cite{maas-etal-2011-learning}, \textit{AGNews dataset}~\cite{zhang2015character}, and \textit{Jigsaw Toxic Comment Classification Challenge Dataset}~\footnote{\url{https://www.kaggle.com/c/jigsaw-toxic-comment-classification-challenge/}}. 
\begin{table*}[ht!]
\resizebox{\textwidth}{!}{
\begin{forest}
for tree={
    grow=east,
    draw,
    rounded corners,
    node options={align=center, font=\small},
    edge={->, >=latex},
    parent anchor=east,
    child anchor=west,
    l sep+=10pt,
    s sep+=5pt,
    anchor=base west,
}
[Latent Semantic Geometry
    [Sparse AutoEncoder, fill=yellow!50
        [Informal Linguistic (or Concept) Attribute Space, fill=yellow!20
            [{\citet{elhage2022superposition,cunningham2023sparse,sae_jack,deng2025unveiling};\\ \citet{gadgil2025ensembling,minder2025robustly,he2025sae}}]
        ]
        [Formal Linguistic Attribute Space, fill=yellow!20
            [{\citet{jing2025sparse}}]
        ]
    ]
    [Vector Quantised VAE, fill=cyan!60
        [Semantic (or Concept) Attribute Space, fill=cyan!30
            [{\citet{roy-grangier-2019-unsupervised,garg2025crosslayerdiscreteconceptdiscovery}*; \citet{zhang-etal-2024-improving}} \textbf{\textcolor{blue}{(our contribution)}}]
        ]
        [Syntax-Semantic Attribute Space, fill=cyan!30
            [{\citet{hosking-lapata-2021-factorising}}]
        ]
        [Hierarchical Syntactic Attribute Space, fill=cyan!30
            [{\citet{hosking-etal-2022-hierarchical}}]
        ]
        [Topic Attribute Space, fill=cyan!30
            [{\citet{yoo2024topic}}]
        ]
    ]
    [Variational AutoEncoder (VAE), fill=green!40
        [Reasoning Attribute Space, fill=green!20
                [Natural Language, fill=green!10
                    [{\citet{zhang2024controllablenaturallanguageinference,zhang2025learningdisentanglelatentreasoning} \textbf{\textcolor{blue}{(our contribution)}}}]
                ]
                [{Others (e.g., Vision)}, fill=green!10
                    [{\citet{pmlr-v80-sun18a, bonnet2024searchinglatentprogramspaces, vankrieken2025neurosymbolicdiffusionmodels}}]
                ]
        ]
        [Semantic Attribute Space, fill=green!20
            [{\citet{carvalho2023learning}; \citet{zhang-etal-2024-learning}; \citet{zhang2024formalsemanticgeometrytransformerbased}} \textbf{\textcolor{blue}{(our contribution)}}]
        ]
        [Syntax-Semantic Attribute Space, fill=green!20
            [{\citet{zhang-etal-2019-syntax-infused,bao2019generating,chen-etal-2019-multi,huang-etal-2021-disentangling,zhang-etal-2024-graph}} \textbf{\textcolor{blue}{(our contribution)}}]
        ]
        [Syntax Attribute Space, fill=green!20
            [\citet{shi2021gtae,huang2021generating,mercatali-freitas-2021-disentangling-generative,felhi2022towards}]
        ]
        [Style-transfer Attribute Space, fill=green!20
            [Single Attribute, fill=green!10
            [{\citet{https://doi.org/10.48550/arxiv.1703.00955,fang2019implicit,john2019disentangled,mai-etal-2020-plug,huang2020cycle}; \\ \citet{duan-etal-2020-pre,shen2020educating,liu2020revision,xu2021vae,li-etal-2022-variational-autoencoder}; \\
                \citet{yoshioka2022spoken,liu-etal-2023-composable}; \citet{zhang2024truthx}*}]
            ]
            [Multiple Attributes, fill=green!10
            [{\citet{hu2021causal,vasilakes-etal-2022-learning,gu-etal-2022-distributional,gu-etal-2023-controllable}}]
            ]
        ]
    ]
]
\end{forest}
}
\caption{Taxonomy for latent semantic geometry via autoencoder architecture. For VAE and VQ-VAE, * indicates configurations that are plug-and-play inside LMs, rather than considering the LM as a single decoder.}
\label{tab:tax}
\end{table*}

\section{Approach} \label{sec:latent_props}

This section reviews the major architectural approaches for learning semantic representation spaces, including VAEs, VQ-VAEs, and SAEs. While VAEs are typically employed to model sentence-level semantic spaces, VQ-VAEs and SAEs are more suited for capturing token-level semantic representations.

\subsection{Sentence-level Semantic Space}

\begin{figure*}[t]
    \includegraphics[width=0.99\linewidth]{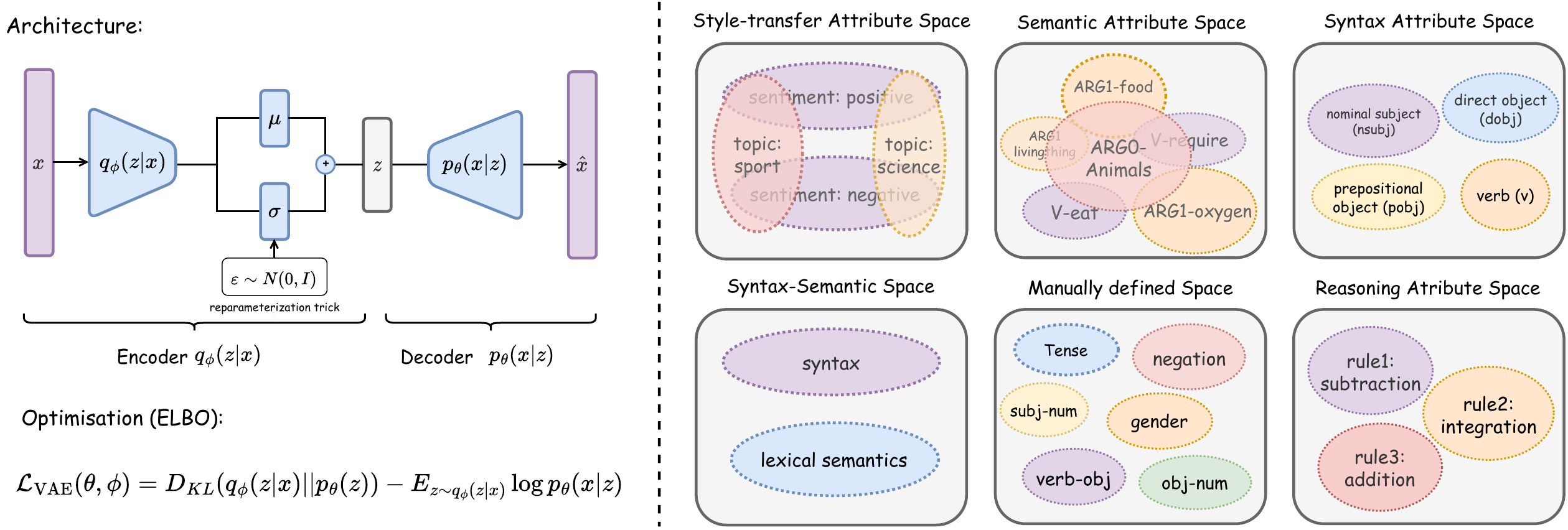}
    \caption{VAE architecture where the latent space $z$ can present different geometrical properties.}
    \label{fig:vae}
\end{figure*}

According to their distinct geometrical properties, we categorise these approaches into the following types of semantic representation spaces:

\textbf{\textit{(1) Style-transfer Attribute Space.}} Style transfer tasks aim to modify sentences with respect to specific informal linguistic attributes, such as \textit{sentiment}, \textit{formality}, or \textit{affirmation/negation}, while preserving the core semantic content. A number of studies have focused on sentiment control, learning latent space to isolate and manipulate emotional tone \cite{john2019disentangled, li-etal-2022-variational-autoencoder, liu2020revision}. Other works target multi-attribute control, where models are trained to simultaneously manage multiple stylistic dimensions within the latent space \cite{liu-etal-2023-composable, gu-etal-2023-controllable, gu-etal-2022-distributional, hu2021causal, vasilakes-etal-2022-learning}. These approaches leverage separated feature representations to enable fine-grained and interpretable text transformations.

Next, we introduce four types of latent space geometries that are designed to encode more formal semantic features (such as syntax, SRL, and reasoning logics).

\textbf{\textit{(2) Syntax Attribute Space.}} Syntactic roles can be considered as generative factors within the latent space of VAEs, where each factor captures a distinct aspect of sentence structure or grammatical function \cite{mercatali-freitas-2021-disentangling-generative,felhi2022towards}. By disentangling these roles in the latent representation, such models aim to isolate syntactic information (e.g., subject, object, modifier) from lexical or semantic content, thereby enabling more structured and interpretable generation and manipulation of language.

\textbf{\textit{(3) Semantic Attribute Space.}} Defining generative factors in terms of semantic roles has recently been explored as a promising direction for structuring latent spaces. \citet{carvalho2023learning} defined the generative factors of natural language definition, through Definition Semantic Roles (DSR)~\cite{silva2018recognizing}, as the prior knowledge to shape the latent space to assist Definition Modelling tasks. In detail, they deploy a DSR-supervised approach to disentangle the latent space, where different dimensions are expected to capture distinct DSR. 

However, focusing solely on structural representations while disregarding lexical content makes latent space learning particularly challenging \textcolor{blue}{\textbf{(Research Gap~1)}}. In this thesis, \textbf{\textcolor{black}{Chapters~\ref{cha:geo} and~\ref{cha:dis}}} provide a novel lens to formalise the latent space geometry from the perspective of \textit{Argument Structure Theory} \cite{jackendoff1992semantic, levin1993english, rappaport2008english}. In this framework, each semantic role–word content combination is represented as a convex cone within the latent space, enabling localised semantic control and enhancing interpretability.

\textbf{\textit{(4) Syntax-Semantic Attribute Space.}} Several studies have explored methods for learning latent spaces in which syntactic structure and lexical semantics are disentangled. For example, \citet{zhang-etal-2019-syntax-infused} used two separate encoders to independently encode semantic and syntactic information. In contrast, \cite{bao2019generating} utilised a single encoder within a VAE framework to disentangle semantic and syntactic representations through multi-task optimisation. 

However, relatively few studies have investigated syntax representation learning using graph-based Transformer VAE architectures, where graph encoders are particularly well-suited for capturing topological structures, thereby yielding latent spaces with improved geometric properties \textcolor{blue}{\textbf{(Research Gap~2)}}. To address this research gap, \textbf{Chapter~\ref{cha:syntax}} introduces a novel dual-encoder VAE model that integrates a graph-based encoder with a semantic encoder. This design enables explicit modelling of syntactic and semantic features while alleviating the information bottleneck associated with single-encoder approaches.

\textbf{\textit{(5) Reasoning Attribute Space.}} In the vision domain, prior research has examined reasoning patterns over visual data \cite{pmlr-v80-sun18a, bonnet2024searchinglatentprogramspaces, vankrieken2025neurosymbolicdiffusionmodels}. These studies demonstrate the potential for capturing structured reasoning in latent representations. In contrast, reasoning patterns in NLI tasks remain comparatively underexplored \textcolor{blue}{\textbf{(Research Gap~3)}}. To address this gap, \textbf{Chapters~\ref{cha:reason} and~\ref{cha:reasondis}} investigate how reasoning structures can be encoded within the latent space of VAEs for NLI~\cite{zhang2024controllablenaturallanguageinference, zhang2025learningdisentanglelatentreasoning}. Our approach aims to organise the latent space to capture inferential relationships, such as mathematical reasoning rules, enabling the model to represent not only sentence-level semantics but also the logical relations that connect them.

\subsection{Token-level Semantic Space}

\subsubsection{Vector Quantised VAEs}
\begin{figure*}[ht!]
    \includegraphics[width=0.99\linewidth]{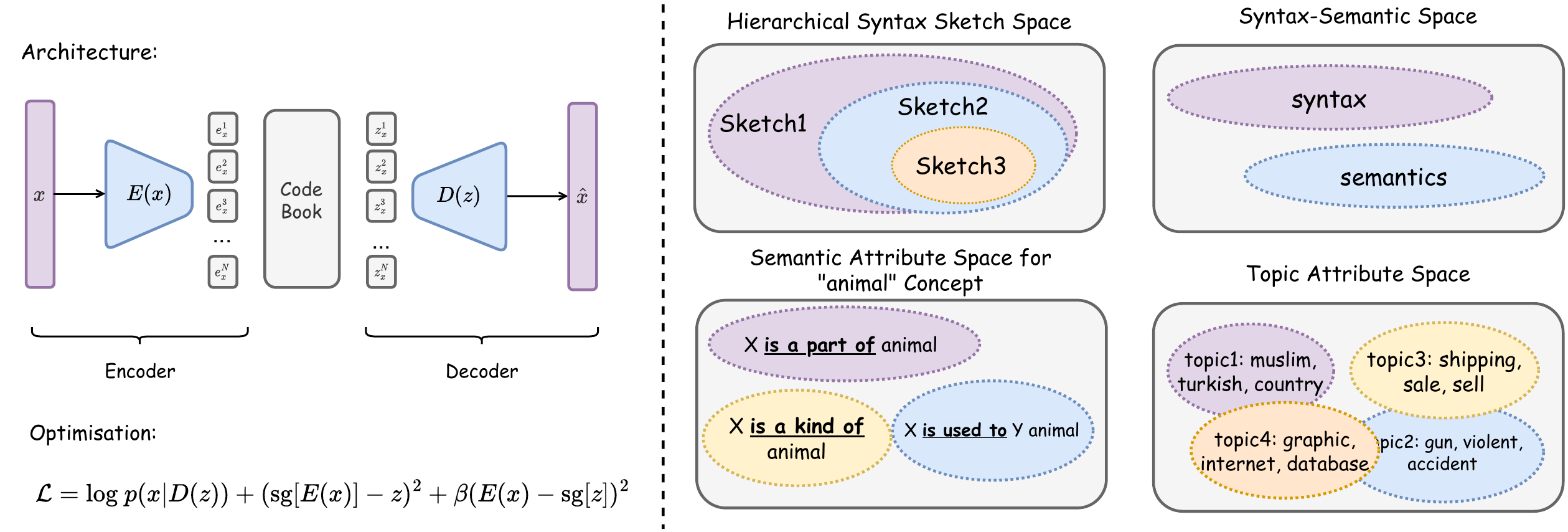}
    \caption{VQ-VAE architecture where the latent space $z$ can present different geometrical properties.}
    \label{fig:vqvae}
\end{figure*}



Recent extensions of the VQ-VAE framework have demonstrated its effectiveness in learning discrete latent semantic spaces tailored to specific NLP tasks, delivering different geometrical properties.

\textbf{\textit{(1) Topic Attribute Space.}} Topic-VQVAE~\cite{yoo2024topic} focuses on document-level generation by learning topic-aware latent codes. It leverages vector quantisation to discover discrete topic representations that guide the generation process, enabling more coherent and thematically consistent document outputs.

\textbf{\textit{(2) Hierarchical Syntactic Attribute Space.}}  
Natural language exhibits a hierarchical structure across multiple linguistic levels, particularly at the syntactic level, where constituents recursively contain sub-constituents. To capture this property, HRQ-VAE \cite{hosking-etal-2022-hierarchical} introduces a hierarchical syntactic sketch space, which represents syntax in a nested, discrete latent structure. This approach enables the model to better reflect the recursive and compositional nature of syntactic patterns in language generation and understanding.

\textbf{\textit{(3) Syntax-Semantic Attribute Space.}}  
\citet{hosking-lapata-2021-factorising} propose a model that factorises sentence meaning into discrete syntactic and continuous semantic components. Focusing on the task of question paraphrasing, their model reconstructs a question by combining the surface form from an exemplar with the semantic content from a paraphrase. This dual-space approach enables more controlled generation, allowing for explicit manipulation of syntax and semantics independently while preserving overall meaning.

\textbf{\textit{(4) Semantic Attribute Space.}} 
To investigate the geometrical properties of discrete latent spaces over TLMs \textcolor{blue}{\textbf{(Research Gap~4)}}, \textbf{\textcolor{black}{Chapter~\ref{cha:discrete}}} integrates the VQ-VAE framework into the T5 encoder-decoder architecture to learn a discrete semantic space. The authors further propose techniques such as traversal and interpolation to explore and manipulate representations within this discrete space, targetting fine-grained semantic control.





\subsubsection{Sparse AutoEncoder} 
\begin{figure*}[ht!]
    \includegraphics[width=0.99\linewidth]{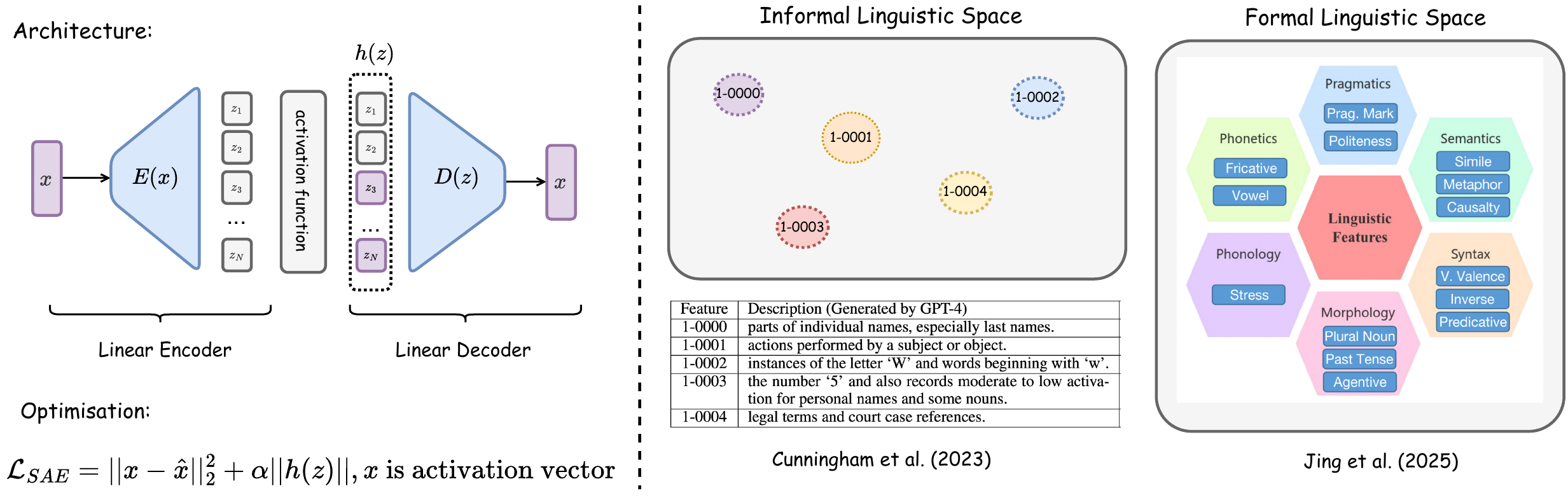}
    \caption{SAE architecture where the latent space $z$ can present different geometrical properties.}
    \label{fig:sae}
\end{figure*}
\paragraph{Overview.} Previous studies have revealed that TLMs often exhibit \textit{polysemanticity}, where a single neuron responds to multiple, semantically unrelated inputs~\cite{elhage2022superposition}. This entanglement of concepts within individual neurons poses a significant challenge to interpretability and fine-grained control. Sparse Autoencoders (SAEs) have emerged as a promising approach to mitigate this issue. SAEs are designed to learn a sparse, linear, and decomposable representation of the internal activations of TTLMs, where each dimension, termed a base vector, ideally corresponds to a distinct, interpretable linguistic feature. By enforcing a sparsity constraint, ensuring that only a small number of features are active at any given time, the model encourages each active feature to align with a distinct and interpretable concept~\cite{yun-etal-2021-transformer,cunningham2023sparse}.

Architecturally, a sparse autoencoder consists of a linear encoder that maps input representation to a higher-dimensional feature space and a linear decoder that reconstructs the input from this representation. To promote sparsity, an additional regularisation term, such as KL divergence or an L1 penalty, is applied to the hidden activations, encouraging most neurons to remain inactive (i.e., output values close to zero) for any given input. This constraint forces the model to learn meaningful, distinct features and prevents trivial identity mappings. A comprehensive survey of SAEs, including architectural variants and optimisation strategies, is provided by \citet{shu2025survey}.

\paragraph{Latent Space Geometry.} Recent studies have shown that SAEs exhibit meaningful linguistic geometrical properties, which can be leveraged to control language model generation. We categorise the geometrical properties into two main types.

\textbf{\textit{(1) Informal Linguistic Attribute Space.}}  
This refers to the organisation of latent dimensions corresponding to loosely defined or stylistic language properties. SAEs have been shown to uncover interpretable features aligned with lexical semantic attributes, enabling controlled generation and feature-level interventions~\cite{elhage2022superposition, cunningham2023sparse}. Additionally, recent work has investigated the behaviour of LLMs in multilingual settings, revealing that SAEs can also capture language-specific features and align them across languages in the latent space~\cite{deng2025unveiling}. Recent work further investigates task-specific attribute separation in open-ended generation control, such as sentiment, truthfulness, and politics polarity steering tasks~\cite{he2025sae}.

\textbf{\textit{(2) Formal Linguistic Attribute Space.}} Recent work investigated the linguistic capabilities of LLMs~\cite{jing2025sparse}. The authors extract fine-grained linguistic features across six dimensions, including phonetics, phonology, morphology, syntax, semantics, and pragmatics, by encoding LLM hidden states into sparse, interpretable representations. They evaluate these features using custom datasets, including minimal contrast and counterfactual sentence pairs, and define two metrics: Feature Representation Confidence (FRC) and Feature Intervention Confidence (FIC) to assess how well features correspond to and control linguistic phenomena. Through targeted interventions, they demonstrate that specific features causally affect LLM outputs, providing insights into deep semantic processing and cross-layer linguistic representation.

\section{Conclusion} \label{sec:concl}

In this survey, we offer a novel perspective on latent space geometry through the lens of compositional semantics, a direction we refer to as \textit{semantic representation learning}. This direction enables a bridge between symbolic and distributional semantics, helping to mitigate the gap between them. By integrating structured semantic knowledge into learned representations, it holds the potential to address core challenges in machine learning, such as mechanistic interpretability, composition and generalisation. We review and compare three mainstream autoencoder architectures, including VAE, VQ-VAE, and SAE, and examine the distinctive latent geometries they induce in relation to semantic structure and interpretability. 




\chapter{Formal Semantic Geometry} \label{cha:geo}
This chapter investigates \questionA{} proposing a novel theoretical framework to bridge the gap between deep latent semantics and formal linguistic representations, and a practical framework based on Language VAE architecture.

\section{Introduction}
Language Models (LMs) have provided a flexible scaling-up foundation for addressing a diverse spectrum of tasks \citep{touvron2023llama}. Nonetheless, the question remains: can we develop language representations or models that offer more granular levels of control and interpretation from the perspective of ``formal/structural'' semantics? Addressing this question will enable us to enhance the controllability, interpretability, and safety of LMs.

Formal semantics, which provides a canonical, granular, and rigid representation, has been investigated for decades, such as Montague Semantics \citep{dowty2012introduction}, Davidsonian Semantics \citep{davidson1967logical}, Abstract Meaning Representation \citep{banarescu2013abstract}, Semantic Role Labelling \citep{palmer2010semantic}, and Argument Structure Theory (AST \citep{jackendoff1992semantic}). One typical characteristic of such formal semantics is the \textit{localisation} or \textit{composition} property. For example, in the sentence: \textit{animals require oxygen for survival}, the words are functionally combined into sentence semantics: 
$\lambda x (\text{animals}(x) \rightarrow \text{require}(x, \text{oxygen}))$
where $x$ is the variable of any entity within a logical structure. In this case, we can localise the sentence semantics by replacing $x$ with \textit{birds}, etc. This localised process indicates the interpretation in Cognitive Science \citep{smolensky2006harmony,lees1957syntactic}. However, such localisation is precisely what current distributional semantics lack, thereby limiting their controllability and interpretability.
\begin{figure}[t]
    \centering
    \includegraphics[width=0.7\linewidth]{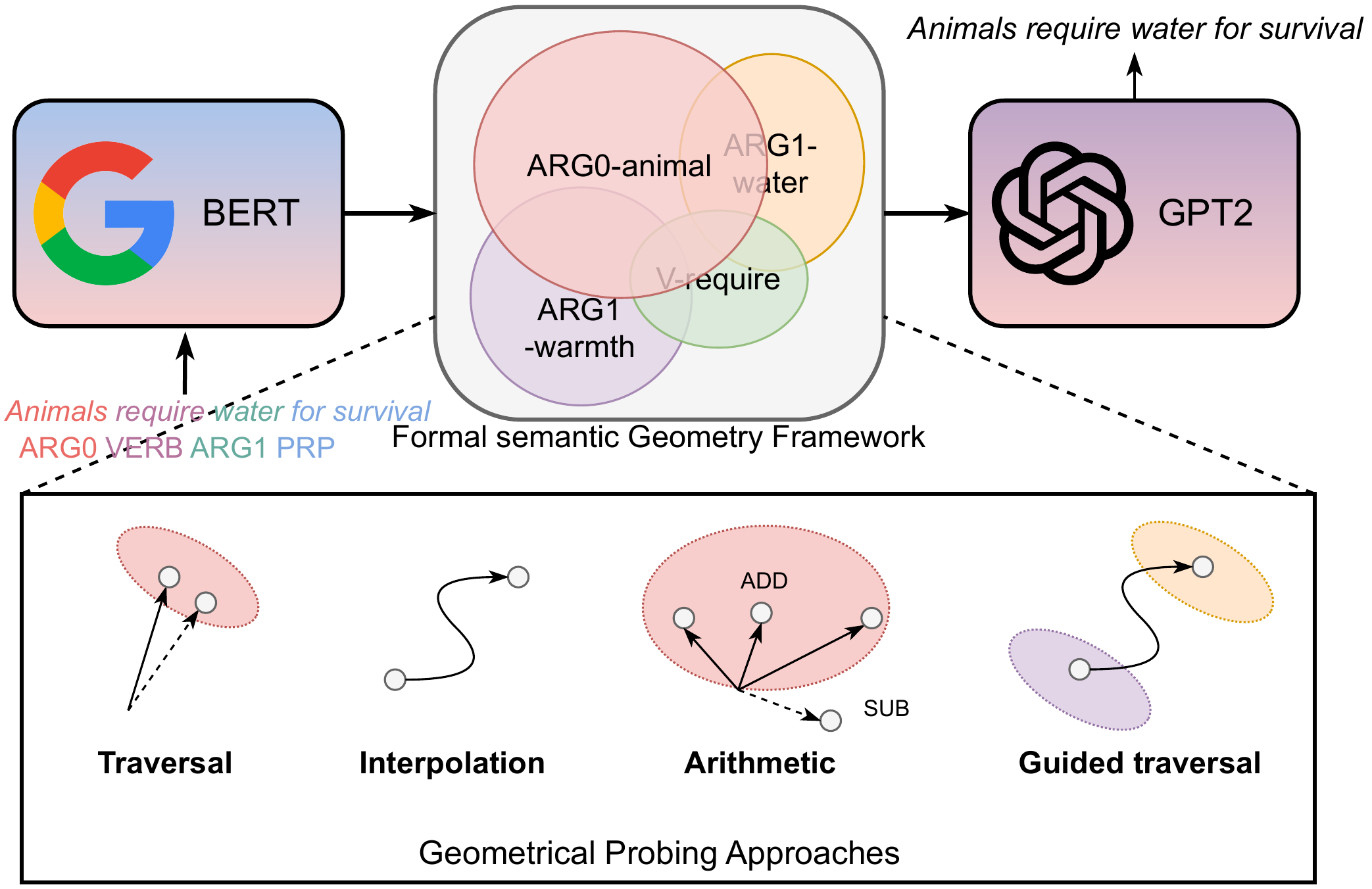}
    \caption{Overview: latent sentence semantics can be decomposed into \textit{semantic role-word content} features.}
    \label{fig:e2e_arch}
\end{figure}

Disentanglement \citep{bengio2013deep}, which refers to the feature-dimension alignment, can potentially provide such localisation, which has been widely investigated to localise image features, such as \textit{nose} in facial images  \citep{esser2020disentangling,jeon2019efficient, liu2021smoothing}. In Transformers \citep{vaswani2017attention}, however, token embeddings, residual stream, and attention are suffered from the polysemanticity phenomenon, meaning that multiple dimensions contribute to a feature. Although some prior studies explored the possibility of language disentanglement, most are focused on coarse-grained/task-specific semantic features, such as sentiment, within the context of style-transfer tasks \citep{john2019disentangled,bao2019generating,hu2021causal,vasilakes-etal-2022-learning,gu-etal-2022-distributional,liu-etal-2023-composable,gu-etal-2023-controllable}. 

 In this work, we focus on the localisation of \textit{general} semantic features of sentences over distributional space to shorten the gap between deep latent semantics and formal linguistic representations \citep{10.3115/1075218.1075283,banarescu2013abstract,mitchell2023we}, integrating the flexibility of distributional-neural models with the properties of linguistically grounded representations, facilitating both interpretability and generative control from the perspective of formal semantics. We specifically choose the conceptually dense explanatory sentences from WorldTree \citep{jansen2018worldtree} due to their clear formal semantic representation designed in the Explanatory Reasoning task.

In the NLP domain, Variational AutoEncoders (VAEs \cite{https://doi.org/10.48550/arxiv.1312.6114}) have been recognised as a prominent foundation for investigating generation control and interpretation through the observable low-dimensional smooth and regular latent spaces (e.g., std Gaussian space) \citep{john2019disentangled,li-etal-2022-variational-autoencoder,bao2019generating,mercatali2021disentangling, felhi2022towards,vasilakes-etal-2022-learning}. Therefore, we probe the localisation property of formal semantics over latent sentence spaces under the VAE architecture. Specifically:

\textbf{(1)} We first propose a geometrical framework to present the formal semantic features of sentences as \textit{semantic role-word content} pairs (denoted as role-content) from the perspective of AST \cite{jackendoff1992semantic} within the compositional distributional model \cite{clark2008compositional}. Subsequently, \textbf{(2)} we introduce a supervised approach for learning the role-content features of explanatory sentences in latent spaces. \textbf{(3)} Additionally, we propose a method to control sentence generation by navigating the sentence vectors across different role-content features within our geometric framework. \textbf{(4)} Our findings reveal that the role-content features are encoded as a convex cone in the latent sentence space (Figure \ref{fig:e2e_arch}). This semantic geometry facilitates the localisation of sentence generation by enabling the manipulation of sentence vectors through traversal and arithmetic operations within the latent space.

\section{Formal Semantic Geometry} \label{sec:latent_props}
In this section, we first define the sentence semantic features as \textit{semantic role - word content} from the perspective of formal semantics. Then, we link the semantic features with distributional vector spaces in which each \textit{semantic role - word content} is encoded as a convex cone, as shown in Figure \ref{fig:e2e_arch}. 
\paragraph{Formal semantic features.} For formal / structural semantics, \textit{Argument Structure Theory (AST)} \cite{jackendoff1992semantic, levin1993english, rappaport2008english} provides a model for representing sentence structure and meaning of sentences in terms of the interface between the their syntactic structure and the associated semantic roles of the arguments within those sentences. It delineates how verbs define the organisation of their associated arguments and the reflection of this organisation in a sentence's syntactic realisation. AST abstracts sentences as predicate-argument structures, where the predicate $p$ (associated with the verb) has a set of associated arguments $arg_i$, where each argument has an associated positional component $i$ and a thematic/semantic roles $r_i$, the latter categorising the semantic functions of arguments in relation to the verb (e.g. agent, patient, theme, instrument). In the context of this work, the AST predicate-argument representation is associated with a lexical-semantic representation of the content $c_i$ of the term $t_i$.
\begin{table}[ht!]
\centering
\resizebox{7cm}{!}{
\begin{forest}
for tree={
    if n children=0{
      inner sep=1pt,
      align=center,
      base=top,
      s sep=0mm,
      l sep=100mm,
      minimum size=1em,
      tier=terminal
    }{},
  }
  [S [NP [NN [ \textit{ARG0 (Agent)} [\textit{animals}] ] ]] [VP [VBZ [ \textit{PRED} [\textit{require}]] ] [NP [ NP [\textit{ARG1 (Patient)} [\textit{oxygen}] ] ] 
  [PP [ARGM-PRP [\textit{PRP} [\textit{for}] ] 
  [\textit{PRP} [\textit{survival}] ]
  ]]]]]
\end{forest}
}
\end{table}

In this work, we simplify and particularise the relationship between the argument structure and the distributional lexical semantic representation as a \textit{role-content} relation, where the structural syntactic/semantic relationship is defined by its shallow semantics, i.e. as the composition of the content of the terms, their position in the predicate-argument (PArg) structure ($arg_i$) and their semantic roles (SRs) ($r_i$: $pred$, $arg$), as described below:
$$\underbrace{animals}_{ARG0}~\underbrace{require}_{PRED}~\underbrace{oxygen}_{ARG1}~\underbrace{for~survival}_{ARGM-PRP}$$
Therefore, we define the semantics of sentences, $sem(s)$, as the compositions between \textit{role-content}, which can be described as follows:
$$
sem(s) = \underbrace{t_1({c_1}, {r_1})}_{i.e., ARG0-animals} \oplus \dots \oplus \underbrace{t_i({c_i}, {r_i})}_{PRP-survival}
$$
Where $t_i({c_i}, {r_i})=c_i \otimes r_i$ represents the semantics of term $t_i$ with content $c_i$ (i.e., \textit{animals}) and SRL $r_i$ (i.e., \textit{ARG0}) in context $s$. $\otimes$: connects the meanings of words with their roles, using the compositional-distributional semantics notation of \cite{smolensky2006harmonic,Clark2007CombiningSA,clark2008compositional}. $\oplus$: connects the lexical semantics (word content + structural role) to form the sentence semantics. To deliver the localisation or composition property, the sentence semantics should be able to present separation or disentanglement under connector $\oplus$. E.g., replacing \textit{ARG0-animals} with \textit{ARG0-fishes}. 

\begin{figure}[t]
    \centering
    \includegraphics[width=0.7\columnwidth]{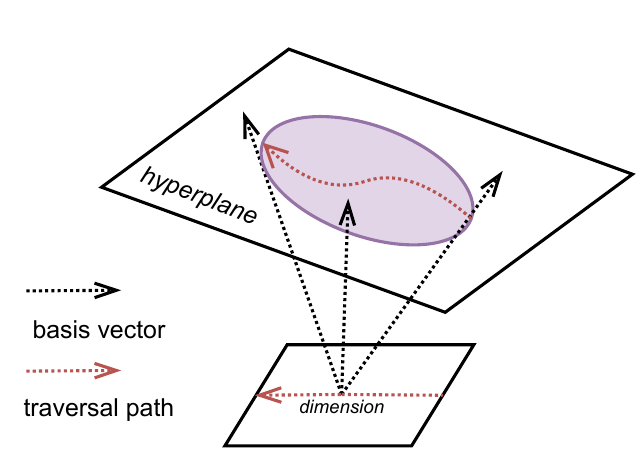}
    \caption{Algorithm \ref{alg:guide}: by modifying the latent dimensions, we can control the movement of latent vectors over latent space.}
    \label{fig:guide_trav}
\end{figure}

\paragraph{Formal semantic features in vector space.} After defining the semantic features of sentences, we propose the concept of a \textit{convex cone of semantic feature}. In linear algebra, a \textit{cone} refers to a subset of a vector space that is convex if any $\alpha \overrightarrow{v_i} + \beta \overrightarrow{v_j}$ if any $\overrightarrow{v_i}$ and $\overrightarrow{v_j}$ belong to it. $\alpha$ and $\beta$ are positive scalars. Formally, the definition of convex cone, $C$, is described as a set of vectors:
$
C = \{ x \in V | x = \sum_{i=1}^n \alpha_i v_i, \alpha_i \geq 0, v_i \in R \}
$
where $x$ is an element vector in vector space $\mathbb{R}$, $v_i$ are the basis vectors. $\alpha_i$ are non-negative scalars. In this context, we consider each \textit{role-content} feature as a convex cone, $C$, corresponding to a hypersolid in high-dimensional vector space:
$
C_{c_i, r_i} = \{ t({c_i}, {r_i}) | t({c_i}, {r_i}) \in sem(s), s \in \textit{corpus} \}
$
where $t({c_i}, {r_i})$ represents the basis vector in $C_{c_i, r_i}$ (Figure \ref{fig:guide_trav}). According to set theory, we can define the formal semantic space as follows:

\textit{\textbf{Assumption1:} The sentence semantic space is the union of all unique $C_{c_i, r_i}$ convex cones:}
$$
C_{c_1, r_1} \cup C_{c_2, r_2} \cup \dots \cup C_{c_{V^{(c)}}, r_{V^{(r)}}}
$$
$V$ is the vocabulary of a corpus. Based on Assumption1, we can establish:

\textit{\textbf{Proposition1:} The geometrical location of sentence semantic vectors, $sem(s)$, can be determined by the intersection of different $C_{c_i, r_i}$:}
\[
\begin{aligned}
    sem(s) & = t_1({c_1}, {r_1}) \oplus \dots \oplus t_i({c_{i}}, {r_{i}}) \\
    &= \{ t_1({c_1}, {r_1})\} \oplus \dots \oplus \{t_i({c_{i}}, {r_{i}}) \} \\
    &= C_{c_1, r_1} \cap C_{c_2, r_2} \cap \dots \cap C_{c_{i}, r_{i}}
\end{aligned}
\]
\section{Geometrical Formal Semantic Control}
In this section, we first show that our formal semantic geometry can interpret sentence generation, such as arithmetic \cite{shen2020educating}, and extend the ``Linear Representation Hypothesis''. Then, we propose a new semantic control approach, which recursively traverses the latent dimensions to probe the semantic geometry over latent spaces.

\paragraph{Geometrical algebra interpretability.} Arithmetic has been considered a common way to control word or sentence semantics over latent spaces \cite{mikolov-etal-2013-linguistic}. E.g., the addition operation can steer the sentence semantics \cite{shen2020educating,mercatali2021disentangling,liu2023context}, or linear interpolation can generate smooth intermediate sentences \cite{hu-etal-2022-fuse}. However, they lack an explanation for these phenomena. We show that our geometrical framework can provide an intuitive explanation for these phenomena. 

For linear interpolation, for example, it takes two sentences $ x_1 $ and $ x_2 $ and obtains latent vectors $ z_1 $ and $ z_2 $, respectively. It interpolates a path $ z_k = z_1 \cdot (1 - k) + z_2 \cdot k$  with $k$ increased from $ 0 $ to $ 1 $ by a step size of $ 0.1 $. 
Given two sentences with one role-content set overlap, $C_{c_{j}, r_{j}}$. We can describe:
\[
\begin{aligned}
    &sem(s_1) \cap sem(s_2) \\
    &=\{ C^{s_1}_{c_1, r_1} \cap \dots \cap C^{s_1}_{c_{i}, r_{i}} \} \cap \{C^{s_2}_{c_1, r_1} \cap \dots \cap C^{s_2}_{c_{i}, r_{i}} \} \\
    &=\{ C^{s_1}_{c_1, r_1} \cap \dots \cap C^{s_2}_{c_{i}, r_{i}} \} \cap C^{s_{1(2)}}_{c_{j}, r_{j}}
\end{aligned}
\]
According to the definition of convex cone, if $z_1$ and $z_2$ are left in $C^{s_{1(2)}}_{c_{j}, r_{j}}$, the weighted sum vector, $z_t$, is also in $C^{s_{1(2)}}_{c_{j}, r_{j}}$. Therefore, the intermediate sentence semantics can be described as:
\[
\begin{aligned}
    &sem(s^t_{1 \rightarrow 2}) \\
    &=(1-k) \times sem(s_1) + k \times sem(s_2) \\
    &=\{ \{z_1 \cdot (1 - k) + z_2 \cdot k \}, \dots \{\dots\}\} \cap C^{s_{1(2)}}_{c_{j}, r_{j}} \\
\end{aligned}
\]
That is, the intermediate sentences will hold the $\{c_{j}, r_{j}\}$ information during interpolation. 

\paragraph{Linear representation hypothesis.} ``Linear representation hypothesis'' refers to high-level concepts being represented linearly as directions in representation space, which has been widely evaluated to interpret Large LMs' mechanism \cite{marks2023geometry,xie2021explanation,wang2024large,jiang2024origins,park2023linear,park2024geometry}. However, a main challenge for this hypothesis is that it’s not clear what constitutes a high-level concept.

Our geometrical framework can further support and extend this hypothesis by answering the questions: What and how are they ``linearly'' encoded? For example, given a set of $N$ atomic sentences: $s_i$: \textit{bird is a kind of living thing} varying the content of arg1. Their semantics can be described below:
\[
\begin{aligned}
& sem(s) =  \{ C^{s_i}_{c_i, arg1}, \dots \} \cap \dots \cap C_{living~ thing, arg2}, \text{where}~ c_i \in \{ \text{tiger}, \text{bird}, \dots \}
\end{aligned}
\]
In this case, the concept \textit{living thing} is encoded as a convex cone where all different $C^{s_i}_{c_i, arg1}$ contribute to its boundary, leading to a direction. The hierarchical relations between \textit{living thing} and \textit{bird, etc.} are determined by the convex cones \textit{is a kind of}.

\paragraph{Guided traversal.} Since we describe different sentence semantic features, $\{c_i, r_i\}$, as distinct convex cones, $C_{c_i, r_i}$, within a $N$-dimensional vector space, $V \in \mathbb{R}^{N}$, we can linearly divide each basis dimension, $i \in {\{1, \dots, N \}}$, into different value regions, $[a, b]^{(i)}$, based on minimal information entropy. Consequently, there is a sequence of dimensional subspaces for each semantic feature. Thus, movement between different $C_{c_i, r_i}$ regions can be achieved by moving out the dimensional regions within this sequence. This process can be implemented via a decision tree. In figure \ref{fig:decision_tree}, for example, we can move the sentence from $C_{pred, causes}$ to $C_{pred, means}$ by modifying the values started from \textit{dim 21 $\le -0.035$}, ..., ending at \textit{dim 10 $\le -1.11$}. By traversing the tree path, we can control the sentence generation by moving between convex cones, detailed in Algorithm \ref{alg:guide}.
\begin{algorithm}[ht!]
\caption{Guided latent space traversal} \label{alg:guide}
\begin{algorithmic}[1]
\State Datasets: $D = \{s_1, \dots, s_n\}$ 
\State Labels: $Y = \{y_1, \dots, y_n \}$, $y_i \in \{0, 1\}$ 
\State \textit{\# \textcolor{blue}{0:pred-causes}, \textcolor{red}{1:pred-means}}
\State Seed: $s=$ \textit{fire \textcolor{blue}{causes} chemical change}
\For{$s_i \in D$}
\State $z_i \leftarrow \text{Encoder}(s_i)$
\EndFor
\State $X \leftarrow \{z_1, \dots, z_n\}$
\State $\text{tree} \leftarrow \text{DecisionTreeClassifier}(X,Y)$
\State $\text{path} \leftarrow \text{filter}(\text{tree})$ \textit{\# choose the shortest path between $C_0$ and $C_1$}
\State $z \leftarrow \text{Encoder}(s)$
\For{$\text{node} \in \text{path}$}
    \State (dim, range, yes/no) $\leftarrow$ node 
    \State \textbf{if} in current branch \textbf{do} 
    \State ~ z[dim] $\leftarrow$ $v \notin \text{range}$ \textbf{if} yes \textbf{else} $v \in \text{range}$
    \State \textbf{else do} 
    \State ~ z[dim] $\leftarrow$ $v \in \text{range}$ \textbf{if} yes \textbf{else} $v \notin \text{range}$
    
\EndFor
\State $s \leftarrow$ \text{Decoder}(z) \textit{\# fire \textcolor{red}{means} chemical change}
\end{algorithmic}
\end{algorithm}
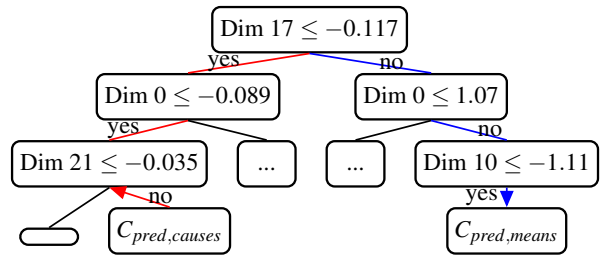
\begin{figure}[ht!]
\centering
\resizebox{7.9cm}{!}{
\begin{forest}
for tree={draw, rounded corners, node options={align=center, minimum width=1cm, line width=0.4mm}, s sep=5mm, l sep=3mm, edge={line width=0.3mm, - }}
[Dim 17 $\leq -0.117$,edge+={draw=blue},edge+={draw=red}
    [Dim 0 $\leq -0.089$,edge+={draw=red},edge+={draw=red}, edge label={node[midway, left, draw=none]{yes}}
        [Dim 21 $\leq -0.035$,edge+={draw=red}, edge label={node[midway, left, draw=none]{yes}}
            []
            [$C_{pred,causes}$, edge+={<-, draw=red}, edge label={node[midway, right, draw=none]{no}}]
        ]
        [...]
    ]
    [Dim 0 $\leq 1.07$,edge+={draw=blue},edge label={node[midway, right, draw=none]{no}}
        [...]
        [Dim 10 $\leq -1.11$,edge+={draw=blue}, edge label={node[midway, right, draw=none]{no}}
            [$C_{pred,means}$, edge+={->, draw=blue}, edge label={node[midway, left, draw=none]{yes}}]]
    ]
]
\end{forest}
}\caption{Traversal between different role-content sets by moving along the tree path.}
\label{fig:decision_tree}
\end{figure}

Based on our algorithm, we can use classification metrics as proxy metrics to evaluate latent space geometry. E.g., accuracy and recall for measuring feature \textit{separability} and \textit{density}. 
\section{SRL-Conditional VAE} \label{sec:disentang} 
In this section, we investigate the architecture of VAE to integrate the latent sentence space with LMs and propose a supervision approach to learn formal semantic geometry (i.e., role-content).
\paragraph{Model architecture.} We consider Optimus \cite{li2020optimus} as the foundation which used BERT and GPT2 as Encoder and Decoder, respectively. In detail, the sentence representation, $\text{Embed(x)}$, encoded from BERT[cls] will first transform into a Gaussian space by learning the parameters $\mu$ and $\sigma$ through multilayer perceptions $W_\mu$, $W_{\sigma}$. The final latent sentence representations can be obtained via: $z = W_{\mu} \times \text{Embed(x)} + W_{\sigma}$, which, as an additional Key and Value, is concatenated into the original Key and Value weights of GPT2, which can be described as:
$
\text{Attention}(Q, K, V) = \text{softmax}( \frac{Q [z;K]^T}{\sqrt{d}})[z;V]
$
where $Q$ has the shape $\mathbb{R}^{\text{seq} \times 64}$, $K, V$ has the shape $\mathbb{R}^{(\text{seq}+1) \times 64}$ (64 is dimension of GPT2 attention, $\text{seq}$ is sequence length). Since $Q$ represents the target, $K$ and $V$ represent the latent representations. By intervening the $KV$ with $z$, we can learn the transformation between the latent space and observation distribution. 

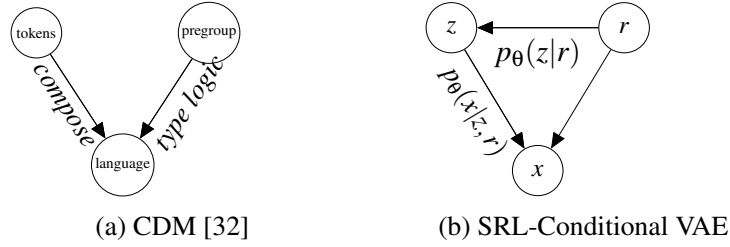
\begin{figure}[ht!]
\centering
\resizebox{10cm}{!}{
\begin{minipage}{4.5cm}
\begin{tikzpicture}
  \node[latent] (r) at (2.4,2) {\tiny pregroup};
  \node[latent] (z) at (0,2) {\tiny tokens};
  \node[latent] (x) at (1.2,0.15) {\tiny language};
  \edge {z, r} {x};
  \path (z) edge[->] node[pos=0.5,below,sloped] {\small \textit{compose}} (x);
  \path (r) edge[->] node[pos=0.4,below,sloped] {\small \textit{type logic}} (x);
\end{tikzpicture}
\subcaption{CDM \cite{clark2008compositional}}
\end{minipage}
\hspace{1cm}
\begin{minipage}{4.5cm}
\begin{tikzpicture}
  \node[latent] (r) at (2.4,2) {$r$};
  \node[latent] (z) at (0,2) {$z$};
  \node[latent] (x) at (1.2,0) {$x$};
  \edge {r} {z};
  \edge {z,r} {x};
  \path (r) edge[->] node[below] {$p_\theta(z|r)$} (z);
  \path (z) edge[->] node[below,sloped] {\small $p_\theta(x|z,r)$} (x);
\end{tikzpicture}
\subcaption{SRL-Conditional VAE}
\end{minipage}
\hspace*{1pt}
}
\caption{Comparison between Compositional Distributional Model (CDM) (left) and SRL-Conditional VAE (right).}
\label{fig:cvae}
\end{figure}

\paragraph{Optimisation.} It can be trained via the evidence lower bound (ELBO) on the log-likelihood of the data $x$ \cite{Kingma2014AutoEncodingVB}. To bind the word content and semantic role information in latent space, we conditionally inject the semantic role sequence into latent spaces where the latent space $z$ and semantic role $r$ are dependent. The joint distribution can be described as:
$$
P_{\theta}(x,r,z)=\underbrace{P_{\theta}(x|z,r)}_{likelihood} \times \underbrace{P_{\theta}(z|r)}_{prior} \times P(r)
$$
Figure~\ref{fig:cvae} depicts the corresponding computational graph, which is conceptually aligned with the Compositional Distributional Model (CDM) framework~\cite{clark2008compositional}.

Specifically, we first model the categorical structures by encoding the semantic roles sequence to learn the prior distribution with parameters $\mu^{(srl)}$ and $\sigma^{(srl)}$. Then, we jointly encode semantic roles and lexical tokens to learn the approximate posterior parameterised by $\mu$ and $\sigma$. By minimising the Kullback-Leibler (KL) divergence between prior and approximate posterior, the semantic features can be encoded in the latent sentence space.
 
Moreover, to avoid the KL vanishing problem, which refers to the KL term in the ELBO becoming very small or approaching zero, we select the cyclical schedule to gradually and cyclically increase weights of KL $\beta$ from 0 to 1 \cite{fu-etal-2019-cyclical} and a KL thresholding scheme \cite{li-etal-2019-surprisingly} that chooses the maximum between KL and threshold $\lambda$. The final objective function can be described as follows:
\begin{align*} 
\mathcal{L}_\text{CVAE} = - \mathbb{E}_{q_\phi(z|r,x)} \Big[ \log p_{\theta} ( x | z,r ) \Big] + \beta \sum_i \max \left[ \lambda , \text{KL} q_\phi(z_i|x, r) || p(z_i|r) \right ]
\end{align*}

Here, $q_\phi$ denotes the approximate posterior (i.e., the encoder). The index $i$ corresponds to the $i$-th latent dimension, for which a threshold is applied in order to prevent values from becoming excessively small.



\section{Empirical Evaluation}
In the experiment, we quantitatively and qualitatively evaluate the latent space geometry via 1. traversal, 2. arithmetic, and 3. guided traversal. 
\subsection{Latent Traversal}

\paragraph{Qualitative evaluation.} Traversal refers to the random walk over latent space. It can be done by decoding the latent vector in which each dimension is resampled and other dimensions are fixed \cite{higgins2016beta,kim2018disentangling,carvalho2022learning}. Given a latent vector from a ``seed'' sentence, we can traverse its neighbours to evaluate the geometry. \uline{As illustrated in Table~\ref{tab:trav_examples_main_geo}, those traversed sentences can hold the same content under different semantic roles as the input, such as \textit{automobile} in \textit{ARG1}, indicating \textit{role-content} feature separation in latent spaces} \textbf{(Finding~1)}.
\begin{table}[h]
\begin{tcolorbox}[fontupper=\small, fontlower=\small, middle=0.3cm, top=1pt, bottom=1pt]
\underline{an automobile is a kind of vehicle} \\ \\
\textcolor{blue}{an automobile} is a kind of moving object  \\
\textcolor{blue}{an automobile} is a kind of object \\
\\
an airplane is \textcolor{blue}{a kind of vehicle} \\
a car is \textcolor{blue}{a kind of vehicle}
\\
\textcolor{blue}{an airplane} is used for carrying passengers \\
\textcolor{blue}{an airplane} is a kind of object
\end{tcolorbox}
\caption{Traversal showing \textcolor{blue}{held} semantic factors in explanations corpus.}
\label{tab:trav_examples_main_geo}
\end{table}

\begin{figure}[ht!]
\centering
    \includegraphics[width=0.8\linewidth]{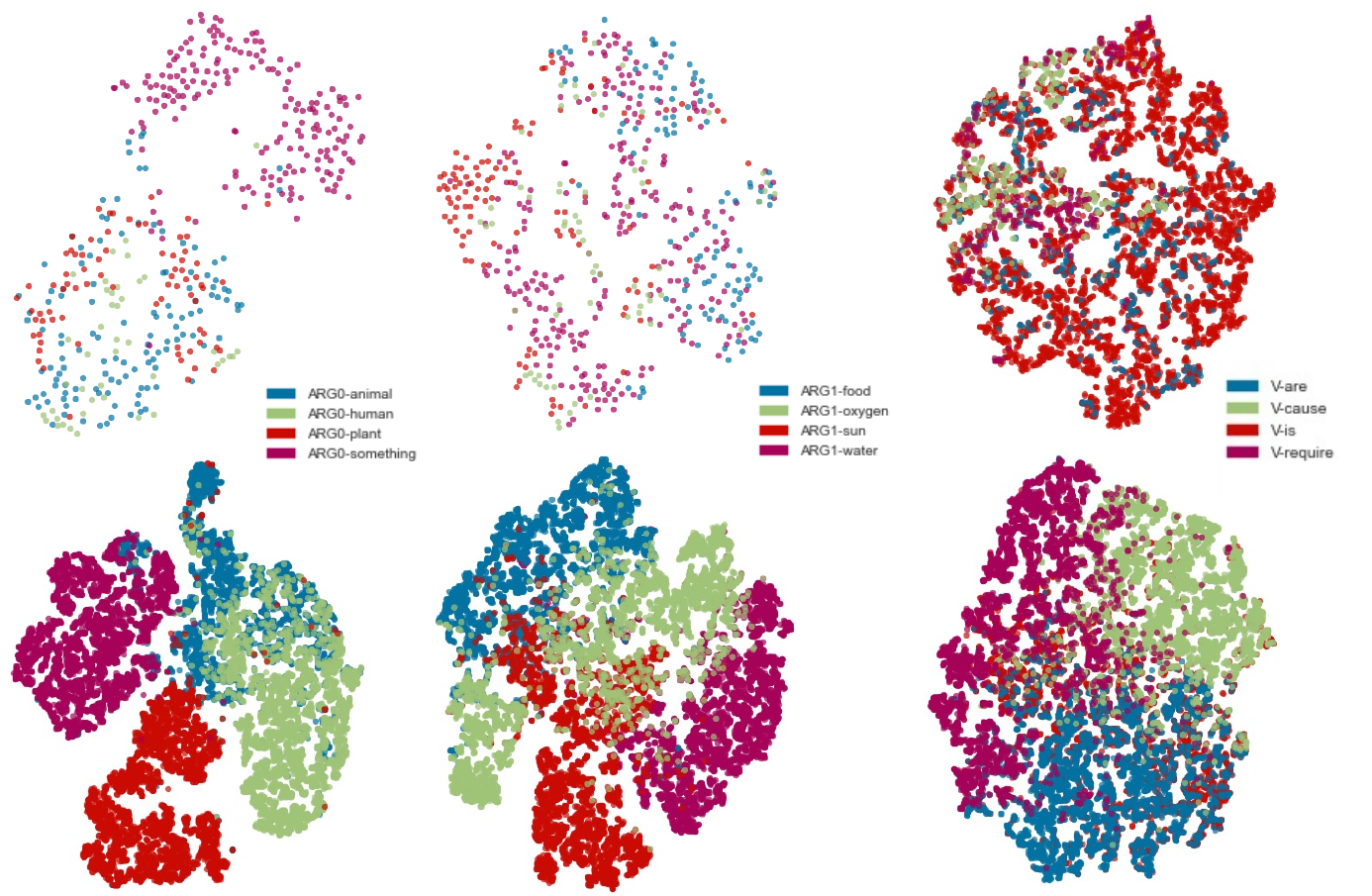}
    \caption{t-SNE plot of role-content distribution before and after traversal. From left to right are ARG0-(animal, human, plant, and something), ARG1-(food, oxygen, sun, and water), and PRED-(are, cause, is, require) (top: original role-cluster distribution, bottom: distribution after traversal).}
    \label{fig:da_arith_cluster}
\end{figure}

\paragraph{Quantitative evaluation.} Next, we employ t-SNE \cite{van2008visualizing} to examine \textit{role-content} features cluster and separation over latent space (i.e., \textit{natural clustering property} \cite{bengio2013deep}). In the corpus, however, due to the small number of data points within each role-content cluster, t-SNE cannot capture the differences between clusters well, resulting in the visualised latent space not displaying good role-content separability (top in figure \ref{fig:da_arith_cluster}). Therefore, we increase the number of data points in different role-content clusters by traversing each and keeping those resulting data points with the same role-content. Then, we visualise the role-content cluster at the bottom of figure \ref{fig:da_arith_cluster}. \uline{We can find that the features are clustered and separated over the latent space. If this was not the case, after traversing the resulting vectors from the same role-content cluster, the visualisation should show the same entanglement as the original datapoints distribution} \textbf{(Finding~2)}.

\subsection{Latent Arithmetic}
\paragraph{Qualitative evaluation.} In addition, we demonstrate the geometric properties via interpolation in Table~\ref{tab:interpol_examples}. 
\begin{table}[ht!]
\begin{tcolorbox}[fontupper=\small, fontlower=\small, top=1pt, bottom=1pt]
\textcolor{blue}{a beach ball is a kind of container} \\
1. a pool table is a kind of object \\
2. a balloon is a kind of object \\
3. a magnet is a kind of object \\
4. a neutron is a kind of particle \\
5. a proton is a kind of particle\\
\textcolor{blue}{an atom is a kind of particle}
\tcblower
\textcolor{blue}{protons are found in the nucleus of an atom} \\
1. protons are found in the nucleus of an atom \\
2. 1 atom is positive 1 in electric charge \\
3. \textcolor{red}{1 in 6000 is equal to 27 in 10 years} \\ 
4. if protons and neutrons have the same number of neutrons then those two particles are physically closer than one another \\
5. if a neutron has a negative -10 electric charge then the atom will not be able to move \\
6. if a neutron has a negative -10 electric charge then the neutron will not have a positive electric charge \\
\textcolor{blue}{if a neutral atom loses an electron then an atom with a positive charge will be formed}
\end{tcolorbox}
\caption{Interpolation examples (top: interpolation between sentences with similar semantic information, bottom: interpolation between sentences with different semantic information). Only unique sentences shown.}
\label{tab:interpol_examples}
\end{table}
For the top-most one, we can observe that sentences are smoothly moved from source to target (e.g., from \textit{beach ball} to \textit{atom} connected by \textit{ballon}, \textit{magnet}, \textit{neutron}, and \textit{proton}) where the same role-content (i.e., \textit{pred-is}) unchanged. In contrast, the second case doesn't display the smooth interpolation path. E.g., the third sentence connecting different semantic structures is unrelated to both source and target due to a discontinuous space gap between different clusters. Both indicate that the explanatory sentences might be clustered according to different semantic role structures.

\begin{table}[ht!]
\begin{tcolorbox}[fontupper=\small, fontlower=\small, middle=0.3cm, top=1pt, bottom=1pt]
\underline{$s_1$: animals require food for survival} \\
\underline{$s_2$: animals require warmth for survival} \\
\textcolor{blue}{animals} eat plants  \\
\textcolor{blue}{animals} produce milk \\
\textcolor{blue}{animals} usually eat plants \\
\textcolor{blue}{animals} require shelter to survive
\tcblower
\underline{$s_1$: water vapor is invisible} \\
\underline{$s_2$: the water is warm} \\
\textcolor{blue}{igneous rocks} are found under the soil \\
\textcolor{blue}{quartz} is usually very small in size \\
\textcolor{blue}{quartz} is formed by magma cooling \\
\textcolor{blue}{sedimentary} is formed by lithosphere collapsing
\end{tcolorbox}
\caption{$s_1 \pm s_2$ (top: addition, bottom: subtraction).}
\label{tab:arith_examples}
\end{table}

Following the definition of convex cone, we next traverse the resulting sentence after adding or subtracting two sentences with the same role-content feature. \uline{As illustrated in Table~\ref{tab:arith_examples}, the adding operation tends to hold the same role-content (e.g., \textit{ARG0-Animals}) as inputs. In contrast, the subtraction loses such control, e.g., from \textit{ARG1-water} to \textit{ARG1-quartz}} \textbf{(Finding~3)}. 


\begin{figure}[ht!]
\centering
\includegraphics[width=0.99\linewidth]{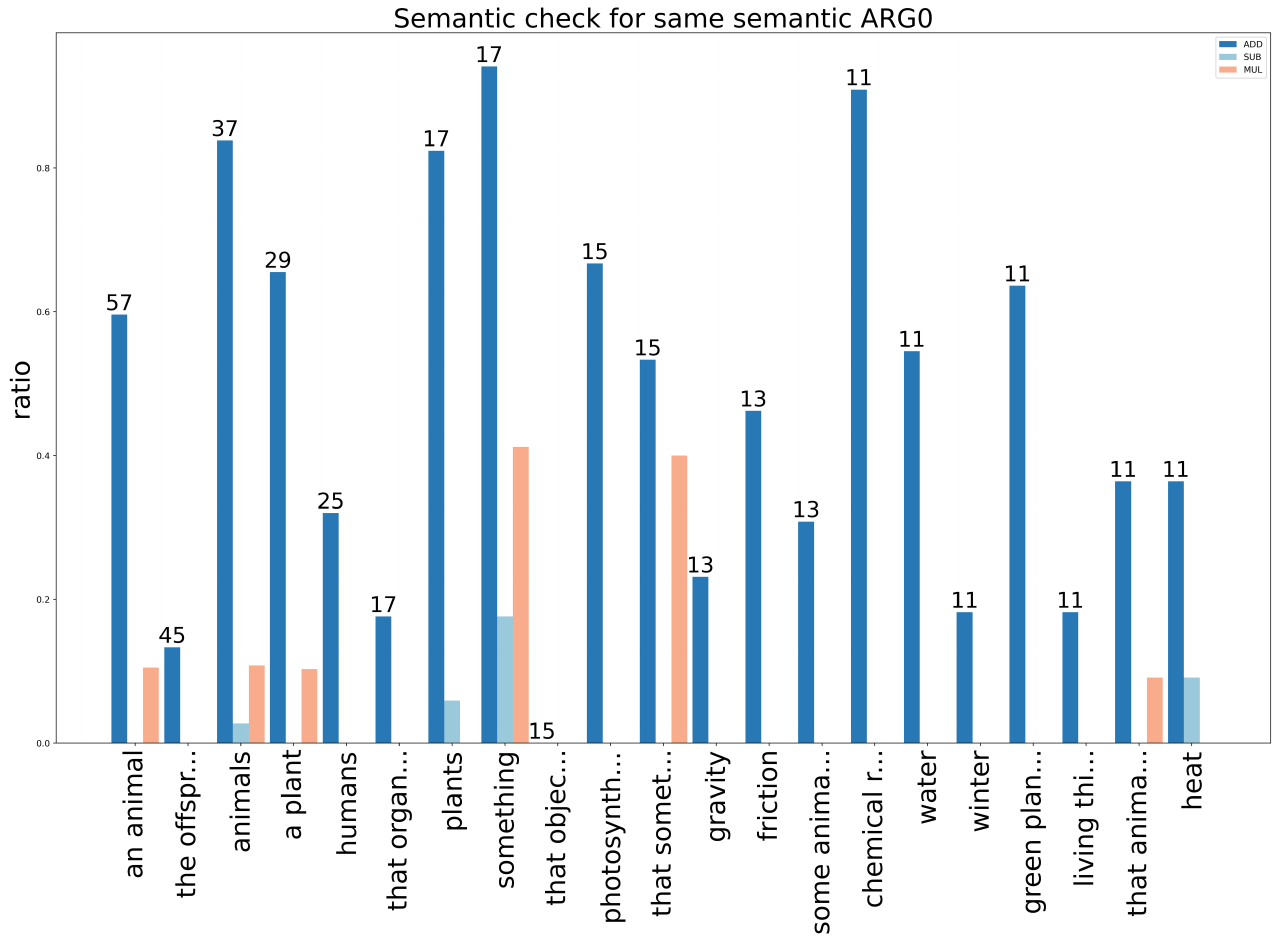}
\caption{Arithmetic, $s_1 \pm s_2$, for ARG0 with contents (dark blue: addition, shallow blue: subtraction, orange: element-wise production).} \label{fig:a0_animal}
\end{figure}
\begin{figure}[ht!]
\centering
\includegraphics[width=0.99\linewidth]{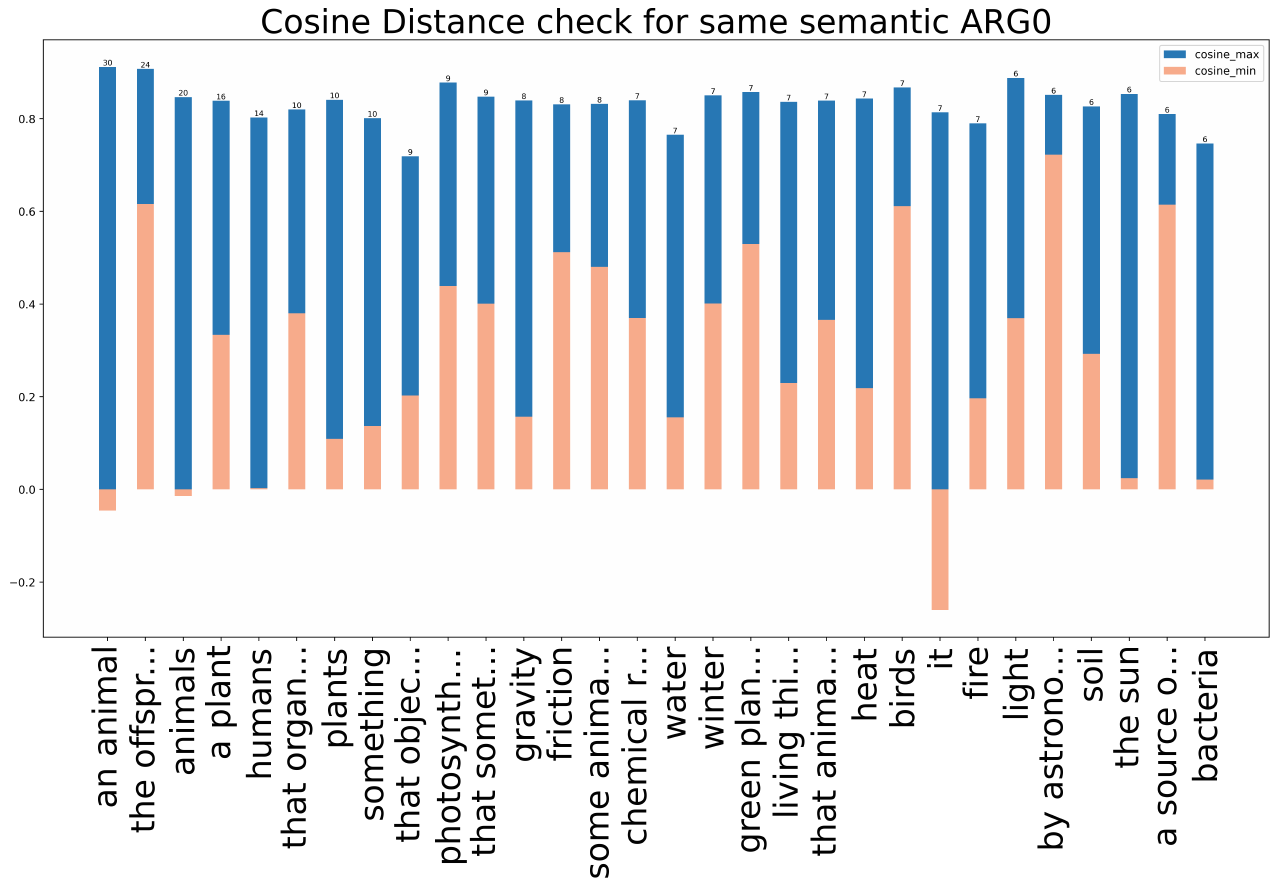}
\caption{Evaluating the geometrical size of role-content clusters (blue: max, orange: min).}
\label{fig:a0_size}
\end{figure}

\paragraph{Quantitative evaluation.} Next, we quantitatively assess our geometrical framework by calculating the ratio of the same role-content results from the vector addition and subtraction for all sentence pairs with a matching role. \uline{As illustrated in Figure \ref{fig:a0_animal}, the ADDed results (dark blue) can greatly hold the same role-content as inputs, indicating our geometrical framework. In contrast, the SUBed results (shallow blue) suffer from semantic shift} \textbf{(Finding~4)}. Similar observations for VERB and ARG1 can be found in Figure \ref{fig:consistency_verb_sem} and \ref{fig:consistency_arg1_sem}.
Besides, we can quantify each role-content cluster's geometrical area by calculating the cosine similarity between randomly selected sentence pairs in this cluster. We report the maximal and minimal distance in Figure \ref{fig:a0_size}. 

\subsection{Guided Latent Traversal} \label{sec:guide}
Finally, we examine the latent space geometry with our algorithm \ref{alg:guide}. The categories mentioned next are chosen based on their frequencies to ensure the balance during the classifier's training.

\begin{table}[ht!]
\begin{tcolorbox}[fontupper=\small, fontlower=\small, middle=0.3cm, top=1pt, bottom=1pt]

\ul{if something receives sunlight it will absorb the sunlight} \\
Dim27: \textcolor{blue}{if} a thing absorbs sunlight \textcolor{blue}{then} that thing is warmer  \\
Dim12: \textcolor{blue}{if} something is eaten \textcolor{blue}{then} that something produces heat \\
Dim08: \textcolor{blue}{if} something gets too hot in sunlight \textcolor{blue}{then} that something is less able to survive \\
Dim03: something \textcolor{blue}{contains} physical and chemical energy \\
Dim21: something \textcolor{blue}{contains} sunlight \\
Dim10: some things \textcolor{blue}{are} made of matter \\
Dim00: something \textcolor{blue}{is} made of atoms \\
Dim17: \textcolor{red}{a forest} \textcolor{blue}{contains} life \\
Dim00: something that is cold \textcolor{blue}{has} a lower temperature \\
Dim21: something \textcolor{blue}{rises} in temperature \\
Dim00: something \textcolor{blue}{is} formed from things dissolved in water \\
Dim30: something that is cold \textcolor{blue}{has} fewer nutrients \\
Dim21: something that is not moved \textcolor{blue}{is} dead
\end{tcolorbox}
\caption{Movement from \textit{conditional} to \textit{atomic} sentences.}
\label{tab:movement_topic_example}
\end{table}

\paragraph{Qualitative evaluation.} Firstly, we evaluate the traversal between different semantic role structures, e.g, conditional and atomic sentences. \uline{Table \ref{tab:movement_topic_example} shows that the cluster of the generated sentence changes as the values of different dimensions change sequentially (e.g., the first three sentences hold the same characteristic \textit{if ... then ...} as the input. The remaining sentences gradually move closer to the target characteristics, such as \textit{is}). Meanwhile, the sentences can hold the subject, \textit{something}, during the movement, corroborating our geometry framework} \textbf{(Finding~5)}.

\begin{table}[ht!]
\begin{tcolorbox}[fontupper=\small, fontlower=\small, middle=0.3cm, top=1pt, bottom=1pt]
\ul{fire causes chemical change} \\
Dim06: fire \textcolor{blue}{causes} chemical changes \\
Dim22: fire \textcolor{blue}{causes} chemical reactions \\
Dim02: fire can \textcolor{blue}{cause} harm to plants \\
Dim27: smoke can \textcolor{blue}{cause} harm to organisms \\
Dim14: fire \textcolor{blue}{causes} physical harm to objects \\
Dim24: fire can \textcolor{blue}{cause} chemical changes \\
Dim08: fire \textcolor{blue}{destroys} material \\
Dim01: fire \textcolor{blue}{means} chemical change \\
Dim14: \textcolor{red}{waste} \textcolor{blue}{means} igneous metal \\
Dim06: \textcolor{red}{combustion} \textcolor{blue}{means} burning \\
Dim00: \textcolor{red}{combustion} \textcolor{blue}{means} chemical changes \\
Dim21: \textcolor{red}{combustion} \textcolor{blue}{means} burning \\
Dim00: fire \textcolor{blue}{is} formed by thermal expansion \\
Dim18: fire chemical \textcolor{blue}{means} chemical energy \\
Dim03: fire \textcolor{blue}{is} corrosive
\tcblower
\ul{winter means cold environmental temperature}\\
Dim03: winter \textcolor{blue}{means} cold - weather\\
Dim18: winter \textcolor{blue}{means} cold weather \\
Dim00: winter \textcolor{blue}{means} weathering \\
Dim21: \textcolor{red}{drought} \textcolor{blue}{means} high temperatures / low precipitation \\
Dim00: winter \textcolor{blue}{means} high amounts of precipitation \\
Dim06: \textcolor{red}{drought} \textcolor{blue}{causes} natural disasters \\
Dim14: \textcolor{red}{drought} \textcolor{blue}{has a negative impact on} crops \\
Dim01: \textcolor{red}{drought} \textcolor{blue}{has a negative impact on} animals \\
Dim08: \textcolor{red}{drought} \textcolor{blue}{causes} animal populations to decrease \\
Dim24: \textcolor{red}{drought} \textcolor{blue}{causes} ecosystem loss \\
Dim14: \textcolor{red}{drought} \textcolor{blue}{causes} animals to have lower natural temperature \\
Dim27: cold climates \textcolor{blue}{causes} wildfires \\
Dim02: \textcolor{red}{climate change} can \textcolor{blue}{cause} low rainfall \\
Dim22: \textcolor{red}{global warming} \textcolor{blue}{causes} droughts \\
Dim06: winter \textcolor{blue}{causes} weather patterns
\end{tcolorbox}
\caption{Movement between \textit{cause} and \textit{mean}.}
\label{tab:movement_verb_example}
\end{table}

Next, we evaluate the traversal between predicates. Table \ref{tab:movement_verb_example} shows the movement between verbs (\textit{cause} and \textit{mean}). \uline{We can observe that the predicate is modified from \textit{causes} to \textit{mean}. In the traversal process, some sentences fall into the \textit{PRED-is} region. The reason is that the \textit{PRED-is} cluster is widely scattered in latent space (shown in Figure \ref{fig:da_arith_cluster}), which leads to a big overlap between \textit{PRED-is} and \textit{PRED-mean}} \textbf{(Finding~6)}. Moreover, we calculate the ratio of the generated sentences that hold the expected predicate, \textit{mean}, from 100 sentences with predicate \textit{cause}. The ratio is 0.71, which indicates that the decision tree is a reliable way to navigate the movement of sentences. Finally, we evaluate the traversal between arguments. Table \ref{tab:movement_arg_example} shows the movement from argument \textit{water} to \textit{something}. Similarly, \uline{the ARG1 can be modified from \textit{water} to \textit{something} following its path. Besides, the final generated explanation still holds a similar semantic structure, \textit{is a kind of}, compared with the input} \textbf{(Finding~7)}.
\begin{table}[ht!]
\begin{tcolorbox}[fontupper=\small, fontlower=\small, middle=0.3cm, top=1pt, bottom=1pt]

\ul{water is a kind of substance} \\
Dim12: \textcolor{blue}{water} is a kind of substance  \\
Dim00: \textcolor{blue}{water} is a kind of liquid \\
Dim23: \textcolor{blue}{liquid} is a kind of material \\
Dim29: \textcolor{blue}{water} has a positive impact on a process \\
Dim17: absorbing \textcolor{blue}{water} is similar to settling \\
Dim06: \textcolor{red}{absorbing} is similar to reducing \\
Dim21: absorbing \textcolor{blue}{something} is similar to absorbing something \\
Dim04: storing \textcolor{blue}{something} means being protected \\
Dim06: producing \textcolor{blue}{something} is a kind of process \\
Dim04: storing \textcolor{blue}{something} is similar to recycling \\
Dim21: absorbing \textcolor{blue}{something} is a kind of process \\
Dim01: absorbing \textcolor{blue}{something} can mean having that something \\
Dim22: folding \textcolor{blue}{something} is similar to combining something \\
Dim07: improving \textcolor{blue}{something} is a kind of transformation \\
Dim11: absorbing \textcolor{blue}{something} is a kind of method \\
Dim07: absorbing \textcolor{blue}{something} is a kind of process
\end{tcolorbox}
\caption{Movement from \textit{water} to \textit{something}.}
\label{tab:movement_arg_example}
\end{table}

\paragraph{Quantitative evaluation.} Finally, we use classification metrics, including accuracy (\textit{separability}) and recall (\textit{density}), as proxy metrics to assess latent space geometry. \uline{As shown in Table \ref{tab:proxy_metrics}, all features show higher separation where argument1 leads to the highest separation, indicating better latent space geometry} \textbf{(Finding~8)}. 
\begin{table}[ht!]
\centering
\begin{tabular}{lll}  \toprule
    \textbf{Formal semantic features}  & \textbf{separation}$\uparrow$ & \textbf{density}$\uparrow$  \\ \hline
    predicate (causes, means) & 0.87 & 0.92 \\
    argument1 (water, something) & 0.95 & 0.48 \\
    structure (condition, atomic) & 0.58 & 0.55 \\ \toprule
\end{tabular}
\caption{Proxy metrics for latent space geometry.} 
\label{tab:proxy_metrics}
\end{table}

\section{Related Work} \label{sec:related}
\paragraph{Formal-distributional semantics.} Integrating distributional semantics with formal / symbolic semantics is challenging in the field of artificial intelligence. In the Reasoning domain, for example, existing approaches usually perform symbolic behaviour via explicitly symbolic representation injection, including graph \citep{khashabi2018question,khot2017answering,jansen2017framing,thayaparan2021explainable}, linear programming \citep{valentino2022case,thayaparan2024differentiable}, adopting iterative methods, using sparse or dense encoding mechanisms \citep{valentino2020explainable,lin2020differentiable,valentino2022hybrid,bostrom-etal-2021-flexible}, or synthetic natural language expression \citep{clark2020transformers,yanaka-etal-2021-sygns,fu2024exploring}, among others. Comparatively, we explore the formal semantic property over distributional semantics via latent sentence geometry, which can potentially deliver better interpretation to current LMs.

\paragraph{Language geometry.} There is a line of work that studies the geometry of word and sentence representations \citep{arora-etal-2016-latent,mimno-thompson-2017-strange,ethayarajh-2019-contextual,reif2019visualizing,li2020sentence,chang2022geometry,jiang2024uncovering}. E.g., $king - man + woman = queen$, which the word vectors can be manipulated with geometric algebra. This phenomenon indicates the linear subspaces in language representations, similar features are encoded as a close direction in latent space, which has been widely explored ranging from word \citep{mikolov2013distributed} to sentences \citep{ushio2021bert}, Transformer-based LMs \citep{merullo2023language,hernandez2023linearity}, and multi-modal models \citep{trager2023linear,huh2024platonic}. Under the linear subspace hypotheses, a significant work explored the interpretability \citep{li2022emergent,geva2022transformer,nanda2023emergent} and controllability \citep{trager2023linear,merullo2023language,turner2023activation} of neural networks. In this work, we emphasise the formal semantic geometry for bridging the distributional and formal semantics, which is currently under-explored.

\paragraph{Language disentanglement.} Disentanglement, refers to separating features along dimensions \citep{bengio2013deep}, leading to clear geometric and linear representations. In the NLP domain, many studies explored the disentanglement between specific linguistic perspectives, such as sentiment-content \citep{john2019disentangled}, semantic-syntax \citep{bao2019generating}, and negation-uncertainty \citep{vasilakes-etal-2022-learning}, or syntactic-level disentanglement \citep{mercatali2021disentangling, felhi2022towards}. However, a fundamental issue has been overlooked: the definition of disentanglement in the image domain \citep{esser2020disentangling} cannot be directly applied to the context of computational linguistics due to the variability and complexity of language expression and high entanglement after current Transformer-based encoders. Therefore, we contribute to a new lens on the disentanglement (separation) of sentence features from the perspective of formal semantics.

\section{Conclusion}
In this study, we propose a \textit{formal semantic geometry} framework to investigate the localisation of formal semantic features, with the goal of enhancing the controllability and interpretability of distributional semantic spaces. Theoretically, a formal semantic feature is defined as a role–content pair, represented geometrically as a convex cone within the latent space. Practically, we introduce a supervised learning approach to induce and align such geometric structures. The learned latent space geometry is extensively evaluated through operations such as traversal, arithmetic, and guided traversal. Experimental results provide empirical evidence for the existence of formal semantic geometry, answering \questionA{}

\paragraph{Reproducibility.} The data and the related codebase are available online: \url{https://github.com/SnowYJ/geometricOptimus}.

\section{Scoping and Limitations}
The geometric analysis indicates that the role-content regions still have significant overlapping over latent spaces. Therefore, a new potential task can be how we can better separate semantic features to provide better localisation or composition behaviour over distributional semantic spaces.

Moreover, as shown in Figure~\ref{fig:a0_size}, \ref{fig:consistency_verb_sem}, and \ref{fig:consistency_arg1_sem}, although the addition operation better preserves the same role–content feature compared with subtraction, the resulting ratio is still not particularly high. E.g., some role-content features are less than 50 per cent. This limitation arises from the relatively small number of samples in the corpus that share identical role–content features, which may in turn influence the experimental results.

\chapter{Formal Semantic Disentanglement} \label{cha:dis}

Chapter~\ref{cha:geo} revealed that the formal semantic features (i.e., \textit{role-content}) can be potentially separated in the latent space. However, those features have significant overlapping. To provide a better feature separability, this chapter investigates \questionB{}

\section{Introduction}
\begin{figure}[ht!]
\centering
    \includegraphics[width=0.8\linewidth]{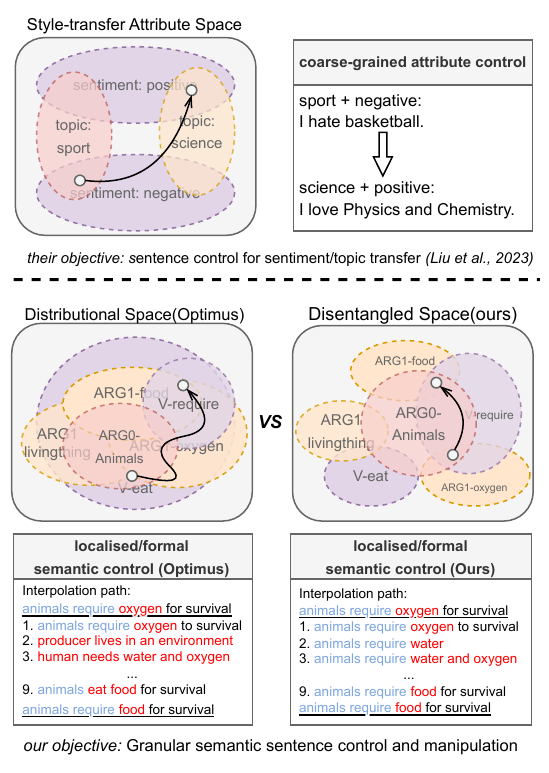}
    \caption{Top: attribute space geometry. Bottom: general semantic geometry, where left: distributional semantic space of Optimus \citep{li2020optimus}, right: our compositionality-induced semantic space where the geometrical location of sentence vectors can be located by the intersection of role-content clusters.}
    \label{fig:comparison}
\end{figure}

Most previous work on controlled text generation have concentrated on style transfer tasks: modifying sentences with regard to markers of sentiment, formality, affirmation/negation \citep{john2019disentangled,bao2019generating,hu2021causal,vasilakes-etal-2022-learning,gu-etal-2022-distributional,liu-etal-2023-composable,gu-etal-2023-controllable} (Figure \ref{fig:comparison} top). Disentanglement of language generative factors over Variational Autoencoder (VAE) spaces has been a key mechanism to deliver this type of generative control \citep{john2019disentangled,bao2019generating,vasilakes-etal-2022-learning}. 

In Chapter~3, we demonstrated that a more general form of semantic control can be achieved in the latent space of Optimus \citep{li2020optimus}, the first standard transformer-based VAE, where a BERT \citep{devlin2019bert} encoder and a GPT2 \citep{Radford2019LanguageMA} decoder are connected within a VAE bottleneck. Using representations of conceptually dense \textit{explanatory sentences} \citep{jansen2018worldtree}, they showed that sentences (e.g. \textit{animals require oxygen for survival}), can be represented within a space which can be organised around the associations between predicate, arguments and their associated token content: \textit{ARG0-animals} or \textit{VERB-require}, is geometrically resolved to a hypersolid over the latent space. Nevertheless, the ability to learn and control such separation is still limited as different semantic factors of the sentence are still overlapped and entangled in the latent space (e.g., \textit{V-eat} and \textit{V-require} in Figure \ref{fig:comparison} bottom left), indicating distributional sentence semantics cannot be currently localised and controlled from the perspective of formal semantics (i.e., \textit{predicate-argument structures}, \textit{compositionality}) \citep{marcus2003algebraic,Nefdt2020-NEFAPC,dankers-etal-2022-paradox}. 

This work aims to improve the localisation and semantic control of latent sentence spaces, by delivering a model which can better separate and control syntactic-semantic features (e.g. predicate-argument) and their associated lexical semantics content. This type of representation can provide the foundation to shorten the gap between deep latent semantics and formal linguistic representations \citep{10.3115/1075218.1075283,banarescu2013abstract,mitchell2023we}, integrating the flexibility of distributional-neural models with the properties of linguistically grounded representations, facilitating both interpretability and generative control.

To deliver this type of semantic control within the distributional sentence space, following the methodological framework introduced by \cite{zhang2022}, we target on improving the semantic separability of sentences by focusing on explanatory sentences. Inspired by the work of \cite{esser2020disentangling}, we integrate a flow-based invertible neural network (INN) \citep{dinh2014nice} as a plug-in control component to learn the bijective transformation between the distributional hidden space of the transformer-based language autoencoder (BERT-GPT2) and the smooth Gaussian space of the INN (Figure \ref{fig:optimus_latent_space}). Specifically, we first pre-train an autoencoder (AE) to learn sentence representations from the transformers' latent spaces. Then, we freeze the AE weights and train the INN to map the AE representations to a Gaussian space. Since INN models define a bijective transformation, we can control the autoencoder generation by manipulating the INN latent spaces, which is more efficient and significantly less resource intensive than re-training a language AE end-to-end.

More importantly, we propose a supervised training strategy within the INN setting to learn a latent space with improved semantic separability, namely: the \textit{semantic role-content pairs} and their associated clusters can be better separated over the latent space modelled by the INN (Section \ref{sec:cluster_sup}). In this case, we can improve localised control over the decoding process due to the reduction of overlapping (ambiguous) regions. \textit{A more separable and geometrically consistent sentence space} can be then operated over to improve the generative control with support of geometric operators, such as interpolation \cite{bowman2016generating} (Section \ref{sec:guide_interpolate}). The contributions of this work are summarised below:

\textbf{1.} We approach sentence disentanglement and generation control from the point of view of \textit{Argument Structure Theory (AST)}, bridging latent space features with a canonical, linguistics-informed, semantic representation of sentences. \textbf{2.} We find that integrating a flow-based INN mechanism into a transformer-based language-AE architecture is an effective mechanism for transforming the hidden space of the autoencoder into a smooth Gaussian latent space for representing sentences. \textbf{3.}  We propose a supervised training strategy for INNs to learn a controllable semantic space with higher disentanglement and separability of semantic features, when compared to previous work. \textbf{4.} Using this mechanism, we systematically employ geometrical data augmentation strategies to assist on sentence representation disentanglement. 


\section{Methodology}
In this section, we first define the target task addressed in this work. We then describe the proposed architecture and training objectives, followed by the data augmentation strategies employed to support model training.
\subsection{Sentence Semantic Disentanglement}

In Chapter~\ref{cha:geo}, we illustrated that a sentence $s$ consists of a sequence PArgs/SRs and word content associations. Upon encoding in latent space, this can be described as:
\[
sem(s) = \underbrace{t_1({c_1}, {r_1})}_{i.e., ARG0-animals} \oplus \dots \oplus \underbrace{t_i({c_i}, {r_i})}_{PRP-survival}
\]
where $t_i({c_i}, {r_i})=c_i \otimes r_i$ represents the semantics of term $t_i$ with content $c_i$ (i.e., \textit{animals}) and SRL $r_i$ (i.e., \textit{ARG0}) in context $s$, $\otimes$: connects the meanings of words with their roles, using the compositional-distributional semantics notation of \cite{smolensky2006harmonic,clark2008compositional}. $\oplus$: connects the lexical semantics (word content + structural role) to form the sentence semantics. This work applies distinct symbols aiming to emphasise the disentanglement aspects associated with the AST structure. If the sentence representation can be semantically disentangled under $\oplus$, the $sem(s)$ can be decomposed into: 
$$sem(s) = \{ t_1({c_1}, {r_1})\} \oplus \dots \oplus \{t_i({c_{i}}, {r_{i}}) \}$$
where each set represents a specific role-content cluster resolved to a hypersolid over the latent space, in this case, given a set of $N$ sentences within the same predicate cluster $t({c}, {r})$ (i.e., \textit{V-require}) but different $sem(s)$, those sentence vectors can represent $t({c}, {r})$ features independently of other features (i.e., \textit{ARG0-animals}), forming the $t({c}, {r})$ cluster:
$$\{sem(s_1), ..., sem(s_N)\} = \{t({c}, {r})\}_{\times N} \oplus \{... \}$$ 
Therefore, we can evaluate the disentanglement (i.e., \textit{natural clustering property} \cite{bengio2013deep}) of sentence semantics by evaluating the density within $\{t({c}, {r})\}$ set(cluster) (classifier recall) and the separation between different $\{t({c}, {r})\}$ set(clusters) (classifier accuracy) with downstream classifiers based on the \textit{manifold hypothesis for classification} \cite{Rifai2011TheMT}, rather than disentanglement metrics, which usually calculate the separation between latent dimensions, commonly used in the image domain \cite{higgins2016beta, kim2018disentangling, chen2018isolating, ridgeway2018learning}. Next, we will introduce the INN-based mechanism, which is used to learn this semantically disentangled space.

\subsection{Training Strategy}
\begin{figure}[ht!]
    \centering
    \includegraphics[width=10cm]{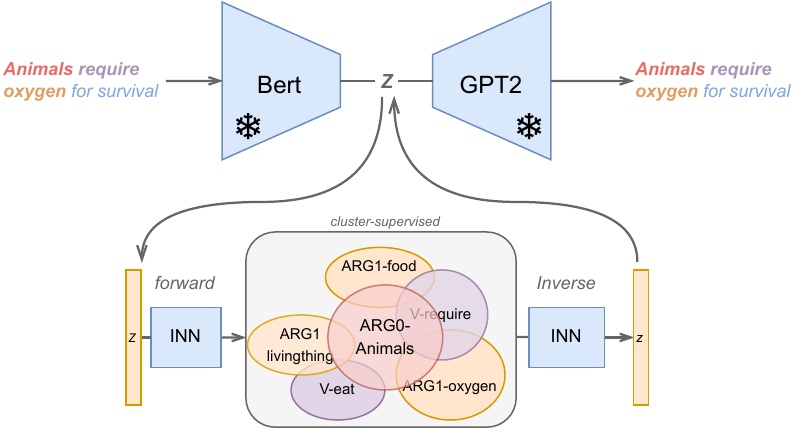}
    \caption{Transforming the representations of explanatory sentences from a language autoencoder (BERT-GPT2), into a semantically separable latent space with the support of the INN mechanism, where a sentence representation can be decomposed into a predicate-argument-level semantics (role-content).}
    \label{fig:optimus_latent_space}
\end{figure}

\paragraph{Architecture overview.} We encode each sentence $x$ with a frozen autoencoder (e.g., BERT-GPT2) and consider its sentence representation $E(x)$ as the input of INNs (Figure \ref{fig:optimus_latent_space}). We propose two training strategies to map the hidden representations into the Gaussian space.

\paragraph{Unsupervised INN.} Firstly, we train the INN-based model unsupervised, which minimises the negative log-likelihood of the marginal distribution of latent representation $z=E(x)$:
$\mathcal{L}_\text{unsup} = 
- \mathbb{E}_{x \sim p(x)} \Big[ T(E(x)) \Big]^2 + \log \left| T'(E(x)) \right| \nonumber$
As the minimisation leads to a bijective mapping between the distributed representation and the disentangled latent representation (multivariate Gaussian space), it allows for a more semantically consistent representation of the geometric (role-content) clustering properties of its latent space, allowing for a more consistent traversal and interpolation \cite{li2020optimus} over the sentence space (Figure~\ref{fig:comparison}).


\paragraph{Cluster-supervised INN.} According to the findings of \cite{zhang2022}, the content of the predicate-argument structure/semantic roles can be disentangled over the latent space approximated to multivariate Gaussian learned using the Language VAE setting. Using the same foundation, we train the INN component to learn the embeddings, by minimising the distance between points in the same role-content regions and maximising the distance between points in different regions, based on the explanation embeddings and their corresponding central point from the language autoencoder model. For example, given a sentence "\textit{\textbf {animals} require food for survival}" and its central vector of \textit{ARG0-animals}, the training moves the sentence representation closer to the \textit{ARG0-animals} region centre in the INN latent space. Specifically, during the calculation of the posterior, we replace the mean and variance of the standard Gaussian distribution by the centre point of its cluster and a hyper-parameter, which should be less than one, respectively. In this case, each role-content cluster in the latent space will be mapped to a space where each cluster will have its embeddings more densely and regularly distributed around its centre. The objective function can be described as follows:
$$
\mathcal{L}_\text{sup} = - \mathbb{E}_{x \sim p_{cluster}(x)} \frac{\Big[ T(E(x)) - \mu_{cluster} \Big]^2}{{1-\sigma^2}} +  \log \left| T'(E(x)) \right|
$$
Where $T(E(x))$ learns the transformation from $x$ to $z \sim N(\mu_{cluster}, 1-\sigma^2)$. $\sigma^2$ is a parameter which can be empirically determined (in this particular context the optimal value was found to be 0.6). 
\subsection{Geometrical Data Augmentation}
Data augmentation, which captures and augments a common or distinct feature across different samples, has been considered a common technique to assist disentanglement, such as in Graph \cite{li2021disentangled} and Image \cite{liu2022learning} representations, but is still limited in the context of sentence generation. In this work, we consider the vector arithmetic and traversal operators as a systematic mechanism to support data augmentation (via semantically controlled sentence generation) for each role-content cluster, described as follows: 
\begin{equation}
    \begin{aligned}
        &(1)\quad \text{v} = average(E'(x_i), E'(x_j)) \\ \nonumber
        &(2)\quad \text{v}_{neighbour} = \text{v}[i] \sim N(0, 1)_{\forall i \in \{0,..,size(\text{v})\}} \\
        &(3)\quad x_{new} = D'(\text{v}_{neighbour})
    \end{aligned}
    \label{eq:data_augmentation}
\end{equation}


\begin{table}[ht!]
    \centering
    \begin{tabular}{ll}
        \toprule
        Role-content & Augmented sentences \\ \hline
        \multirow{3}{*}{ARG0-animal} 
        & \textcolor{blue}{an animal} requires energy to move \\
        & \textcolor{blue}{animals} produce offspring \\
        & \textcolor{blue}{an animal} requires shelter \\
        & \textcolor{blue}{an animal} can use its body to breathe \\ \hline
        \multirow{3}{*}{ARG0-human} 
        & \textcolor{blue}{humans} travel sometimes \\
        & \textcolor{blue}{humans} usually use gasoline \\
        & \textcolor{blue}{humans} use coal to make food \\
        & \textcolor{blue}{humans} depend on pollinators for survival \\ \hline
        \multirow{3}{*}{PRED-are} 
        & wheels \textcolor{blue}{are} a part of a car \\
        & toxic chemicals \textcolor{blue}{are} poisonous \\
        & green plants \textcolor{blue}{are} a source of food for animals \\
        & copper and zinc \textcolor{blue}{are} two metals \\ \hline
        \multirow{3}{*}{PRED-mean} 
        & summit \textcolor{blue}{mean} the top of the mountain \\
        & colder \textcolor{blue}{mean} a decrease in heat energy \\
        & cleaner \textcolor{blue}{mean} ( less ; lower ) in pollutants \\
        & friction \textcolor{blue}{mean} the product of a physical change \\ \toprule
    \end{tabular}
    \caption{Augmented explanations.} \label{tab:aug_data}
\end{table}

\noindent where $x_k \in S$ (original corpus), $E'$ and $D'$ are the encoder and decoder of Optimus fine-tuned over $S$. $average$ operation aims to modify the sentence while maintaining the target role-content common to both $x_i$ and $x_j$ \cite{zhang2022}. The term $\text{v}[i] \sim N(0, 1)$ is introduced to resample each dimension of $\text{v}$ in the latent space (i.e., traverse its neighbour) and $x_{new} = D'(\text{v}_{neighbour})$ generates a new sentence. Finally, we only keep the sentences holding the target role-content, where the PArgs/SRs of $x$ are annotated via the \textit{AllenNLP} \cite{gardner2018allennlp} semantic role labeller. Table \ref{tab:aug_data} lists randomly selected examples from augmented explanations. 

\section{Empirical Evaluation}
For the experiments, we start by focusing on the effect of the supervised INN mechanism to examine its impact on the sentence semantic separability of the distributional latent space, detailed in Section \ref{sec:cluster_sup}. Next, we examine the localised semantic generation control enabled by such semantic separability via latent interpolation in Section \ref{sec:guide_interpolate}.
\subsection{Disentanglement Encoding Evaluation} \label{sec:cluster_sup}
We examine the latent space separability (i.e., \textit{natural clustering property} \citep{bengio2013deep}) of our supervision approach on different predicate-argument/semantic roles. In the context of this work, the semantic role labels are not used to control the generation. Instead, we use the predicate argument position markers, e.g. including \textit{ARG0}, \textit{ARG1}, \textit{PRED(V)}, where each category has \textit{a)} four possible word contents ($c_i$), or \textit{b)} the same content (i.e., \textit{animal}) with different argument/roles, including \textit{ARG0,1,2}.

\begin{figure}[ht!]
\centering
    \includegraphics[width=0.8\linewidth]{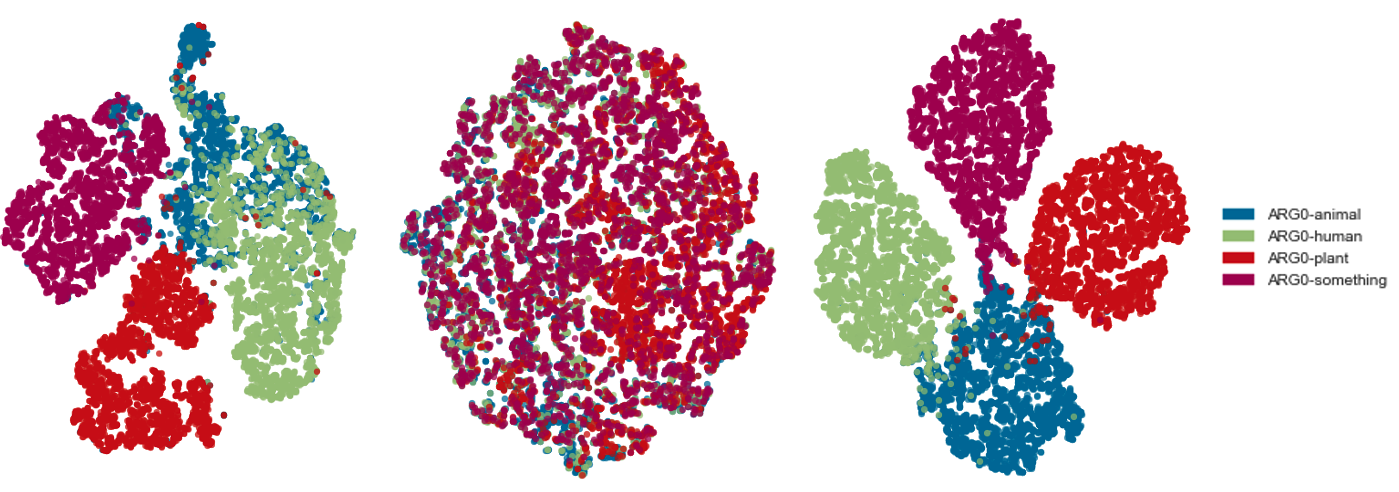}
    \caption{ARG0: t-SNE plot, different colour represents different content regions (blue: animal, green: human, red: plant, purple: something) (left: Optimus, middle: unsupervised, right: cluster supervised), same order for remaining visualisations. This visualisation shows that supervised embeddings concentrate on the respective cluster centre.}
    \label{fig:a0_sup}
\end{figure}

\paragraph{Disentanglement between \textit{ARG0} clusters.} For \textit{ARG0}, we choose \textit{human}, \textit{animal}, \textit{plant}, and \textit{something} due to having the highest frequency in the original dataset, and evaluate model performance from two directions, including forward and backward mapping. Within forward mapping, we assess the disentanglement of the latent space of the INN model from two perspectives (visualisation and classification metrics). Figure \ref{fig:a0_sup} displays the distributions of four role-content clusters over the latent space. As we observe, \uline{after the cluster-supervised training strategy, the embeddings are more concentrated at the centre of their cluster, and there is a clear boundary between clusters, indicating a better disentanglement compared to Optimus and unsupervised INNs} \textbf{(Finding~1)}.

We then quantitatively evaluate the disentanglement of ARG0-content clusters. We consider classification task metrics (\textit{accuracy}, \textit{precision}, \textit{recall}, \textit{f1}) as proxies for evaluating region separability, effectively testing cluster membership across different clusters. We choose a non-parametric downstream classifier (i.e., kNN) to quantitatively evaluate the separation of clusters and parametric downstream classifiers, including Naive Bayes (NB) and Support Vector Machine (SVM), to assess both separability and representation capability of latent sentence spaces \citep{Rifai2011TheMT, conneau-etal-2018-cram}.

\begin{table}[ht!]
\small
\centering
\setlength\tabcolsep{2.5pt}
\begin{tabular}{cccccc}
\toprule
\multicolumn{6}{c}{ARG0: disentanglement proxy metrics} \\ \hline

classifier & train & accuracy & precision & recall  & f1 score \\ \hline
\multirow{3}{*}{KNN} & O & 0.972 & 0.973 & 0.972  & 0.972 \\
 & U & 0.938 & 0.938 & 0.938  & 0.938 \\
 & C & \textbf{0.979} & \textbf{0.979} & \textbf{0.979}  & \textbf{0.979} \\ \hline
 
\multirow{3}{*}{NB} & O & 0.934 & 0.934 & 0.933  & 0.933 \\
& U & 0.958 & 0.958 & 0.958  & 0.958 \\
 & C & \textbf{0.978} & \textbf{0.978} & \textbf{0.978}  & \textbf{0.978} \\ \hline
 
\multirow{3}{*}{SVM} & O & 0.970 & 0.970 & 0.970  & 0.970 \\
 & U & 0.972 & 0.972 & 0.972  & 0.972 \\
 & C & \textbf{0.980} & \textbf{0.980} & \textbf{0.980}  & \textbf{0.980} \\ \toprule
\end{tabular}

\caption{Disentanglement of ARG0 between Optimus (O), unsupervised INN (U), and cluster-supervised INN (C) where KNN: k-neighbours, NB: naive bayes, SVM: support vector machine. The abbreviations are the same for the remaining tables.} \label{tab:arg0_exp}
\end{table}

As shown in table \ref{tab:arg0_exp}, all classifiers trained over supervised latent representations outperformed the unsupervised INN (U) and Optimus (O), indicating that the cluster-supervised approach leads to better disentanglement and representation. Moreover, \uline{(O) demonstrates superior performance compared to (U) for the KNN-based evaluation. However, it exhibits lower performance than (U) in NB and SVM. This suggests that the INN-AutoEncoder configuration can more effectively capture sentence semantics (from the point-of-view of AST+distributional content), in the reconstruction task, since the VAEs' training process is prone to experiencing posterior collapse} \textbf{(Finding~2)}.

As for the evaluation of the backwards mapping, we calculate the ratio of generated sentences that hold the same role-content as the inputs (henceforth called the invertibility ratio). We randomly selected 100 embeddings as inputs and showed the corresponding ratios in Table \ref{tab:arg0_inv_exp}. We can observe that \uline{both unsupervised and supervised cases can achieve high invertibility ratios, indicating that the INN mechanism provides stable invertibility with or without cluster supervision} \textbf{(Finding~3)}.

\begin{table}[ht!]
\small
\centering
\begin{tabular}{ccccc}
\toprule 
\multicolumn{5}{c}{ARG0: invertibility ratio (backward: $T'$)} \\ \hline
train & human & animal & plant & something \\ \hline
U & 0.980 & 0.890 & 0.990 & 1.000 \\
C & 1.000 & 0.860 & 0.990 & 0.950 \\ \toprule
\end{tabular}
\caption{Invertibility test for ARG0, both INNs with AutoEncoder setup can achieve high ratios, indicating stable invertibility with or without cluster supervision.} \label{tab:arg0_inv_exp}
\end{table}

\begin{figure}[ht!]
\centering
    \includegraphics[width=0.75\linewidth]{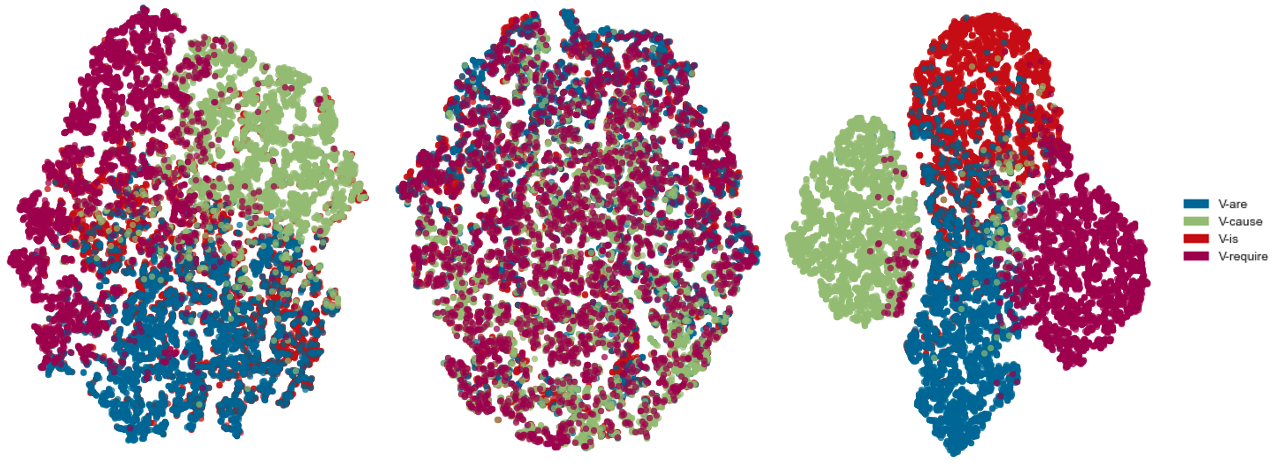}
    \caption{PRED: t-SNE plot (blue: are, green: cause, red: is, purple: require).}
    \label{fig:v_sup}
\end{figure}

\paragraph{Disentanglement between \textit{PRED} clusters.} Next, we analyze the disentanglement between \textit{predicate (PRED)} clusters. \uline{In Figure \ref{fig:v_sup}, although the disentanglement of PRED clusters is not as high as ARG0, the latent space with cluster supervision still performs better than both the unsupervised case and the Optimus model. In Table \ref{tab:v_exp}, the supervised INN model achieves better disentanglement, and both unsupervised and supervised models could obtain a higher ratio} \textbf{(Finding~4)}.


 
 

\begin{table}[ht]
\small
\centering
\setlength\tabcolsep{2.5pt}
\begin{tabular}{cccccc}
\toprule
\multicolumn{6}{c}{PRED: disentanglement proxy metrics (forward: $T$)} \\ \hline

classifier & train & accuracy & precision & recall  & f1 score \\ \hline
\multirow{3}{*}{KNN}  & O & 0.911 & 0.914 & 0.910  & 0.911 \\
& U & 0.869 & 0.873 & 0.865  & 0.868 \\ 
& C & \textbf{0.922} & \textbf{0.927} & \textbf{0.918}  & \textbf{0.922} \\ \hline
 
\multirow{3}{*}{NB}  & O & 0.865 & 0.866 & 0.866  & 0.865 \\
& U & 0.873 & 0.874 & 0.871  & 0.872 \\ 
& C & \textbf{0.903} & \textbf{0.903} & \textbf{0.902}  & \textbf{0.903}  \\ \hline
 
\multirow{3}{*}{SVM}  & O & 0.902 & 0.902 & 0.903  & 0.902 \\
 & U & 0.905 & 0.906 & 0.902  & 0.904 \\ 
 & C & \textbf{0.910} & \textbf{0.912} & \textbf{0.909}  & \textbf{0.910} \\ \toprule
\end{tabular}
\caption{Forward evaluation for predicate clusters. The invertibility test for PRED can be found in the work \cite{zhang2023learning}.} \label{tab:v_exp}
\end{table}

\begin{figure}[ht!]
\centering
    \includegraphics[width=0.8\linewidth]{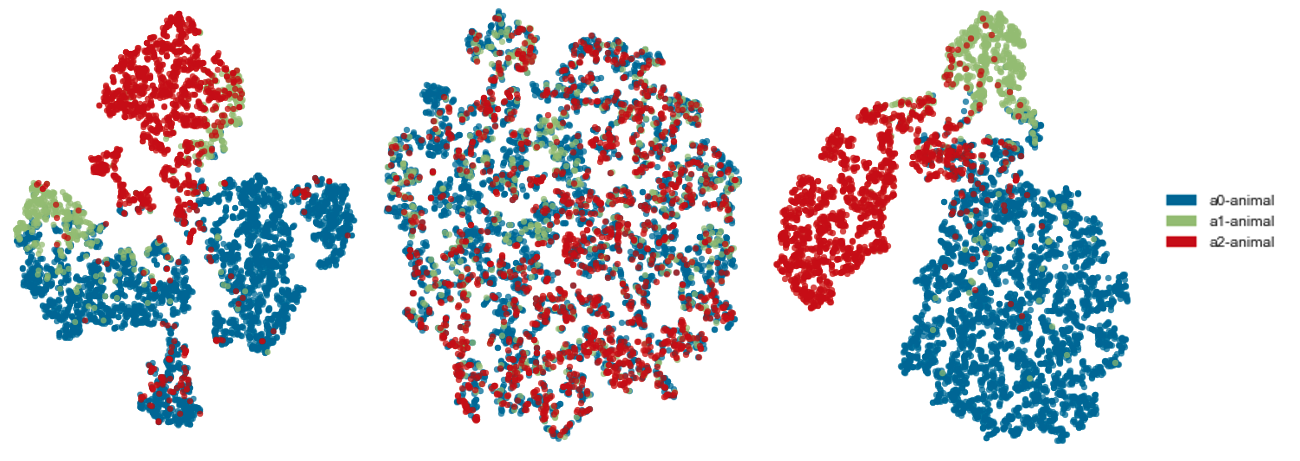}
    \caption{Animal: t-SNE plot (blue: ARG0-animal, green: ARG1-animal, red: ARG2-animal).}
    \label{fig:a012_animal}
\end{figure}

\paragraph{Disentanglement between \textit{ARG0,1,2} clusters.}
The experiments up to this point investigated the separation between the same semantic role type but different content clusters. Next, we explore the separability of different pred-argument types with the same content. We thus focus on the \textit{animal} cluster, and investigate the disentanglement between \textit{ARG0-animal}, \textit{ARG1-animal}, and \textit{ARG2-animal}. \uline{As illustrated in Figure \ref{fig:a012_animal}, the animal clusters with different pred-argument types can be separated after cluster-supervised training, which indicates that the INN model can capture the difference between the same content with different pred-argument types in the case of a similar topic, indicating the INN-based approach could jointly learn separable embeddings w.r.t. role-content and content alone} \textbf{(Finding~5)}. 

\begin{table}[ht!]
\centering
\begin{tabular}{cccc}
\toprule
\multicolumn{4}{c}{Animal: disentanglement metrics (\textit{f1 score})} \\ \hline

train & KNN & NB & SVM  \\ \hline
O & 0.960 & 0.928 & 0.946 \\
U & 0.958 & 0.930 & 0.947 \\
C & \textbf{0.967} & \textbf{0.937} & \textbf{0.950} \\ \toprule
\end{tabular}
\caption{Forward evaluation for Animal, we only show \textit{f1} since the same value across different metrics. Results indicate improved separation across role clusters. The invertibility test for PRED can be found in the work \cite{zhang2023learning}.} \label{tab:animal_exp}
\end{table}

Table \ref{tab:animal_exp} shows the disentanglement metrics. Similarly to the previous experiment, the supervised case outperforms both the unsupervised and the Optimus models.

\subsection{Disentanglement Decoding Evaluation} \label{sec:guide_interpolate}
Finally, we evaluate the disentangled sentence geometry from the perspective of sentence generation. We specifically focus on linear interpolation as it can provide more efficient traversal between sentences and clusters than other traversal approaches (e.g., \textit{Ornstein-Uhlenbeck}), commonly used in the NLP domain \citep{li2020optimus} and in the evaluation of disentanglement \citep{bengio2013deep}.

\paragraph{Interpolation localisation.} Firstly, we evaluate the localisation of latent interpolation that interpolates a path $z_t = z_1 \cdot (1 - t) + z_2 \cdot t$  with $t$ increased from $0$ to $1$ by a step size of $0.1$, where $z_1$ and $z_2$ represent the latent representations of source and target sentences. As a result, $9$ sentences are generated on each interpolation step. On a latent space with better token-level role-content separation, given two sentences with the same role-content as endpoints, we can observe that the intermediate sentence can hold the same role-content during interpolation.

In terms of qualitative evaluation, Table \ref{tab:guide_generation} provides the interpolation paths of cluster-supervised INN and Optimus, as for Optimus, we can observe that the intermediate explanations could transition smoothly from source to target for \textit{argument}. However, the \textit{predicate} is more abruptly changed, indicating lower \textit{predicate-content} disentanglement (e.g., \textit{predicate-require} and \textit{predicate-eat}). Instead, the supervised INN can fix the \textit{predicate-require} during interpolation, \uline{indicating better separability between different predicate-content results in better generation control} \textbf{(Finding~6)}.
\begin{table}[ht!]
\begin{tcolorbox}[fontupper=\small, fontlower=\small, title=interpolation localisation: \textit{predicate-require} ]
\underline{source: humans \textcolor{blue}{require} freshwater for survival}\\

Optimus: \\
1. humans \textcolor{blue}{require} water and food through fossil fuels \\
2. humans \textcolor{blue}{require} water for survival \\
3. humans \textcolor{red}{produce} small amounts of consumer food \\
4. human \textcolor{red}{has} a positive impact on a plant's survival \\
5. humans \textcolor{red}{convert} food into animal prey \\
6. humans \textcolor{red}{make} food for themselves by eating \\
7. animals \textcolor{blue}{require} food for survival \\
8. animals \textcolor{blue}{require} nutrients from the air \\
9. humans \textcolor{red}{eat} plants for food \\
10. animals \textcolor{blue}{require} food for survival \\


Cluster-supervised INN: \\
1. humans \textcolor{blue}{require} water for survival \\
2. nonhumans \textcolor{blue}{require} water for survival \\
3. animals \textcolor{blue}{require} water and food \\
4. animals \textcolor{blue}{require} water to survive \\
5. animals \textcolor{blue}{require} water to live \\
6. animals \textcolor{blue}{require} food for survival \\
7. animals \textcolor{blue}{require} food for survival \\
8. animals \textcolor{blue}{require} food for survival \\
9. animals \textcolor{blue}{require} food for survival \\
10. animals \textcolor{blue}{require} food to survive \\


\underline{target: animals \textcolor{blue}{require} food to survive}
\end{tcolorbox}
\caption{Interpolation examples, indicating the cluster-supervised INN can provide better localised/symbolic semantic control.}
\label{tab:guide_generation}
\end{table}
\begin{figure}[ht!]
\centering
    \includegraphics[width=12cm]{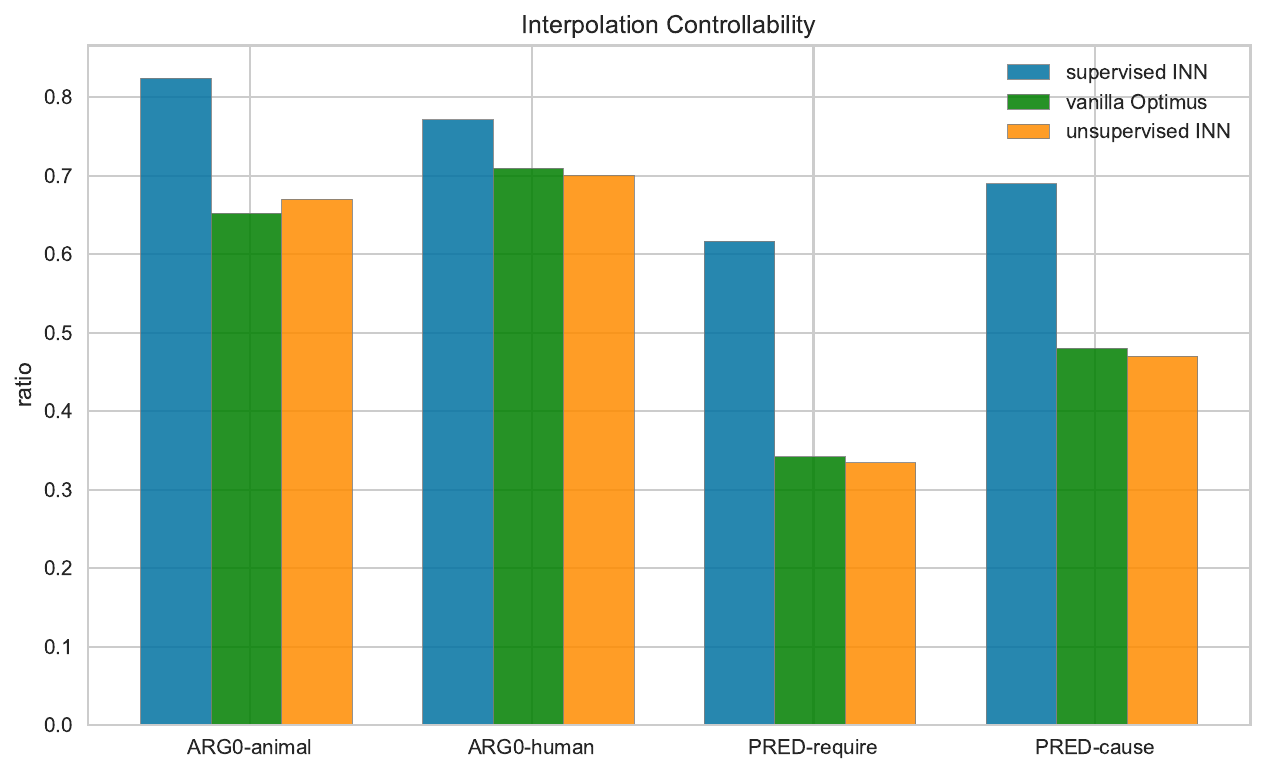}
    \caption{Interpolation control evaluation, we can observe that supervised INN with better semantic separability can lead to better localised semantic control.}
    \label{fig:interpolate_ratio}
\end{figure}

We then quantitatively evaluate the localisation of interpolation. We randomly select 200 sentence pairs from the dataset holding the same role-content and report the ratio of intermediate sentences with the same role-content as inputs. As illustrated in Figure \ref{fig:interpolate_ratio}, the intermediate sentences from the supervised INN can better hold the same role-content as inputs, especially for \textit{predicate} which usually has a lower effect on distributional sentence semantics \citep{zhang2022}, indicating that our supervision can lead to better latent space separability and generation control.

\paragraph{Interpolation smoothness.} Moreover, we quantitatively evaluate the latent space geometry via interpolation smoothness metrics (IS), which calculates the ratio between the ideal semantic distance (i.e., the aligned semantic distance between source and target sentences) and the actual semantic distance (i.e., the sum of semantic distance between adjacent sentences during interpolation). A higher ratio indicates that the actual path aligns better with the ideal path, suggesting better semantic-geometric properties. The metric is defined as:
$$\text{IS} = \mathbb{E}_{(s_0, ..., s_T) \sim P} \frac{\delta(\text{align}(s_0, s_T))}{\sum^T_{t=0} \delta(\text{align}(s_t, s_{t+0.1}))}$$
where $s_0, ..., s_T$ is the sequence of sentences during interpolation, $\delta$ and $\text{align}$ are sentence similarity and alignment functions, respectively, which are performed via Word Mover’s Distance \citep{zhao-etal-2019-moverscore}.
We choose the standard language VAE baselines (i.e., the prior is the std. Gaussian distribution). We randomly sample 200 sentence pairs and report the IS metric. \uline{As illustrated in Table \ref{tab:interpolation_smoothness_inn}, our model can deliver smoother interpolations compared to the baselines, indicating that semantic disentanglement can lead to better latent space geometry} \textbf{(Finding~7)}. 

\begin{table}[ht!]
\setlength\tabcolsep{2.5pt}
\small
\centering
\renewcommand\arraystretch{1}
\begin{tabular}{llll}
\toprule
Evaluation Metrics & avg IS$\uparrow$ & max IS$\uparrow$ & min IS$\uparrow$ \\ \hline
DAE \citep{10.1145/1390156.1390294} & 0.144 & 0.330 & 0.055 \\
AAE \citep{makhzani2016adversarial} & 0.142 & 0.284 & 0.054 \\
LAAE\citep{rubenstein2018latent} & 0.172 & 0.347 & 0.056 \\ 
DAAE \citep{shen2020educating} & 0.055 & 0.061 & 0.023 \\ 
$\beta$-VAE \citep{Higgins2016betaVAELB} & 0.198 & 0.379 & 0.041 \\ 
AdaVAE \citep{tu2022adavae} & 0.085 & 0.105 & 0.050 \\
Della \citep{hu-etal-2022-fuse} & 0.253 & 0.416 & 0.155 \\ 
Optimus \citep{li2020optimus} & 0.220 & 0.525 & 0.130 \\
AutoEncoder (BERT-GPT2) & 0.259 & 0.585 & 0.165 \\
INN (U) (our) & 0.251 & 0.540 & 0.159 \\
INN (C) (our) & \underline{\textbf{0.282}} & \underline{\textbf{0.607}} & \underline{\textbf{0.206}} \\ 
\toprule
\end{tabular}
\caption{Geometrical examination via IS metric.} \label{tab:interpolation_smoothness_inn}
\end{table}

\section{Related Work}

\paragraph{Sentence disentanglement.} In the Natural Language Generation domain, most previous investigations explored the disentanglement between two specific linguistic perspectives, such as sentiment-content \citep{john2019disentangled,li-etal-2022-variational-autoencoder}, semantic-syntax \citep{bao2019generating}, and negation-uncertainty \cite{vasilakes-etal-2022-learning}, or syntactic disentanglement \citep{mercatali2021disentangling, felhi2022towards}. In this work, we provide a formal-geometrical lens, with the support of \textit{argument structures} as a sentence representation model, for sentence disentanglement targeting for localised semantic control. This work is the first integration of flow-based INN mechanisms to improve disentanglement, separation and semantic control of sentence spaces. 


\paragraph{INNs in NLP.} \citet{csahin2020two} concentrate on modelling morphological inflection and lemmatisation tasks, utilising INN to learn a bijective transformation between the word surface and its morphemes. \citet{li2020sentence} proposed BERT-flow, transforming sentences from a BERT sentence space to a standard Gaussian space. \citet{ding-gimpel-2021-flowprior} deployed flow-based INN to enrich VAE prior distribution, while \citet{gu-etal-2023-controllable} use flow mechanisms to control attributes in style transfer tasks. This work focused on semantic separability, geometrical operations and control over the distributed representation of sentences. Moreover, this work is the first to explore geometrical data augmentation to support semantic disentanglement.

\section{Conclusion}
This work addresses the task of Sentence Semantic Disentanglement. We propose a flow-based Invertible Neural Network (INN) mechanism within the Optimus model to achieve more disentangled and well-separated latent sentence spaces. Our aim is to bridge formal and distributional perspectives on sentence semantics. Experimental results demonstrate that the proposed approach transforms the distributed hidden space of an autoencoder into a latent space in which formal semantic features are more effectively localised, interpolated, and controlled. This finding answers the \questionB{}

\paragraph{Reproducibility.} The data and the related codebase are available online: \url{https://github.com/SnowYJ/disentangledINN}.

\section{Scoping and Limitations}
This work focuses on disentangling shallow semantic structures, with particular emphasis on semantic role, as formal semantic representation is grounded in syntactic structure. A promising future direction is toward deeper structures by modelling syntactic trees within the latent space, or non-shallow semantic representation paradigms.

Moreover, while the language autoencoder with unsupervised INN exhibits a distinct learning pattern with regard to semantic distribution, further understanding is required in terms of the semantic distribution of unsupervised INNs in language modelling tasks. Furthermore, this study exclusively focused on a corpus of explanations. The exploration of its performance on other types of sentences, including sentences with complex clausal-phrasal constructions, or sentences with non-compositional idioms, is yet to be undertaken.

\chapter{Syntax Representation} \label{cha:syntax}
In Chaper~\ref{cha:geo} and \ref{cha:dis}, we propose formal semantic geometry and disentanglement to bridge the formal and distributional semantics. In formal semantics theory, syntactic tree information serves as an essential first step in determining the semantic roles and relational structure of a sentence. Therefore, this chapter investigates \questionC{}
\section{Introduction}

Injecting explicit syntactic information in Variational AutoEncoders (VAEs) \citep{kingma2013auto} has led to improved performance on several language generation tasks, such as paraphrasing and translation \citep{dai2018syntax,chen-etal-2017-improved,felhi-etal-2022-exploiting,yang-etal-2021-syntactically}. Among existing techniques, a line of research explores syntactic injection via sentence-level semantics-syntax disentanglement, which consists in the explicit separation of distributional semantic and structural syntactic features through the optimisation of heterogeneous latent spaces \citep{bao-etal-2019-generating, chen-etal-2019-multi, zhang-etal-2019-syntax-infused}. Such methods, implemented under multi-task learning or dual encoder architectures, have been demonstrated to improve: (i) generation controllability and interpretability \citep{bao-etal-2019-generating}, (ii) robustness and generalisation, (iii) fine-grained representation and latent space organisation \citep{chen-etal-2019-multi}, and more importantly (iv) injecting syntactic features into VAEs can allow for optimization in low-dimensional and regularized latent Gaussian space, rather than \textit{complex discrete sequence spaces} as investigated in previous work \citep{pouran-ben-veyseh-etal-2020-graph,zanzotto-etal-2020-kermit,li-etal-2023-explicit,mohammadshahi-henderson-2023-syntax}, which represents an efficient to improve text generation \citep{qin-etal-2020-back,kumar2021controlled}. However, most of these methods focus on LSTM-based VAEs, and their effectiveness for larger architectures based on Transformers, such as Optimus \citep{li2020optimus}, is still under-explored.

To combine the benefits of larger pre-trained VAEs and latent separation methods, this paper focuses on the injection of structural syntactic information in Transformer-based VAEs (i.e., Optimus \citep{li2020optimus}). Specifically, we investigate a first overarching research question: \textit{``RQ1: How can we best capture explicit syntactic information in the latent space of Transformer-based VAEs?''} 
we address this question by directly intervening on the Optimus architecture to induce a latent space separation via graph-based \citep{kipf2016semi} and sequential neural encoders \citep{devlin2018bert}. Specifically, our hypothesis is that Graph Neural Networks (GNNs) \citep{kipf2016semi,hamilton2017inductive,yun2020graph} can induce specialised and complementary latent representations that can better capture structural syntactic relations and alleviate the information bottleneck in VAEs' semantic encoder \citep{alemi2016deep,tenney2019you} (i.e. trade-off between semantics and syntax). 

\begin{figure}[ht!]
    \centering
    \includegraphics[width=12cm]{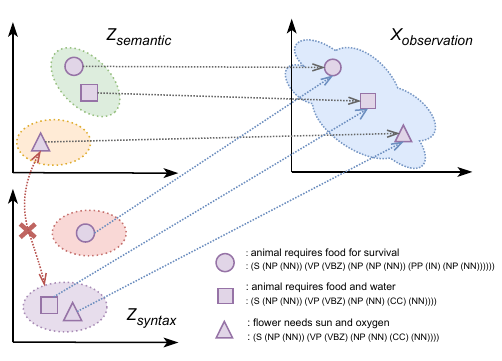}
    \caption{Decoding under heterogeneous syntactic-semantic spaces can result in two main challenges: (i) The syntactic representations may not possess a one-to-one mapping with the semantic representations (i.e., one syntactic structure can correspond to multiple sentence representations), (ii) the optimisation of heterogeneous latent spaces can result in different latent distributions, making generation hard to control.}
    \label{fig:latent_space_geometry}
\end{figure}

Subsequently, we focus on the problem of leveraging multiple, specialised latent spaces derived from the dual encoder architecture for decoding. This leads to several challenges (Figure \ref{fig:latent_space_geometry}) since (i) the syntactic representations may not possess a one-to-one mapping with the semantic representations (i.e., one syntactic structure can correspond to multiple sentence representations), (ii) the optimisation of heterogeneous latent spaces can result in different latent distributions, a feature that can affect decoding and language generation performance, and (iii) compared with an LSTM decoder, Transformer-based decoders (e.g., GPT2) are typically larger and contain information acquired during pre-training, being more difficult to control. 

Those challenges lead to our second research question: \textit{``RQ2. How can multiple, specialised latent spaces be effectively injected into the VAE decoder?''} To answer it, we investigate injection mechanisms for Transformer-based VAEs via the following methods: (i) we separately inject syntax and semantic representations into the attention weights of the decoder (i.e., Query and Key-Value), and (ii) consider low-rank injections, including \textit{addition}, \textit{memory} \citep{li2020optimus}, and \textit{tensor fusion} \citep{liu-etal-2018-efficient-low}, which directly operate over the attention matrices and potentially reduce information redundancy \citep{hu-etal-2022-fuse}. 


We perform extensive experiments to evaluate the resulting VAE architectures on both mathematical expressions \citep{valentino2023multioperational,meadows2023symbolic} and natural language explanatory sentences \citep{jansen2018worldtree}. Overall, our contributions can be summarised as follows: 

\textbf{1.} We propose a dual encoder architecture for Transformer-based VAEs integrating graph-based and sequential models to better capture and disentangle semantic and structural syntactic features in multiple, specialised latent spaces. 

\textbf{2.} We explore the injection of such representations into the decoder of Transformer-based VAEs via low-rank vector operations to better guide the generation process. 

\textbf{3.} We perform extensive experiments showing that the adoption of a graph-based encoder coupled with a transformer encoder can reduce the loss of information in the sentence bottleneck, resulting in improved reconstruction and language modelling. Overall, we found that the proposed VAE architecture can significantly improve performance and generalisation when compared to sentence-level VAE baselines. 


\section{Methodology}

\begin{figure*}[t]
    \centering
    \includegraphics[width=\textwidth]{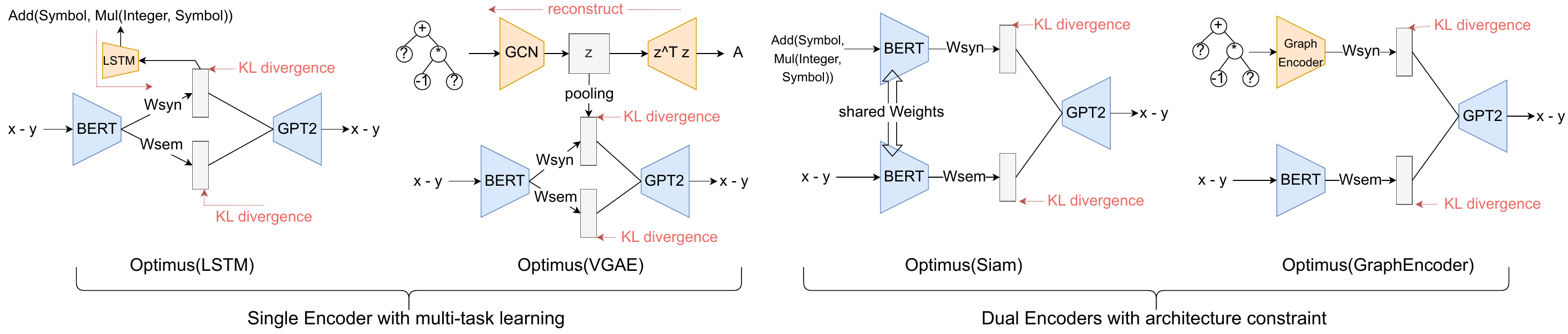}
    \caption{Overview of different methods to explicitly represent and disentangle syntactic information in the latent space of Transformer-based VAEs.}
    \label{fig:sem_syn_baselines_main}
\end{figure*}

Our methodology consists of two main phases. First, we investigate different encoding strategies to explicitly capture syntactic and structural information in a separate latent space. Subsequently, we explore techniques to fuse syntactic and semantic features and inject them into the decoder model.
Regarding the encoding phase, we explore four architectures based on two different configurations (i.e., \textit{multi-task learning} and \textit{dual encoder}) integrating both \textit{sequential} and \textit{graph-based} models under Optimus (BERT-GPT2) \textit{memory} setup (see Figure \ref{fig:sem_syn_baselines_main}). Regarding the decoding phase, we consider the best encoding configuration in terms of syntactic representation and propose different injection mechanisms via low-rank operations over attention-weight matrices of GPT2. The following sections describe each phase in detail (Sections \ref{sec:encoding_phase} and \ref{sec:decoding_phase}), including how the encoding and decoding stages are integrated into an end-to-end VAE architecture (Section \ref{sec:arc}).

\subsection{Graph-based Syntax Encoding} \label{sec:encoding_phase}

\paragraph{Multi-Task Learning.}
\citet{bao-etal-2019-generating} proposed a multi-task learning strategy to achieve such a goal in LSTM-based VAEs via learning and fusing two distinct latent representations. They adopt a separate space for encoding explicit syntactic dependencies through the adoption of an LSTM decoder used to reconstruct flattened constituency parse trees. Here, we build upon this setup to enrich the latent representation in Optimus \cite{li2020optimus}. Specifically, given a separate latent syntax representation, $z_{syn}$, encoded via BERT \cite{devlin2018bert}, we explore the following mechanisms (see Figure \ref{fig:sem_syn_baselines_main}):

\textbf{1.} Similarly to \cite{bao2019generating}, we adopt an LSTM \cite{10.1162/neco.1997.9.8.1735} decoder to generate linearised syntactic trees, where $z_{syn}$ is fed into the first hidden state of the LSTM. We refer to this configuration as \textit{Optimus (LSTM)}.

\textbf{2.} We jointly train a Variational Graph AutoEncoder (VGAE, \citet{kipf2016variational}) on syntactic trees, where the latent node embeddings are mean-pooled into a sentence-level syntax representation $z^{gcn}_{syn}$.  We refer to this configuration as \textit{Optimus (VGAE)}.
Here, the syntactic representations $z^{gcn}_{syn}$ and $z_{syn}$ can be optimised via MSE in a multi-task setting. Specifically, the general objective function can be formalised as:
\[
\begin{aligned}
\mathcal{L}_{\text{VAE}} = & \mathbb{E}_{q_\phi(z_{sem},z_{syn}|x)} \Big[ \log p_{\theta}( x | z_{sem}, z_{syn}) \Big] \\
& - \text{KL}(\phi(z_{sem}|x)||p(z)) - \text{KL}(\phi(z_{syn}|x)||p(z)) + \mathcal{L}_{\text{syn}}(z_{syn})
\end{aligned}
\]
Where $q_\phi, p_{\theta}$ represent the encoder and decoder. The objective functions for optimising the syntactic spaces $\mathcal{L}_{\text{syn}}(z_{syn})$ can be specialised according to the model:
LSTM: 
$$\mathcal{L}^{lstm}_{\text{syn}}(z_{syn})=\sum^n_{i=1} \log p(s_i|s_1, \dots, s_{i-1}, z_{syn})$$
and VGAE:
$$\mathcal{L}^{vgae}_{\text{syn}}(z_{syn})=\sum^{dim}_{j=1}(z^{j}_{gcn} - z_{syn}^j)^2 + \mathcal{L}^{vgae}(A, N)$$
Where $s_i$ represents the token of a flattened syntax tree, while $A$ and $N$ are the Adjacent matrix and Node embeddings of the syntax tree. Additional details for the VGAE model and the optimisation of $\mathcal{L}^{vgae}$ can be found in the original paper \cite{kipf2016variational}. 

\paragraph{Dual Encoder.} In addition to the multi-task learning setup, we build upon \citet{zhang-etal-2019-syntax-infused,huang-chang-2021-generating} which propose two distinct language encoders to induce syntactic disentanglement. Specifically, we experiment with: 

\textbf{1.} Two distinct BERT encoders via a Siamese neural network. We refer to this configuration as \textit{Optimus (Siam)}.

\textbf{2.} A Graph encoder, such as GCN \cite{kipf2016semi}, GraphSAGE \cite{hamilton2017inductive}, and Graph Transformer (TransCONV, \citet{yun2020graph}), coupled with a BERT encoder. We refer to this configuration as \textit{Optimus (GraphEncoder)}.
Here, the general objective function can be formalised as:
\[
\begin{aligned}
& \mathbb{E}_{q^{sem}_\phi(z_{sem}|x), q^{syn}_\phi(z_{syn}|x_{syn})} \Big[ \log p_{\theta}( x | z_{sem}, z_{syn}) \Big]  \\
& - \text{KL}(\phi(z_{sem}|x)||p(z)) - \text{KL}(\phi(z_{syn}|x)||p(z))
\end{aligned}
\]
Where $q^{sem}_\phi, q^{syn}_\phi$ represent semantic and syntax encoders respectively, while $x_{syn}$ represents the input for the syntax encoder. For graph encoders, we represent $x_{syn}$ using an adjacency matrix and node embedding pairs. For the language syntax encoder, on the other side,  we represent $x_{syn}$ as a flattened syntactic tree without word content.

As our experiments revealed that the dual graph-sequential encoder configuration (i.e., \textit{Optimus (GraphEncoder)}) can achieve the best results in terms of syntactic representation (see Table \ref{tab:enoding_syntax}), we consider this setup for integration into an end-to-end VAE architecture (see Section \ref{sec:arc}).

\subsection{Heterogeneous Spaces Decoding} \label{sec:decoding_phase}
\begin{figure*}[ht!]
    \centering
    \includegraphics[width=\textwidth]{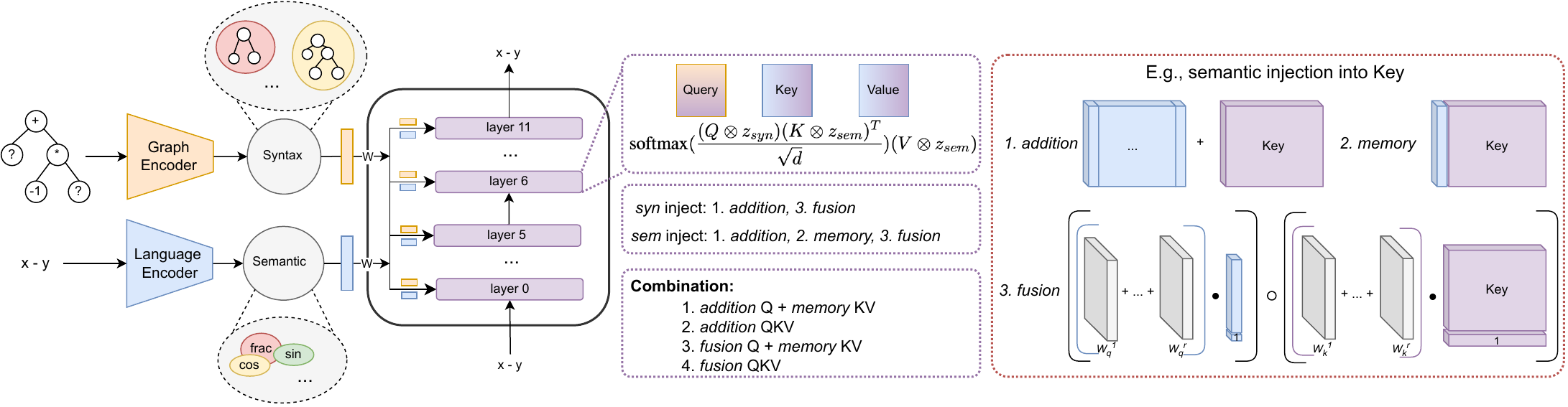}
    \caption{Architectural overview. Semantic and syntactic features are encoded into heterogeneous latent spaces via graph-based and sequential encoders. The resulting latent spaces are injected into the GPT2 decoder via low-rank operations.}
    \label{fig:overview}
\end{figure*}
To preserve the separation of the latent spaces and, at the same time, leverage heterogeneous representations during decoding, we explore methods to inject semantic (i.e., $z_{sem}$) and syntactic space (i.e., $z_{syn}$) directly into the attention mechanism of GPT2 (via QKV). Specifically, we inject different latent representations to different attention weights:
\begin{equation}
\begin{split}
    \text{softmax}( \frac{(Q \otimes z_{syn}) (K \otimes z_{sem})^T}{\sqrt{d}})(V \otimes z_{sem}) \nonumber
\end{split}
\end{equation}
Where $\otimes$ represents the latent injection operation.
As for syntactic injection ($z_{syn}$), we consider two kinds of low-rank operations $\otimes$, \textit{addition}, and \textit{fusion} \cite{liu-etal-2018-efficient-low}, which directly work on attention weights. As for \textit{addition}, we inject $z_{syn}$ into each low-rank token representation in Q, which can be formalised as follows: $\tilde{Q} = \sum_{i=1}^{seq} Q[i, :] + z_{syn}$ where $\tilde{Q}$ represents the new Q values obtained after syntax injection. 

As for \textit{fusion}, we adapt the tensor fuse operation \cite{liu-etal-2018-efficient-low, hu-etal-2022-fuse}. In more detail, given a hyper-parameter, rank $r=4$, the $\tilde{Q}$ can be described as: 
$$\tilde{Q} = (\sum_{i=1}^r W_q^i [Q;\mathbbm{1}]) \circ (\sum_{i=1}^r W_z^{i, syn} [z_{syn};\mathbbm{1}])$$ 
Where $\mathbbm{1}$ is the matrix of ones, $W_z^{i, syn}$ and $W_q$ are the trainable linear transformations. 

As for semantic injection ($z_{sem}$), we consider three operations: \textit{addition}, \textit{memory}, and \textit{fusion}, where \textit{addition} and \textit{fusion} operations are the same as before but works on KV. \textit{Memory} is the same as Optimus \textit{memory} injection \cite{li2020optimus}. 
We refer \cite{liu-etal-2018-efficient-low} for an in-depth description of tensor fusion. 
\subsection{End-to-End VAE Architecture} \label{sec:arc}

\paragraph{Encoder.} At the encoding stage, we consider the dual graph-sequential encoding mechanism adopting BERT as a sequential encoder and experimenting with two different graph-based encoders, including GraphSAGE \cite{hamilton2017inductive}, and TransCONV \cite{yun2020graph}. The dual graph-sequential encoding can alleviate the information bottleneck derived from the encoding stage. To derive the syntactic space, $z_{syn}$, we use a mean pooling operation to obtain a sentence-level representation from the node embeddings $N$ and the adjacency matrix $A$: $$\text{Embed}_{syn} = \text{MeanPool}(\text{GraphEnc}(A, N))$$

For the semantic space, $z_{sem}$, we consider the special token [CLS] in BERT as the input of a linear transformation ($W$) to obtain a sentence-level representation: $\text{Embed}_{sem} = W(\text{LanguageEnc}(x)_{\text{[CLS]}})$
Where $x$ is the input sentence. Both spaces are constrained to follow a Gaussian distribution by learning the parameters $\mu$ and $\sigma$ through multilayer perceptrons $W_\mu^{sem}$, $W_{\sigma}^{sem}$, $W_\mu^{syn}$, and $W_{\sigma}^{syn}$. The final latent representations can be obtained via: $$z_{sem(syn)} = W_{\mu}^{sem(syn)} \times \text{Embed}_{sem(syn)} + W_{\sigma}^{sem(syn)}$$

\paragraph{Decoder.} Since the architecture constraint, $z_{sem}$ and $z_{syn}$ have the potential to capture diverse features with a high level of disentanglement. To this end, we experiment with different decoding injection setups and low-rank operations: (1) \textit{addition} for QKV (i.e., addition QKV), (2) \textit{fusion} for QKV (fusion QKV), (3) \textit{addition} for Q and \textit{memory} for KV (addition Q), and (4) \textit{fusion} for Q and \textit{memory} for KV (fusion Q).

\paragraph{Optimisation.} Our model can be trained via Evidence Lower Bound (ELBO) $x$ \cite{kingma2013auto}. To avoid the KL vanishing issue, which refers to the Kullback-Leibler (KL) divergence term in the ELBO becomes very small or approaches zero, we select the cyclical schedule to increase weights of KL $\beta$ from 0 to 1 \citep{fu-etal-2019-cyclical} and a KL thresholding scheme \citep{li-etal-2019-surprisingly} that chooses the maximum between KL and threshold $\lambda$. The final objective function can be described as follows:
\begin{align*} \label{eq:elbo_loss}
& \mathcal{L}_\text{VAE} = \mathbb{E}_{q^{sem}_\phi(z_{sem}|x), {q^{syn}_\phi(z_{syn}|A, N)}} \Big[ \log p_{\theta}( x | z_{sem} , z_{syn}) \Big] \\
& - \beta \max \left[ \lambda , \text{KL} q^{sem}_\phi(z_{sem}|x) || p(z) \right ] - \beta \max \left[ \lambda , \text{KL} q^{syn}_\phi(z_{syn}|x) || p(z) \right ]
\end{align*}

\section{Empirical Evaluation} \label{sec:empirical}

Following the stages in our methodology, we first evaluate different encoding setups for injecting syntactic information into VAEs (Section \ref{sec:encoding_phase}). Subsequently, we consider the best encoding configuration to examine which decoding strategy (Section \ref{sec:arc}) can lead to better language modelling performances. Finally, we evaluate the best architectural setup for downstream tasks. To experiment, we focus on both \textit{explanatory sentences} and \textit{mathematical expressions}. The rationale behind this choice is that (1) explanatory sentences \citep{jansen2018worldtree,valentino2022hybrid,thayaparan-etal-2021-explainable, zhang2023type} provide a semantically challenging yet sufficiently well-scoped scenario to evaluate the syntactic and semantic organisation of the space; (2) mathematical expressions \citep{valentino2023multioperational,meadows2023symbolic} follow a well-defined syntactic structure and set of symbolic rules that are notoriously difficult for neural models. 




\subsection{Encoding: Latent Representations} \label{sec:enc_syn}
\paragraph{Evaluation.} Firstly, we evaluate different encoding setups to the effect of semantic-syntax distribution in latent space from three perspectives: (i) latent space geometry: whether the latent space can capture the corresponding features -- i.e., sentences with the same/different features are clustered/separated accordingly in the latent space. In this case, we can evaluate the organisation of the latent space via MSE of k-means \cite{zhang2023learning, michlo2023overlooked}; (ii) syntactic features: following the probing method \cite{conneau-etal-2018-cram}, we train a linear classifier to predict tree depth. Here, better classification performances indicate a higher separability of syntactic features in the latent space; and (iii) semantic and syntax space alignment: we adopt statistical metrics to compare latent distributions such as Mutual Information (MI), Kullback–Leibler divergence (KL), and Wasserstein distance (Wass). \uline{As illustrated in Table \ref{tab:enoding_syntax}, we can observe that (1) the Optimus(GraphEncoder) can better capture the syntactic structures and induce a better latent space separation, (2) It can lead to a better organisation of the semantic space $\text{MSE}(sem)$} \textbf{(Finding~1)}. We will further explore this phenomenon in subsequent sections.
\begin{table*}[ht!]
\resizebox{\columnwidth}{!}{
\small
\centering
\begin{tabular}{lcccc|cccccccc}
\toprule
Corpus & \multicolumn{4}{c|}{\textit{Mathematical expression}} & \multicolumn{5}{c}{\textit{Explanatory sentences}} \\
Proxy metrics & $\text{MSE}$(sem)$\downarrow$ & MSE(syn)$\downarrow$ & $\text{Acc}_{dep}$(syn)$\uparrow$ & $\text{Acc}_{dep}$(sem)$\downarrow$  & $\text{MSE}$(sem)$\downarrow$ & MSE(syn)$\downarrow$ & $\text{Acc}_{dep}$(syn)$\uparrow$ & $\text{Acc}_{dep}$(sem)$\downarrow$ & $\text{F1}_{dep}$(sem)$\downarrow$\\ \midrule 
LSTM & 079.02 & 070.48 & 000.74 & 000.74 & 176.39 & 158.03 & 000.40 & 000.40 & 000.41\\
VGAE & 125.68 & 434.52 & 000.81 & 000.82 & 169.42 & 110.30 & 000.40 & 000.38 & 000.45 \\
Siam & 191.97 & 053.90 & 000.85 & 000.52 & 074.86 & 031.95 & 000.43 & 000.35 & 000.42 \\
GraphEncoder & -- & -- & -- & -- & -- & -- & -- & -- & -- \\
 + GCN & \underline{\textbf{\textcolor{black}{004.31}}} & 065.79 & 000.72 & \underline{\textbf{\textcolor{black}{000.27}}} & 069.77 & 091.94 & 000.49 & \underline{\textbf{\textcolor{black}{000.12}}} & \underline{\textbf{\textcolor{black}{000.30}}} \\
 + GraphSAGE & 208.21 & 053.20 & \underline{\textbf{\textcolor{black}{000.98}}} & 000.52 & 058.12 & 004.10 & 000.50 & 000.39 & 000.46  \\
 + TransConv & 249.00 & \underline{\textbf{\textcolor{black}{038.30}}} & \underline{\textbf{\textcolor{black}{000.98}}} & 000.57 & \underline{\textbf{\textcolor{black}{058.10}}} & \underline{\textbf{\textcolor{black}{003.35}}} & \underline{\textbf{\textcolor{black}{000.51}}} & 000.38 & 000.47 \\ \midrule
$\text{F1}_{dep}^*$(sem)$\downarrow$ & $\text{F1}_{dep}$(syn)$\uparrow$ & MI(sem,syn)$\downarrow$ & KL(sem||syn)$\uparrow$ & Wass(sem,syn)$\uparrow$ & $\text{F1}_{dep}$(syn)$\uparrow$ & MI(sem,syn)$\downarrow$ & KL(sem||syn)$\uparrow$ & Wass(sem,syn)$\uparrow$\\ \midrule
\multicolumn{1}{c}{000.71} & 000.70 & 004.88 & 005.74 & 000.53 & 000.43 & 004.87 & 001.01 & 000.78\\
\multicolumn{1}{c}{000.84} & 000.84 & 004.85 & 026.12 & 000.32 & 000.44 & 004.66 & 007.04 & 000.90 \\
\multicolumn{1}{c}{000.41} & 000.87 & 004.85 & 011.95 & 000.69 & 000.44 & 004.96 & 008.72 & 000.80\\
\multicolumn{1}{c}{--} & -- & -- & -- & -- & -- & -- & -- & --\\
\multicolumn{1}{c}{\underline{\textbf{\textcolor{black}{000.24}}}} & 000.79 & 004.82 & 024.05 & 000.72 & \underline{\textbf{\textcolor{black}{000.54}}} & 004.78 &  011.77 & 000.30 \\
\multicolumn{1}{c}{000.42} & \underline{\textbf{\textcolor{black}{000.98}}} & 005.04 & 005.12 & 000.69 & 000.44 & 004.45 & \underline{\textbf{\textcolor{black}{043.45}}} & \underline{\textcolor{black}{\textbf{001.92}}} \\
\multicolumn{1}{c}{000.52} & \underline{\textbf{\textcolor{black}{000.98}}} & \underline{\textbf{\textcolor{black}{004.80}}} & \underline{\textbf{\textcolor{black}{031.63}}} & \underline{\textbf{\textcolor{black}{001.19}}} & 000.48 & \underline{\textbf{\textcolor{black}{003.54}}} & 012.78 & 000.75\\ \bottomrule
\end{tabular}
}
\caption{Proxy metrics for evaluating the organisation of the latent syntactic and semantic space for different encoding configurations of Optimus. The \underline{\textbf{\textcolor{black}{best}}} results indicate that the graph-language encoding setup can effectively capture syntactic information and maintain separation.} \label{tab:enoding_syntax}
\end{table*}

\begin{figure}[ht!]
    \centering
    \includegraphics[width=\linewidth]{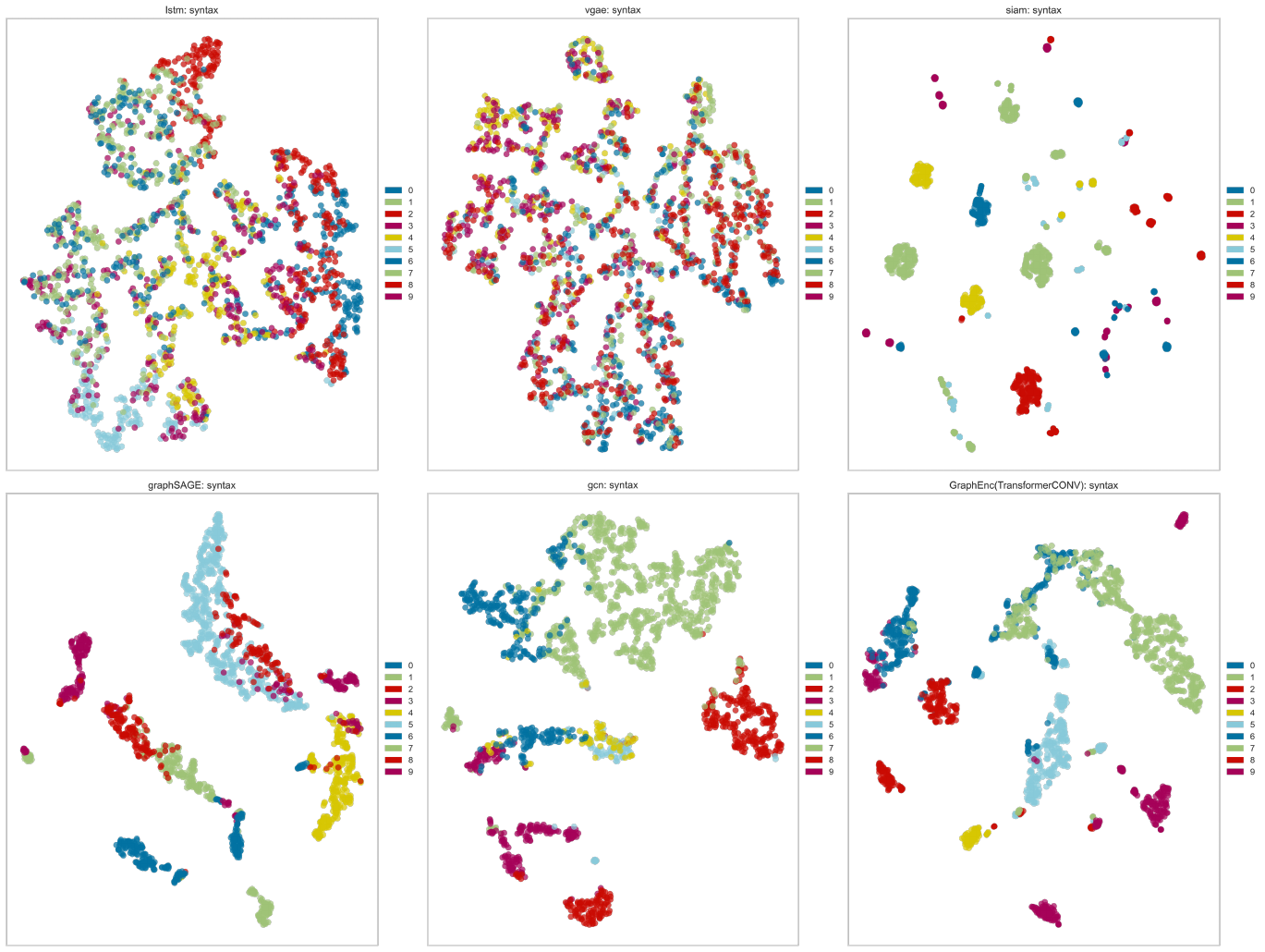}
    \caption{Visualising the syntactic clusters for mathematical expressions reveals that graph encoders can better represent syntactic information in latent space (top: LSTM, VGAE, Siam, bottom: graph encoders with GraphSAGE, GCN, TransformerCONV).}
    \label{fig:syntax_latent_space}
\end{figure}

\paragraph{Visualisation.} Next, we visualise the cluster separation of latent space via t-SNE (see Figure \ref{fig:syntax_latent_space}). From the visualisation, we can observe that the Optimus injection with a separated GraphEncoder can induce a better separation between different syntactic clusters. \uline{These results reveal that the integration of graph-based and sequential models in a dual-encoder setup can better capture structural syntactic information while maintaining a separation between latent spaces} \textbf{(Finding~2)}.

\subsection{Decoding: Language Modelling} \label{sec:dec_lm}
\begin{table*}[ht!]
\scriptsize
\setlength\tabcolsep{2.5pt}
\resizebox{\columnwidth}{!}{
\small
\centering
\renewcommand\arraystretch{1}
\begin{tabular}{lccccccccccccccc}
\toprule
\multicolumn{1}{c}{Corpus} & \multicolumn{10}{c}{\textit{Mathematical expression}} & \multicolumn{5}{c}{\textit{Explanatory sentences}} \\ 
\multicolumn{1}{c}{Metrics} & \multicolumn{2}{c}{EVAL} & \multicolumn{2}{c}{VAR-SWAP} & \multicolumn{2}{c}{EASY} & \multicolumn{2}{c}{EQ-CONV} & \multicolumn{2}{c}{LEN} & BLEU & BLEURT & Cosine & Loss$\downarrow$ & PPL$\downarrow$ \\ \midrule

\multicolumn{16}{c}{\textit{sentence VAE baselines}} \\
01. AAE(768) & 0.10 & 0.75 & 0.00 & 0.25 & 0.02 & 0.53 & 0.00 & 0.54 & 0.00 & 0.51 & 0.35 & -0.95 & 0.80 & 3.35 & 28.50\\
02. LAAE(768) & 0.00 & 0.43 & 0.00& 0.25 & 0.00 & 0.27 & 0.00 & 0.29 & 0.00 & 0.44 & 0.26 & -1.07& 0.78& 3.71 & 40.85\\
03. DAAE(768) & 0.00 & 0.24 & 0.00& 0.21 & 0.00 & 0.21 & 0.00 & 0.22 & 0.00 & 0.42 & 0.22 & -1.26& 0.76& 4.00 & 54.59\\
04. $\beta$-VAE(768) & 0.00 & 0.14 & 0.00& 0.15 & 0.00 & 0.13 & 0.00 & 0.14 & 0.00 & 0.35 & 0.06& -1.14& 0.77& 3.69 & 40.04\\ 
05. Optimus(768) & 0.99 & 0.99 & 0.00& \underline{\textbf{\textcolor{black}{0.38}}} & 0.81 & 0.93 & 0.00& 0.81 & \underline{\textbf{\textcolor{black}{0.14}}} & 0.76 & 0.35 & -0.59 & 0.83 & 0.98 & 2.66 \\ \midrule 
\multicolumn{16}{c}{\textit{different encoding setups with memory injection}} \\
06. LSTM & \underline{\textbf{\textcolor{black}{1.00}}} & \underline{\textbf{\textcolor{black}{1.00}}} & 0.00& 0.35 & 0.73 & 0.94 & 0.00& 0.77 & 0.06 & 0.74 & 0.41 & -0.41 & 0.85 & 1.04 & 2.82 \\ 
07. VGAE & 0.98 & 0.99 & 0.00& 0.34 & 0.72 & 0.93 & 0.00& 0.74 & 0.04 & 0.71 & 0.26 & -0.91 & 0.78 & 1.14 & 2.55 \\
08. Siam & \underline{\textbf{\textcolor{black}{1.00}}} & \underline{\textbf{\textcolor{black}{1.00}}} & 0.00& 0.30 & 0.22 & 0.80 & 0.00& 0.78 & 0.03 & 0.75 & 0.49 & -0.15 & 0.88 & 0.94 & 2.55 \\ 
GraphEncoder \\ 
09. + GCN & 0.00 & 0.40 & 0.00& 0.22 & 0.00 & 0.27 & 0.00& 0.37 & 0.00 & 0.43 & 0.15 & -1.19 & 0.75 & 1.24 & 3.45 \\
10. + GraphSAGE & 0.88 & 0.96 & 0.00& 0.28 & 0.06 & 0.46 & 0.00& 0.69 & 0.00 & 0.60 & 0.45 & -0.28 & 0.87 & 1.00 & 2.71 \\
11. + TransCONV & 0.89 & 0.95 & 0.00& 0.28 & 0.14 & 0.53 & 0.00 & 0.67 & 0.00 & 0.61 & 0.17 & -1.16 & 0.75 & 1.21 & 3.35 \\ \midrule
\multicolumn{16}{c}{\textit{Graph-language encoders: injecting syntax into Q, semantic into KV}} \\
BERT-GraphSAGE & \\
12. + addition Q & 0.99& 0.99 & 0.00 & 0.27 & 0.23& 0.63& 0.00& 0.71 &0.02&0.66& 0.60 &0.22&0.92 &0.74 & 2.09 \\
13. + addition QKV & \underline{\textbf{\textcolor{black}{1.00}}} & \underline{\textbf{\textcolor{black}{1.00}}} & 0.00 & 0.35 & 0.65 & 0.90 &0.00& 0.80 & 0.06 & 0.75 & 0.63 & 0.31 & 0.93 & 0.65 & 1.91 \\
14. + fusion Q & 0.94 & 0.97 & 0.00 & 0.29 & 0.08 & 0.63 & 0.00 & 0.71 & 0.00 & 0.62 & 0.55 & 0.03 & 0.91 & 0.90 & 2.45 \\ 
15. + fusion QKV & \underline{\textbf{\textcolor{black}{1.00}}} & \underline{\textbf{\textcolor{black}{1.00}}} &0.00& \underline{\textbf{\textcolor{black}{0.38}}} & 0.37 & 0.84 & 0.00 & 0.80 & 0.02 & 0.73 & 0.46 & -0.23 & 0.88 & 1.10 & 3.00 \\
BERT-TransCONV & \\
16. + addition Q & 0.98& 0.99 &0.00& 0.26&0.31&0.69&0.00&0.67&0.01&0.63& 0.59 & 0.18 & 0.92 & 0.76 & 2.13 \\
17. + addition QKV & \underline{\textbf{\textcolor{black}{1.00}}} & \underline{\textbf{\textcolor{black}{1.00}}} & 0.00 & \underline{\textbf{\textcolor{black}{0.38}}} & \underline{\textbf{\textcolor{black}{0.90}}} & \underline{\textbf{\textcolor{black}{0.98}}} &0.00 & \underline{\textbf{\textcolor{black}{0.82}}} & 0.10 & \underline{\textbf{\textcolor{black}{0.78}}} & \underline{\textbf{\textcolor{black}{0.65}}} & \underline{\textbf{\textcolor{black}{0.35}}} & \underline{\textbf{\textcolor{black}{0.94}}} & \underline{\textbf{\textcolor{black}{0.62}}} & \underline{\textbf{\textcolor{black}{1.85}}} \\
18. + fusion Q & 0.96 & 0.98 & 0.00 & 0.29 & 0.18 & 0.60 & 0.00 & 0.74 & 0.00 & 0.64 & 0.53 & -0.02 & 0.90 & 0.98 & 2.66 \\ 
19. + fusion QKV & 0.99 & 0.99 & 0.00 & 0.35 &0.45 & 0.82 & 0.00 & 0.80 & 0.01 & 0.74 & 0.46 & -0.16 & 0.88 & 1.13 & 3.09\\ \bottomrule
\end{tabular}
}
\caption{Results on language modelling. Regarding mathematical expressions, we adopt exact match (left) and bleu (right) as evaluation metrics for each test set. The best results are highlighted in \underline{\textbf{\textcolor{black}{bold}}}.} \label{tab:enoding_recon}
\end{table*}
\paragraph{Baselines.} We assess performances on language modelling using a different set of baselines\footnote{We choose the standard transformer-based VAE (Optimus) with single latent space (i.e., with the prior being a standard Gaussian distribution) for a fair comparison. Some variants, such as Della \cite{hu-etal-2022-fuse}, DPrior \cite{fang-etal-2022-controlled}, \cite{li-etal-2022-variational-autoencoder}, etc., were not selected.}. Specifically, we evaluate the performance of vanilla Optimus \cite{li2020optimus} and four LSTM-based autoencoders (AEs), including $\beta$-VAE \cite{Higgins2016betaVAELB}, adversarial AE (\citet{makhzani2016adversarial}, AAE), label adversarial AE (\citet{rubenstein2018latent}, LAAE), and denoising adversarial autoencoder (\citet{shen2020educating}, DAAE). All baselines have a latent size of 768. For semantic-syntax separated VAE setups, we evenly split the latent space for both. Moreover, we compare the proposed injection mechanism via low-rank operations with a standard memory injection \cite{li2020optimus}.
\paragraph{Metrics.} As for mathematical latex expressions, we use Exact Match and Bleu to evaluate the robustness of models on five different test sets, where four of them include out-of-distribution examples, (1) EVAL: mathematical expressions following the training set's distribution (like $U + cos{(n)}$), (2) VAR: mathematical expressions composed of a different set of variables (like $U + cos{(beta)}$), (3) EASY: mathematical expressions with a lower number of variables (like $cos{(n)}$), (4) EQ: mathematical derivations with equality insertions (like $E=U+cos{(n)}$), (5) LEN:  mathematical derivations with a higher number of variables (like $U + cos{(n)}) + A + B$). For explanatory sentences, we use five metrics, including BLEU \citep{Papineni02bleu:a}, BLEURT \citep{https://doi.org/10.48550/arxiv.2004.04696}, cosine similarity from pre-trained sentence T5 \citep{https://doi.org/10.48550/arxiv.2108.08877}, cross-entropy (Loss), and perplexity (PPL).
\paragraph{Results.} Firstly, we evaluate the performance of baselines with different syntactic injection setups. In the middle of Table \ref{tab:enoding_recon}, most configurations lead to lower performance, especially when using graph encoders, compared to vanilla Optimus, indicating that a standard memory injection mechanism for leveraging heterogeneous latent space is not effective. Conversely, by comparing line 05 to lines 12,14,16,18, \uline{injecting only syntactic information in Q can improve reconstruction performances on explanatory sentences} \textbf{(Finding~3)}. Moreover, we evaluate whether injecting heterogeneous latent representations into different attention components (Q,K,V) can further improve the results. In the bottom of Table \ref{tab:enoding_recon}, \uline{injecting semantic and syntax spaces into different attention components can additionally improve model performance} \textbf{(Finding~4)} (lines 9-11 vs 12,14,16,18), demonstrating that semantic and syntax spaces possess complementary features. Finally, we evaluate which injection strategies can achieve the best results. We found that \uline{\textit{addition} injection with BERT-TransCONV (line 17) can achieve the best overall results on both corpora} \textbf{(Finding~5)}. Next, we further analyse why syntax injection can improve model performance in natural language sentences. 

\begin{figure}[ht!]
    \centering
    \includegraphics[width=\columnwidth]{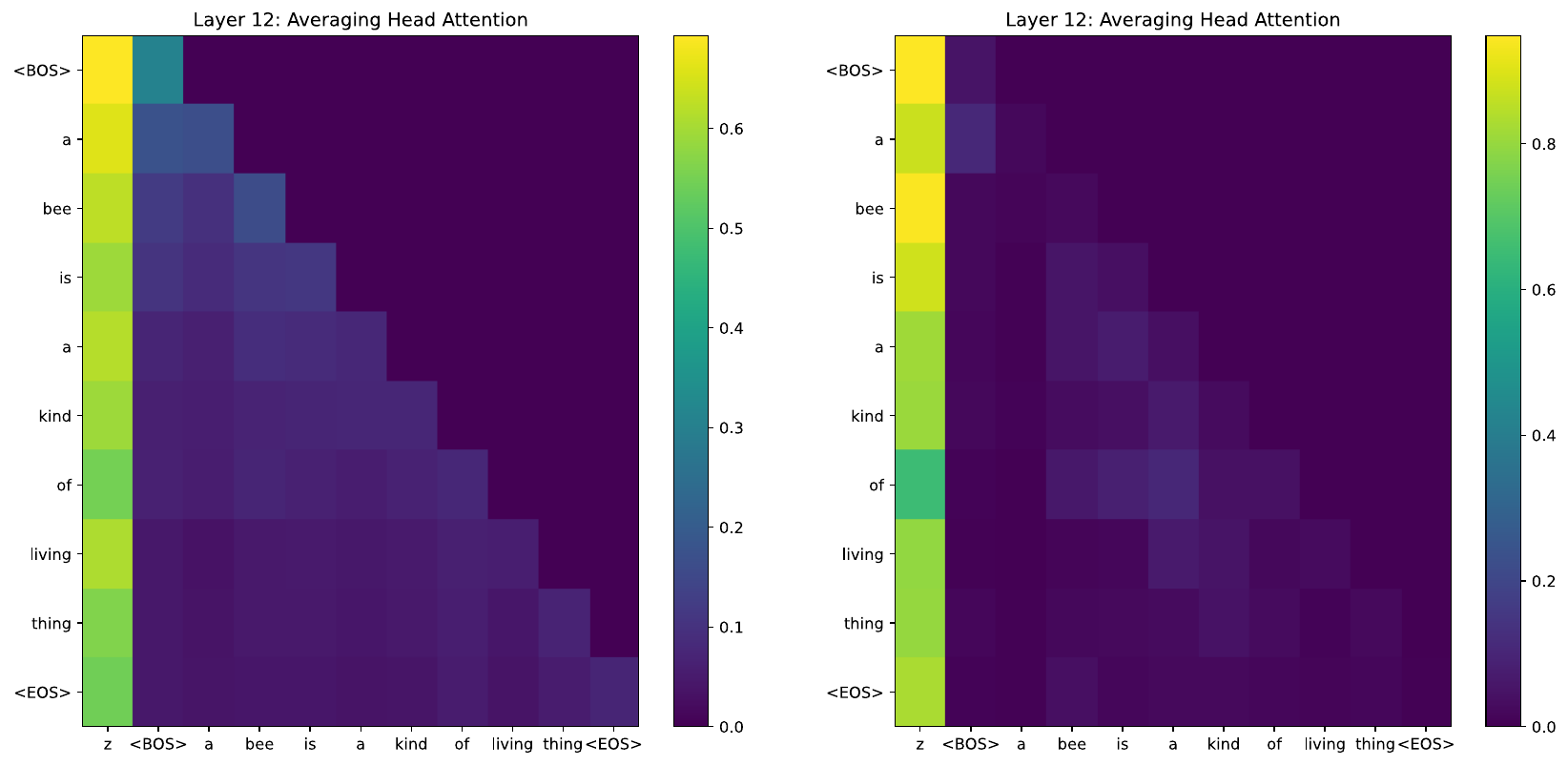}
    \caption{Visualising attention weighs (left: vanilla Optimus, right: BERT-TransCONV with \textit{addition Q} setup) where \textit{bee}: 0.58 $<$ 0.94, \textit{living thing}: (0.27, 0.15) $<$ (0.80, 0.80).}
    \label{fig:attn}
\end{figure}

\paragraph{Analysis.} We conjecture that \uline{the syntax and semantics separation allows the BERT encoder to capture and represent more fine-grained lexical information, alleviating the information loss in the sentence bottleneck} \textbf{(Finding~6)}. 
Given an input: \textit{a bee is a kind of living thing}, we found the reconstruction of vanilla Optimus to be \textit{a frog is a kind of amphibian}. This shows that Optimus is distracted by syntactic features, (\textit{x is a kind of y}) that are highly frequent in the training set and struggles in the reconstruction of specific lexical content (i.e., \textit{frog} and \textit{amphibian}). In contrast, the proposed architecture allows the semantic space to specialise in lexical content since the graph encoder already captures the syntax. To additionally support our claim, we visualise the attention weights of GPT2. In Figure \ref{fig:attn}, the first column of each heatmap represents the lexical information carried by the latent representation. \uline{We can observe that the proposed architecture with BERT-TransCONV + \textit{addition Q} setup (right) pays more attention to specific lexical elements (i.e., \textit{bee}) compared to vanilla Optimus (left). This also explains how the integration of a graph-based encoder can indirectly improve organisation for the semantic space (MSE in Table \ref{tab:enoding_syntax})}. 



\subsection{Downstream Evaluation}
\paragraph{Guided Generation.} One advantage of the VAE architecture is that it allows controlling sentence generation by manipulating latent representations via traversal and interpolation. In this experiment, we quantitatively assess the controllability of the decoding via latent traversal. Specifically, given an initial point in Gaussian space, we perform an \textit{Ornstein-Uhlenbeck} random walk \cite{pinsky2010introduction} \footnote{$\tilde{z}_{t+1}=-\gamma \tilde{z}_{t}+\sigma W_t$ where $t$ is the index, $W_t \in N(0,1)$, $\gamma$ and $\sigma$ are scalar hyper-parameters.} for semantic space and fix syntax space. In detail:

\textbf{1.} We set the traversal radius (r) - a predefined hyper-parameter, and sample an initial point/vector (sampled from Gaussian space).
\textbf{2.} We traverse the semantic latent space using \textit{Ornstein-Uhlenbeck} random walk and calculating the Euclidean distance between the traversed vectors and the initial point.
\textbf{3.} We keep only the samples whose distance is $> r_{t-1}$ and $< r_t$ when $t=1, r_{t-1} = 0$.
\textbf{4.} We generate the sentences from the latent spaces using the model and then compute the syntax tree edit distance (i.e., the distance between the syntactic trees) of the samples retrieved in step 3 and calculate the average distance.
\textbf{5.} Repeat 2 - 5.

\begin{figure}[ht!]
    \centering
    \includegraphics[width=0.7\columnwidth]{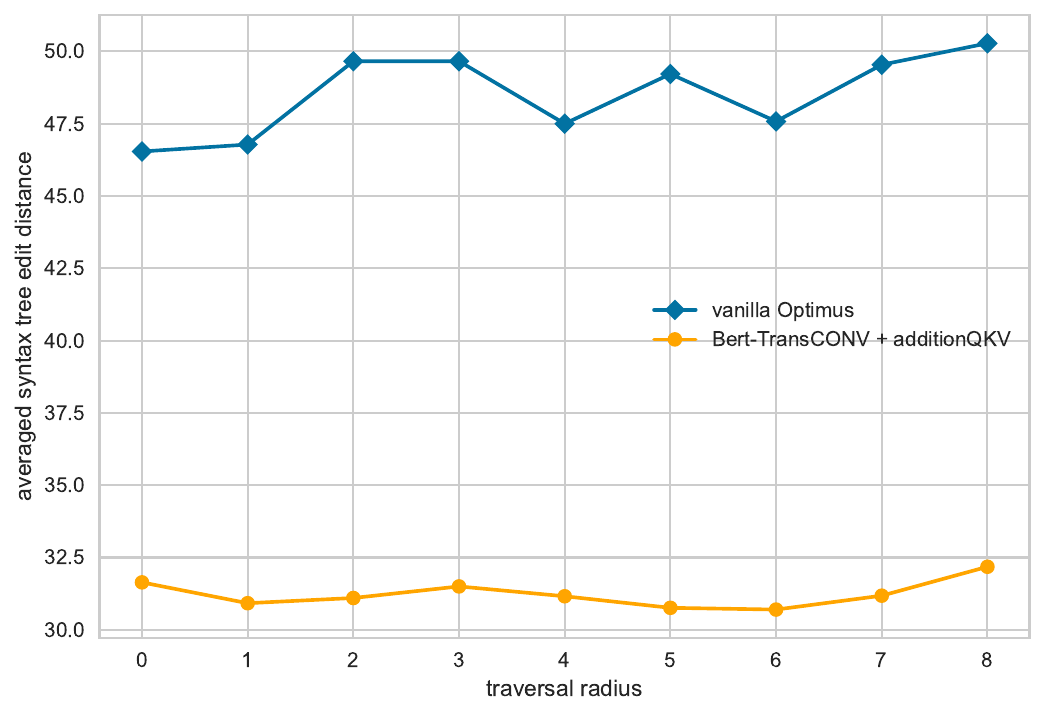}
    \caption{Traversing semantic space with increasing traversal radius while keeping syntax space fixed. We can observe improved syntax control in decoding by separating syntax and semantics.}
    \label{fig:trav}
\end{figure}

If the model can learn semantic-syntactic separation, the generated sentence can be syntactically controlled. To experiment, we quantitatively evaluate the similarity of syntactic structures between initial and traversed sentences via syntax tree edit distance. We gradually increase the radius of the random walk to perform a comparison between vanilla Optimus and BERT-TransCONV(addition QKV). In Figure \ref{fig:trav}, we can conclude that \uline{the proposed architecture can better hold the syntax structure, indicating better separation} \textbf{(Finding~7)}. 


\paragraph{Mathematical Derivations.} Finally, we explore the representation quality for mathematical expressions on downstream equational inference tasks \cite{meadows2023symbolic}. Specifically, we focus on expression derivation, where, given a premise $x$ and a mathematical operation $t$ (i.e., differentiation, integration) the goal is to predict whether a target mathematical expression $y$ can be derived from $x$ via $t$. Here, we adopt the dataset from \citep{valentino2023multioperational} and examine whether a linear probing classifier \citep{ferreira2021does} trained on latent representations encoded from frozen pre-trained models can predict the correct operation $t$ in a multi-label classification problem (i.e., given premise $x$ and target result $y$) and whether the classifier can predict a valid conclusion $y$  (i.e. Conclusion Classification) given a premise $x$ in a binary classification setting (using random negative examples). Experiments reveal that \uline{separately injecting latent semantic and syntactic representations can provide complementary information and improve performance on both tasks} \textbf{(Finding~8)}.

\begin{table*}[ht!]
\centering
\setlength\tabcolsep{2.5pt}
\small
\centering
\begin{tabular}{lcccc}
\toprule
Inference Type & \multicolumn{2}{c}{Operation Class.} & \multicolumn{2}{c}{Conclusion Class.} \\ 
Metrics & Acc & F1  & Acc & F1\\ \midrule
Optimus(768) & 0.89 & 0.89 & 0.68 & 0.68 \\
\midrule
LSTM & 0.89 & 0.89 & 0.59 & 0.62 \\
VGAE & 0.79 & 0.80 & 0.56 & 0.62 \\
Siam & \underline{\textbf{\textcolor{black}{0.92}}} & \underline{\textbf{\textcolor{black}{0.92}}} & 0.59 & 0.59 \\
GraphEncoder &  &  &  &  \\ 
+ GCN & 0.73 & 0.74 & 0.57 & 0.55  \\
+ GraphSAGE & 0.87 & 0.87 & 0.64 & 0.63 \\
+ TransCONV & 0.88 & 0.89 & 0.63 & 0.62 \\ \midrule
Bert-GraphSAGE & \\
+ addition QKV & 0.88 & 0.88 & 0.69 & 0.69 \\
+ fusion QKV & 0.90 & 0.90 & \underline{\textbf{\textcolor{black}{0.71}}} & \underline{\textbf{\textcolor{black}{0.71}}}  \\
Bert-TransCONV & \\
+ addition QKV & \underline{\textbf{\textcolor{black}{0.92}}} & \underline{\textbf{\textcolor{black}{0.92}}} & 0.68 & 0.68 \\ 
+ fusion QKV & 0.91 & 0.91 & 0.59 & 0.59 \\ \bottomrule
\end{tabular}
\caption{Results for the mathematical derivations probing task reveal that separately injecting latent semantic and syntactic representations can provide complementary information, resulting in enhanced performance.} \label{tab:classification}
\end{table*}

\section{Related Work} \label{sec:related}
\paragraph{Language VAE.} Most previous language VAE works are based on LSTM instantiated on different generation tasks, including dialogue generation \citep{zhao-etal-2017-learning}, text style transfer \citep{john-etal-2019-disentangled, shen2020educating}, text paraphrasing \citep{bao-etal-2019-generating}, etc. The development of Optimus \citep{li2020optimus} led to more research focusing on how to control the generation of Transformer-based architectures by latent space geometry \citep{zhang2023learning} or pre-defined priors \citep{fang-etal-2022-controlled,li-etal-2022-variational-autoencoder,hu2021causal}. Comparatively, we focused on the semantic-syntax separation with the help of a graph-based encoder. To our knowledge, the combination of graph encoders and VAEs for text generation is underexplored.

\paragraph{Learning Syntactic Representations.} From the perspective of model architecture, three kinds of encoders can learn syntactic representations, including graph-based encoders \citep{wu2023graph}, sequential encoders (i.e., LSTM \citep{10.1162/neco.1997.9.8.1735} and Transformers \citep{vaswani2017attention}), and tree-based encoders \citep{harer2019tree} (i.e., using Recursive Neural Networks \citep{harer2019tree, mrini-etal-2021-recursive}), with the latter two commonly used in the natural language generation domain \citep{Raffel2020t5}. Nevertheless, whether these models truly capture structural information or just the lexical combination of tokens is not fully clarified \citep{shi2016does}. This work uses graph-based encoders \citep{kipf2016semi} to better capture topological relations in syntactic trees. Graph Neural Networks \citep{zhou2020graph} have been effective for encoding syntactic and relational structures in various NLP tasks \citep{wu2023graph,sachan-etal-2021-syntax,veyseh2020graph}.
\section{Conclusion}
This work focused on the semantic-syntax separation through Optimus. We first implement several encoding baselines and reveal that language-graph encoding setups can better capture syntax information and maintain semantic-syntax separation. However, the language-graph encoding setup leads to low reconstruction performance. To solve this problem, we explored the integration of heterogeneous latent spaces via injection mechanisms. Experimental results showed that our setup can greatly improve language modelling performance, and revealed that the semantic-syntax separation can assist the language modelling task since the language encoder pays more attention to fine-grained lexical semantics, avoiding the distraction of syntax information captured by the separated syntax encoder, which can alleviate the information bottleneck of the language encoder. Those findings answer the \questionC{}


\paragraph{Reproducibility.} The data, models, and the related codebase are available online: \url{https://github.com/SnowYJ/sem_syn_separation}

\section{Scoping and Limitations}
Although the semantic-syntax separated latent space can provide better latent space geometry, it is still challenging to efficiently control the decoding stage through latent geometry itself, due to the discrete nature of the latent sentence space. Therefore, future direction can investigate better localisation in the discrete latent space.

\chapter{Semantic Discretisation} \label{cha:discrete}
In addition to the VAE architecture with its continuous latent sentence space, natural language representation often exhibits an inherently discrete structure, reflected in symbolic elements such as words. To better align with this discretised nature of language, this chapter investigates \questionD{}
\section{Introduction}
The emergence of deep generative neural networks supported by Variational AutoEncoders (VAEs) \citep{kingma2013auto} enables the localisation of syntactic and semantic properties within complex sentence latent spaces. By localising and manipulating these generative factors within the latent spaces, one can better control the properties of the textual output, enhancing performance on downstream tasks \citep{carvalho2023learning, john-etal-2019-disentangled}, and providing mechanisms for representing and disentangling syntactic and semantic features within natural language \citep{zhang2023learning, https://doi.org/10.48550/arxiv.2109.07169}.
\begin{figure}[t]
\centering
    \includegraphics[width=\columnwidth]{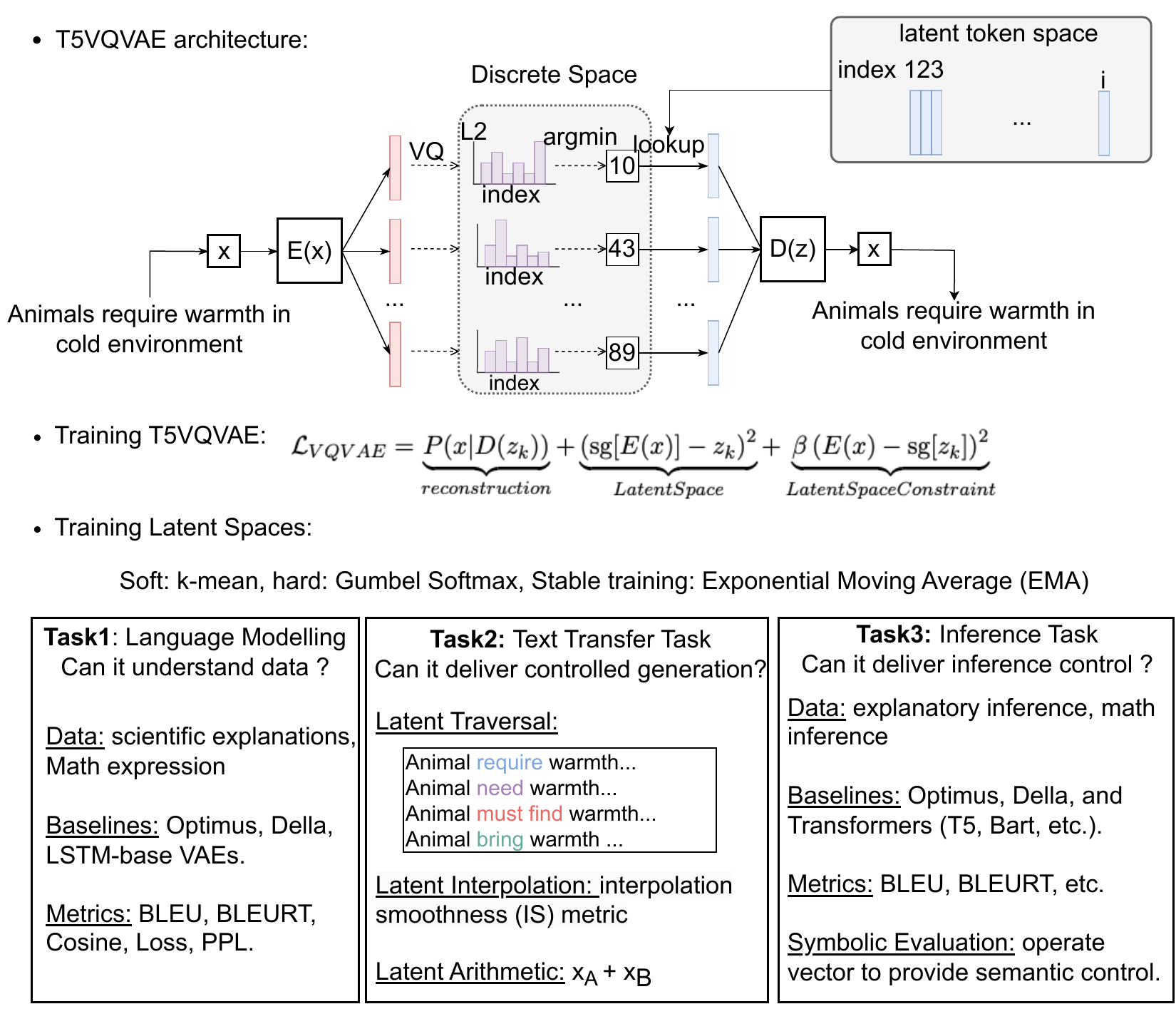}
    \caption{By controlling the token-level discrete latent space in VAEs, we aim to explicitly guide the cross-attention mechanism in T5 to improve the generation process. We focus on three challenging tasks to assess precise semantic control and inference.}
    \label{fig:overview}
\end{figure}

Recent work \citep{carvalho2023learning, zhang2023learning} investigated controllable text generation via latent sentence geometry based on the canonical Optimus architecture \citep{li2020optimus} (the first large pre-trained language VAE). However, the Optimus architecture brings its associated challenges since (i) the Optimus setup does not allow for a fine-grained (i.e., token-level) semantic control as sentence-level representation features are ignored by most attention heads especially in lower layers, where lexical-level semantics is captured \citep{hu-etal-2022-fuse}; (ii) the sentence bottleneck in the VAE architecture leads to inevitable information loss during inference \citep{zhang2023llamavae,zhang2023graph}.


This work concentrates on addressing these architectural limitations by aiming to minimise the information loss in the latent space and effectively control the decoder and its attention mechanism. The Vector Quantized Variational AutoEncoder (VQVAE) \citep{van2017neural}, as a discrete latent variable model, can be considered an ideal mechanism to alleviate these issues since it preserves and closely integrates both a coarse-grained continuous latent sentence space and a fine-grained latent token space that can preventinformation loss. More importantly, its latent token space can directly work on the cross-attention module \citep{vaswani2017attention} to guide the generation in seq2seq models, such as T5 \citep{raffel2020exploring}. Therefore, we hypothesise that such a mechanism can enable better generalisation and semantic control in Transformer-based VAEs.

Following these insights, we propose a novel approach named T5VQVAE, a model that leverages the controllability of VQVAE to guide the token-level self-attention mechanism during the generation process. We evaluate T5VQVAE on three challenging and diverse downstream tasks including (1) language modelling, (2) text transfer (guided text generation via the movement of latent vectors), and (3) natural language and symbolic inference tasks. An illustration of the complete model architecture and experimental setup can be found in Figure~\ref{fig:overview}. The overall contribution of the paper can be summarised as follows: 

\begin{enumerate}
    \item  We propose T5VQVAE, the first pre-trained language Vector-Quantised variational Autoencoder, bridging the gap between VAEs and token-level representations, improving sentence-level localisation, controllability, and generalisation under VAE architectures. The experiments reveal that the proposed model outperforms previous state-of-the-art VAE models, including Optimus \citep{li2020optimus}, on three target tasks, as well as delivering improved semantic control when compared to the previous state-of-the-art.
    \item We propose the Interpolation Smoothness (IS) metric for quantitatively evaluating sentence interpolation performance, a fundamental proxy for measuring the localisation of syntactic and semantic properties within sentence latent spaces. The experimental results indicate that T5VQVAE can lead to better interpolation paths (suggesting better interpretability and control). 
    \item Experiments on syllogistic-deductive NLI and mathematical expression derivation reveal that a quasi-symbolic behaviour may emerge in the latent space of T5VQVAE, and that the model can be explicitly controlled to achieve superior reasoning capabilities. 
\end{enumerate}

\section{Methodology} \label{sec:latent_props}
\subsection{T5VQVAE Architecture}
In this section, we first present our model, T5VQVAE, whose primary goal is to learn a latent space by reconstructing input sentences. Next, we illustrate its objective function, which consists of three parts designed to improve semantic control: reconstruction term, latent space optimization term, and encoder constraint term. Finally, we highlight the architectural advantages of T5VQVAE compared to Transformer-based VAEs.

\paragraph{Model architecture.} \citet{van2017neural} first proposed the VQVAE architecture for learning a discretised latent space of images, showing that it can alleviate the issue of \textit{posterior collapse}, in which the latent representations produced by the Encoder are ignored by the Decoder \citep{kingma2013auto}. In this work, we propose to integrate T5 encoder/decoder into the VQVAE architecture for representation learning with natural language. T5 was selected due to its consistent performance across a large range of NLP tasks and its accessibility. To cast T5 into a VQVAE model, we first establish a latent token embedding space, denoted as the codebook, represented by $z \in \mathbb{R}^{K \times I}$. Here, $K$ refers to the number of tokens in the codebook, and $I$ represents the dimensionality of each token embedding. When given a token $x$, the Encoder $E$ maps it into a vector representation, denoted as $E(x)$. Then, the nearest latent representation $z_k$ from the codebook $z$ is selected based on the $L2$ distance. The input of the cross-attention module can then be formalised as follows:
$\hat{x} = \text{MultiHead}\left(D(x)W^q, z_kW^k, z_kW^v\right)$
Here, $z_k$ is the key and value and $D(x)$, which represents the input token embedding of the decoder, is the query. $\hat{x}$ represents the reconstructed token, while $W^q$, $W^k$, and $W^v$ are trainable weights of \textbf{q}uery, \textbf{k}ey, and \textbf{v}alue.

\paragraph{Training T5VQVAE} The training of T5VQVAE can be then considered as the optimisation of three independent parts, including $D(z_k)$, $z_k$, and $E(x)$. Starting from $D$, the model can be trained by maximising the reconstruction probability $P(x|D(z_k))$ via the teach-forcing scheme. Next, the $z_k$ is optimised by minimising the $L2$ distance between $E(x)$ and $z_k$, which can be described as $(\text{sg}[E(x)] - z_k)^2$ where $\text{sg}$ is the stop gradient operation. Finally, $E(x)$ can be trained via the $L2$ distance. By ensuring that $E(x)$ can learn the latent embedding under the constraint of $R^{K \times I}$ rather than learning an embedding directly, we can guide the model to achieve better performance. A commitment weight $\beta < 1$ is used to constraint the $E$ close to $z_k$, which can be described as: $\beta (E(x) - \text{sg}[z_k])^2$. $\beta$ is set to 0.25 following the same setup as \citep{van2017neural} to preserve a behaviour consistent with their findings. The final objective function of T5VQVAE can be formalised as follows:
\[
\begin{aligned}
\mathcal{L}_{VQVAE} = \underbrace{P(x|D(z_k))}_{(1) reconstruction} + \underbrace{\left ( \text{sg}[E(x)] - z_k \right )^2}_{(2) Latent Space}
+ \underbrace{ \beta \left ( E(x) - \text{sg}[z_k] \right )^2}_{(3) Latent Space Constraint}
\end{aligned}
\]

\paragraph{Training the latent space.} There are two possible strategies to update the latent space: \textit{i.} k-means and \textit{ii.} Gumbel softmax. Regarding k-means, for each token embedding $w_i$ in a sentence, it selects the nearest latent token embedding, $z_k$, to its token embedding $e^{w_i}$. This process is equivalent to classifying $e^{w_i}$ using k-means and then choosing the corresponding central point $z_k$ as the input for $D(z_k)$. This can be expressed as follows:
\[
\begin{aligned}
z_{w_i} = z_k, \enspace \text{where} \enspace k = \operatorname{argmin}_j\left\| e^{w_i} - z^j \right\|_2
\end{aligned}
\]
To improve the stability of latent space training (term 2), we adapted the Exponential Moving Average (EMA) training scheme to update $z$ \cite{roy2018theory}. 
Figure \ref{fig:loss_curve} displays the training and testing loss curves of T5VQVAE with EMA or not.
Instead of using k-means, which performs a soft selection of the index $k$, we can utilize the Gumbel softmax trick \citep{https://doi.org/10.48550/arxiv.1611.01144} for a hard sampling of the index $k$. This trick involves sampling a noise value $g_k$ from the Gumbel distribution and then using the softmax function to normalize the output, resulting in a probability distribution. By selecting the index with the highest probability, we obtain a discrete choice. This entire process can be described as follows:
\[
\begin{aligned}
z_{w_i} &= z_k, \text{where} ~~
k = \operatorname{argmax}_k \frac{\exp (\log(t_k) + g_k)/ \tau}{\sum_{k=1}^K \exp (\log(t_k) + g_k)/ \tau} \\
\end{aligned}
\]
\noindent In this context, $t_k$ represents the probability of the $k$-th token, which can be obtained through a linear transformation before being fed into the Gumbel softmax. The parameter $\tau$ serves as a temperature hyper-parameter that controls the closeness of the new distribution to a discrete distribution. As $\tau$ approaches zero, the distribution becomes one-hot, while a non-zero value of $\tau$ leads to a more uniform distribution. In our experiments, we experienced convergence issues when using the Gumbel softmax scheme, and therefore decided to adopt the k-means mechanism which generally leads to better results.

\begin{figure}[ht!]
\centering
    \includegraphics[scale=0.65]{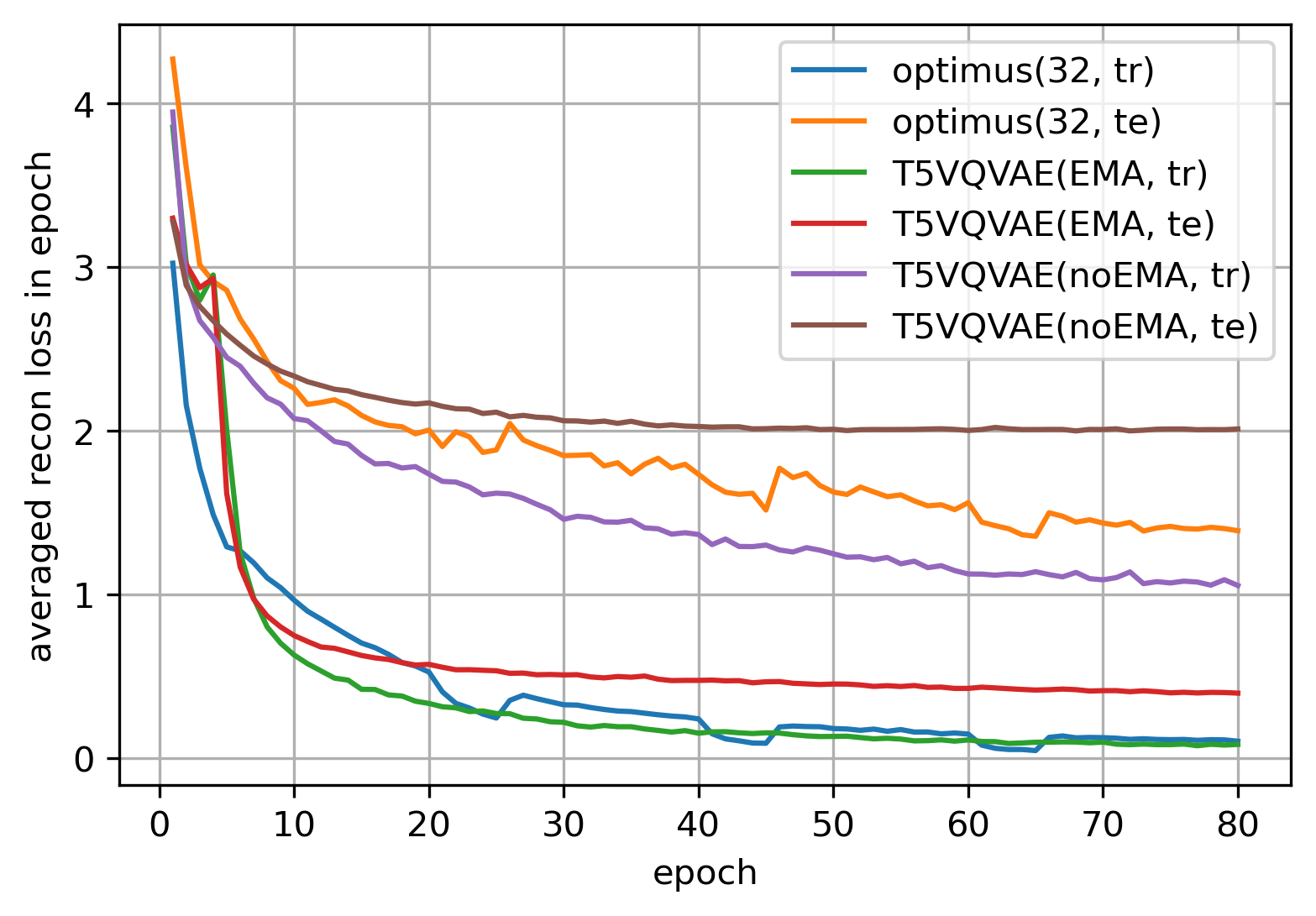}
    \caption{Loss curves of T5VQVAEs (base) with and without EMA and Optimus on the WorldTree corpus.}
    \label{fig:loss_curve}
\end{figure}

\paragraph{Advantages of T5VQVAE.} Compared with state-of-the-art Transformer VAEs such as Optimus \citep{li2020optimus}, our model has the following architectural advantages: (i) efficient and stable latent space compression. During the training of Optimus, in fact, the KL term in ELBO is regularized cyclically \citep{fu-etal-2019-cyclical} to avoid KL vanishing and posterior collapse, which leads to an unstable training process (figure \ref{fig:loss_curve}). In contrast, T5VQVAE avoid the KL regularization term since it becomes a constant value:
\[
\begin{aligned}
\text{KL}\left(q(z_k|x)||p(z_k)\right) &= \sum_{k} q(z_k|x) \log \frac{q(z_k|x)} {p(z)} = 1 \times \log \frac{1}{1/K} = \log K
\end{aligned}
\]
where 
the prior $p(z) = 1/K$ is a uniform distribution. (ii) Better controllability. \citet{hu-etal-2022-fuse} revealed that in Optimus \citep{li2020optimus}, the latent representation is concatenated into key and value which is more likely to be ignored by most attention heads especially in lower layers where lexical-level semantics is captured. In contrast, the latent representations of T5VQVAE are designed to act on the attention heads directly.


\subsection{Controllability Evaluation} \label{sec:eval}
Next, we put forward two metrics for quantitatively evaluating the controllability of the proposed model (T5VQVAE), which we refer to as \textit{semantic disentanglement} and \textit{interpolation smoothness}. The former evaluates the controllability from the perspective of the disentanglement of semantic factors (e.g., arguments and associated semantic roles). The latter evaluates the smoothness and coherence of the latent space geometry during interpolation.

\paragraph{Semantic Disentanglement.} \label{sec:sem_dis} Recent studies have attempted to adapt metrics from the image domain to evaluate the semantic disentanglement of sentences \citep{carvalho2023learning}. Semantic information in a sentence is more likely to be entangled, especially in the context of stacked multi-head self-attention models. As mentioned in \citep{carvalho2023learning}, conceptually dense sentences are clustered according to role-content combination over the VAE latent space. Each semantic role is jointly determined by multiple dimensions rather than one single dimension. Therefore, calculating the importance of one dimension to that semantic role as a disentanglement metric is unreliable. In this work, we quantitatively evaluate the disentanglement of the semantic roles by: (1) calculating the averaged Euclidean distance between different content under that role, such as the distance between \textit{PRED-is} and \textit{PRED-are}, and (2) counting the number of different indices of the same role-content after the vector quantisation. The smaller the distance or the less the number of indices, the more concentrated the distribution of this semantic role in the latent space, indicating better disentanglement.

\paragraph{Interpolation Smoothness.} \label{sec:interpolate_smooth} Interpolation is a standard process for evaluating the geometric properties of a latent space in both image and language domains \citep{li2020optimus, https://doi.org/10.48550/arxiv.2106.09016}. It aims to generate a sequence of sentences following a spatial trajectory from source to target via latent arithmetics. For example, in the VAE latent space, the interpolation path can be described as $ z_t = z_1 \cdot (1 - t) + z_2 \cdot t  $  with $ t $ increased from $ 0 $ to $ 1 $ by a step size of $ 0.1 $ where $ z_1 $ and $ z_2 $ represent latent vectors of source and target sentences, respectively. In this case, each intermediate output $D(z_t)$ should change fewer semantic concepts at each step if the latent space is smooth and regular. In this work, we employ a similar strategy, however follow the more granular token level within the VQVAE. We directly manipulate the interpolation within the latent token space. At each step $t$, we obtain the intermediate latent token embedding $z^{w_i}_t$ within a sentence by calculating the weighted minimal distance between its preceding token embedding $z^{w_i}_{t-0.1}$ and the target token embeddings $z^{w_i}_2$. This process can be described as follows:
\[
\begin{aligned}
z_1^{w_i} &= e^{k_1}, z_2^{w_i} = e^{k_2}, \text{where} \enspace i = [1, ..., L] \\ \nonumber
z^{w_i}_t &= z^k, \text{where} \\
k &= \operatorname{argmin}_j \enspace (1-t) \times \left\| z_{t-0.1}^{w_i} - z^j \right\|_2 \\
&+ t \times \left\| z_2^{w_i} - z^j \right\|_2 \\
s_t &= \left[z^{w_1}_t; \dots; z^{w_L}_t\right]
\end{aligned}
\]
where $s_t$ represents the sentence embeddings at step $t$. The final generated sentence can be decoded as $s_t = D(s_t)$. Once we have obtained the interpolation path, we introduce the interpolation smoothness (IS) metric to quantitatively evaluate its smoothness. This metric involves calculating the aligned semantic distance between the source and the target (referred to as the ideal semantic distance). Subsequently, we calculate the sum of the aligned semantic distances between each pair of adjacent sentences in the path (referred to as the actual semantic distance). Finally, by dividing the ideal semantic distance by the actual semantic distance, we obtain a measure of smoothness. If the result is 1, it indicates that the actual path aligns perfectly with the ideal path, suggesting better geometric properties. Conversely, it suggests a less coherent transformation path, indicating poorer geometric properties. The metric is defined as follows:
\[
\begin{aligned}
\text{IS} = \mathbb{E}_{(s_0, ..., s_T) \sim P} \frac{\delta(\text{align}(s_0, s_T))}{\sum^T_{t=0} \delta(\text{align}(s_t, s_{t+0.1}))}
\end{aligned}
\]
\noindent where $\delta$ and $\text{align}$ are sentence similarity and alignment functions, respectively. In this experiment, sentence similarity and alignment are performed via Word Mover’s Distance \citep{zhao-etal-2019-moverscore} since it can softly perform the semantic alignment.




\section{Empirical Evaluation}
\subsection{AutoEncoding Task}
\paragraph{{Pre-training Data.}} In this work, we focus on the use of conceptually dense explanatory sentences \citep{https://doi.org/10.48550/arxiv.2104.08661}  and mathematical latex expressions \citep{meadows2023symbolic} to evaluate model performance. The rationale behind this choice is that (1) explanatory sentences provide a semantically challenging yet sufficiently well-scoped scenario to evaluate the syntactic and semantic organisation of the space \citep{thayaparan2020survey,valentino2022hybrid,valentino-etal-2022-case}; (2) mathematical expressions follow a well-defined syntactic structure and set of symbolic rules that are notoriously difficult for neural models \citep{meadows2023generating}. Moreover, the set of rules applicable to a mathematical expression fully determines its semantics, allowing for an in-depth inspection and analysis of the precision and level of generalisation achieved by the models \citep{welleck2022symbolic}. Firstly, we conduct a pre-training phase, evaluating the performance of T5VQVAE in reconstructing scientific explanatory sentences from WorldTree \citep{jansen2018worldtree} and mathematical latex expressions from the dataset proposed by \citet{meadows2023symbolic}. 


\paragraph{Baselines.} We consider both \textit{small} and \textit{base} versions of pretrained T5 to initialise the T5VQVAE, where the codebook size is 10000. 
As for the large VAE model, we consider Optimus with random initial weights and pre-trained weights \citep{li2020optimus} and Della \citep{hu-etal-2022-fuse}. We chose two different latent dimension sizes (32 and 768) for both of them. Moreover, we also select several LSTM language autoencoders (AE), including denoising AE (\citet{10.1145/1390156.1390294}, DAE), $\beta$-VAE \citep{Higgins2016betaVAELB}, adversarial AE (\citet{makhzani2016adversarial}, AAE), label adversarial AE (\citet{rubenstein2018latent}, LAAE), and denoising adversarial autoencoder (\citet{shen2020educating}, DAAE). 
\begin{table}[ht!]
\scriptsize
\small
\centering
\renewcommand\arraystretch{1}
\begin{tabular}{llllll} 
\toprule
\multicolumn{6}{c}{\textit{Explanatory sentences}} \\
Evaluation Metrics & BLEU & BLEURT & Cosine & Loss $\downarrow$ & PPL $\downarrow$ \\ \hline
DAE(768) & \textbf{0.74} & \textbf{0.03} & \textbf{0.91} & \textbf{1.63} & \textbf{5.10} \\ %
AAE(768) & 0.35 & -0.95 & 0.80 & 3.35 & 28.50 \\ 
LAAE(768) & 0.26& -1.07& 0.78& 3.71 & 40.85 \\ 
DAAE(768) & 0.22& -1.26& 0.76& 4.00 & 54.59 \\ 
$\beta$-VAE(768) & 0.06& -1.14& 0.77& 3.69 & 40.04 \\ \hdashline 
Optimus(32, rand) & 0.54 & 0.14 & 0.92 & 1.08 & 2.94 \\
Optimus(32, pre) & 0.61 & 0.29 & 0.93 & 0.86 & 2.36 \\
Optimus(768, rand) & 0.49 & -0.04 & 0.90 & 1.32 & 3.74 \\
Optimus(768, pre) & 0.68 & 0.48 & 0.95 & 0.65 & 1.91 \\
DELLA(32, rand) & 0.71 & 0.06 & 0.92 & 0.50 & 1.65 \\ 
DELLA(768, rand) & 0.72 & 0.21 & 0.95 & \textbf{\textcolor{black}{0.41}} & \textcolor{black}{\textbf{1.51}} \\
T5VQVAE(small, soft) & 0.81 & \textcolor{black}{\textbf{0.62}} & \textcolor{black}{\textbf{0.97}} & 0.46 & 1.58 \\
T5VQVAE(base, soft) & \textcolor{black}{\textbf{0.82}} & \textcolor{black}{\textbf{0.62}} & \textcolor{black}{\textbf{0.97}} & 0.75 & 2.11 \\ \hline \hline
\multicolumn{6}{c}{\textit{Mathematical expressions}} \\ 
Evaluation Datasets & EVAL & VAR & EASY & EQ & LEN \\ \hline
DAE(768) & \textbf{0.94} & \textbf{0.50} & \textbf{0.80} & \textbf{0.74} & \textbf{0.58} \\
AAE(768) & 0.41 & 0.41 & 0.39 & 0.41 & 0.52 \\
LAAE(768) & 0.41 & 0.45 & 0.39 & 0.39 & 0.49 \\ 
DAAE(768) & 0.38 & 0.48 & 0.35 & 0.38 & 0.49 \\ 
$\beta$-VAE(768) & 0.39 & 0.48 & 0.37 & 0.39 & 0.50 \\ \hdashline 
Optimus(32, rand) & 0.95 & 0.59 & 0.75 & 0.71 & 0.50 \\
Optimus(768, rand) & 0.96 & 0.61 & 0.79 & 0.75 & 0.54 \\
DELLA(32, rand) & \textcolor{black}{\textbf{1.00}} & 0.55 & 0.89 & 0.72 & 0.63 \\
DELLA(768, rand) & \textcolor{black}{\textbf{1.00}} & 0.55 & 0.93 & 0.79 & 0.64 \\
T5VQVAE(small, soft) & 0.97 & \textcolor{black}{\textbf{0.65}} & \textcolor{black}{\textbf{0.95}} & \textcolor{black}{\textbf{0.90}} & \textcolor{black}{\textbf{0.69}} \\
T5VQVAE(base, soft) & 0.98 & 0.62 & \textcolor{black}{\textbf{0.95}} & 0.85 & 0.68
\\ \toprule
\end{tabular}
\caption{AutoEncoding task evaluation on the test set (soft: k-means). The highest scores of large VAE models and LSTM-based VAE models are highlighted in blue and in bold separately.} \label{tab:semantic_similarity_autoencoding}
\end{table}
\paragraph{Quantitative Evaluation.} As for modelling explanatory sentences, we quantitatively evaluate the performance of the models using five metrics, including BLEU \citep{Papineni02bleu:a}, BLEURT \citep{https://doi.org/10.48550/arxiv.2004.04696}, cosine similarity from pre-trained sentence T5 \citep{https://doi.org/10.48550/arxiv.2108.08877}, cross-entropy (Loss), and perplexity (PPL). As for modelling mathematical expressions, we use BLEU to evaluate the robustness of models on the 5 test sets proposed by \citet{meadows2023symbolic}, one designed to assess in-distribution performance, and four designed to assess out-of-distribution generalisation. Here we provide a full characterisation of the test sets: (1) EVAL: contains mathematical statements following the same distribution of the training set (like $U + cos{(n)}$), including expressions with similar lengths and set of symbols (2) VAR: full mathematical statements with variable perturbations (like $U + cos{(beta)}$), designed to test the robustness of the models when dealing with expressions containing variables never seen during training; (3) EASY: simpler mathematical expressions with a lower number of variables, designed to test length generalisation (like $cos{(n)}$), (4) EQ: full mathematical statements with equality insertions (like $E = U + cos{(n)}$), designed to test the behaviour of the model on equivalent mathematical expressions with minimal perturbations (5) LEN: mathematical statements with a higher number of variables (like $U + cos{(n)}) + A + B$), designed to test generalisation on more complex expressions. As shown in Table \ref{tab:semantic_similarity_autoencoding}, the highest scores for large VAE models and LSTM-based VAE models are highlighted in blue and bold, respectively. Among them, \uline{T5VQVAEs with the k-means scheme outperforms Optimus and LSTM-based VAEs in both corpora and compared with Della, it can deliver better generation and generalisation} \textbf{(Finding~1)}. 

Next, we quantitatively evaluate the disentanglement of T5VQVAE following the semantic disentanglement reference metric \ref{sec:sem_dis}. As displayed in Table \ref{tab:stat_disentanglement}, the number of central points for \textit{PRED} is higher than the remaining role-content, being 24 in \textit{PRED-is} and 6 in \textit{PRED-are}. \uline{This indicates that the semantic information of \textit{PRED} is more widely distributed in the latent space when compared to other roles. This behaviour might be attributed to the fact that the aforementioned predicates are widely used across sentences in the corpus} \textbf{(Finding~2)}. 
\begin{table}[ht!]
\scriptsize
\setlength\tabcolsep{2.5pt}
\small
\centering
\renewcommand\arraystretch{1}
\begin{tabular}{lllll}
\toprule
Role-content & NUM centers & AVG dis & MAX dis & MIN dis \\ \hline
ARG0-animal & 3 & 0.28 & 0.52 & 0.35 \\ 
ARG1-animal & 3 & 0.28 & 0.52 & 0.35 \\
ARG2-animal & 4 & 0.33 & 0.55 & 0.35 \\
PRED-is & 24 & 0.60 & 1.08 & 0.22 \\
PRED-are & 6 & 0.31 & 0.64 & 0.21 \\
MOD-can & 5 & 0.40 & 0.82 & 0.28 \\ 
NEG-not & 2 & 0.25 & 0.51 & 0.51 \\ \toprule
\end{tabular}
\caption{Semantic role disentanglement.} \label{tab:stat_disentanglement}
\end{table}

\begin{table*}[t]
\centering
\begin{tcolorbox}[fontupper=\small, fontlower=\small] 
\begin{multicols}{2}
    \underline{\textbf{an animal requires warmth in cold environments}} \\ \\
    dim0: \textcolor{blue}{an} animal requires warmth in cold environments \\
    dim0: \textcolor{blue}{a} animal requires warmth in cold environments \\
    dim0: \textcolor{blue}{the} animal requires warmth in cold environments \\
    
    dim1: an \textcolor{blue}{organism} requires warmth in cold environments \\
    dim1: an \textcolor{blue}{animal} requires warmth in cold environments \\
    dim1: an \textcolor{blue}{object} requires warmth in cold environments \\
     
    dim2: an animal \textcolor{blue}{needs} warmth in cold environments \\
    dim2: an animal \textcolor{blue}{must find} warmth in cold environments \\ 
    dim2: an animal \textcolor{blue}{brings} warmth in cold environments \\
    dim2: an animal \textcolor{blue}{wants} warmth in cold environments
   \columnbreak
    \\ \\ \\
    dim4: an animal requires warmth \textcolor{blue}{during} cold temperatures \\
    dim4: an animal requires warmth \textcolor{blue}{in} cold environments \\
    dim4: an animal requires warmth \textcolor{blue}{to} cold environments \\
    
    dim5: an animal requires warmth in temperatures \\
    dim5: an animal requires warmth in \textcolor{blue}{warm} environments \\ 
    dim5: an animal requires warmth in \textcolor{blue}{a warm} environment \\
    
    dim6: an animal requires warmth in cold \textcolor{blue}{temperatures} \\
    dim6: an animal requires warmth in cold \textcolor{blue}{climates} \\
    dim6: an animal requires warmth in cold \textcolor{blue}{systems}
\end{multicols}
\end{tcolorbox}
\caption{T5VQVAE(base): traversals showing \textcolor{blue}{controlled} semantic concepts in explanations.}
\label{tab:trav_examples_main}
\end{table*}

\subsection{Text Transfer Task} 
Next, we investigate the controllability of T5VQVAE by manipulating the latent space via geometric transformations. This is referred to as the Text Transfer task. We compare the performance of T5VQVAE (base, soft) and Optimus (32, pretrain) - both trained in the AutoEncoding task - as baselines. We evaluate the latent space using latent traversal, interpolation, and vector arithmetic.

\paragraph{Latent Traversal.} The traversal is inspired by the image domain, only changing the feature interpretation \citep{higgins2016beta,kim2018disentangling}. Specifically, if the vector projection within the latent space can be modified when traversing (re-sampling) one dimension, the output should only change well-defined semantic features corresponding to that dimension. In this experiment, the traversal is set up from a starting sentence. As illustrated in Table \ref{tab:trav_examples_main}, \uline{the T5VQVAE can provide localised semantic control by operating the discrete latent space. Different dimensions in the discrete sentence space can control different parts of the sentence} \textbf{(Finding~3)}. 

\paragraph{Latent Interpolation.} As described in section \ref{sec:interpolate_smooth}, interpolation aims to generate a sequence of sentences from source to target via latent vector arithmetic. An ideal interpolation should lead to reasonable semantic controls at each step. In Table \ref{tab:interpolation}, we can observe that compared with Optimus's interpolation (bottom) where the semantics are changed redundantly, e.g., from \textit{some birds} to \textit{some species mammals} to \textit{most birds} and from \textit{have} to \textit{don't have} to \textit{have}, T5VQVAE (top) leads to a more reasonable (coherent/smoother) pathway. E.g., from \textit{speckled brown color} to \textit{speckled brown feathers} to \textit{speckled wings} to \textit{wings}. 
\begin{table}[t]
\centering
\begin{tcolorbox}[fontupper=\small, fontlower=\small] 
\underline{\textbf{Source: some birds have a speckled brown color}} \\ \\
1. \textcolor{blue}{some birds} \ul{have} \textcolor{orange}{a speckled brown color} \\
2. \textcolor{blue}{some birds} \ul{do not have} \textcolor{orange}{speckled brown feathers} \\
3. \textcolor{blue}{some species mammals} \ul{do not have} \textcolor{orange}{speckled wings} \\
4. \textcolor{blue}{most species mammals} \ul{do not have} \textcolor{orange}{wings} \\

1. \textcolor{blue}{some birds} \ul{have} \textcolor{orange}{scales} \\
2. \textcolor{blue}{some birds} \ul{have} \textcolor{orange}{a speckled brown color} \\
3. \textcolor{blue}{some species mammals} \ul{have} \textcolor{orange}{wings} \\
4. \textcolor{blue}{most birds} \ul{don't have} \textcolor{orange}{wings} \\
5. \textcolor{blue}{most insects} \ul{have} \textcolor{orange}{wings} \\
6. \textcolor{blue}{most species mammals} \ul{don't have} \textcolor{orange}{wings} \\

\underline{\textbf{Target: most species mammals do not have wings}}

\end{tcolorbox}
\caption{Interpolation for T5VQVAE (top) and Optimus (bottom) where \textcolor{blue}{blue}, \ul{underline}, and \textcolor{orange}{orange} represent subject, verb, and object, respectively. Only unique sentences are shown.}
\label{tab:interpolation}
\end{table}

More importantly, we quantitatively evaluate the interpolation behaviour via the IS metric. We randomly select 100 (source, target) pairs and interpolate the path between them. Then, we calculate the averaged, maximal, and minimal ISs. As shown in Table \ref{tab:interpolation_smoothness}, T5VQVAE outperforms Optimus by over 43\% in average, which indicates that \uline{T5QVAE induces a latent space which can better separate the syntactic and semantic factors when contrasted to Optimus}  \textbf{(Finding~4)}.
\begin{table}[t]
\setlength\tabcolsep{2.5pt}
\small
\centering
\renewcommand\arraystretch{1}
\begin{tabular}{llll}
\toprule
Evaluation Metrics & avg IS & max IS & min IS \\ \hline
Optimus(32, pretrain) & 0.22 & 0.53 & 0.13 \\ 
Optimus(768, pretrain) & 0.21 & 0.50 & 0.10 \\
T5VQVAE(base, soft) & \textcolor{black}{\textbf{0.65}} & \textcolor{black}{\textbf{1.00}} & \textcolor{black}{\textbf{0.18}} \\ \toprule
\end{tabular}
\caption{Interpolation smoothness.} \label{tab:interpolation_smoothness}
\end{table}

\paragraph{Latent Vector Arithmetics.} Inspired by word embedding arithmetics, e.g., $king - man + woman = queen$, we explore the compositional semantics via latent arithmetic with the target of sentence-level semantic control. After adding two latent vectors corresponding to two sentences $s_c = s_A + s_B$, we expect the resulting sentence to express the semantic information of both sentences. \uline{From Table \ref{tab:arithmetic_examples}, we can observe that T5VQVAE can generate outputs containing both inputs' semantic information compared with Optimus. E.g., the output contains \textit{are likely to} and \textit{their environment} from $s_A$ and \textit{to survive} and \textit{/} from $s_B$} \textbf{(Finding~5)}. 
\begin{table}[t]
\centering
\begin{tcolorbox}[fontupper=\small, fontlower=\small] 
    \ul{\textbf{$s_A$: animals are likely to have the same color as their environment}}\\
    \ul{\textbf{$s_B$: animals require respiration to survive / use energy}} \\
    
    T5VQVAE: \textcolor{blue}{animals} \textcolor{orange}{are likely to} \textcolor{cyan}{survive / to survive} \textcolor{orange}{in their environment} \\
    Optimus: \textcolor{blue}{animals} have evolved from animals with traits that have an animal instinct
\end{tcolorbox}
\caption{Latent arithmetic $s_A + s_B$ for T5VQVAE(base) and Optimus(32). \textcolor{blue}{blue}, \textcolor{orange}{orange}, and \textcolor{cyan}{shallow blue} indicate the semantic information from both $s_A$ and $s_B$, from $s_A$ only, from $s_B$ only, respectively.}
\label{tab:arithmetic_examples}
\end{table}

\subsection{Inference Task}
Lastly, we move to inference tasks, in which we aim to explore the controllability of T5VQVAE for reasoning with natural and symbolic languages. Specifically, we focus on two tasks, including syllogistic-deductive explanatory inference in EntailmentBank \cite{https://doi.org/10.48550/arxiv.2104.08661}, where a natural language conclusion has to be inferred from two premises, and mathematical expression derivation \cite{meadows2023symbolic}, where the goal is to predict the result of applying a mathematical operation to a given premise expression.

\paragraph{Quantitative Evaluation.} We quantitatively evaluate several baselines following the same procedure as the AutoEncoding task. \uline{Table \ref{tab:metrics_inf} shows that T5VQVAE outperforms all VAE models on both benchmarks} \textbf{(Finding~6)}.
\begin{table}[t]
\setlength\tabcolsep{2.5pt}
\small
\centering
\renewcommand\arraystretch{1}
\begin{tabular}{llllll}
\toprule
\multicolumn{6}{c}{\textit{Explanatory Inference (EntailmentBank)}} \\
Evaluation Metrics & BLEU & Cosine & BLEURT & Loss $\downarrow$ & PPL $\downarrow$ \\ \hline
T5(small) & 0.54 & 0.96 & 0.22 & 0.69 & 1.99 \\
T5(base) & \textbf{0.57} & \textbf{0.96} & \textbf{0.33} & \textbf{0.61} & \textbf{1.84} \\
Bart(base) & 0.54 & 0.96 & 0.17 & 0.63 & 1.87 \\
FlanT5(small) & 0.22 & 0.89 & -1.33 & 0.99 & 2.69 \\ 
FlanT5(base) & 0.32 & 0.89 & -0.31 & 0.95 & 2.58 \\ 
T5bottleneck(base) & 0.35 & 0.91 & -0.20 & 1.24 & 3.45 \\ \hdashline
Optimus(32) & 0.07 & 0.74 & -1.20 & 1.13 & 2.31 \\
Optimus(768) & 0.08 & 0.74 & -1.21 & 0.82 & 2.27 \\
DELLA(32) & 0.08 & 0.85 & -1.23 & 1.69 & 5.41 \\
DELLA(768) & 0.09 & 0.87 & -1.09 & 1.54 & 4.66 \\
T5VQVAE(small) & 0.11 & 0.73 & -1.23 & 0.85 & 2.33 \\
T5VQVAE(base) & \textcolor{black}{\textbf{0.46}} & \textcolor{black}{\textbf{0.94}} & \textcolor{black}{\textbf{0.10}} & \textcolor{black}{\textbf{0.84}} & \textcolor{black}{\textbf{2.31}} \\ \hline \hline
\multicolumn{6}{c}{\textit{Mathematical Expression Derivation}} \\
Evaluation Datasets & EVAL & SWAP & EASY & EQ & LEN \\ \hline
T5(small) & 0.69 & 0.48 & 0.57 & 0.60 & 0.63 \\
T5(base)  & 0.97 & 0.65 & 0.90 & 0.72 & 0.81 \\ \hdashline
Optimus(32) & 0.72 & 0.50 & 0.59 & 0.23 & 0.40 \\
Optimus(768) & 0.79 & 0.56 & 0.63 & 0.29 & 0.44 \\
DELLA(32)  & 0.12 & 0.16 & 0.13 & 0.13 & 0.13 \\
DELLA(768) & 0.13 & 0.18 & 0.12 & 0.13 & 0.14 \\
T5VQVAE(small) & 0.75 & \textcolor{black}{\textbf{0.57}} & 0.77 & \textcolor{black}{\textbf{0.48}} & \textcolor{black}{\textbf{0.50}} \\
T5VQVAE(base) & \textcolor{black}{\textbf{0.76}} & 0.56 & \textcolor{black}{\textbf{0.78}} & 0.47 & \textcolor{black}{\textbf{0.50}} \\
\toprule
\end{tabular}
\caption{Quantitative evaluation on inference tasks.} \label{tab:metrics_inf}
\end{table}

\paragraph{Qualitative Evaluation.} Next, we focus on the NLI task to explore the controllability of T5VQVAE for sentence-level inference traversing the latent space. 
As illustrated in Table \ref{tab:trav_inf_examples}, traversing the dimension corresponding to an individual word (e.g., \textit{object} from premise 1 (P1)) cannot preserve the target word during the traversal along with the semantic coherence of the transitions, indicating that the inference is done entirely in the Encoder. Therefore, we next explore how to manipulate the latent representation to deliver a more controllable inference behaviour. 
\begin{table}[t]
\begin{tcolorbox}[fontupper=\small, fontlower=\small] 
\underline{\textbf{P1: a human is a kind of \textcolor{blue}{object}}} \\
\underline{\textbf{P2: a child is a kind of young human}} \\
\underline{\textbf{C: a child is a kind of object}}\\ \\
dim6: a young object is a kind of child \\
dim6: a boy is a kind of young object \\
dim6: a little boy is a kind of young human
\end{tcolorbox}
\caption{T5VQVAE (base): traversed conclusions.}
\label{tab:trav_inf_examples}
\end{table}
We input two premises into the Encoder to derive the latent token embeddings of individual arguments and guide the generation of the conclusion via the Decoder. For example, for \textit{argument substitution} and \textit{verb substitution}, which refers to the process of obtaining a conclusion by substituting one argument/verb from the first premise to an argument/verb of the second premise, we substitute the respective token embeddings in the latent space and feed the resulting representation to the decoder. Table \ref{tab:quasi_1} shows that by substituting the embeddings of the arguments, we can control the behaviour of the model and elicit a systematic inference behaviour.
\uline{These results show that the latent embeddings can be manipulated to deliver inference control. In particular, we demonstrate that the distributed semantic information in the latent space contains information about co-occurring tokens within the sentence that can be systematically localised and manipulated to generate a sound conclusion. This behaviour can be potentially leveraged as a foundation to build an interpretable and controllable NLI model} \textbf{(Finding~7)}. 
\begin{table}[t]
\begin{tcolorbox}[fontupper=\small, fontlower=\small]
P1: a \textcolor{blue}{shark} is a kind of \underline{fish} \\
P2: a \underline{fish} is a kind of aquatic animal \\
Pred: a \textcolor{blue}{shark} is a kind of aquatic animal
\tcblower
P1: to \underline{move} something can mean to \textcolor{blue}{transfer} something \\
P2: flowing is a kind of \underline{movement} for energy \\
Pred: flowing is a kind of \textcolor{blue}{transfer} of energy
\end{tcolorbox}
\caption{T5VQVAE(base): quasi-symbolic inference examination in AutoEncoder (Top: argument substitution, Bottom: Verb substitution).}
\label{tab:quasi_1}
\end{table}

\section{Related Work} \label{sec:related}

\paragraph{VQVAE in NLP.} In contrast to the Image domain \cite{van2017neural, razavi2019generating}, the integration of VQ-VAEs is relatively less explored in natural language. \citet{https://doi.org/10.48550/arxiv.2110.05999} explored long text generation via VQVAE, where its discrete latent representations capture the global structure of the text. Before that, \citet{huang-ji-2020-semi} utilised a VQVAE-based model to perform semi-supervised event-type induction. Furthermore, \citet{https://doi.org/10.48550/arxiv.1905.12752} proposed the vector quantization as a residual network in the encoder of the transformer, which can perform paraphrasing after trained with an unlabeled monolingual corpus only. In contrast, this work focuses on addressing the question of whether VQ-VAEs can be used as a mechanism to deliver improved semantic control by bridging the gap between language models and VAE architectures.

\section{Conclusion}
In this work, we build a model for improving the semantic and inference control for VAE-enabled language model. We propose a new model (i.e., T5VQVAE) which is based on the close integration of a VQ-VAE and a consistently accessible and high-performing Encoder-Decoder TLM (T5). The proposed model was extensively evaluated with regard to its fine-grained and inference controls using three downstream tasks (autoencoding, text transfer, and inference task). Our experimental results indicate that the T5VQVAE can outperform the canonical state-of-the-art models in those tasks and can deliver a quasi-symbolic behaviour in the inference task (via the direct manipulation of the latent space). These results answer \questionD{}


\paragraph{Reproducibility.} The data, models, and the related codebase are available online: \url{https://github.com/SnowYJ/T5VQVAE}

\section{Scoping and Limitations}

Although this work takes initial steps toward exploring inference manipulation in latent space, achieving full inference control requires systematic annotation of inference rules, an important direction for future research.

Besides, we still observe limitations in out-of-distribution generalisation in the mathematical expressions corpus despite the improvement over existing VAE models in terms of robustness. This, in particular, is highlighted by the decrease in performance obtained on the length generalisation split (LEN) for both autoencoding and expression derivation tasks.
\chapter{Explanatory Inference Control via Inference Types} \label{cha:reason}
Before that, we aim to explore the fine-grained semantic control of explanations generation over the latent space. This chapter investigates the reasoning control by answering \questionE{}

\section{Introduction}
Explanatory sentences from EntailmentBank \cite{jansen2018worldtree,dalvi2021explaining} can encode hierarchical, taxonomic, and causal relations between concepts \citep{gardenfors2015applications}. By understanding and reasoning over these concepts expressed by explanations, humans can make intricate decisions, which is significant in scientific, cognitive, and AI domains. In this work, we focus on the Explanation-based Natural Language Inference (NLI) task where two explanations are provided to derive a single conclusion. Within this task, a central challenge involves achieving localised and quasi-symbolic inference behaviour. E.g., given the two premises: \textit{milk is a kind of liquid} and \textit{liquids can flow}, one may derive the conclusion \textit{milk can flow} by localising and substituting the concept \textit{liquids} with \textit{milk}.

This work investigates a pivotal question: How can vanilla Transformer-based NLI models learn and generalise such quasi-symbolic inference in distributional representation spaces? Resolving this question allows us to shorten the gap between deep latent semantics and symbolic representations \cite{10.3115/1075218.1075283,banarescu2013abstract}, integrating the flexibility of distributional-neural models with the properties of symbolic, compositional representations, facilitating interpretability, compositionality \cite{dankers-etal-2022-paradox,marcus2003algebraic}, and reasoning control. Therefore, this work provides a complete initial step in investigating the quasi-symbolic NLI over the distributional latent semantic space.
\begin{figure}[t]
    \centering
    \includegraphics[width=0.7\columnwidth]{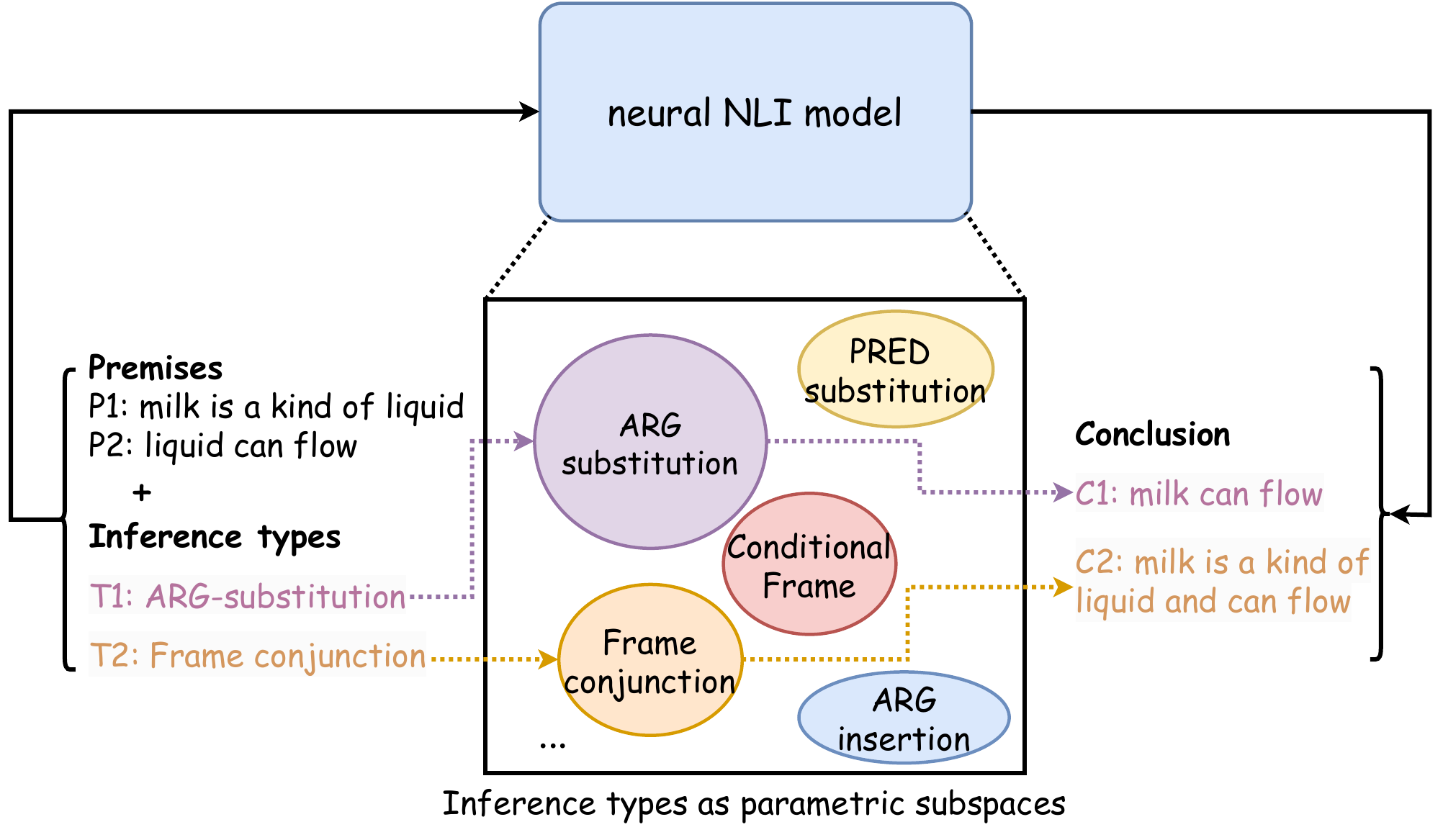}
    \caption{Quasi-symbolic NLI framework. Inference types can be encoded as functional subspaces, which are separated in parametric space. Thus, by manipulating the inference types, we can deliver localised, symbolic inference control.}
    \label{fig:concep_vis}
\end{figure}

\textbf{First,} we systematically annotate the quasi-symbolic NLI behaviour, grounded on linguistic, formal semantic theory. \citet{valentino2021natural} have demonstrated that explanation-based NLI cannot be directly framed as pure logical reasoning, which commonly has subtler incompleteness and consistency problems from a logical point of view. Meanwhile, explanatory NLI corresponding to definable inference patterns and symbolic operations can be localised over the sentence structure. Under the Argument Structure Theory (AST) \cite{jackendoff1992semantic}, the predicate-argument structure and semantic roles from explanatory sentences can be effectively represented, localised, and disentangled in the latent space of transformer-based models \cite{zhang2023learning,zhang-etal-2024-graph}. A particular instance of an AST representation is the Abstract Meaning Representation (AMR) \cite{banarescu2013abstract}. 

Motivated by this middle ground between logical representations and lexico-semantic inference patterns, we introduce granular inference types based on explanatory sentences, using AMR to define the symbolic operations. Specifically, \textit{we leverage the AMR to systematically characterise quasi-symbolic inference behaviours, named inference types, grounded on AMR symbolic graphs.} Using the explanation-based NLI dataset (EntailmentBank), we identify ten categories of symbolic transformations and provide annotations for 5,134 premise-conclusion pairs.

\textbf{Second,} to enable localised, quasi-symbolic NLI control, a vanilla neural NLI model, such as T5 \cite{raffel2020exploring}, should function as a soft rule-based system in which inference types are explicitly encoded and separated within the latent space. Then, providing the same input premises with different inference type specifications should yield conclusions that preserve the same lexical semantics but differ in sentence-level semantics corresponding to different inference types, reflecting the localised, quasi-symbolic inference control. Motivated by this, \textit{we propose a quasi-symbolic NLI framework under Neural Tangent Kernel (NTK) theory} \cite{jacot2018ntk} where different functions (input-output inference type transformation) are encoded as separated subspaces in the parametric space via gradient descent optimisation.

To assess this framework, we prefix premises with inference-type labels to condition model behaviour. After training, the model should exhibit the behaviours, including: (1) Training Dynamics: During training, explicit supervision on inference types aligns the model’s reasoning trajectory with target inference behaviours, improving conclusion prediction accuracy. (2) Inference Composition: By varying inference type during inference, the model can separate the semantics of the premises from the inference behaviour. This enables localised, quasi-symbolic NLI control, allowing for flexible and interpretable reasoning. (3) Inference-type Separation: The inference-type can be clustered and separated within the latent space.


In summary, this work provides a foundation in investigating the quasi-symbolic NLI over distributional semantic spaces, with the following contributions: 

\textbf{Linguistic Formalisation:} Systematic characterisation of inference types via AMR, bridging symbolic operations and explanatory NLI.

\textbf{Theoretical Framework:} A quasi-symbolic NLI framework based on NTK theory where inference types govern subspace formation in latent representations.

\textbf{Empirical Validation:} Demonstrated improvements in training efficiency, inference accuracy, and localised control, suggesting future direction of rule-based learning and generalising in neural spaces. The experimental pipelines are released. The annotations are planned for public release.
\begin{table*}[ht!]
    \small
    \centering
    \resizebox{15.7cm}{!}{
    \begin{tabular}{p{3.5cm}p{3cm}p{0.5cm}p{8cm}}
    \toprule
        Original type & Symbolic type & Prop. & Example entailment relation \\ \hline
        \multirow{9}{*}{Substitution} & \multirow{3}{*}{\shortstack{ARG substitution \\ (ARG-SUB)}} & \multirow{3}{*}{19\%} & P1: \textcolor{blue}{a scar on the knee} is a kind of \textcolor{red}{scar} \\
        &&& P2: a \textcolor{red}{scar} is an acquired characteristic \\
        &&& C: \textcolor{blue}{a scar on the knee} is an acquired characteristic \\
        & \multirow{3}{*}{\shortstack{PRED substitution \\ (PRED-SUB)}} & \multirow{3}{*}{5\%} & P1: food \textcolor{red}{contains} nutrients and energy for living things \\
         &&& P2: to \textcolor{red}{contain} something can mean to \textcolor{blue}{store} something \\
          &&& C: food \textcolor{blue}{stores} nutrients and energy for living things \\
        & \multirow{3}{*}{\shortstack{Frame substitution \\ (FRAME-SUB)}} & \multirow{3}{*}{20\%}  & P1: the \textcolor{blue}{formation of diamonds} requires \textcolor{red}{intense pressure} \\
         &&& P2: the \textcolor{red}{pressure is intense} deep below earth 's crust \\
          &&& C: the \textcolor{blue}{formation of diamonds} occurs deep below the crust of the earth \\ \hline
        \multirow{3}{*}{\shortstack{Inference from Rule}} & \multirow{3}{*}{\shortstack{Conditional frame \\ insertion/substitution \\ (COND-FRAME)}} & \multirow{3}{*}{12\%}  & P1: if \textcolor{blue}{something is renewable} then \textcolor{red}{that something is not a fossil} \\ 
        &&& P2: \textcolor{blue}{fuel wood is a renewable resource} \\
        &&& C: \textcolor{red}{wood is not a fossil fuel} \\ \hline
        \multirow{6}{*}{\shortstack{Further Specification \\ or Conjunction}} & \multirow{3}{*}{\shortstack{ARG insertion \\ (ARG-INS)}} & \multirow{3}{*}{18\%}  & P1: solar energy \textcolor{blue}{comes from the sun} \\ 
        &&& P2: \textcolor{red}{solar energy is a kind of energy} \\
        &&& P3: \textcolor{red}{solar energy is a kind of energy} that \textcolor{blue}{comes from the sun} \\
        & \multirow{3}{*}{\shortstack{Frame conjunction \\ (FRAME-CONJ)}} & \multirow{3}{*}{6\%} & P1: \textcolor{blue}{photosynthesis stores energy} \\
        &&& P2: \textcolor{red}{respiration releases energy} \\
        &&& C: \textcolor{blue}{photosynthesis stores energy} and \textcolor{red}{respiration releases energy} \\ \hline
        \multirow{3}{*}{\shortstack{Infer Class \\ from Properties}} & \multirow{3}{*}{\shortstack{ARG/PRED \\ generalisation \\ (ARG/PRED-GEN)}} & \multirow{3}{*}{1\%}  & P1: \textcolor{blue}{rock} is a hard material \\ 
        &&& P2: \textcolor{red}{granite} is a hard material \\
        &&& C: \textcolor{red}{granite} is a kind of \textcolor{blue}{rock} \\ \hline
        \multirow{3}{*}{\shortstack{Property Inheritance}} & \multirow{3}{*}{\shortstack{ARG substitution \\ (Property Inheritance) \\ (ARG-SUB-PROP)}} & \multirow{3}{*}{0.4\%}  & P1: \textcolor{blue}{blacktop} is made of asphalt concrete \\ 
        &&& P2: asphalt \textcolor{red}{has a smooth surface} \\
        &&& C: a \textcolor{blue}{blacktop} \textcolor{red}{has a smooth surface} \\ \hline

\multirow{3}{*}{Causal Expression} & \multirow{3}{*}{Causality (IFT)} & \multirow{3}{*}{0.8\%} & an optical telescope requires visible light for human to use \\ 
&&& clouds / dusts block visible light \\
&&& \textcolor{blue}{if} there is clouds or dusts, \textcolor{blue}{then} the optical telescope cannot be used  \\ \hline
\multirow{3}{*}{\shortstack{Example-based \\ Inference}} & \multirow{3}{*}{\shortstack{Example \\ (EXAMPLE)}} & \multirow{3}{*}{0.9\%} & a shelter can be used for living in by raccoons \\ 
&&& some raccoons live in hollow logs \\
&&& \textcolor{blue}{an example of} a shelter is a raccoon living in a hollow log \\
\toprule
\end{tabular}
    }
\caption{Examples of symbolic inference types, with their corresponding abbreviations provided in parentheses and used consistently throughout the paper. The EntailmentBank utilised for this task comprises 5,134 instances, with our annotations covering 84\% of the (premises, conclusion) cases. These annotations are planned for public release.} \label{tab:inference_type_example}
\end{table*}
\section{Defining Inference Types} \label{sec:bg}
In this section, we introduce a set of granular inference types derived from explanatory sentences, using AMR to define the symbolic operations. These operations capture the transformations from premises to conclusions at a semantic level. It is important to note that AMR is not employed as a representational component within the proposed model architecture. Rather, it serves as a formal semantic framework to precisely ground and characterise the symbolic inference operations. Table~\ref{tab:inference_type_example} presents the AMR-grounded inference types alongside illustrative examples from the EntailmentBank corpus. In what follows, we formally define each lexico-semantic inference type and its corresponding symbolic transformation.
\begin{figure}[ht!]
    \centering
    \includegraphics[scale=0.45]{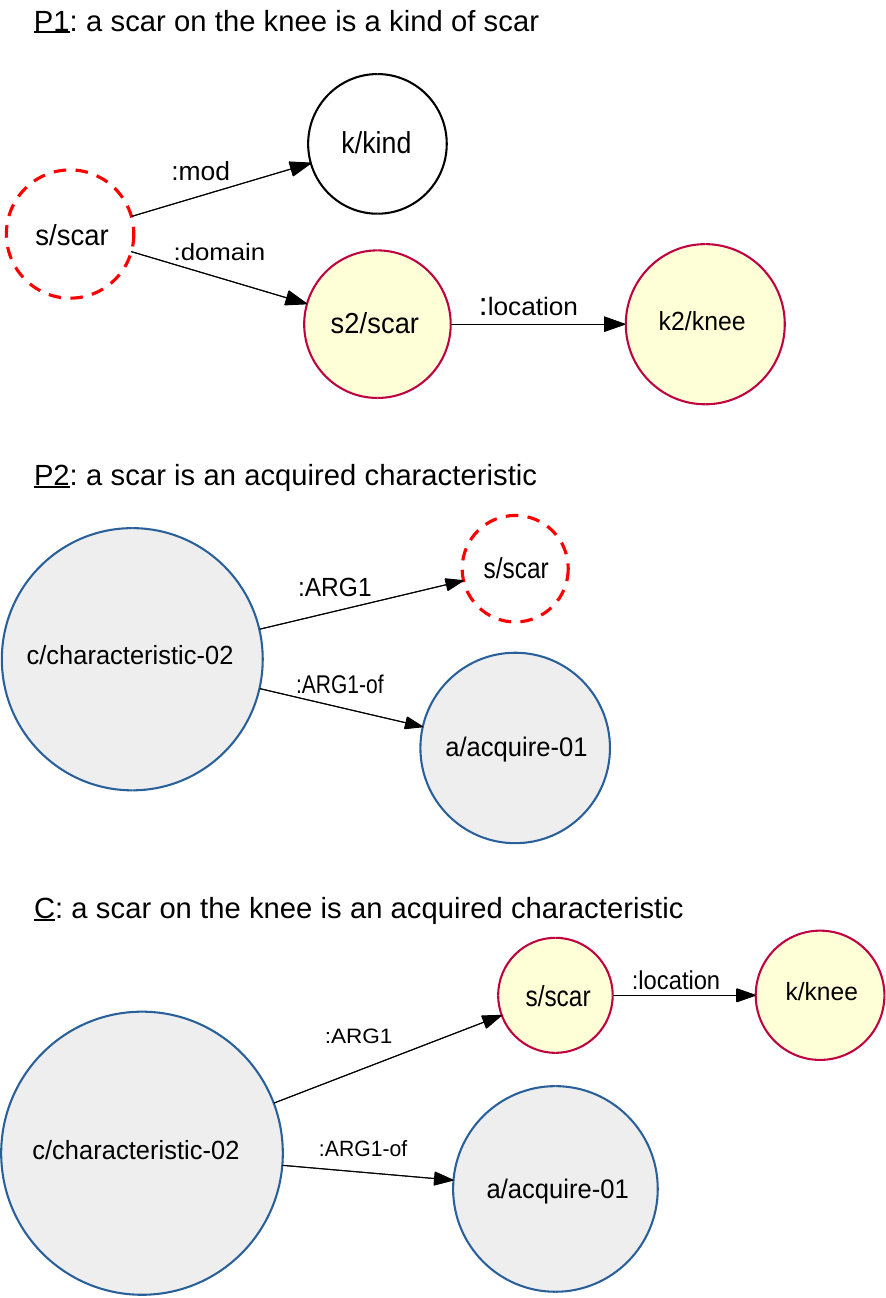}
    \caption{AMR argument substitution: the inference behaviour is defined as subgraph substitution.}
    \label{fig:amr_argsub}
\end{figure}

The \textit{substitution} category refers to obtaining a conclusion by replacing a predicate/argument term from one premise with a predicate/argument term from the other premise. Possible variations of this category include (1) \textit{argument (ARG) substitution}, (2) \textit{predicate (PRED) substitution}, and (3) \textit{frame (PRED+ARG) substitution}. In this category, one premise is used to connect two terms which are usually connected by \textit{is a kind of, is a source of}, etc. Conceptualising the AMR representation as a graph, this can be symbolically represented as a subgraph substitution operation over the premise graphs, as illustrated in Figure~\ref{fig:amr_argsub}. The \textit{PRED substitution} category works in a similar manner, but replacing a predicate term. The two predicates are usually linked by the following patterns: ``\textit{$v_1$ is a kind of $v_2$}'', ``\textit{to $v_1$ something means to $v_2$ something}'', etc. The \textit{frame (PRED+ARG) substitution} category combines both previous categories by replacing a frame (predicate subgraph) of one of the premises with one from the other premise.

The \textit{further specification or conjunction} category allows for obtaining a conclusion by joining both premises. It includes (4) \textit{ARG insertion} and (5) \textit{frame conjunction}. For \textit{ARG insertion}, the conclusion is obtained by connecting an argument from one of the premises to a frame of the other. As for \textit{frame conjunction/disjunction}, the conclusion is obtained by joining the premises graphs through a conjunction/disjunction node (\textit{and}) or (\textit{or}). 

The \textit{inference from rule} category from \citet{dalvi2021explaining} encompasses a specific instance of insertion or substitution, identified as (6) \textit{conditional frame insertion/substitution}. In this category, a frame is either inserted or replaced as an argument of a premise, following a conditional pathway present in the other premise. 

The inference type \textit{infer class from properties} has been re-categorised as (7) \textit{ARG or PRED generalisation}, where a new \textit{:domain} relation frame is created if both premise graphs differ by a single predicate/argument term.
(8) \textit{Property inheritance}, on the other hand, is a special case of \textit{ARG substitution}, where one of the premises describes a \textit{is made of} relationship between the entity in the other premise and its replacement. 

Finally, (9) \textit{Causal Expression} and (10) \textit{Example-based Inference} categories are defined according to the key lexical characteristic of the conclusion, as systematic AMR transformations which could be applied without rephrasing the underlying explanatory sentences could not be determined. 

Thus far, we have defined the explanatory NLI behaviours grounded on the symbolic AMR graph from the perspective of formal semantic theory. In the next section, we aim to introduce the quasi-symbolic NLI framework.

\section{Quasi-symbolic NLI Framework} \label{sec:latent_props}


\subsection{Quasi-symbolic NLI Formalisation}
In this study, we formalise the ``quasi-symbolic NLI behaviour'' as rule-based reasoning over neural representation, where discrete inference behaviours are implemented through differentiable transformations over continuous neural representations. This is achieved by characterising and manipulating quasi-symbolic inference behaviours, denoted by $\pi \in \Pi$, where $\Pi$ represents the space of all possible inference rules. The process involves three key stages: (i) Neural Encoding: The premises $p_1$ and $p_2$ are encoded into continuous vector representations ($\overrightarrow{p_1}$ and $\overrightarrow{p_2}$) in the latent space. (ii) Rule-Based Reasoning: The encoded representations are transformed using a reasoning function guided by the inference behaviour $\pi$. (iii) Neural Decoding: The resulting vector, $\overrightarrow{c}$, is decoded into a natural language conclusion $c$. Formally, the process can be described as follows:
\[
\begin{aligned}
&1.~ \overrightarrow{p_1}, \overrightarrow{p_2} = f_{encode}(p_1, p_2) \\
&2.~ \overrightarrow{c} = f_{reason}(\overrightarrow{p_1}, \overrightarrow{p_2}; \pi) \\
&3.~ c = f_{decode}(\overrightarrow{c})
\end{aligned}
\]
Here, $f_{encode}$, $f_{reason}$, $f_{decode}$ represent the encoding, reasoning, and decoding functions in a neural NLI model. The injection of $\pi$ should exhibit two advantages:

\textit{\textbf{1. Training Dynamics:}} During training, explicit supervision on $\pi$ aligns the model’s reasoning trajectory with target inference behaviours, improving prediction accuracy.

\textit{\textbf{2. Inference Composition:}} By varying $\pi$ during inference, the model can separate the semantics of the premises from the inference behaviour. This enables localised, quasi-symbolic NLI control.

\subsection{Quasi-symbolic NLI Representation}
We focus on standard neural NLI models, with particular - but not exclusive - attention to encoder-decoder architectures such as T5, due to their inherent separation of reasoning and decoding phases, which naturally accommodates quasi-symbolic reasoning. However, this framework can also be adapted into the decoder-only architecture, where the rules are captured, such as through in-context learning \cite{liu2024incontextvectorsmakingcontext}. From a representational perspective, we propose the concepts of latent rule space and feature space to align with the function of the neuro-symbolic NLI model.

\paragraph{Latent rule space.}
The latent rule space refers to the functional parametric space (i.e., models' weights), which captures the structured, rule-based reasoning behaviours $\pi \in \Pi$. 

\textit{\textbf{Proposition1:}} \textit{The inference-types can be encoded and separated within the parametric space.}

We further propose that rule-based reasoning is primarily materialised in the encoder of the NLI model.

\textit{\textbf{Proposition2:}} \textit{The inference behaviour is instantiated at the encoder and can be controlled by the injection of the associated inference type labels.}

\paragraph{Latent feature space.} \label{sec:latent_var_NLI_arc}
The latent feature space refers to the output embedding space. To evaluate the feature representation capability, we next describe the methodological framework behind the construction of the abstract sentence representation within T5 (named T5 bottleneck). As for the encoder's final layer output embedding, we compute dimension-wise mean pooling over token embeddings, followed by a multi-layer perceptron to obtain sentence embeddings. As for the decoder's input embedding, we reconstruct token embeddings via linear projection, feeding them into the decoder’s cross-attention mechanism. Here, we only describe the optimal setup.

\subsection{Proposition1: Formal Proof and Evaluation} \label{sec:proof}





\paragraph{NTK interpretation of inference-type subspaces.}


Each symbolic inference type $\pi$ is explicitly embedded as part of the model input, for example, as a token prefix (EP). As a result, the model effectively learns a function $f_\theta(x, \pi)$, where $x = (p_1, p_2)$ are the premises and $\pi$ is the symbolic inference type. The function $f_\theta$ thus jointly depends on both the content of the premises and the nature of the symbolic operation to be performed.

Within the Neural Tangent Kernel (NTK) framework, the similarity between two input examples of the same inference type $\pi$ is captured by the NTK as follows:
\begin{equation}
\Theta_{\pi}(x, x') = \nabla_\theta f_\theta(x, \pi)^\top \nabla_\theta f_\theta(x', \pi)
\end{equation}
where $\nabla_\theta f_\theta(x, \pi)$ denotes the gradient of the model output with respect to its parameters, evaluated at the input $(x, \pi)$. This kernel quantifies how a parameter update from one input-output pair would affect another pair, conditioned on the shared inference type.

According to NTK theory \cite{jacot2018ntk}, in the infinite-width limit, the evolution of the model’s predictions under gradient descent training can be described by a linear kernel regression in the RKHS (Reproducing Kernel Hilbert Space) associated with $\Theta_{\pi}$. Specifically, the prediction at time $t$, $f_t(x, \pi)$, evolves as:
\begin{equation}
f_t(x, \pi) = f_0(x, \pi) - \Theta_{\pi}(x, \cdot) \left[\Theta_{\pi} + \lambda I\right]^{-1} (f_0 - c)
\end{equation}
where $f_0(x, \pi)$ is the model's output at initialisation for each training input, $\lambda$ is a regularisation parameter, and $c$ is the vector of ground truth conclusions.

Crucially, this formulation implies that each symbolic inference type $\pi$ induces a distinct kernel $\Theta_{\pi}$, which in turn defines a unique RKHS $\mathcal{H}_{\pi}$—that is, a function space within which the model's solutions for inference-type $\pi$ reside. As the symbolic type $\pi$ is varied, the structure of the kernel and the corresponding function space changes, reflecting the distinct reasoning behaviours or transformations associated with different inference operations. Thus, the model encodes different symbolic inference patterns in distinct, kernel-induced subspaces.

\paragraph{Quantitative evaluation.}
For two different inference types, $\pi_i \neq \pi_j$, we examine the relationship between their corresponding neural tangent kernels (NTKs), $\Theta_{\pi_i}$ and $\Theta_{\pi_j}$. Specifically, we are interested in the interaction between the parameter gradients induced by inputs associated with different inference types.

Consider two data points $x$ and $x'$, possibly corresponding to different premise pairs. 
When considering cross-type similarities, we are interested in the inner product between the gradients for different types:
\begin{equation}
G_{ij}(x, x') := \langle \nabla_\theta f_\theta(x, \pi_i), \nabla_\theta f_\theta(x', \pi_j) \rangle
\end{equation}



If the symbolic inference types $\pi_i$ and $\pi_j$ encode fundamentally different reasoning operations (e.g., ARG-SUB vs. PRED-SUB), the gradients with respect to $\theta$ for inputs labelled with $\pi_i$ and those labelled with $\pi_j$ will tend to point in different directions (i.e., orthogonality) in parameter space. This is because each type imposes a distinct task or transformation pattern on the model, causing it to utilise different portions of its capacity. Therefore, by measuring the cosine similarity between gradient vectors associated with different inference types, we can quantify the separability between different inference-type subspaces.

\section{Empirical Evaluation}
The experiment addresses four key questions: (i) Do symbolic inference types enhance model training and inference performance? (ii) Can these inference types be used for prescriptive inference control? (iii) Does the incorporation of a sentence bottleneck contribute to improved feature representation? (iv) Whether the inference-type can be separated and clustered in the latent space?

\subsection{Training and Inference Evaluation} \label{sec:tr_inf_eval}
\begin{table}[ht!]
\scriptsize
\centering
\resizebox{8.5cm}{!}{
\centering
\renewcommand\arraystretch{1}
\begin{tabular}{ccccc} 
\toprule
Baseline & INJ & BLEU & Cosine & BLEURT \\ \hline
\multicolumn{5}{c}{\textit{seq2seqLM: encoder-decoder architecture}}  \\ \hline
\multirow{4}{*}{\shortstack{T5 \\ original \\ (small)}} & DE & 0.55 & 0.96 & 0.30 \\ 
& DP & 0.59 & 0.96 & 0.34 \\ 
& EP & \textcolor{black}{\textbf{0.65}} & \textcolor{black}{\textbf{0.97}} & \textcolor{black}{\textbf{0.45}} \\
& NO & 0.54 & 0.96 & 0.22 \\ \hline

\multirow{4}{*}{\shortstack{T5 \\ original \\ (base)}} & DE & 0.46 & 0.96 & 0.23 \\ 
& DP & 0.53 & 0.96 & 0.25 \\ 
& EP & \textcolor{black}{\textbf{0.61}} & \textcolor{black}{\textbf{0.97}} & \textcolor{black}{\textbf{0.39}} \\ 
& NO & 0.57 & 0.96 & 0.33 \\ \hline
\multirow{4}{*}{\shortstack{T5 \\ original \\ (large)}} & DE & 0.60 & 0.97 & 0.46 \\ 
& DP & 0.64 & 0.97 & 0.44 \\ 
& EP & \textcolor{black}{\textbf{0.67}} & \textcolor{black}{\textbf{0.97}} & \textcolor{black}{\textbf{0.50}} \\ 
& NO & 0.57 & 0.96 & 0.31 \\ \hline
\multicolumn{5}{c}{\textit{CausalLM: decoder only architecture}}  \\ \hline

\multirow{2}{*}{\shortstack{GPT2(xl)}} & DP & \textcolor{black}{\textbf{0.28}} & \textcolor{black}{\textbf{0.91}} & \textcolor{black}{\textbf{-0.90}} \\ 
& NO & 0.27 & 0.90 & -0.97 \\ \hline

\multirow{2}{*}{\shortstack{Qwen2.5(0.5B)}} & DP & \textcolor{black}{\textbf{0.65}} & \textcolor{black}{\textbf{0.97}} & \textcolor{black}{\textbf{0.48}} \\ 
& NO & 0.63 & 0.97 & 0.45 \\ \hline

\multirow{2}{*}{\shortstack{Llama3.2(1B)}} & DP & \textcolor{black}{\textbf{0.63}} & \textcolor{black}{\textbf{0.97}} & \textcolor{black}{\textbf{0.46}} \\ 
& NO & 0.60 & 0.96 & 0.42 \\ \hline

\multicolumn{5}{c}{\textit{seq2seqLM with sentence bottleneck}}  \\ \hline
\multirow{4}{*}{\shortstack{T5 \\ bottleneck \\ (base)}} & DE & 0.35 & 0.91 & -0.15 \\ 
& DP & 0.39 & 0.91 & -0.13 \\ 
& EP & \textcolor{black}{\textbf{0.42}} & \textcolor{black}{\textbf{0.92}} & \textcolor{black}{\textbf{-0.07}} \\
& NO & 0.35 & 0.91 & -0.20 \\ \hline
\multirow{4}{*}{\shortstack{Optimus \\ (BERT-GPT2)}} & DE & \textcolor{black}{\textbf{0.26}} & \textcolor{black}{\textbf{0.80}} & \textcolor{black}{\textbf{-1.11}} \\ 
& DP & 0.25 & 0.79 & -1.14 \\ 
& EP & 0.09 & 0.74 & -1.17 \\
& NO & 0.07 & 0.74 & -1.20 \\
\toprule

\end{tabular}
}
\caption{Quantitative evaluation on testset, where best results are highlighted in \textbf{\textcolor{black}{bold}}. Specification for abbreviation. INJ: ways for injecting the information of inference types into the model, it includes DE: decoder end, DP: decoder prefix, EP: encoder prefix, NO: no inference type.} \label{tab:metrics_with_inference_type}
\end{table}
First, we evaluate (i) if symbolic inference types enhance model training and inference performance. We consider three mechanisms to conditionally inject the symbolic inference types into the model. 
\textbf{i.} The inference type as the prefix for the premises at the Encoder;
\textbf{ii.} The inference type as the prefix for the conclusion in the Decoder; 
\textbf{iii.} The inference type at the end of the conclusion in the Decoder.

\paragraph{Training dynamics.} We first evaluate generative performance after training based on three metrics: BLEURT \cite{https://doi.org/10.48550/arxiv.2004.04696}, BLEU \cite{Papineni02bleu:a}, and cosine similarity against sentenceT5 \cite{https://doi.org/10.48550/arxiv.2108.08877}. By comparing the predicted and golden conclusions, we can fairly evaluate the accuracy of the NLI performance. For the baseline, we choose the T5, GPT2 \cite{Radford2019LanguageMA}, Qwen2.5 \cite{qwen2025qwen25technicalreport}, Llama3.2 \cite{grattafiori2024llama3herdmodels}, our T5 bottleneck and Optimus \cite{li2020optimus} with 768 latent dimensions as testbed. The performances are measured from the Entailment test set.

In Table \ref{tab:metrics_with_inference_type}, across all baseline models, incorporating inference types into the encoder consistently results in improved performance as measured by BLEU, Cosine, and BLEURT metrics, \uline{indicating the inference behaviour is instantiated at the encoder} (\textit{Proposition}). This finding also suggests that \uline{during training, explicit supervision on inference types aligns the model’s reasoning trajectory with target inference behaviours, improving conclusion prediction accuracy} \textbf{(Finding~1)}. A similar observation is reflected in the test loss curve shown in Figure \ref{fig:loss_curve} in the supplementary material.

Furthermore, previous studies have revealed that the LLM evaluation can be consistent with the results obtained by expert human evaluation \citep{chiang-lee-2023-large, liu-etal-2023-g, wang-etal-2023-chatgpt, Huang_2023}. Thus, we also conduct a quantitative analysis to measure whether the generated conclusion contradicts the premises through LLM evaluators, including ChatGPT4o as the baseline and GPT4o-mini for comparison. \uline{Table \ref{tab:manual} indicates that EP can consistently result in improved LLM agreement scores} \textbf{(Finding~2)}. A manual check is presented in the supplementary material (Tables~\ref{tab:non_consist} and \ref{tab:ep_no}). 

\begin{table}[ht!]
\scriptsize
\centering
\resizebox{8.5cm}{!}{
\renewcommand\arraystretch{1}
\begin{tabular}{ccccc} 
\toprule
Baseline & INJ & ChatGPT4o & GPT4o-mini \\ \hline
\multirow{4}{*}{\shortstack{T5 \\ original \\ (large)}} 
& DE & 0.85 & 0.83 \\
& DP & 0.86 & 0.83 \\
& EP & \textbf{\textcolor{black}{0.91}} & \textbf{\textcolor{black}{0.85}} \\
& NO & 0.84 & 0.82 \\

\toprule
\end{tabular}
}
\caption{Agreement scores for the quantitative analysis using LLMs on the test set. 
} \label{tab:manual}
\end{table}

\paragraph{In-context learning.} Next, we quantitatively evaluate the inference types within in-context learning (ICL) in contemporary large language models (LLMs). As illustrated in Table \ref{tab:metrics_llm}, prompting with inference types can improve the performance of ICL in all tested LLMs. Besides, within the context of causal LLMs, an increase in few-shot examples\footnote{We randomly sample the examples with the same inference type as the current test example from the training set. We perform ten times and calculate the average for each metric.}, improves the performance. This finding indicates that \uline{our inference types can be generalised across various checkpoints and architectures, ultimately enhancing the reasoning capabilities of LLMs} \textbf{(Finding~3)}.
\begin{table}[ht!]
\centering
\resizebox{10.5cm}{!}{
\begin{tabular}{cccccc}
\toprule
Baseline & INJ & Num & BLEU & Cosine & BLEURT \\ \hline
\multicolumn{6}{c}{\textit{Seq2seqLLM: encoder-decoder architecture}}  \\ \hline 
\multirow{4}{*}{\shortstack{CoT-T5 (11b) \\ \cite{kim2023cot}}}
& Yes & 10 & 0.51 & 0.97 & 0.39 \\
& Yes & 5 & 0.51 & 0.97 & 0.39 \\
& Yes & 0 & 0.50 & 0.97 & 0.36 \\ 
& NO & 0 & \textbf{\textcolor{black}{0.46}} & \textbf{\textcolor{black}{0.96}} & \textbf{\textcolor{black}{0.31}} \\ \hline
\multirow{4}{*}{\shortstack{Flan-T5 (xxl)}} 
& Yes & 10 & 0.51 & 0.97 & 0.41 \\
& Yes & 5 & 0.53 & 0.97 & 0.43\\
& Yes & 0 & 0.50 & 0.96 & 0.37 \\ 
& NO & 0 & \textbf{\textcolor{black}{0.48}} & \textbf{\textcolor{black}{0.96}} & \textbf{\textcolor{black}{0.36}} \\ \hline
\multicolumn{6}{c}{\textit{CausalLLM: decoder only architecture}}  \\ \hline
\multirow{4}{*}{\shortstack{GPT-3.5-turbo-0125}} 
& Yes & 10 & 0.52  & 0.96 & 0.40\\
& Yes & 5 & 0.48  & 0.96 & 0.35\\
& Yes & 0 &  0.46  & 0.96 &  \textbf{\textcolor{black}{0.31}}\\ 
& NO & 0 & \textbf{\textcolor{black}{0.42}} & 0.96 & 0.33 \\ \hline
\multirow{4}{*}{\shortstack{GPT-4-0613}} 
& Yes & 10 & 0.53 & 0.97 & 0.50 \\
& Yes & 5 & 0.52 & 0.97 & 0.47 \\
& Yes & 0 & 0.52 & 0.97 & 0.50 \\ 
& NO & 0 & \textbf{\textcolor{black}{0.47}} & \textbf{\textcolor{black}{0.96}} & \textbf{\textcolor{black}{0.40}} \\ \hline
\multirow{4}{*}{\shortstack{llama3-70b-8192}} 
& Yes & 10 & 0.54 & 0.97 & 0.54 \\
& Yes & 5 & 0.52 & 0.97 & 0.52 \\
& Yes & 0 & 0.51 & 0.97 & 0.47 \\ 
& NO & 0 & \textbf{\textcolor{black}{0.44}} & \textbf{\textcolor{black}{0.96}} & \textbf{\textcolor{black}{0.40}} \\
\toprule
\end{tabular}
}
\caption{ICL evaluation of test cases, where worst results are highlighted in \textbf{\textcolor{black}{bold}}. The prompt is \textit{``performing natural language inference [where the inference type is type, description], $[p1; p2; c]_{\times \text{num}}$. p1, p2, what is the conclusion?"}. $num$ is the number of examples. The \textit{description} is based on the definition of inference types in Section \ref{sec:bg}.} \label{tab:metrics_llm}
\end{table}

\subsection{Quasi-symbolic NLI Evaluation} \label{sec:quasi_sym_inf_eval}
Second, we evaluate (ii) if these inference types can be used for prescriptive inference control.
\paragraph{Qualitative evaluation.} We qualitatively evaluate the quasi-symbolic NLI behaviour on the generation of conclusions by systematically intervening on the inference type prior to the encoder. As illustrated in Table \ref{tab:control_generation}, we can observe that the associated linguistic properties of the conclusion can be controlled consistently with the inference type modifications and this inference control is independent of the semantics of premises, which indicates that \uline{the representation mechanisms can improve inference control with regard to symbolic/lexico-semantic properties} \textbf{(Finding~4)}. For example, when the type is ARG substitution (ARG-SUB), the model can generate \textit{the blacktop is made of a smooth surface} by replacing the argument \textit{asphalt concrete} with \textit{smooth surface}. The conclusions are changed to \textit{asphalt and blacktop have the same surface} when the inference type is the conjunction (FRAME-CONJ). Additional examples are provided in Table \ref{tab:more_example_3}.
\begin{table}[ht!]
\begin{tcolorbox}[fontupper=\small, fontlower=\small] 
\underline{P1: \textcolor{blue}{blacktop} is made of \textcolor{red}{asphalt concrete}} \\
\underline{P2: \textcolor{red}{asphalt} has a \textcolor{orange}{smooth surface}}\\ \\
ARG-SUB: the \textcolor{blue}{blacktop} is made of \textcolor{orange}{smooth surface} \\
ARG-SUB-PROP: \textcolor{blue}{blacktop} has a \textcolor{orange}{smooth surface} \\
ARG/PRED-GEN: a \textcolor{blue}{blacktop} is a kind of \textcolor{red}{asphalt} \\
ARG-INS: \textcolor{red}{asphalt concrete} \textcolor{blue}{blacktop} has a \textcolor{orange}{smooth surface} \\
FRAME-CON: \textcolor{red}{asphalt} and \textcolor{blue}{blacktop} have the same surface \\
IFT: if the \textcolor{red}{asphalt} has a \textcolor{orange}{smooth surface} then the \textcolor{blue}{blacktop} will have a \textcolor{orange}{smooth surface}
\end{tcolorbox}

\caption{Controllable generation over original T5 (base) (ARG-SUB: argument substitution, ARG/PRED-GEN: argument/predicate generalisation. ARG-SUB-PROP: property inheritance. ARG-INS: argument insertion, FRAME-CON: frame conjunction, IFT: casual expression.). 
}
\label{tab:control_generation}
\end{table}

\paragraph{Quantitative analysis.} Next, we perform an automated quantitative analysis using LLMs, including ChatGPT4o and GPT4o-mini. For each pair of premises in the EntailmentBank test set, we apply various inference types to generate a diverse set of conclusions using the fine-tuned T5 (base) model. We then assess the resulting (premises, conclusion, inference type) tuples based on two criteria: (i) whether the generated conclusion contradicts the premises, and (ii) whether the (premises, conclusion) pair is consistent with the specified inference type. Utilising the prompt detailed in Table \ref{tab:prompt} (supplementary material), we report the model agreement score for each criterion. As illustrated in Table \ref{tab:llm_eval}, \uline{the T5 (base) model with controlled symbolic inference types achieves consistency and alignment scores exceeding 60\% for both evaluated dimensions} \textbf{(Finding~5)}.
\begin{table}[ht!]
\centering
\resizebox{7.5cm}{!}{
\begin{tabular}{lcc} \toprule
Evaluators & consistency & alignment \\ \hline
ChatGPT4o & 0.67 & 0.63 \\ 
GPT4o-mini & 0.65 & 0.62 \\ \bottomrule
\end{tabular}
}
\caption{Automated quantitative analysis scores.} \label{tab:llm_eval}
\end{table}

\subsection{Latent Feature Space Evaluation} \label{sec:latent_sent_eval}
Third, we evaluate (iii) whether the incorporation of feature space (i.e., abstract sentence bottleneck) contributes to improved feature, concept representation in the NLI task. We especially select the VAE baselines due to their analogous encoder-bottleneck-decoder architecture, wherein the compressed sentence bottleneck captures high-level, generalised semantics (concepts) \citep{mercatali-freitas-2021-disentangling-generative,zhang2023learning} and evaluate the abstract sentence embedding using as an associated explanation retrieval task (explanation-regeneration - i.e. retrieving the associated explanatory facts relevant to a claim) \cite{valentino2021hybrid}. Given a question-and-answer pair, it reconstructs the entailment tree by searching the explanations from a fact bank (i.e., WorldTree \cite{jansen-etal-2018-worldtree}) in an iterative fashion using a dense sentence encoder. In this framework, we can replace the sentence embeddings with the proposed T5 bottleneck baseline to evaluate its abstract sentence embeddings. We compare the T5 bottleneck with abstract sentence representation baselines: Optimus and four LSTM VAEs, and evaluate them via mean average precision (MAP). \uline{As illustrated in Table \ref{tab:scar}, the T5 bottleneck outperforms all baselines, indicating that it can deliver a better abstract representation of explanatory sentences and entailment relations in a retrieval setting} \textbf{(Finding~6)}.
\begin{table}[ht!]
\centering
\resizebox{8.5cm}{!}{
\begin{tabular}{ccccc} \toprule
depth & t=1 & t=2 & t=3 & t=4 \\ \hline
DAE\cite{10.1145/1390156.1390294} & 30.27 & 31.74 & 30.65 & 30.74 \\
AAE\cite{makhzani2016adversarial} & 29.13 & 30.47 & 29.33 & 29.14 \\
LAAE\cite{rubenstein2018latent} & 19.13 & 20.86 & 18.32 & 18.01 \\ 
DAAE\cite{shen2020educating}  & 13.16 & 15.42 & 14.30 & 13.97 \\
$\beta$-VAE\cite{Higgins2016betaVAELB} & 10.03 & 10.07 & 10.05 & 10.05 \\
Optimus(768) & 28.21 & 29.35 & 28.35 & 28.27  \\ 
T5 bottleneck(768) & \textcolor{black}{\textbf{34.47}} & \textcolor{black}{\textbf{35.28}} & \textcolor{black}{\textbf{34.50}} & \textcolor{black}{\textbf{34.47}}  \\ \toprule
\end{tabular}
}
\caption{Explanatory inference retrieval task where t represents the depth of entailment tree.} \label{tab:scar}
\end{table}







\subsection{Latent Inference-type Separation}
\begin{figure}[ht!]
    \centering
    \includegraphics[width=0.99\columnwidth]{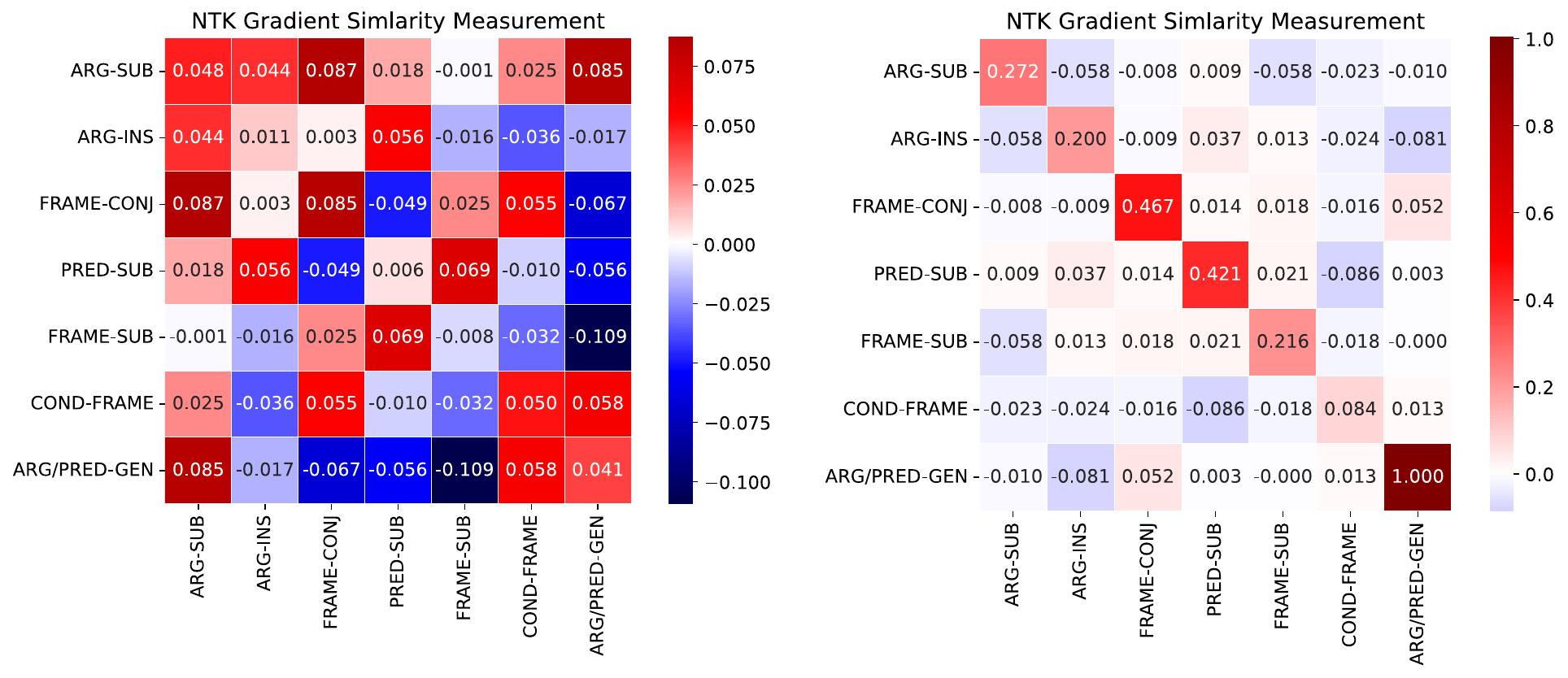}
    \caption{Quantitative measuring the separability between different inference-type subspaces in T5 (small) where left: NO, right: encoder prefix (EP).}
    \label{fig:heatmap_type}
\end{figure}

Finally, we evaluate the separability of inference-types in the latent space. For the latent parametric space, we measure the cosine similarity between gradient vectors associated with different inference types. As shown in Figure \ref{fig:heatmap_type}, we can observe that when injecting inference-type categories into the model during training, the diagonal values exhibit higher values, indicating \uline{the inference-type subspaces can be better separated in the parameter space} \textbf{(Finding~7)}.

\begin{figure}[ht!]
    \centering
    \includegraphics[width=0.99\columnwidth]{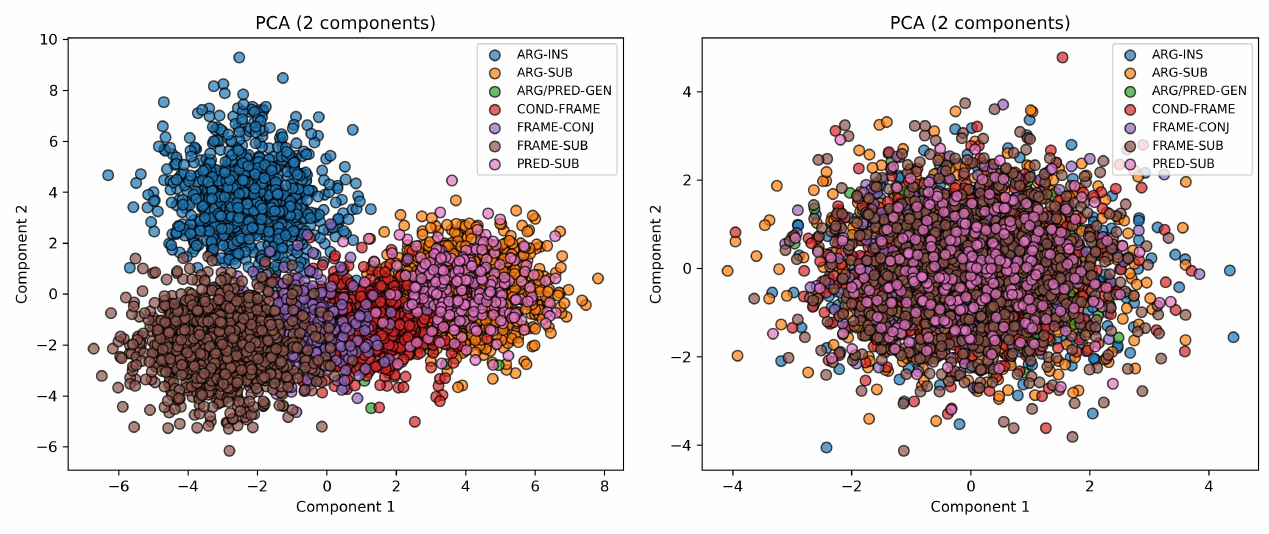}
    \caption{PCA visualisation: inference types cluster and separation, where left: EP, right: NO.}
    \label{fig:vis}
\end{figure}

Next, we evaluate whether inference rules exhibit separability within the latent sentence space. We jointly train the latent space with a linear classifier to predict the inference type categories. As shown in Figure \ref{fig:vis}, our results indicate that \uline{inference types can be partially clustered and separated within this latent space, suggesting the feasibility of rule-based learning through appropriate optimisation strategies \cite{xie2025logicrlunleashingllmreasoning} or architectural designs, such as disentangling rules from lexical semantics} \textbf{(Finding~8)}.
\section{Related Work} \label{sec:related}
In this section, we review the related work around two topics: \textit{neuro-symbolic representations} and \textit{semantic control over latent spaces}, to highlight the current research limitation and elucidate the motivation underlying our work.

\paragraph{Neuro-symbolic representations.} A longstanding goal in NLP is to blend the representational strengths of neural networks with the interpretability of symbolic systems to build more robust NLI models. Current methods usually inject symbolic behaviour through explicit symbolic representations, including graph \cite{khashabi2018question,khot2017answering,jansen2017framing,kalouli-etal-2020-hy,thayaparan2021explainable}, linear programming \cite{valentino2022case,thayaparan2024differentiable}, adopting iterative methods, using sparse encoding mechanisms \cite{valentino2020explainable,lin2020differentiable}, synthetic quasi-natural language expression \cite{clark2020transformers,yang2021learning,yanaka-etal-2021-sygns,fu2024exploring,weir-etal-2024-enhancing}, symbolic-refined LLMs \cite{olausson-etal-2023-linc,quan2024verification}, etc. 
Those studies ignore the underlying neuro-symbolic behaviour in neural representation space. 

From an Explainable AI perspective, many studies have shown that neural networks can encode sparse neural-symbolic concepts without explicit symbolic injection across areas like image embedding \cite{ren2022towards,deng2021discovering,li2023does}, word embedding \cite{ethayarajh2018towards,allen2019vec,ri2023contrastive}, contextual embedding \cite{gurnee2023finding,nanda-etal-2023-emergent,li2024inference}, and LLM interpretation \cite{pmlr-v235-park24c,templeton2024scaling}. By understanding the symbolic behaviour within neural networks, their decision-making logic can be better interpreted and controlled \citep{chen2024alignment}.
In this work, we draw on quasi-symbolic NLI objectives within distributional neural models, targetting better controllability and interpretability.

\paragraph{Semantic control over latent spaces.} Latent variable models, such as VAE \cite{kingma2013auto} and Diffusion \cite{dhariwal2021diffusion}, have shown the capability of symbolic representation, control, and interpretation over the distributional space, which are widely deployed in the NLP domain, such as disentangled representation learning \cite{zhang2023learning} and style-transfer \cite{liu-etal-2023-composable,gu-etal-2023-controllable,zhang-etal-2024-improving}. Guided by semantic annotation, such as labels \cite{carvalho2023learning} and classifiers \citep{ho2022classifier}, distinct semantic features can be geometrically separated and composed in the latent space, enhancing localisation and interpretability. 
However, this concept remains under-explored in the NLI domain. Thus, we propose the quasi-symbolic NLI representation conceptual framework and inference types as an initial step to probe the localised, quasi-symbolic NLI behaviour.

\section{Conclusion} \label{sec:concl}
This study serves as a foundational step in exploring the quasi-symbolic NLI behaviour within distributional semantic spaces. We establish the connection between natural and symbolic language inferences by characterising quasi-symbolic inference behaviours based on AMR graphs. Then, we propose the quasi-symbolic NLI representation framework. Experimental results reveal that integrating symbolic inference types enhances training dynamics, inference precision, and explanation retrieval, suggesting the potential for neuro-symbolic NLI. These results answer the \questionE{}

\paragraph{Reproducibility.} The data is available online: \url{https://drive.google.com/drive/folders/1l7Vx-nfDBWX0CVUC8YkUF7bdz6NutcCX?usp=sharing}

\section{Scoping and Limitations}


In the domain of LLM automatic evaluation, the prevailing strategy is to select the most advanced LLM as the automatic evaluator \citep{chiang-lee-2023-large, liu-etal-2023-g, wang-etal-2023-chatgpt, Huang_2023}. We perform a quantitative analysis of the inference consistency in the deductive reasoning process of LLMs, such as ChatGPT-4o. However, this assessment may be unreliable due to the inherent limitations of LLMs in logical reasoning. Human evaluation presents a potential alternative, yet the rigorous design of a protocol to systematically verify the logicality of NLI remains an under-explored area in this field. Although we perform a qualitative manual check for LLM evaluation, this assessment is not systematic or rigorously structured. 

Furthermore, this work hypothesises that the parametric space can be separated from the perspective of NTK theory, and we evaluate this hypothesis by measuring gradient vector similarity. However, NTK theory relies on the assumption of infinitely wide neural networks. This discrepancy between theoretical assumptions and practical model architectures may limit the reliability of the proposed framework.

\chapter{Explanatory Inference Disentanglement} \label{cha:reasondis}

Chapter~\ref{cha:reason} illustrated that the annotated inference types can help model training and deliver inference control. To learn the reasoning behaviours in the latent space, this chapter investigates \questionF{}
\section{Introduction}







\begin{figure*}[t]
    \centering
    \includegraphics[width=\linewidth]{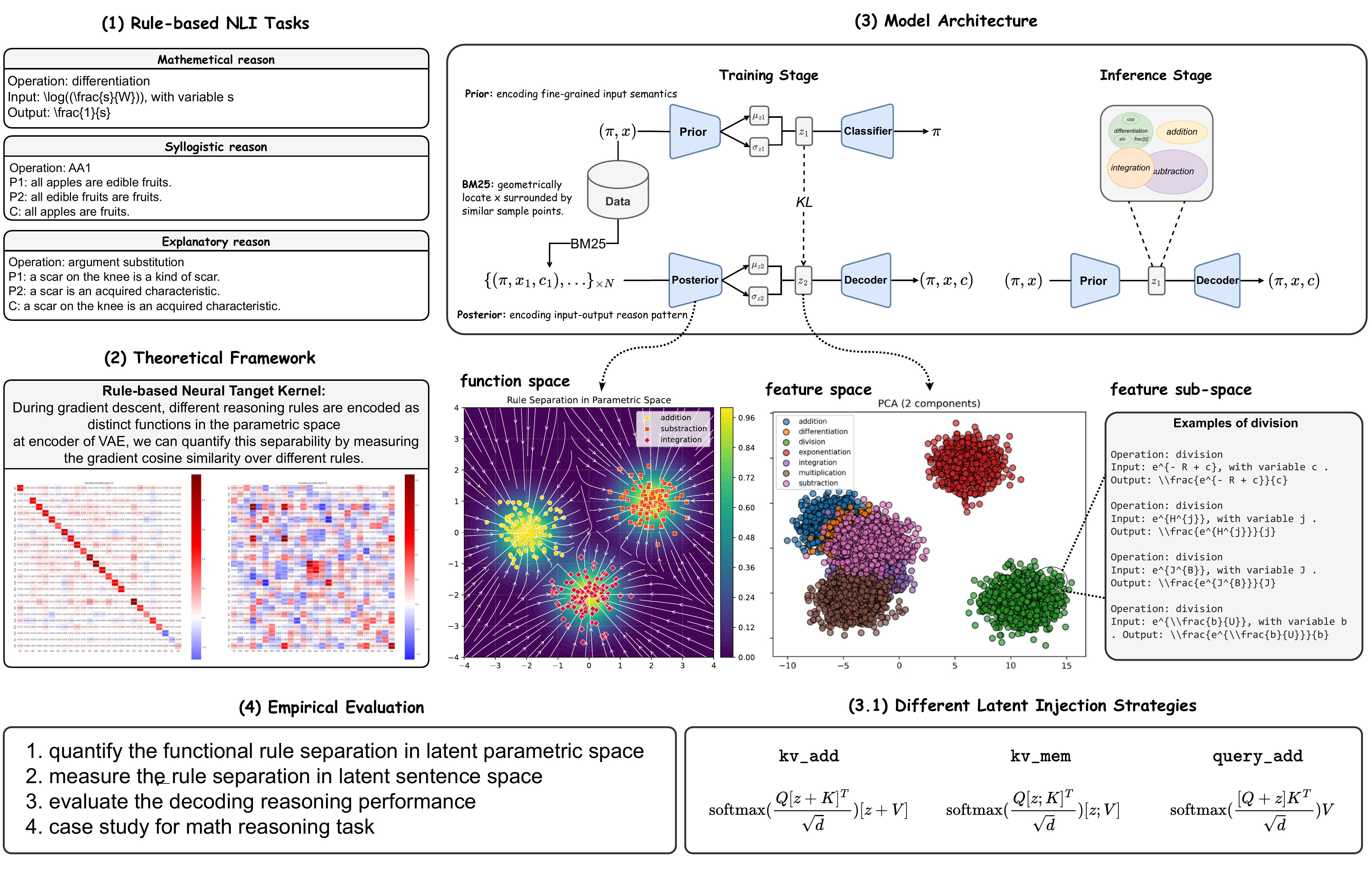}
    \caption{Overview, where $(\pi, x, c)$ represents the \textit{(rule, input premise(s), conclusion)}. To systematically evaluate rule-based learning within a VAE framework, first, we examine three rule-based NLI tasks. Second, we formalise the hypothesis that reasoning rules can be functionally and separately encoded within the encoder’s parametric space, enabling rule learning in the latent space, grounded in the theoretical framework of neural tangent kernels. Third, we introduce an end-to-end VAE architecture, with three different latent injection setups, designed to capture coarse-grained reasoning patterns in its latent space while remaining sensitive to the lexical semantics of the input.}
    \label{fig:method}
\end{figure*}

Encoding reasoning patterns as explicit rules within latent representations holds significant promise for enhancing the generalisation, interpretability, and controllability of neural models, with high downstream impact on AI safety and regulatory compliance \cite{bonnet2024searchinglatentprogramspaces,yu2025grammarbasedordinarydifferentialequation}. Over recent years, Transformer-based language models (TLMs) have achieved notable success across a variety of Natural Language Inference (NLI) tasks \cite{yang2024qwen2technicalreport,qwen2025qwen25technicalreport}. Nonetheless, a growing body of research has demonstrated that in many instances these models often rely on memorisation rather than generalisation and that rule-based control mechanisms fail to be fully enforced \cite{yan2025phdlevelllmstrulygrasp}.

Therefore, this investigation focuses on the question: \textit{How can reasoning rules be explicitly encoded within the latent space of language models?} 
In the context of this work, a reasoning rule is defined as an input–output pattern that reflects a specific transformation of inference behaviour. It is important to note that this study does not aim to explore rule composition or generalisation. Rather, the primary objective is to explicitly encode reasoning input–output constraints within the model’s latent space, targeting better interpretability and controllability in the latent space.

Variational Autoencoders (VAEs) \cite{kingma2013auto} provide a compelling framework for this direction, where the integration of prior distribution serves as an inductive bias, enabling the model to leverage existing knowledge and providing a principled way to incorporate domain constraints \cite{papamarkou2024position}. This work focuses on lightweight LMs ($<1B$) as VAE decoders, allowing accessible evaluation of training dynamics, memory capacity, and model updates of the Transformer architecture \cite{zhong2025understanding,morris2025languagemodelsmemorize}. Additionally, because our primary motivation is to study latent rule learning within the encoder latent space, we fully re-train the decoder. As a result, the geometrical properties of the latent space are not constrained by the model’s architecture, enabling a more  controlled analysis of representation learning.

Accordingly, we propose a complete pipeline for learning NLI reasoning rules within Transformer-based language VAEs \cite{li2020optimus,zhang-etal-2024-graph}: 

\textbf{First,} we focus on three distinct rule-based reasoning tasks, each characterised by unique syntax, inference patterns, and levels of granularity: Mathematical Reasoning \cite{meadows2023symbolic}, Syllogistic Reasoning \cite{valentino2025mitigatingcontenteffectsreasoning}, and Explanatory Reasoning \cite{zhang2024controllablenaturallanguageinference}.

\textbf{Second,} following the Neural Tangent Kernel (NTK) theory \cite{jacot2018ntk}, we claim that by supervising the inference rule information, inference rules can be encoded in the parametric space (i.e., weight matrix) of the encoder, which subsequently leads to the rule separation in the latent sentence space. This formalisation allows us to explicitly encode reasoning rules in the latent space.

\textbf{Third,} we propose an end-to-end VAE architecture where the latent space can encode both coarse-grained reasoning rules and fine-grained reasoning patterns that arise from semantic differences in the inputs, and evaluate three ways to inject the latent space into the LM decoder by intervening in the attention network.

We conduct extensive experiments to assess the effectiveness of rule encoding and reasoning generation capabilities which elicits the following results and findings:

\textbf{Disentangled Reasoning:} Under explicit signal supervision, reasoning rules, as functional mappings, can be disentangled within the encoder’s parametric space. This separation results in distinct clustering of rules in the output feature space, suggesting the potential for employing rule-based NTK theory to better understand the training dynamics and internal geometry of the gradient-based neural NLI model.

\textbf{Prior Knowledge Injection:} Different latent space injection setups result in varying levels of reasoning performance. The optimal configuration is achieved by injecting the latent space into the Query of the attention network. Intuitively, injecting rule-based reasoning constraints into the Query enables the model to more effectively retrieve the stored Value from memory based on Key. This approach offers a simple method for integrating prior knowledge into LMs.

\textbf{Information Bottleneck:} A case study on the mathematical reasoning task reveals a performance bottleneck in decoder-only LMs, where increasing the number of samples per operation fails to yield performance improvements beyond a certain threshold. Additionally, we observe that FFN networks are more effective than attention in maintaining the separation of reasoning rules within the learned parametric space. Both provide additional insights on understanding Transformer architectures through memorisation.

To the best of our knowledge, this is the first study to explore the explicit encoding of NLI rules within language VAE latent spaces, targeting better latent space geometry to support rule-based inferences.

\section{Rule-based Learning for NLI} \label{sec:rule_def}
\subsection{Natural Language Inference Rules}
In this study, we investigate three distinct types of reasoning tasks: mathematical derivation \cite{meadows-etal-2024-symbolic}, syllogistic reasoning \cite{valentino2025mitigatingcontenteffectsreasoning}, and explanatory reasoning \cite{zhang2024controllablenaturallanguageinference}. By spanning a range of reasoning granularities, these tasks facilitate a systematic evaluation. 

\paragraph{Mathematical reasoning.} Mathematical expressions \cite{valentino2023multioperational,meadows2023symbolic} follow a well-defined syntactic structure and set of symbolic rules that are notoriously difficult for neural models. The dataset \cite{meadows2023symbolic} includes seven human-annotated symbolic rules, encompassing operations such as \textit{differentiation} and \textit{integration}.

\paragraph{Syllogistic reasoning.} Syllogistic reasoning involves classical categorical logic, including four standard forms: Universal Affirmative (A) — “All A are B,” Universal Negative (E) — “No A are B,” Particular Affirmative (I) — “Some A are B,” and Particular Negative (O) — “Some A are not B.” The dataset encodes 24 valid syllogistic inference patterns, such as AA1 (Barbara).

\paragraph{Explanatory reasoning.} Explanatory sentences \cite{jansen2018worldtree,valentino2022hybrid} provide a semantically challenging yet sufficiently well-scoped scenario to evaluate the syntactic and semantic organisation of the space. Based on the EntailmentBank corpus \cite{dalvi2021explaining}, \citet{zhang2024controllablenaturallanguageinference} defined ten inference types based on the transformation over Abstract Meaning Representation (AMR) graph \cite{banarescu2013abstract}.

\subsection{Rule-based Neural Tangent Kernel} \label{sec:ntk}

This work aims to encode latent reasoning rules within a low-dimensional latent space to guide the LM decoder’s reasoning generation, leveraging the VAE framework. We frame rule-based learning as learning the transformation from input premise(s) to output conclusion within the encoder. We formalise this framework by introducing a novel method grounded in the \textit{Neural Tangent Kernel (NTK) theory} \cite{jacot2018ntk}.

\paragraph{Latent subspace separation.}
Let $\mathcal{M}$ be a VAE model parameterised by $\theta = (\theta_{\text{enc}}, \theta_{\text{dec}})$. Suppose the encoder can represent a set of symbolic inference rules $\Pi = \{\pi_1, \pi_2, \dots, \pi_n\}$. Then, under supervised training with rule annotations:

\textit{\textbf{Proposition:} the encoder's parameters $\theta_{\text{enc}}$ induce a parametric structure in which each inference rule $\pi_i$ corresponds to a distinct subspace $S_{\pi_i} \subseteq \mathbb{R}^D$ of the encoder representation space.}
\begin{equation}
    \forall \pi_i, \pi_j \in \Pi,\, \pi_i \neq \pi_j \quad \Rightarrow \quad S_{\pi_i} \cap S_{\pi_j} \approx \emptyset
\end{equation}

This parametric separation (Figure~\ref{fig:heatmap}) directly leads to clearly delineated latent feature subspaces (Figure~\ref{fig:pca_main}), each uniquely encoding a symbolic inference rule.

\paragraph{Connection to NTK theory.}
NTK theory provides a rigorous theoretical framework for understanding neural network training dynamics by examining the kernel induced by gradients of the network's outputs with respect to its parameters. It suggests that during gradient descent, the network effectively learns linear approximations within distinct functional subspaces, each approximating discrete symbolic reasoning rules within the target reasoning task. Formally, let $f_{\text{encode}}$ be the encoder function such that:
\begin{equation}
    f_{\text{encode}}: (\pi, x, c) \mapsto \mathbf{z}_{(\pi, x, c)} \in \mathcal{Z} \subseteq \mathbb{R}^D
\end{equation}

where $x$ is the input premise(s), c is the conclusion, $\pi \in \Pi$ is the reasoning rule, and $\mathcal{Z}$ is the latent representation space of dimension $D$. 
Each rule $\pi$ is explicitly embedded as part of the model input. As a result, the model effectively learns a function $f_\theta(\pi, x, c)$. The function $f_\theta$ thus jointly depends on both the content of the transformation from premise(s) to conclusion and the nature of the symbolic operation to be performed.
Within the NTK framework, the similarity between two input examples of the same inference type $\pi$ is captured by the NTK as follows:
\begin{equation}
\Theta_{\pi}(x, x') = \nabla_\theta f_\theta(x, \pi)^\top \nabla_\theta f_\theta(x', \pi)
\end{equation}
\noindent where $x$ represents $(x,c)$ pair for concision. $\nabla_\theta f_\theta(x, \pi)$ denotes the gradient of the model output with respect to its parameters, evaluated at the input $(x, \pi)$. This kernel quantifies how a parameter update from one input-output pair would affect another pair, conditioned on the shared rule.

According to the NTK theory, the evolution of the model’s predictions under gradient descent training can be described by a linear kernel regression in the RKHS (Reproducing Kernel Hilbert Space) associated with $\Theta_{\pi}$. Specifically, the prediction at time $t$, $f_t(x, \pi)$, evolves as:
\begin{equation}
f_t(x, \pi) = f_0(x, \pi) - \Theta_{\pi}(x, \cdot) \left[\Theta_{\pi} + \lambda I\right]^{-1} (f_0 - c)
\end{equation}
\noindent where $f_0(x, \pi)$ is the model's output at initialisation for each training input, $\lambda$ is a regularisation parameter, and $c$ is the vector of ground truth conclusions.

Crucially, this formulation implies that each rule $\pi$ induces a distinct kernel $\Theta_{\pi}$, which in turn defines a unique RKHS $\mathcal{H}_{\pi}$, that is, a function space within which the model's solutions for rule $\pi$ reside. As the $\pi$ is varied, the structure of the kernel and the corresponding function space changes, reflecting the distinct reasoning behaviours associated with different inference operations. Thus, the model encodes different symbolic inference patterns in distinct, kernel-induced subspaces.


For two different rules, $\pi_i \neq \pi_j$, we examine the relationship between their corresponding NTKs, $\Theta_{\pi_i}$ and $\Theta_{\pi_j}$. Specifically, we are interested in the interaction between the parameter gradients induced by inputs associated with different inference types. Given two data points $x$ and $x'$, possibly corresponding to different premise pairs, the NTK entry for each rule is:
\begin{equation}
\Theta_{\pi_i}(x, x') = \nabla_\theta f_\theta(x, \pi_i)^\top \nabla_\theta f_\theta(x', \pi_i)
\end{equation}
and
\begin{equation}
\Theta_{\pi_j}(x, x') = \nabla_\theta f_\theta(x, \pi_j)^\top \nabla_\theta f_\theta(x', \pi_j)
\end{equation}
When considering cross-rule similarities, we are interested in the inner product between the gradients for different rules.
\begin{equation}
G_{ij}(x, x') := \langle \nabla_\theta f_\theta(x, \pi_i), \nabla_\theta f_\theta(x', \pi_j) \rangle
\end{equation}
If the rules $\pi_i$ and $\pi_j$ encode fundamentally different reasoning operations (e.g., \textit{addition} vs. \textit{subtraction} in Math Derivation), the gradients with respect to $\theta$ for inputs labeled with $\pi_i$ and those labeled with $\pi_j$ will tend to point in different directions in parameter space of the encoder.

Under idealised training, where the data for each rule is sufficiently distinct and the network has enough capacity, the gradients for one rule will have minimal overlap with those of the other. This can be formalised by observing that:
\begin{equation} \label{eq9}
\langle \nabla_\theta f_\theta(x, \pi_i), \nabla_\theta f_\theta(x', \pi_j) \rangle \approx 0 \qquad \text{for}~ \pi_i \neq \pi_j
\end{equation}
This property implies that the parameter updates driven by examples from different rules are approximately orthogonal, meaning that training on one type will not interfere with or alter the function learned for the other type. In the language of NTK and kernel regression, this corresponds to the induced RKHS for each type, $\mathcal{H}{\pi_i}$ and $\mathcal{H}{\pi_j}$, being approximately disjoint:
\begin{equation}
\mathcal{H}{\pi_i} \cap \mathcal{H}{\pi_j} \approx \emptyset
\end{equation}

In the next section, we will use this foundation to introduce the proposed VAE architecture and its supporting optimisation function.


\section{Approach} \label{sec:latent_props}
\paragraph{Latent space properties.} We posit that the latent space should satisfy two essential geometrical properties:

\textit{Property 1: Rule-level encoding.} The latent space must capture the transformation defined by the reasoning rule $\pi$, which maps an input $x$ to a conclusion $c$, i.e., $\pi: x \rightarrow c$.

\textit{Property 2: Semantic-level encoding.} The latent space should also encode the lexical semantics of the $x$, accounting for fine-grained variations in reasoning patterns that arise from semantic differences in the inputs. For example, within the Math Derivation task, under \textit{differentiation}, different inputs yield distinct transformations:

$\pi: x_1 \rightarrow c_1$: 8 \textbackslash sin\{(u)\} $\rightarrow$ 8 \textbackslash cos\{\(u\)\}

$\pi: x_2 \rightarrow c_2$: \textbackslash log\{\(K\)\} $\rightarrow$ \textbackslash frac\{1\}\{K\}

By ensuring both properties, the latent space is capable of capturing both coarse-grained and fine-grained reasoning patterns.

\paragraph{Architecture.} We adopt a Transformer-based VAE framework, employing two distinct encoders, both instantiated with BERT~\cite{devlin2019bert}, to model the prior and posterior Gaussian distributions. The posterior encoder is optimised to capture the reasoning rule transformation, thereby satisfying \textit{Property 1}. Simultaneously, the prior encoder serves as a regulariser, encouraging the posterior to encode fine-grained sub-rule variations grounded in the lexical semantics of the input, thereby satisfying \textit{Property 2}. 

Concretely, given a target $(\pi, x)$ pair, we first retrieve a set of semantic similar examples $\{(\pi, x_1, c_1), \dots, (\pi, x_N, c_N)\}$ using BM25 retrieval (N is 12 during the experiment). These examples are input to the posterior encoder, which learns a latent representation of the reasoning transformation by averaging the latent vectors of the retrieved instances. Geometrically, averaging the latent sample vectors positions the target $x$ at the centroid of the surrounding samples within the latent space, which is naturally aligned with how information is encoded in the latent space \cite{zhang-etal-2024-learning,zhang2024formalsemanticgeometrytransformerbased}.

The decoder then utilises this aggregated representation to generate the corresponding conclusion $c$ for the target triplet $(\pi, x, c)$. In parallel, the prior encoder processes the target $(\pi, x)$ pair to predict the rule $\pi$ within the latent space via a linear classifier. By minimising the Kullback–Leibler (KL) divergence between the posterior and prior distributions, the latent space is encouraged to satisfy both geometric properties.

\paragraph{Latent injection.} We adopt three strategies to inject the latent variable $z$ into the decoder (e.g., Qwen2.5~\cite{qwen2025qwen25technicalreport}). Each approach modifies the attention mechanism to incorporate information from the latent space.

\paragraph{1. \texttt{kv\_add}} Following~\cite{zhang-etal-2024-graph}, the latent vector $z$ is added to both the Key ($K$) and Value ($V$) matrices in the attention network:
\begin{equation}
\text{softmax}\left(\frac{\text{Q}[\text{z} + \text{K}]^\top}{\sqrt{d}}\right)[\text{z} + \text{V}]
\end{equation}

\paragraph{2. \texttt{kv\_mem}} Following~\cite{li2020optimus}, the latent vector $z$ is concatenated to the Key and Value matrices:
\begin{equation}
\text{softmax}\left(\frac{\text{Q}[\text{z}; \text{K}]^\top}{\sqrt{d}}\right)[\text{z}; \text{V}]
\end{equation}

\paragraph{3. \texttt{query\_add}} In this setup, the latent vector $z$ is added to the Query ($Q$) matrix:
\begin{equation}
\text{softmax}\left(\frac{[\text{Q} + \text{z}] \text{K}^\top}{\sqrt{d}}\right)\text{V}
\end{equation}

Here, $Q$, $K$, and $V$ denote the query, key, and value matrices in the attention mechanism, each with dimensions $\mathbb{R}^{\text{dim} \times \text{seq}}$, where $\text{dim}$ is the attention dimensionality and $\text{seq}$ is the sequence length.

\paragraph{Optimisation.} Finally, the model can be trained end-to-end via the evidence lower bound (ELBO) on the log-likelihood of the data $x$ \cite{kingma2013auto}. To avoid the KL vanishing issue, we select the cyclical schedule to increase weights of KL $\beta$ from 0 to 1 \cite{fu-etal-2019-cyclical} and a KL thresholding scheme \cite{li-etal-2019-surprisingly} that chooses the maximum between KL and threshold $\lambda$. The final objective function can be described as follows:
\begin{equation}
\begin{aligned} \label{eq:elbo_loss}
\mathcal{L}_\text{VAE} = & \mathbb{E}_{q_\phi(z|\pi,x,c)} \Big[ \log p_{\theta} (\pi,x,c | z ) \Big]  \\
& - \beta \max \left[ \lambda , \text{KL} q_\phi(z|\pi,x,c) || p(z|\pi,x) \right ] \\
& + \texttt{cls\_weight} \times \mathcal{L}_\text{classifier}
\end{aligned}
\end{equation}
Where $q_\phi$, $p$, and $p_{\theta}$ represent the posterior encoder, prior encoder, and decoder, respectively. \texttt{cls\_weight} controls the strength of the classification loss. In our experiments, \texttt{cls\_weight} is evaluated at 1.0, 0.5, and 0.1.
\section{Empirical Analysis} \label{sec:empirical}
\subsection{Decoding Evaluation}
\paragraph{Base model.} First, we evaluate the reasoning performance of different decoder-only models, including Qwen2 \cite{yang2024qwen2technicalreport}, Qwen2.5 \cite{qwen2025qwen25technicalreport}, GPT2 \cite{radford2019language}, and trimmed Llama3 \cite{grattafiori2024llama3herdmodels}\footnote{https://huggingface.co/andrijdavid/Llama3-1B-Base}. Due to the scale of the dataset and limitations in computational resources, we restrict our fine-tuning to smaller models with fewer than 1 billion parameters. During training, the input format is described as:
$\text{operation:}~\pi,~\text{premise:}~x,~\text{conclusion:}~c$
During inference, we omit $c$, allowing the model to generate the conclusion. To evaluate generation quality, we employ both the BLEU score \cite{Papineni02bleu:a} and accuracy (acc) as performance metrics. For the Math Reasoning task, in addition to the in-distribution test set, we also assess model performance on an out-of-distribution (OOD) test set, where the mathematical expressions are composed using a different set of variables.
\begin{table*}[ht!]
\centering
\renewcommand{\arraystretch}{1}
\resizebox{\linewidth}{!}{
\begin{tabular}{cccccccc}
\toprule
\textbf{Base Model} &
\multicolumn{2}{c}{\textbf{Math Derivations}} & 
\multicolumn{2}{c}{\textbf{Math Derivations (OOD)}} &
\multicolumn{2}{c}{\textbf{Syllogistic Reasoning}} &
\textbf{Explanatory Reasoning} \\
&
bleu & acc & bleu & acc & bleu & acc & bleu \\
\midrule
GPT2-medium & 0.2019 & 0.0171 & 0.0206 & 0.0018 & 0.0108 & 0.0000 & 0.5947 \\
Llama3-1B & 0.3412 & 0.0257 & 0.0625 & 0.0114 & 0.3612 & 0.1200 & 0.3160 \\
Qwen2-0.5B & 0.4200 & 0.1171 & 0.0869 & 0.0128 & 0.8130 & 0.4600 & 0.5372 \\
Qwen2.5-0.5B & \textbf{\textcolor{black}{0.6260}} & \textbf{\textcolor{black}{0.3800}} & \textbf{\textcolor{black}{0.1293}} & \textbf{\textcolor{black}{0.0185}} & \textbf{\textcolor{black}{0.9014}} & \textbf{\textcolor{black}{0.7000}} & \textbf{\textcolor{black}{0.6566}} \\
\bottomrule
\end{tabular}
}
\caption{Quantitative evaluation for decoder-only models. We can observe that Qwen2.5-0.5B demonstrates strong performance across various baselines, making it a suitable choice as the decoder component in our VAE architecture.} \label{tab:decode_lm}
\end{table*}

As shown in Table \ref{tab:decode_lm}, Qwen2.5 demonstrates consistently strong performance across all tasks. Based on these results, we select Qwen2.5 as the decoder model for subsequent experiments.

\paragraph{VAE model.} Next, we evaluate the performance of VAE on downstream reasoning tasks, where the decoder is Qwen2.5-0.5B, the latent dimension is 32 following the same setup as Optimus \cite{li2020optimus}. In addition, we include results from the base model trained via fine-tuning (denoted as FT) as well as inference using few-shot examples. The input format remains consistent with previous settings, with examples repeated accordingly. For few-shot selection, we employ BM25 to retrieve the most relevant samples.
\begin{table*}[htbp]
\centering
\resizebox{\linewidth}{!}{
\begin{tabular}{ccccccccc}
\toprule
\textbf{Injection} & \textbf{Train Setup} &
\multicolumn{2}{c}{\textbf{Math Reason}} & 
\multicolumn{2}{c}{\textbf{Math Reason (OOD)}} &
\multicolumn{2}{c}{\textbf{Syllogistic Reason}} &
\textbf{Explanatory Reason} \\
& &
bleu & acc & bleu & acc & bleu & acc & bleu \\
\midrule

\multirow{3}{*}{Base model}
    & Zero-shot & 0.6260 & 0.3800 & \textbf{0.1293} & 0.0185 & 0.9014 & 0.7000 & \underline{\textbf{0.6566}} \\
    & Few-shot & 0.1596 & 0.0614 & 0.1011 & 0.0057 & 0.1534 & 0.0000 & 0.0765 \\
    & Few-shot (FT) & 0.1601 & 0.0557 & 0.1022 & 0.0028 & 0.1712 & 0.0000 & 0.0801 \\
    \midrule
    \midrule
\multirow{3}{*}{kv\_add} 
    & prior=False & \textcolor{black}{\textbf{0.7285}} & \textbf{0.5285} & 0.1055 & 0.0200 & 0.4839 & 0.3400 & 0.6127 \\
    & weight=1.0 & 0.4986 & 0.2128 & 0.0869 & 0.0157 & 0.0719 & 0.0000 & 0.2049 \\
    & weight=0.5 & 0.4925 & 0.2814 & 0.0504 & 0.0185 & 0.4870 & 0.0900 & 0.2623 \\
\midrule
\multirow{3}{*}{kv\_mem} 
    & prior=False & 0.6430 & 0.4185 & 0.0985 & 0.0200 & 0.4672 & 0.3300 & 0.6313 \\
    & weight=1.0 & 0.5255 & 0.2300 & 0.0956 & 0.0185 & 0.0000 & 0.0000 & 0.6022 \\
    & weight=0.5 & 0.6808 & 0.4857 & 0.1130 & 0.0185 & 0.9452 & 0.8500 & 0.5987 \\
\midrule
\multirow{3}{*}{query\_add}
    & prior=False & 0.6501 & 0.3885 & 0.1130 & \textbf{0.0200} & \textbf{\underline{0.9681}} & \textbf{\underline{0.9200}} & \textbf{0.6449} \\
    & weight=1.0 & 0.7262 & 0.5057 & 0.1223 & \underline{\textbf{0.0214}} & 0.4373 & 0.3300 & 0.6118 \\
    & weight=0.5 & \underline{\textbf{0.8130}} & \underline{\textbf{0.6642}} & \underline{\textbf{0.1441}} & 0.0185 & \textbf{0.9461} & \textbf{0.8600} & 0.6220 \\
\bottomrule
\end{tabular}
}
\caption{Quantitative evaluation, where the base model is Qwen2.5-0.5B. Top two values are highlighed in \textbf{\underline{bold}} and \textbf{\textcolor{black}{bold}}. Prior=False is the setup of VAE without trainable prior. We can observe that query\_add leads to the best performance in general.} \label{tab:decode}
\end{table*}

As illustrated in Table \ref{tab:decode}, we can observe that \uline{injecting the latent space into the query can generally result in better performance compared with other setups. Intuitively, injecting reasoning information into the Query enables the model to more effectively retrieve the stored Value from memory based on Key} \textbf{(Finding~1)}.

Furthermore, when the base model is trained using few-shot examples, its performance declines significantly. We observe that the model repeats to generate more examples. To ensure a fair comparison across methods, we did not apply any filtering to remove these redundant outputs.

Additionally, we evaluate the reasoning performance of the VAE model trained with and without BM25-based sample selection. In the absence of BM25, training samples are randomly drawn from the corpus. \uline{As presented in Table \ref{tab:decode_bm25}, integrating BM25-based selection during training yields enhanced reasoning performance, indicating the effectiveness of relevance-guided sampling. From a geometric perspective, BM25 retrieves samples that exhibit the highest lexical similarity, effectively selecting the nearest neighbours in the latent space} \textbf{(Finding~2)}.

In the following section, we evaluate the model’s capability to learn and represent latent reasoning rules within the latent space.
\begin{table}[ht]
\centering
\small
\renewcommand{\arraystretch}{1.2}
\begin{tabular}{ccccc}
\hline
\textbf{Injection} & \textbf{Prior} & \textbf{BM25} & \textbf{bleu} & \textbf{acc} \\ \hline
kv\_add & False & False & 0.3691 & 0.0614 \\
kv\_mem & False & False & 0.5525 & 0.2371 \\
query\_add & False & False & 0.5914 & 0.3021 \\ \hline
kv\_add & False & True & 0.7285 & 0.5285 \\
kv\_mem & False & True & 0.6430 & 0.4185 \\
query\_add & False & True & 0.6501 & 0.3885 \\ \hline
\end{tabular}
\caption{Quantitative evaluation for VAE model with or without BM25 in Math Reason task. We can observe that selecting the training sample using BM25 can help model performance.} \label{tab:decode_bm25}
\end{table}

\subsection{Encoding Evaluation}
\begin{figure*}[ht!]
    \centering
    \includegraphics[width=0.9\linewidth]{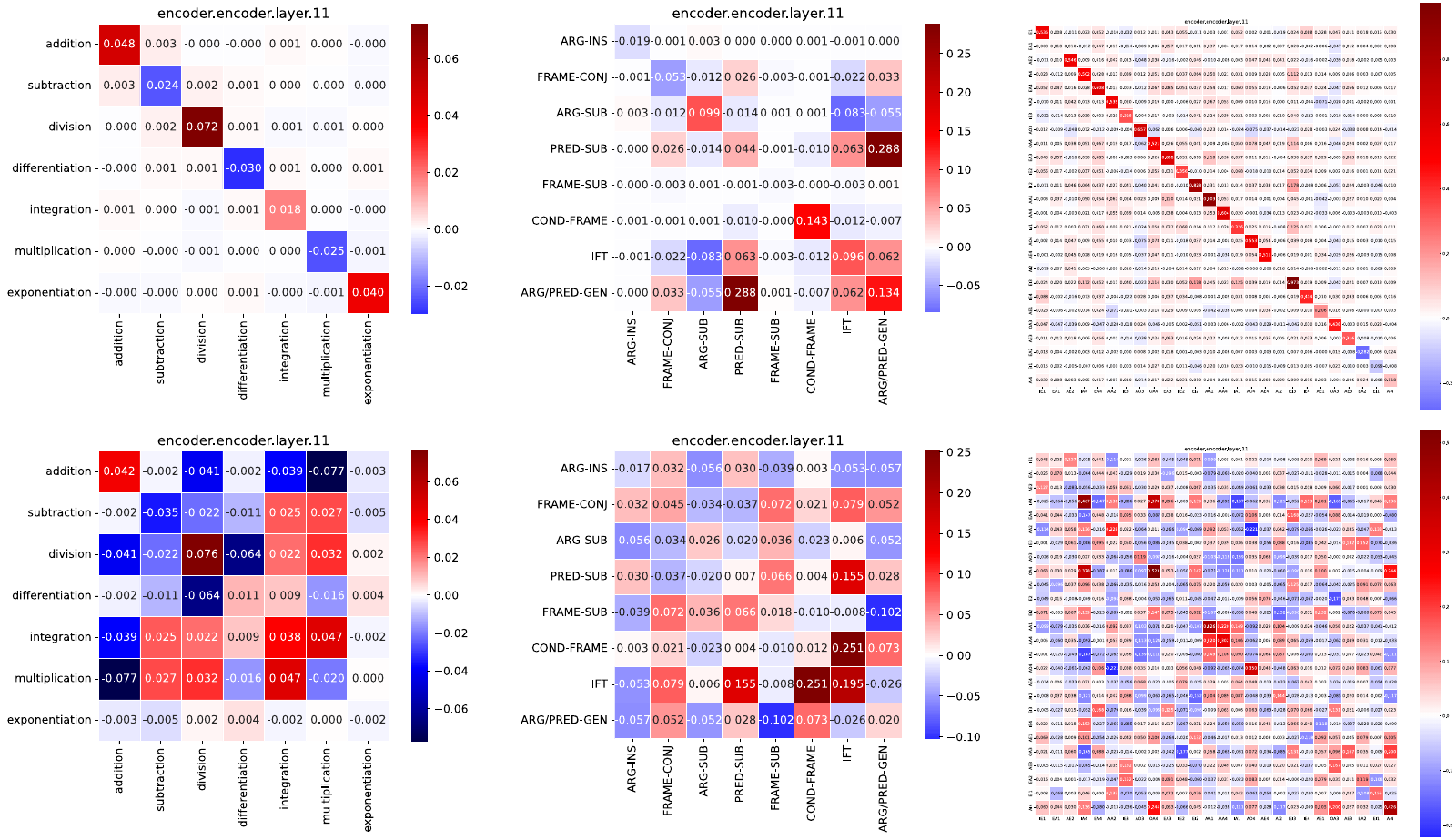}
    \caption{Gradient heatmap for the last posterior encoder layer (query\_add setup), where the left: math derivation, middle: explanatory reasoning, right: syllogistic reasoning, Top: \texttt{cls\_weight} is 1.0, bottom: \texttt{cls\_weight} is 0.1. We can observe that the non-diagonal values are notably close to 0 when providing higher \texttt{cls\_weight} (the red colour elements are less scattered), suggesting that incorporating rule information during training enhances the separation of rule subspaces in the encoder's parameter space. 
    }
    \label{fig:heatmap}
\end{figure*}

\paragraph{Latent parametric space.} As we illustrated in Section \ref{sec:ntk} (Equation \ref{eq9}), by measuring the cosine similarity between gradient vectors associated with different rules, we can quantify the separability between different rule subspaces in the encoder, comparing settings with strong \texttt{cls\_weight} = 1.0 (left) and weak \texttt{cls\_weight} =0.1 (right) during training. As illustrated in Figure \ref{fig:heatmap}, when the classification weight (\texttt{cls\_weight}) is set to 1.0, most non-diagonal values are close to zero (orthogonality). In contrast, with \texttt{cls\_weight} set to 0.1, a greater number of non-diagonal elements exhibit higher values (the red colour elements are much more scattered). This observation suggests that \uline{explicit supervision facilitates the separation of reasoning rules within the encoder’s parameter space} \textbf{(Finding~3)}.
\begin{figure*}[ht!]
    \centering
    \includegraphics[width=\linewidth]{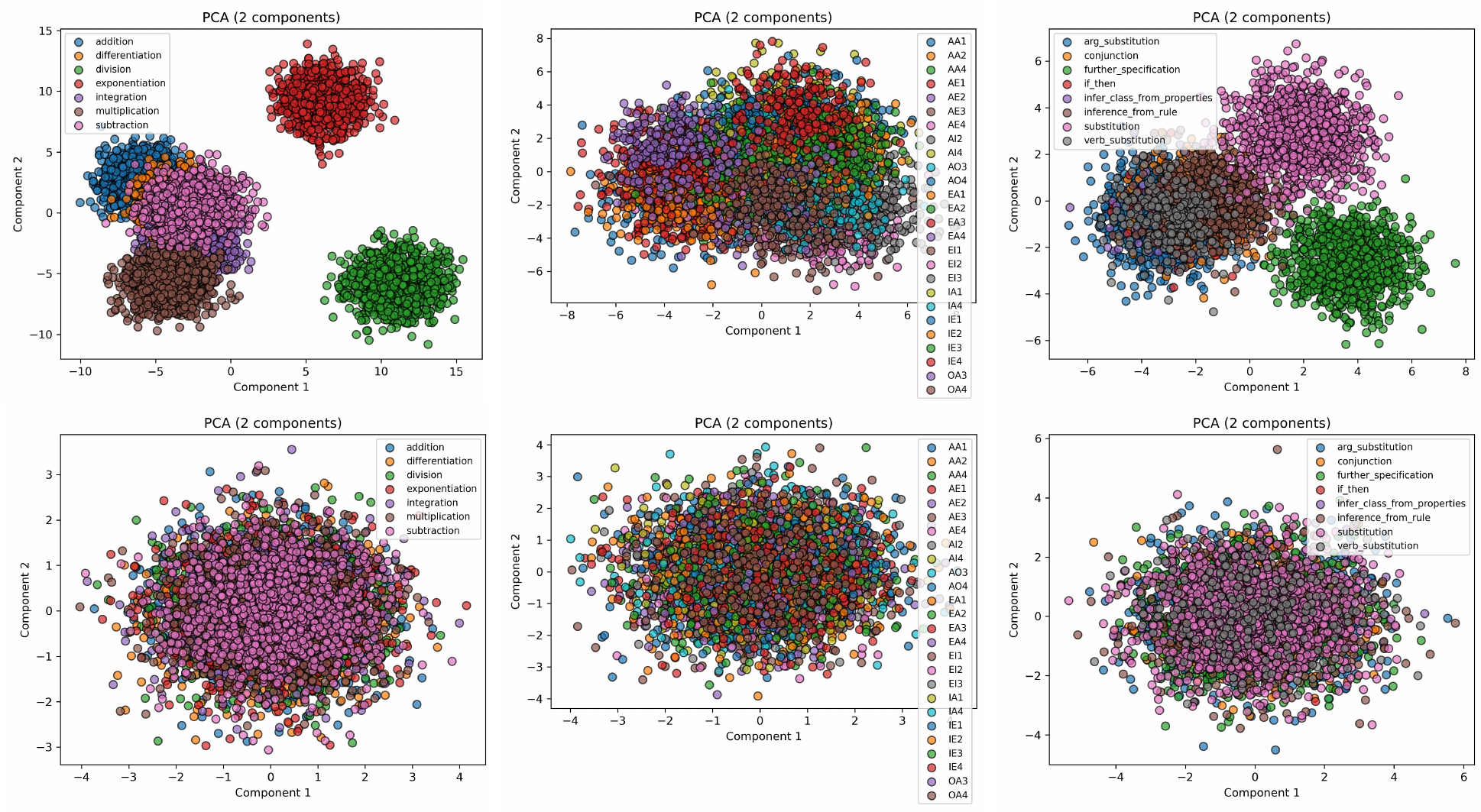}
    \caption{PCA visualisation for \texttt{query\_add} injection setup, where top: \texttt{cls\_weight} is 1.0, bottom: \texttt{cls\_weight} is 0.1. We can observe that the model struggles to learn the rules when the weight is close to zero, indicating the neural network tries to deliver reason behaviour via memorisation, rather than rule-based learning. 
    }
    \label{fig:pca_main}
\end{figure*}

\paragraph{Latent sentence space.} In Section \ref{sec:ntk}, we illustrated that rule separation in the parametric space leads to corresponding separation in the feature space. As shown in Figure \ref{fig:pca_main}, the sentence representations tend to form distinct clusters that reflect rule information when the classifier is given a higher \texttt{cls\_weight}. However, when the \texttt{cls\_weight} approaches zero, these rule-based clusters disappear. \uline{This suggests that the neural network relies more on the memorisation of lexical combinations than on rule-based learning} \textbf{(Finding~4)}.

\begin{figure*}[ht!]
    \centering
    \includegraphics[width=\linewidth]{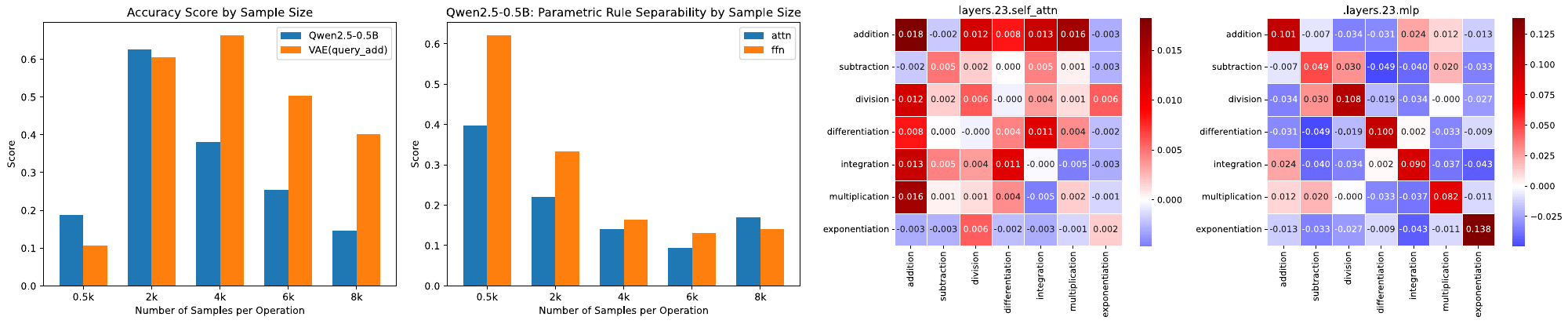}
    \caption{Case study for Math Reasoning task, where left: analysing how varying the number of training samples for each operation affects the reasoning capabilities. Right: comparing the parametric rule separation between attn and ffn at the last layer in Qwen2.5-0.5B, a pretrained checkpoint with a training sample size of 4k.}
    \label{fig:cases}
\end{figure*}

\subsection{Case Study for Math Reasoning}

\paragraph{Information bottleneck.} First, we analyse how varying the number of training samples for each operation affects the reasoning capabilities of the decoder-only LM, i.e., Qwen2.5 (0.5B). As illustrated in Figure \ref{fig:cases} (left bar plot), it can be observed that when the number of samples per category exceeds 2,000, there is a noticeable decline in accuracy. This suggests that increasing the sample size may introduce greater variability or complexity, potentially disrupting the consistency of each operation. \uline{The observation also highlights a limitation of current autoregressive LMs: rather than engaging in rule-based reasoning, they tend to rely on retrieving memorised training instances embedded in their parameters \cite{zhong2025understanding}. Explicitly injecting latent reasoning representations can help mitigate this issue} \textbf{(Finding~5)}.

\paragraph{Parametric rule separation.} Second, we assess the parametric separation of rule-based information across different model components, specifically the attention (attn) and feed-forward network (ffn) layers, using the same methodology outlined previously. We report the sum of the average diagonal values across all layers. \uline{As illustrated in Figure \ref{fig:cases} (right bar plot), the ffn layers (right) exhibit a greater tendency to encode rule-based information compared to the attention layers, indicating a more prominent role in capturing structured reasoning patterns} \textbf{(Finding~6)}.

\section{Related Work} \label{sec:related}
In this section, we review the related work around two topics: \textit{rule-based representation learning} and \textit{language VAEs}, to highlight the current research limitation and elucidate the motivation underlying our work.

\paragraph{Rule-based Representation Learning.}
Encoder-only models have been employed for tasks such as mathematical operations, where the encoder learns structured transformations, as demonstrated by \cite{valentino2024multioperationalmathematicalderivationslatent}. In addition, VAE-based approaches have shown promise in tasks requiring structured reasoning, such as visual program synthesis, where the goal is to generate programs that fulfill a specified task \cite{pmlr-v80-sun18a, bonnet2024searchinglatentprogramspaces, vankrieken2025neurosymbolicdiffusionmodels}. Similarly, grammar-based approaches using VAEs have been applied to infer ordinary differential equation (ODE) formulas from data \cite{yu2025grammarbasedordinarydifferentialequation}. Moreover, Decoder-only models, particularly large language models (LLMs), leverage in-context learning by using demonstrations to infer and apply underlying reasoning patterns \cite{liu2023context,bhattamishraunderstanding}.

Despite these advancements, relatively few studies have investigated rule-based learning in the context of NLI. To address this gap, we present a foundational exploration of rule-based learning applied to the NLI task.

\paragraph{Language VAEs.} Language VAEs have been widely applied in NLP tasks, such as style transfer tasks: modifying sentences with regard to markers of sentiment, formality, affirmation/negation \cite{ shen2020educating,john2019disentangled,bao2019generating,hu2021causal,vasilakes-etal-2022-learning,gu-etal-2022-distributional,liu-etal-2023-composable,gu-etal-2023-controllable}, story generation \cite{fang2021transformerbased}, dialogue generation \cite{zhao-etal-2017-learning}, text paraphrasing \cite{bao-etal-2019-generating}, and textual, syntactic, semantic representation learning domain, such as syntax disentanglement \cite{mercatali-freitas-2021-disentangling-generative}, semantic-syntax separation \cite{zhang-etal-2024-graph}, semantic disentanglement \cite{silva-de-carvalho-etal-2023-learning,zhang-etal-2024-learning}, etc. Comparatively, we focused on the Natural Language Inference (NLI) task. To our knowledge, the application of VAEs for NLI is underexplored.
\section{Conclusion}
This study serves as a foundational step in exploring the rule-based representation learning under the language VAE architecture for NLI tasks. We propose a complete pipeline for learning reasoning rules within Transformer-based language VAEs. This pipeline encompasses three rule-based reasoning tasks, a supporting theoretical framework, and a practical end-to-end architecture. The experiment illustrates the following findings: 

\textit{Disentangled reasoning:} Under explicit signal supervision, reasoning rules, viewed as functional mappings, can be disentangled within the encoder’s parametric space. This separation results in distinct clustering of rules in the output feature space. \textit{Prior knowledge injection:} injecting reasoning information into the Query enables the model to more effectively retrieve the stored Value from memory based on the Key. This approach offers a simple method for integrating prior knowledge into decoder-only language models. \textit{Performance bottleneck:} In mathematical reasoning tasks using Qwen2.5 (0.5B), increasing sample count doesn’t improve performance beyond a point. Moreover, ffn layers are better than attention layers at preserving the separation of reasoning rules in the model's parameters. These results answer the \questionF{}

\paragraph{Reproducibility.} The data and code are available online: \url{https://github.com/SnowYJ/LatentReasonVAE}
\section{Scoping and Limitations}

\paragraph{Composition and generalisation.} A key limitation of the current approach lies in the use of Transformer-based auto-regressive decoders, which typically struggle with compositionality and generalisation. These decoders tend to memorise input-output reasoning patterns observed during training, rather than demonstrating genuine compositional reasoning capabilities. As a result, while our framework enables learning memorised reasoning behaviours within the latent space, it remains constrained in its ability to extrapolate beyond seen combinations or perform novel inference.
To address this limitation and explore compositional reasoning beyond memorisation, particularly in NLI tasks, future work can focus to emerging \textit{diffusion-based language models}, such as Mmada~\citep{yang2025mmadamultimodallargediffusion} and Gemini Diffusion. These models offer more flexible decoding mechanisms that allow for structured constraint integration and optimisation during generation~\citep{vankrieken2025neurosymbolicdiffusionmodels}. We hypothesise that such frameworks may provide a more principled foundation for enabling compositional generalisation in controlled and interpretable latent spaces.
\chapter{Conclusion}\label{cha:con}

\section{Summary and Conclusion}

In this chapter, we provide a summary of how the research questions stated in Chapter 1 have been answered across the thesis, together with the main findings and a discussion on some of the main limitations of the presented work.

\begin{itemize}[leftmargin=1.5em]
    \item \questionA
\end{itemize}

\paragraph{Contribution.} Chapter~\ref{cha:geo} introduced a theoretical framework, \textit{Formal Semantic Geometry}, designed to bridge the divide between formal (symbolic) semantics and distributional semantics. Building upon this formal-distributional framework, we propose a practical methodology for learning geometric properties within the latent space of Transformer-based variational autoencoders (VAEs) for language.

\paragraph{Findings.} The experiments in Chapter~\ref{cha:geo} demonstrate that, within the latent semantic space, sentence meaning can be decomposed into \textit{formal semantic features} of the type \textit{role--content}, with each feature represented as a convex cone. The position of a sentence vector in this space is determined by the intersection of the cones corresponding to the \textit{role--content} components it contains. This theoretical foundation enables the semantic localisation of sentences by traversing the latent space across various \textit{role--content} regions.

\paragraph{Scope and Limitations.} The geometric analysis reveals that \textit{role--content} regions exhibit significant overlap within the distributional latent space. This suggests a promising direction for future work: developing methods to better disentangle or separate semantic features in order to achieve improved localisation and compositional behaviour in distributional semantic representations.

\begin{itemize}[leftmargin=1.5em]
    \item \questionB
\end{itemize}

\paragraph{Contribution.} Building on the \textit{Formal Semantic Geometry} framework, Chapter~\ref{cha:dis} introduces a novel natural language learning task, \textit{sentence semantic disentanglement}, and proposes a flow-based model for disentangling \textit{role--content} semantic features.

\paragraph{Findings.} The experiments in Chapter~\ref{cha:dis} show that the proposed method effectively separates semantic features. This separability is evaluated through both visualisation techniques and performance on downstream classification tasks. Furthermore, the disentangled representation enables improved symbolic semantic control via interpolation. We present both quantitative and qualitative evaluations to assess the smoothness and localisation of the interpolation trajectories during sentence generation.

\paragraph{Scope and Limitations.} The \textit{Formal Semantic Geometry} framework is grounded in formal semantic theory, particularly Argument Structure Theory, which assumes sentence-level semantics can be captured through shallow semantic roles. Since semantic role labels are derived from deeper syntactic structures, further research is necessary to extend the approach to encompass a broader range of syntactic phenomena beyond shallow roles.

\begin{itemize}[leftmargin=1.5em]
    \item \questionC
\end{itemize}

\paragraph{Contribution.} Chapter~\ref{cha:syntax} presents a dual-encoder VAE framework designed to disentangle syntactic and semantic information within the latent space. In this framework, syntactic structure is captured via a graph encoder, while semantic content is encoded using a language encoder (i.e., BERT). To effectively decode these heterogeneous latent representations, we introduce an injection strategy that distributes the separated latent spaces into distinct components of the attention mechanism, thereby mitigating generation degradation.

\paragraph{Findings.} Experimental results show that: (1) the graph encoder effectively captures syntactic structure in the latent space; (2) combining graph and language encoders enhances the separation between syntax and semantics; (3) the proposed injection strategy reduces generation degradation caused by latent heterogeneity.

\paragraph{Scope and Limitations.} Despite the effectiveness of continuous latent spaces in capturing global properties, their capacity for fine-grained and localised semantic control remains limited due to the inherently discrete nature of natural language.

\begin{itemize}[leftmargin=1.5em]
    \item \questionD
\end{itemize}




\paragraph{Contribution.} To bridge discrete latent semantic spaces with distributional semantics, in Chapter~\ref{cha:discrete}, we propose a novel model to enhance semantic and inference control in VAE-based language modelling architectures. The proposed model, T5VQVAE, integrates a vector-quantised variational autoencoder (VQ-VAE) with a high-performing, pretrained language model (T5), enabling consistent access to powerful generative capabilities while maintaining symbolic-like latent control.

\paragraph{Findings.} We extensively evaluate T5VQVAE across three downstream tasks, including autoencoding, text transfer, and explanatory inference, to assess its semantic controllability in latent space. Experimental results show that T5VQVAE outperforms strong baselines, achieving improved performance and exhibiting localised, quasi-symbolic behaviour in downstream tasks.

\paragraph{Scope and Limitations.} While T5VQVAE demonstrates potential inference control, enabling systematic reasoning behaviour annotation and learning remains an open challenge. Further research is required to enable the model to learn and encode reasoning rules.

\begin{itemize}[leftmargin=1.5em]
    \item \questionE
\end{itemize}




\paragraph{Contribution.} Chapter~\ref{cha:reason} presents a foundational investigation into quasi-symbolic inference within distributional semantic spaces. The key contributions are as follows: \textbf{(1)} We establish a theoretical connection between natural language and symbolic inference from a linguistic perspective by systematically characterising quasi-symbolic inference behaviours, termed \textit{symbolic inference types}, grounded in Abstract Meaning Representation (AMR) and Argument Structure Theory (AST). \textbf{(2)} We introduce the \textit{quasi-symbolic NLI representation conceptual framework}, which bridges symbolic and distributional semantics by guiding the formation of inference behaviours in neural latent spaces using the defined symbolic inference types.

\paragraph{Findings.} Experimental results indicate that supervision based on symbolic inference types improves model training efficiency, inference quality, and semantic localisation. These findings suggest that distinct inference types may correspond to functional subspaces that are separable or disentangled within the neural parametric space. We support this hypothesis through quantitative evaluation using the Neural Tangent Kernel (NTK) theory.

\paragraph{Scope and Limitations.} The \textit{quasi-symbolic NLI representation conceptual framework} posits that reasoning behaviours are encoded in complex, high-dimensional parametric spaces. However, enabling models to learn and encode such reasoning patterns within compressed latent sentence representations remains a significant challenge for future work.

\begin{itemize}[leftmargin=1.5em]
    \item \questionF
\end{itemize}

\paragraph{Contribution.} We propose a complete pipeline for learning reasoning rules within Transformer-based language VAEs. This pipeline encompasses three rule-based reasoning tasks, a supporting theoretical framework from the perspective of the NTK theory, and a practical end-to-end architecture.

\paragraph{Findings.} The experiment illustrates the following findings: \textit{Disentangled reasoning:} Under explicit signal supervision, reasoning rules, viewed as functional mappings, can be disentangled within the encoder’s parametric space. This separation results in distinct clustering of rules in the output feature space. \textit{Prior knowledge injection:} injecting reasoning information into the Query enables the model to more effectively retrieve the stored Value from memory based on Key. This approach offers a simple method for integrating prior knowledge into decoder-only language models.

\paragraph{Scoping and Limitations.} A key limitation of the current approach lies in the use of Transformer-based auto-regressive decoders, which typically struggle with compositionality and generalisation. These decoders tend to memorise input-output reasoning patterns observed during training, rather than demonstrating genuine compositional reasoning capabilities. As a result, while our framework enables learning memorised reasoning behaviours within the latent space, it remains constrained in its ability to extrapolate beyond seen combinations or perform novel inference.

\section{Opportunities for Future Research}

Here, we present a list of possible opportunities for future work:

\begin{itemize}
    \item How can semantic representation learning improve compositional generalisation in NLI? Recent work by \citet{fu-frank-2024-exploring} explores how models can better generalise across compositional structures, but a more principled semantic grounding in the latent space remains an open challenge.
    \item Whether autoregressive LLMs can exhibit human-interpretable compositional behaviours in their latent space, particularly in the context of NLI remains an under-explored research area. While these models have demonstrated impressive performance on NLI benchmarks, their internal mechanisms for handling compositional reasoning are often opaque.
    \item Emerging diffusion-based models, such as Neuro-symbolic diffusion~\cite{nie2025large, van2025neurosymbolic}, offer new opportunities to combine the generative strengths of large-scale models with symbolic-level controllability. However, the design of semantically interpretable latent priors within these frameworks is still in its early stages. Incorporating the compositional semantic representation can potentially improve the sampling efficiency during the denoising process. Developing structured latent representations that align with formal semantics could enable more transparent reasoning and controllable generation in future generative architectures.
\end{itemize}

\section{Ethical Implications}

While the research presented in this thesis is primarily on the fundamental development of interpretable and controllable natural language representation learning, we recognise that the deployment of such models in real-world scenarios carries potential ethical implications that extend beyond the technical domain.

Data and Bias: In the thesis, part of the data considered is synthetic, and synthetic math derivations. Naturally, there is a disparity between synthetic data and real-world data, and it is difficult to characterise all differences between the datasets. This leads to unknown biases in experiments and fine-tuned models.


\bibliography{refs}    
\bibliographystyle{plain}



\appendix
\chapter{Formal Semantic Geometry}
\section{Experiment Setting} \label{sec:dsr_labels}
\paragraph{Dataset.} Table \ref{tab:stats_data_geo} displays the statistical information of the datasets used in the experiment. The data of the two datasets partially overlap, so only the unique explanations are selected as the experimental data. 
\begin{table}[ht!]
    \small
    \centering
    \renewcommand\arraystretch{1.3}
      \resizebox{7.6cm}{!}{
    \begin{tabular}{|c|cc|}
        \hline
        Corpus & Num data. & Avg. length \\ \hline
        WorldTree \cite{jansen2018worldtree} & 11430 & 8.65 \\
       EntailmentBank \cite{https://doi.org/10.48550/arxiv.2104.08661} & 5134 & 10.35 \\ \hline
    \end{tabular}
    }
    \caption{Statistics from explanations datasets.} \label{tab:stats_data_geo}
\end{table}

Table \ref{tab:visua_details} illustrates the semantic, structure, and topic information of explanatory sentences over the latent space. The explanatory sentences are automatically annotated using the semantic role labelling (SRL) tool, which can be implemented via the AllenNLP library \cite{Gardner2017ADS}. We report in Table~\ref{tab:srl_silva} the semantic roles from the explanations corpus. 
\begin{table*}[ht!]
    \small \setlength\tabcolsep{4.5pt}
    \centering
    \resizebox{\columnwidth}{!}{
\renewcommand\arraystretch{1.1}
    \begin{tabular}{p{1cm}p{14cm}}  \toprule
        \textbf{Cluster}         & \textbf{Theme and Pattern}                       \\ \hline
        0 & Theme: physics and chemistry. Pattern: \textit{if then} and \textit{as}. E.g., if a substance is mixed with another substance then those substances will undergo physical change.   \\ \hline
        1 & Theme: country, astronomy, and weather. E.g., new york state is on earth \\ \hline
        2 & Theme: physics and chemistry. Pattern: \textit{is a kind of}. E.g., light is a kind of wave. \\ \hline
        3 & Theme: biology. E.g., a mother births offspring. \\ \hline
        4 & Theme: synonym for verb. Pattern: \textit{means} and \textit{is similar to}. E.g., to report means to show. \\ \hline
        5 & Theme: astronomy. E.g., the solar system contains asteroids.\\ \hline
        6 & Theme: animal/plant. Pattern: \textit{is a kind of}. E.g., a seed is a part of a plant. \\ \hline
        7 & Theme: item. E.g., a telephone is a kind of electrical device for communication.\\ \hline
        8 & Theme: synonym for life. Pattern: \textit{means} and \textit{is similar to}. E.g., shape is a kind of characteristic.  \\ \hline
        9 & Theme: geography. Pattern: \textit{is a kind of}. E.g., a mountain is a kind of environment.\\ \hline
        10 & Theme: animal and plant. Pattern: \textit{if then} and \textit{as}. E.g., if a habitat is removed then that habitat is destroyed.  \\ \hline
        11 & Theme: scientific knowledge. Pattern: \textit{(;)}, \textit{number} and \textit{/}. E.g., freezing point is a property of a ( substance ; material ). \\ \hline
        12 & Theme: item. Pattern: \textit{is a kind of object}. E.g., a paper is a kind of object. \\ \hline
        13 & Theme: chemistry and astronomy. E.g., oxygen gas is made of only oxygen element.\\ \hline
        14 & Theme: general about science. Pattern: \textit{(;)}. E.g., seed dispersal has a positive impact on ( a plant ; a plant 's reproduction). \\ \hline
        15 & Theme: item. Pattern: \textit{is a kind of}. E.g., fertilizer is a kind of substance. \\ \hline
        16 & Theme: physics and chemistry. Pattern: \textit{(;)}. E.g., the melting point of oxygen is -3618f ; -2188c ; 544k. \\ \hline
        17 & Theme: animal. E.g., squirrels live in forests. \\ \hline
        18 & Theme: nature. E.g., warm ocean currents move to cooler ocean regions by convection.\\ \hline
        19 & Theme: life. E.g., pond water contains microscopic living organisms.\\ \bottomrule
    \end{tabular}
    }
    \caption{K-means cluster information in explanations corpus.} 
    \label{tab:visua_details}
\end{table*}

\begin{table*}[ht!]
    \small
    \centering
\resizebox{\columnwidth}{!}{
\renewcommand\arraystretch{1.2}
    \begin{tabular}{p{2cm}p{2cm}p{11cm}}  \toprule
        \textbf{SRL}     & \textbf{Prop. \%}    & \textbf{Description and Example}                       \\ \hline
        ARGM-DIR & 0.80 & Directionals. E.g. all waves transmit energy \textbf{from one place to another}  \\ \hline
        ARGM-PNC & 0.08 & Purpose. E.g. many animals blend in with their environment \textbf{to not be seen by predators} \\ \hline
        ARGM-CAU & 0.05 & Cause. E.g. cold environments sometimes are white in color \textbf{from being covered in snow} \\ \hline
        ARGM-PRP & 1.30 & Purpose. E.g. a pot is made of metal \textbf{for cooking} \\ \hline
        ARGM-EXT & 0.04 & Extent. E.g. as the amount of oxygen exposed to a fire increases the fire will burn \textbf{longer} \\ \hline
        ARGM-LOC & 4.50 & Location. E.g. a solute can be dissolved \textbf{in a solvent} when they are combined \\ \hline
        ARGM-MNR & 2.00 & Manner. E.g. fast means \textbf{quickly} \\ \hline
        ARGM-MOD & 9.80 & Modal verbs. E.g. atom \textbf{can} not be divided into smaller substances \\ \hline
        ARGM-DIS & 0.07 & Discourse. E.g. if something required by an organism is depleted \textbf{then} that organism must replenish that something \\ \hline
        ARGM-GOL & 0.20 & Goal. E.g. We flew \textbf{to Chicago} \\ \hline
        ARGM-NEG & 1.20 & Negation. E.g. cactus wrens building nests in cholla cacti does \textbf{not} harm the cholla cacti \\ \hline
        ARGM-ADV & 6.70 & Adverbials \\ \hline
        ARGM-PRD & 0.20 & Markers of secondary predication. E.g. \\ \hline
        ARGM-TMP & 7.00 & Temporals. E.g. a predator \textbf{usually} kills its prey to eat it \\ \hline
        O & - & Empty tag. \\ \hline
        V & 100 & Verb. \\ \hline
        ARG0 & 32.0 & Agent or Causer. E.g. \textbf{rabbits} eat plants \\ \hline
        ARG1 & 98.5 & Patient or Theme. E.g. rabbits eat \textbf{plants} \\ \hline
        ARG2 & 60.9 & indirect object / beneficiary / instrument / attribute / end state. E.g. animals are \textbf{organisms} \\ \hline
        ARG3 & 0.60 & start point / beneficiary / instrument / attribute. E.g. sleeping bags are designed \textbf{to keep people warm} \\ \hline
        ARG4 & 0.10 & end point. E.g. when water falls from the sky that water usually returns \textbf{to the soil} \\ \bottomrule
        
    \end{tabular}
    }
    \caption{SRL that appears in explanations corpus.} 
    \label{tab:srl_silva}
\end{table*}



\paragraph{Hyperparameters.} The training process of the decision tree binary classifier can be implemented via scikit-learn packages with default hyperparameters. As for Optimus, the latent space size is 32 in the experiment. The training details are following the original experiment from Optimus \cite{li2020optimus}.

\section{Further Experimental Results} \label{sec:apd_exp_res}




\paragraph{Qualitative evaluation for arithmetic.} Table~\ref{tab:arith_other_examples} lists the traversed explanations after addition (blue) and subtraction (red) on different semantic role information. We can observe that the resulting sentences after addition can hold the same role-content as inputs, revealing latent space geometry.

\begin{table*}[ht!]
\begin{tcolorbox}[fontupper=\small, fontlower=\small, middle=0.3cm, title=ADD and SUB arithmetic]
ARGUMENT1: \\
\underline{a needle is a kind of object} \\
\underline{a tire is a kind of object} \\
a wire \textcolor{blue}{is a kind of object} \\
a stick \textcolor{blue}{is a kind of object} \\
a ball \textcolor{blue}{is a kind of object} \\
a serotype is \textcolor{red}{similar to intersex egg} \\
a zygote contains \textcolor{red}{many cell types} \\
an xylem is made \textcolor{red}{of two clumps} \\
VERB: \\
\underline{chromosomes are located in the cells} \\
\underline{Australia is located in the southern hemisphere} \\
stars are \textcolor{blue}{located} in the solar system \\
Jupiter is \textcolor{blue}{located} in the milky way galaxy \\
aurora is \textcolor{blue}{located} in the constellation of Leo \\
a crystal is \textcolor{red}{made} of metal \\
an alloy is \textcolor{red}{made} of iron and zinc \\
an aluminum plug \textcolor{red}{is} nonmagnetic \\
LOCATION: \\
\underline{volcanoes are often found under oceans} \\
\underline{mosquitos can sense carbon dioxide in the air} \\
polar ice sheets are located \textcolor{blue}{along rivers} \\
hurricanes occur frequently along the coast \textcolor{blue}{in Africa} \\
tide waves cause flooding \textcolor{blue}{in coastal waters} \\
\textcolor{red}{valley is a kind of location} \\
\textcolor{red}{shape is a property of rocks} \\
\textcolor{red}{desert is a kind of place} \\
TEMPORAL: \\
\underline{as the population of prey decreases competition} \underline{between predators will increase} \\
\underline{as competition for resources decreases the ability} \underline{to compete for resources will increase} \\
\textcolor{blue}{as the population of an environment decreases} ecosystem function will decrease \\
\textcolor{blue}{as the spread of available air mass increases} the population will increase \\
\textcolor{blue}{as the number of heavy traffic required increases} the traffic cycle will decrease \\
\textcolor{red}{some types of lizards live in water} \\
\textcolor{red}{a rose is rich in potassium} \\
\textcolor{red}{a fern grass roots foot trait means a fern grass} \\
NEGATION: \\
\underline{pluto has not cleared its orbit} \\
\underline{sound can not travel through a vacuum} \\
radio waves \textcolor{blue}{don't} have electric charge \\
electromagnetic radiation \textcolor{blue}{does not} have a neutral electric charge \\
electromagnetic radiation contains \textcolor{blue}{no} electric charge \\
Mars \textcolor{red}{is} a kind of moon / planet \\
Anothermic rock \textcolor{red}{is} a kind of metamorphic rock \\
Anal Cetus's skeleton \textcolor{red}{is} a kind of fossil
\end{tcolorbox}
\caption{Latent sapce arithmetic for five semantic tags (blue: addition, red: subtraction).}
\label{tab:arith_other_examples}
\end{table*}




\paragraph{Quantitative evaluation for arithmetic.} \label{sec:apd_consistency} Quantitative evaluation for our hypotheses via latent arithmetic. Both VERB and Object can perform high ratio after addition, indicating role-content separability.







\begin{figure*}[ht]
\begin{center}
    \includegraphics[width=\linewidth]{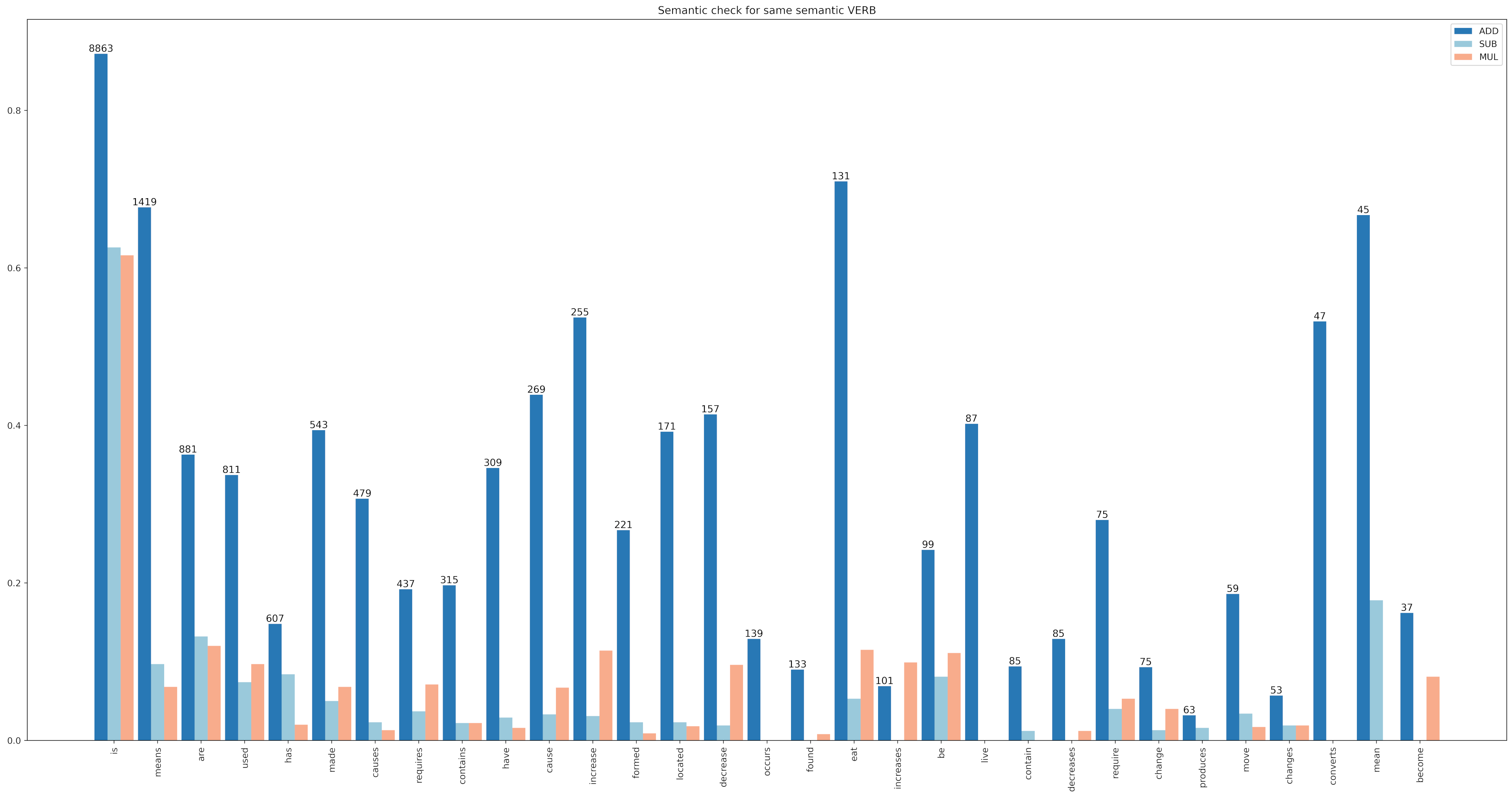}
    \caption{Predicate (VERB). The content \textit{is} shows the high ratio after subtraction, indicating that the \textit{V-is} is widely distributed over the latent space.}
    \label{fig:consistency_verb_sem}
    \end{center}
\end{figure*}

\begin{figure*}[ht]
\begin{center}
    \includegraphics[width=\linewidth]{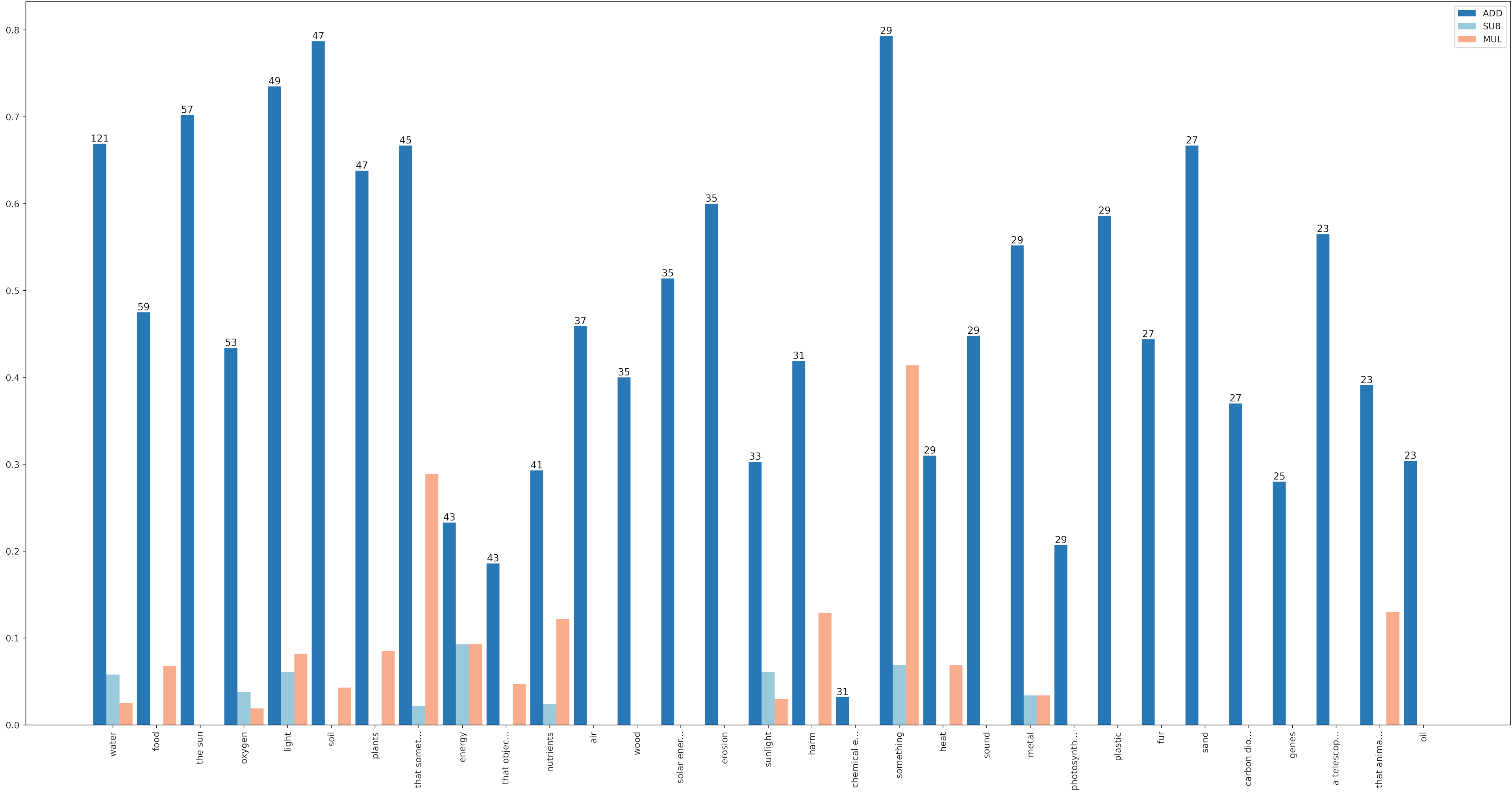}
    \caption{Object (ARG1).}
    \label{fig:consistency_arg1_sem}
\end{center}
\end{figure*}

\begin{figure*}[ht]
\begin{center}
    \includegraphics[width=\linewidth]{geometry/figs/consistency/dist_verb_same_sem.png}
    \caption{Cosine distance of sentence pairs in VERB-content clusters.}
    \label{fig:cos_dist_v}
    \end{center}
\end{figure*}

\begin{figure*}[ht]
\begin{center}
    \includegraphics[width=\linewidth]{geometry/figs/consistency/dist_arg1_same_sem.png}
    \caption{Cosine distance of sentence pairs in ARG1-content clusters.}
    \label{fig:cos_dist_a1}
\end{center}
\end{figure*}




\chapter{Formal Semantic Disentanglement}
\section{Experiment setting} \label{sec:train_detail}

\begin{figure}[ht!]
\begin{center}
    \includegraphics[width=0.7\columnwidth]{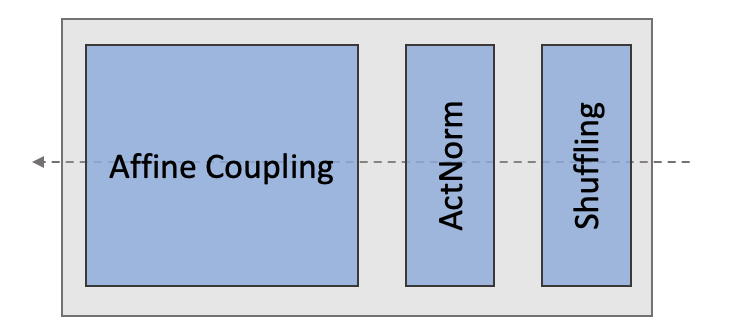}
    \caption{INN one single block.}
    \label{fig:inn_block}
 \end{center}
\end{figure}

\paragraph{INN.} The INN consists of 10 invertible blocks. Each is built from three layers, including an affine coupling \cite{dinh2016density}, permutation layer, and ActNorm \cite{kingma2018glow}. Figure \ref{fig:inn_block} displays one single invertible block. The model was implemented using the FrEIA library \footnote{\url{https://github.com/VLL-HD/FrEIA}}. As for training hyperparameters of INN, firstly, both input and output have the same dimensions as the latent space dimension of the autoencoder. Secondly, inside the affine coupling block, the sub-network is MLP with 512 as the hidden dimension. Thirdly, we use AdamW to optimise the model where the learning rate is 5e-04 in the experiment. 

\noindent The forward process of the affine coupling layer can be described as follows: 
\begin{equation}
\begin{split}
x_a, x_b = \text{split}(x) \\
\log s, t = m_{\theta}(x_b) \\
s = \exp (\log s) \\
y_a = s \odot x_a + t \\
y_b = x_b \\
y = \text{concat}(y_a, y_b)
\end{split}
\end{equation}
Where $m_{\theta}$ is a two-layer neural network. $x$ and $y$ are the input and output. The reversed process is:
\begin{equation}
\begin{split}
y_a, y_b = \text{split}(y) \\
\log s, t = m_{\theta}(y_b) \\
s = \exp (\log s) \\
x_a = (y_a - t) / s \\
x_b = y_b \\
y = \text{concat}(x_a, x_b)
\end{split}
\end{equation}

\section{Additional Supervision Results} \label{sec:dis_pred}
\begin{figure}[ht!]
\centering
    \includegraphics[scale=0.16]{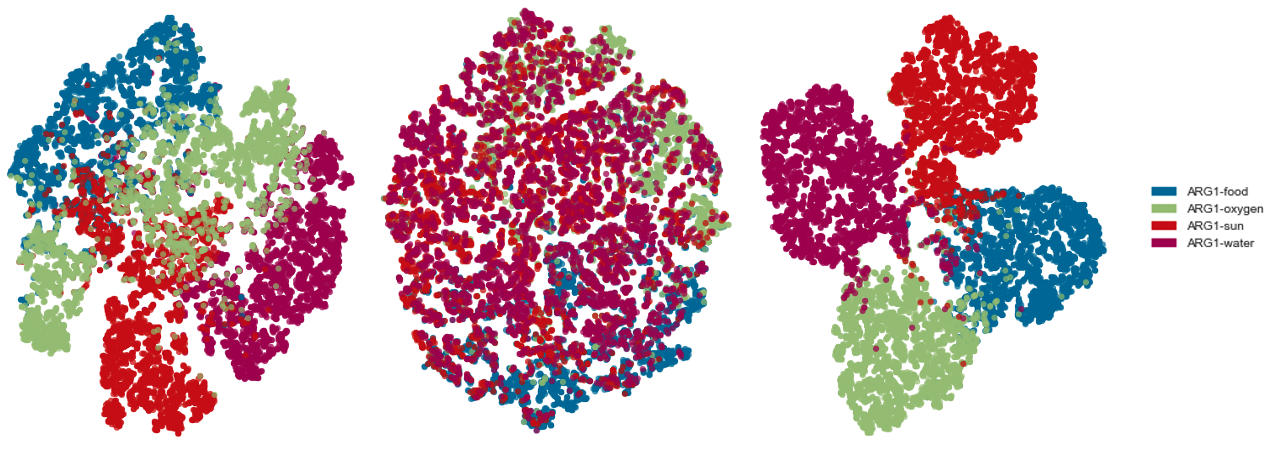}
    \caption{ARG1: t-SNE plot (blue: \textit{food}, green: \textit{oxygen}, red: \textit{sun}, purple: \textit{water}). Supervision (right) induces separability comparable with ARG0. PCA plot is provided in Figure \ref{fig:a1_pca}.}
    \label{fig:food_water}
\end{figure}

\paragraph{Disentanglement between \textit{ARG1} clusters} We consider four \textit{ARG1} clusters, including \textit{ARG1-food}, \textit{ARG1-oxygen}, \textit{ARG1-sun}, \textit{ARG1-water}, and evaluate model performance following the same procedure. Figure \ref{fig:food_water} displays the distributions of four role-content clusters over the latent space. With similar observations as before, the INN cluster-supervised training strategy can learn better disentanglement between ARG1 clusters. 
Table \ref{tab:arg1_exp} and \ref{tab:arg1_invert_ratio} show the disentanglement metrics and invertibility ratio, respectively. With similar observations as the previous experiment: all classifiers trained over the supervised latent representation outperform both the unsupervised INN model and Optimus, and both unsupervised and supervised cases can achieve higher ratios (at least 0.95).

 
 


\begin{table}[ht!]
\small
\centering
\setlength\tabcolsep{2.5pt}
\resizebox{10cm}{!}{
\renewcommand\arraystretch{1}
\begin{tabular}{cccccc}
\toprule
\multicolumn{6}{c}{ARG1: disentanglement proxy metrics (forward: $T$)} \\ \hline

classifier & train & accuracy & precision & recall  & f1 score \\ \hline
\multirow{3}{*}{KNN}  & O & 0.934 & 0.934 & 0.933  & 0.933 \\
& U & 0.914 & 0.914 & 0.914  & 0.913 \\
& C & \textbf{0.954} & \textbf{0.954} & \textbf{0.954}  & \textbf{0.954}  \\ \hline
 
\multirow{3}{*}{NB} & O & 0.904 & 0.910 & 0.902  & 0.904 \\
& U & 0.922 & 0.922 & 0.922  & 0.922 \\
& C & \textbf{0.957} & \textbf{0.957} & \textbf{0.957}  & \textbf{0.957} \\ \hline
 
\multirow{3}{*}{SVM} & O & 0.951 & 0.951 & 0.951  & 0.950 \\
& U & 0.953 & 0.953 & 0.952 & 0.953 \\
& C & \textbf{0.959} & \textbf{0.959} & \textbf{0.959}  & \textbf{0.959} \\ \toprule

\end{tabular}
}
\caption{Forward evaluation for ARG1, consistent results on different classifiers indicate that supervision can perform better semantic disentanglement.} \label{tab:arg1_exp}
\end{table}

\paragraph{Invertibility ratio.} Table \ref{tab:arg1_invert_ratio}, \ref{tab:v_invert_ratio}, and \ref{tab:animal_invert_ratio} report the invertibility test for \textit{ARG1}, \textit{PRED}, and \textit{ARG0,1,2} clusters, respectively. We can observe that INN with both training approaches can perform stable invertibility. 

\begin{table}[ht]
\small
\centering
\setlength\tabcolsep{2.5pt}
\begin{tabular}{cccccc}
\toprule 
\multicolumn{6}{c}{ARG1: invertibility ratio (backward: $T'$)} \\ \hline
\multicolumn{2}{c}{train} & food & oxygen & sun & water \\ \hline
\multicolumn{2}{c}{U} & 0.990 & 0.980 & 0.950 & 1.000 \\ 
\multicolumn{2}{c}{C} & 0.960 & 0.950 & 0.960 & 1.000 \\\toprule
\end{tabular}
\caption{backward evaluation for ARG1 clusters. unsupervised INN (U), and supervised INN (C).} \label{tab:arg1_invert_ratio}
\end{table}
\begin{table}[ht]
\small
\centering
\setlength\tabcolsep{2.5pt}
\begin{tabular}{cccccc}
\toprule 
 \multicolumn{6}{c}{PRED: invertibility test (backward: $T'$)} \\ \hline
\multicolumn{2}{c}{train} & is & are & cause & require \\ \hline
\multicolumn{2}{c}{U} & 1.000 & 0.950 & 0.970 & 0.800 \\
\multicolumn{2}{c}{C} & 1.000 & 0.880 & 0.900 & 0.820 \\ \toprule
\end{tabular}
\caption{backward evaluation for predicate clusters. unsupervised INN (U), and supervised INN (C).} \label{tab:v_invert_ratio}
\end{table}

\begin{table}[ht!]
\small
\centering
\setlength\tabcolsep{2.5pt}
\renewcommand\arraystretch{1}
\begin{tabular}{cccc}
\toprule
\multicolumn{4}{c}{Animal: invertibility ratio (backward: $T'$)} \\ \hline
\multicolumn{1}{c}{train} & ARG0 & ARG1 & ARG2 \\ \hline
\multicolumn{1}{c}{U} & 0.990 & 0.990 & 0.900 \\
\multicolumn{1}{c}{C} & 0.970 & 0.960 & 0.920 \\ \toprule
\end{tabular}
\caption{Backward evaluation for Animal.} \label{tab:animal_invert_ratio}
\end{table}

\begin{figure}[ht!]
    \centering
    \includegraphics[width=0.9\columnwidth]{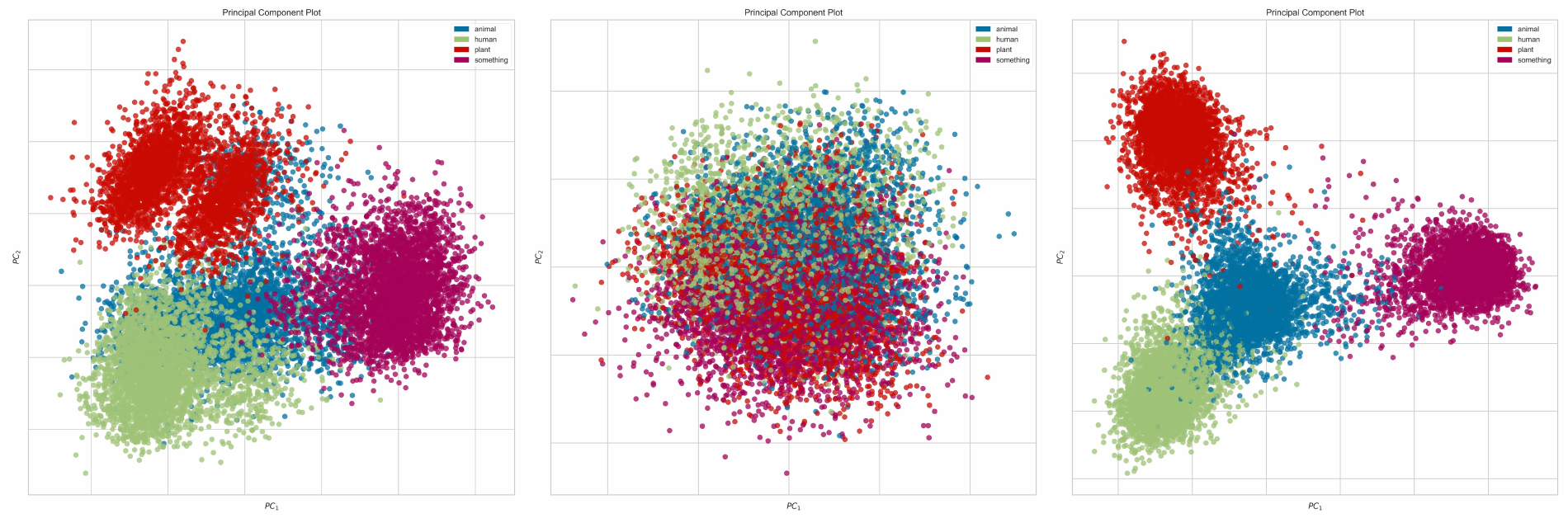}
    \caption{PCA visualization for \textit{ARG0}.}
    \label{fig:a0_pca}
\end{figure}

\begin{figure}[ht!]
    \centering
    \includegraphics[width=0.9\columnwidth]{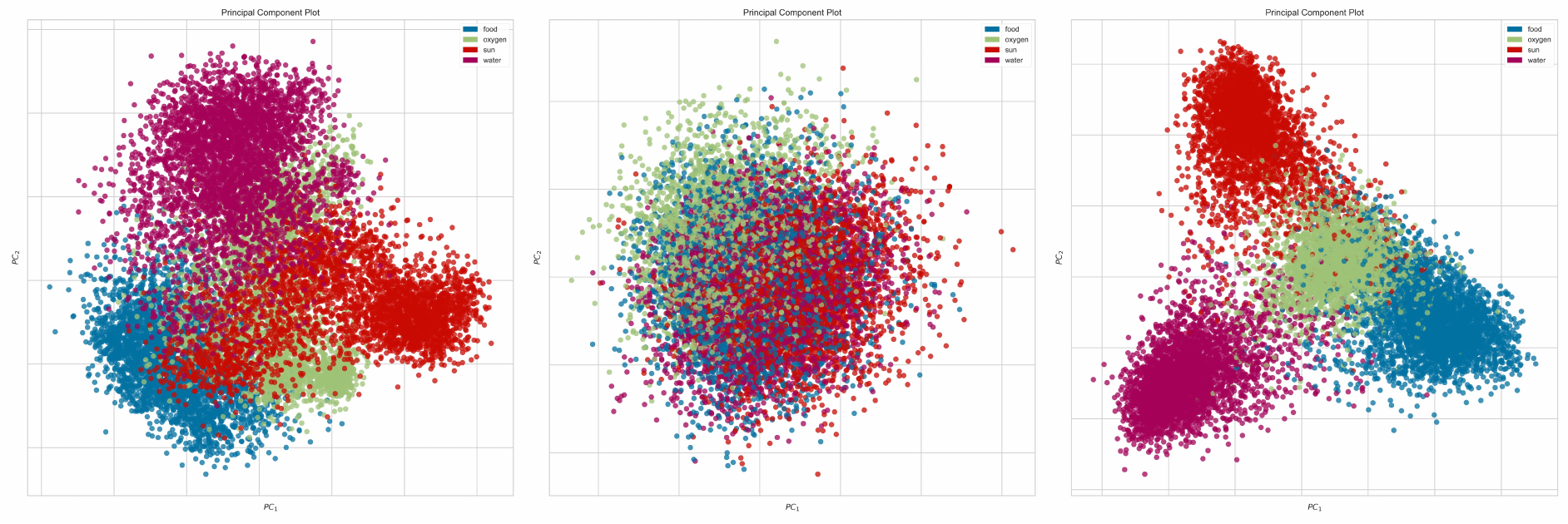}
    \caption{PCA visualization for \textit{ARG1}.}
    \label{fig:a1_pca}
\end{figure}

\begin{figure}[ht!]
    \centering
    \includegraphics[width=0.9\columnwidth]{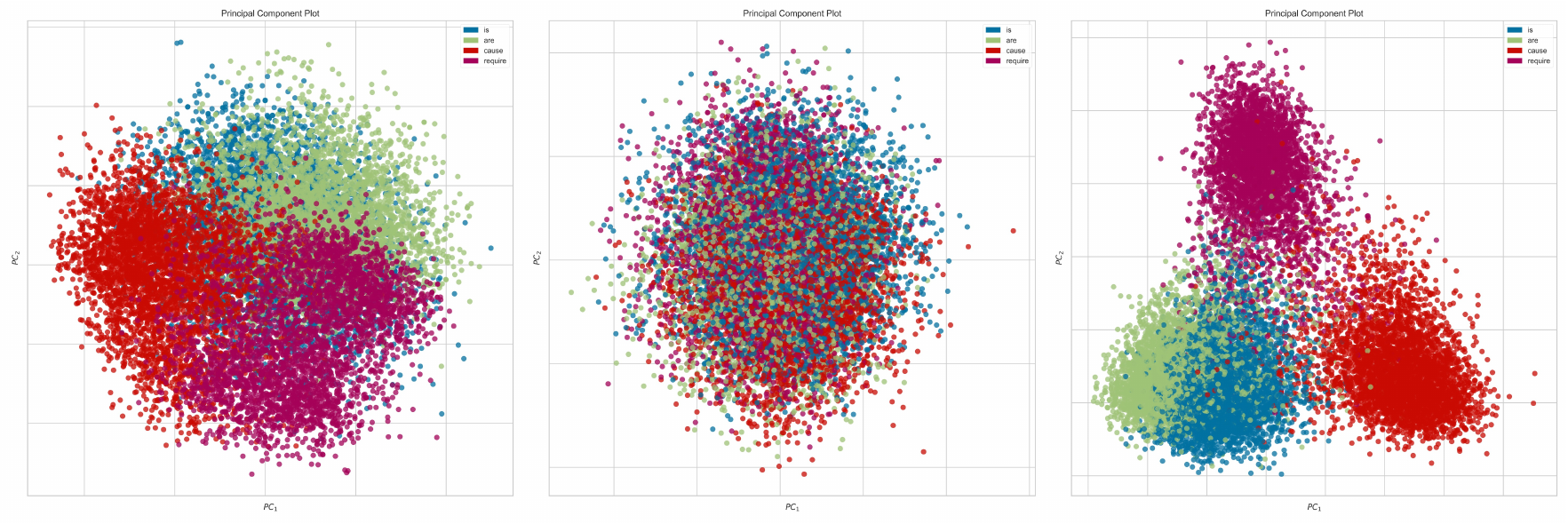}
    \caption{PCA visualization for \textit{PRED}.}
    \label{fig:verb_pca}
\end{figure}

\begin{figure}[ht!]
    \centering
    \includegraphics[width=0.9\columnwidth]{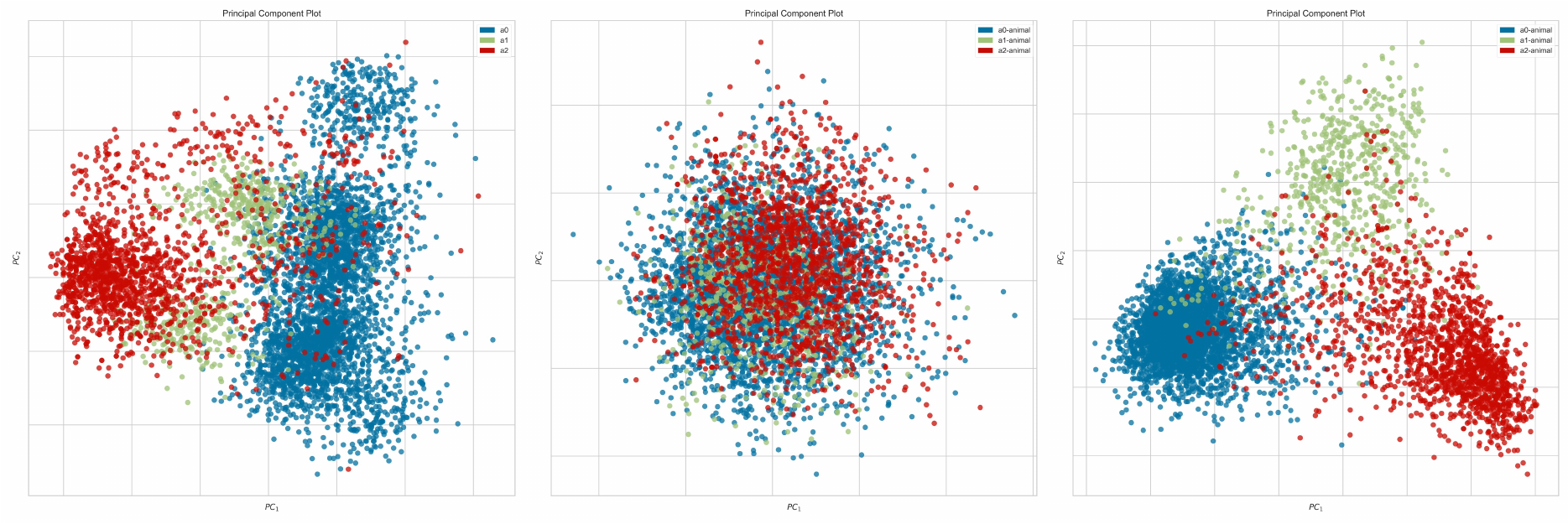}
    \caption{PCA visualization for \textit{Animal}.}
    \label{fig:animal_pca}
\end{figure}

\paragraph{Principal component analysis (PCA) visualisation.} In addition to the non-linearised t-SNE plot, we also provide linearised visualisation via PCA. Figure \ref{fig:a0_pca},\ref{fig:a1_pca},\ref{fig:verb_pca}, and \ref{fig:animal_pca} visualize the separation of \textit{ARG0}, \textit{ARG1}, \textit{PRED}, and \textit{animal}. Similar to the observation before, cluster supervision can lead to better separation and cluster.

\section{Ablation of Data Augmentation} \label{sec:ablation}
\begin{figure}[ht!]
    \centering
    \includegraphics[width=0.9\columnwidth]{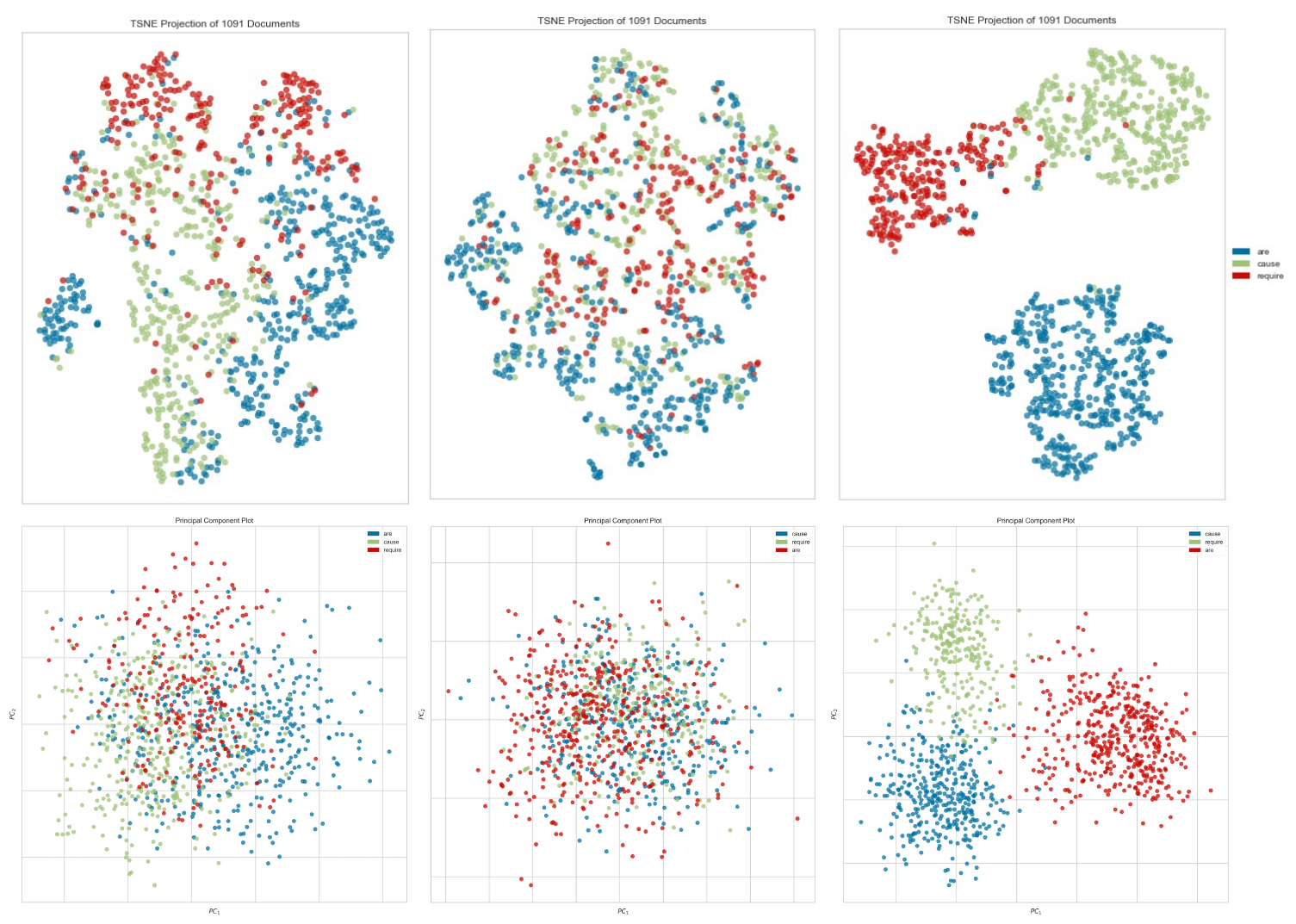}
    \caption{Ablation: t-SNE plot (top), PCA plot (bottom) (left: Optimus, middle: unsupervised, right: cluster-supervised) where blue: \textit{PRED-are}, green: \textit{PRED-cause}, red: \textit{PRED-require}.}
    \label{fig:ablation}
\end{figure}

\paragraph{\textit{PRED} semantic role.} Firstly, we analyse the effect of our supervision approach on \textit{PRED} semantic role with three lexical contents without data augmentation, including \textit{are} ($\times449$), \textit{cause} ($\times380$), and \textit{require} ($\times262$). The rationale for their selection is that they are less frequent in corpus and partially overlap in latent space. Moreover, the contents under \textit{PRED} usually have less effect on the contextual semantics \cite{zhang2022}. Those difficulties allow us to fairly analyse the effect of our supervision approach. Following a similar order, we first visualise the t-SNE and PCA plots in Figure \ref{fig:ablation}. As we can observe, the cluster-supervised approach can better represent the cluster and separation for different contents under \textit{PRED} semantic role label without data augmentation. Next, we apply downstream classifiers to evaluate cluster separation. As illustrated in Table \ref{tab:ablation_metrics}, our cluster-supervised approach results in better classification performance, indicating better disentanglement. 

\begin{table}[ht]
\small
\centering
\setlength\tabcolsep{2.5pt}
\resizebox{10cm}{!}{
\begin{tabular}{cccccc}
\toprule
\multicolumn{6}{c}{\textit{PRED}: disentanglement proxy metrics} \\ \hline

classifier & train & accuracy & precision & recall  & f1 score \\ \hline
\multirow{3}{*}{KNN} & O & 0.858 & 0.847 & 0.844  & 0.846 \\
 & U & 0.837 & 0.849 & 0.827  & 0.830 \\
 & C & \textbf{0.965} & \textbf{0.963} & \textbf{0.961}  & \textbf{0.962} \\ \hline
 
\multirow{3}{*}{NB} & O & 0.839 & 0.823 & 0.833  & 0.826 \\
& U & 0.901 & 0.895 & 0.891  & 0.893 \\
 & C & \textbf{0.977} & \textbf{0.974} & \textbf{0.975}  & \textbf{0.974} \\ \hline
 
\multirow{3}{*}{SVM} & O & 0.876 & 0.863 & 0.866  & 0.865 \\
 & U & 0.954 & 0.953 & 0.949  & 0.950 \\
 & C & \textbf{0.967} & \textbf{0.965} & \textbf{0.967}  & \textbf{0.966} \\ \toprule
\end{tabular}
}
\caption{Ablation: disentanglement proxy metrics for \textit{PRED-are}, \textit{PRED-cause}, and \textit{PRED-require}.} \label{tab:ablation_metrics}
\end{table}

\section{Linear Interpolation}
In tables \ref{tab:guide_generation_app2} and \ref{tab:guide_generation_app1}, we provide more controllable interpolation examples. Those examples reveal that the latent space with better role-content separation from supervised INN can provide better interpolation control, indicating better latent space geometry.
\begin{table}[ht!]
\begin{tcolorbox}[fontupper=\small, fontlower=\small, title=Interpolation localisation: \textit{predicate-require} ]
\underline{source: humans \textcolor{blue}{require} freshwater for survival}\\ \\
AutoEncoder: \\
1. humans \textcolor{blue}{require} water to survive \\
2. marine mammals \textcolor{blue}{require} great amounts of water \\
3. animals \textcolor{blue}{require} oxygen to survive  \\
4. animals \textcolor{blue}{require} water for survival \\
5. animals \textcolor{red}{must eat} water to survive \\
6. animals \textcolor{blue}{require} water and food  \\
7. animals \textcolor{blue}{require} water for survival \\
8. animals \textcolor{red}{must eat} to survive  \\
9. animals \textcolor{blue}{require} food for survival  \\
10. animals \textcolor{red}{must eat} food to survive \\

Unsupervised INN: \\
1. nonhumans \textcolor{blue}{require} water to survive \\
2. marine animals \textcolor{blue}{require} food for survival  \\
3. animals \textcolor{red}{must breath} to survive  \\
4. animals \textcolor{blue}{require} water for survival \\
5. animals \textcolor{blue}{require} water from their ecosystems  \\
6. animals \textcolor{blue}{require} water for survival \\
7. animals \textcolor{red}{must eat} food for survival \\
8. animals \textcolor{blue}{require} food for survival \\
9. animals \textcolor{blue}{require} food for survival \\
10. animals \textcolor{blue}{require} food for survival \\


\underline{target: animals \textcolor{blue}{require} food to survive}
\end{tcolorbox}

\caption{Interpolation examples where top and bottom sentences are source and target, respectively.}
\label{tab:guide_generation_app}
 \end{table}

\begin{table*}[ht!]
\centering
\begin{tcolorbox}[fontupper=\small, fontlower=\small,title=Interpolation localisation: \textit{predicate-is}]
\underline{source: the sun \textcolor{blue}{is} in the northern hemisphere}\\ \\
1. the sun \textcolor{blue}{is} located in the northern hemisphere \\
2. the sun \textcolor{blue}{is} in the northern hemisphere \\
3. the sun \textcolor{blue}{is} made of air around the sun \\
4. the sun \textcolor{blue}{is} a source of sunlight for organisms \\
5. the sun \textcolor{blue}{is} a source of sunlight for birds \\
6. the sun \textcolor{blue}{is} a source of energy for organisms living in an arctic environment \\
7. the sun \textcolor{blue}{is} a source of food for plants \\
8. food \textcolor{blue}{is} a source of oxygen ; water for plants \\
9. food \textcolor{blue}{is} a source of energy for plants by producing heat \\
10. food \textcolor{blue}{is} a source of energy for a plant or animal / living thing \\

1. the sun \textcolor{blue}{is} the dominant star in the night sky \\
2. the sun \textcolor{blue}{is} closer to the earth than it is to the sun \\
3. the sun \textcolor{blue}{is} a star in the night sky \\
4. the sun \textcolor{blue}{is} good for the environment by providing sunlight to plants \\
5. the atmosphere \textcolor{blue}{is} an environment for intensive farming \\ 
6. the respiratory system \textcolor{red}{carries} oxygen to the rest of the body \\
7. food \textcolor{red}{contains} nutrients ; water ; food energy \\
8. food \textcolor{blue}{is} the nutrient for ( plants ; animals ) \\
9. producers \textcolor{red}{are} a source of energy for producers by weathering \\
10. food \textcolor{blue}{is} a part of a plant / animals / living things \\

\underline{target: food \textcolor{blue}{is} a source of energy for animals / plants}
\end{tcolorbox}
\caption{Interpolation examples (top: supervised INN, bottom: Optimus).}
\label{tab:guide_generation_app2}
\end{table*}

\begin{table*}[ht!]
\begin{tcolorbox}[fontupper=\small, fontlower=\small, title=Interpolation localisation: \textit{argument-animals} and \textit{predicate-require}]
\underline{source: \textcolor{blue}{animals require} food to survive}\\ \\
1. \textcolor{blue}{animals require} water to survive \\
2. \textcolor{blue}{animals require} food for survival \\
3. \textcolor{blue}{animals require} food for survival \\
4. \textcolor{blue}{animals require} nutrients from food \\
5. \textcolor{blue}{an animal requires} food for survival \\
6. \textcolor{blue}{an animal requires} food for survival \\
7. \textcolor{blue}{an animal requires} nutrients from producers \\
8. \textcolor{blue}{an animal requires} nutrients for survival \\
9. \textcolor{blue}{an animal requires} nutrients from food \\
10. \textcolor{blue}{an animal requires} nutrients from producers \\

1. \textcolor{blue}{animals} \textcolor{red}{need} sunglasses for protection  \\
2. \textcolor{blue}{animals} \textcolor{red}{live} in an environment \\
3. \textcolor{blue}{animals} \textcolor{red}{need} food to thrive \\
4. \textcolor{blue}{animals require} energy for survival \\
5. \textcolor{red}{a consumer} \textcolor{red}{uses} some of the food that is available \\
6. only \textcolor{red}{a producer} \textcolor{red}{eats} plants \\
7. \textcolor{red}{a human} \textcolor{red}{produces} its own food \\
8. \textcolor{blue}{an animal requires} nutrients in a source of food to survive \\
9. \textcolor{blue}{an animal requires} energy to perform photosynthesis \\
10. \textcolor{blue}{an animal requires} nutrients to grow \\ \\
\underline{target: \textcolor{blue}{an animal requires} nutrients from producers}
\end{tcolorbox}
\caption{Interpolation examples (top: supervised INN, bottom: Optimus).}
\label{tab:guide_generation_app1}
\end{table*}



\chapter{Syntax Representation}
\begin{figure*}[t]
    \centering
    \includegraphics[width=\textwidth]{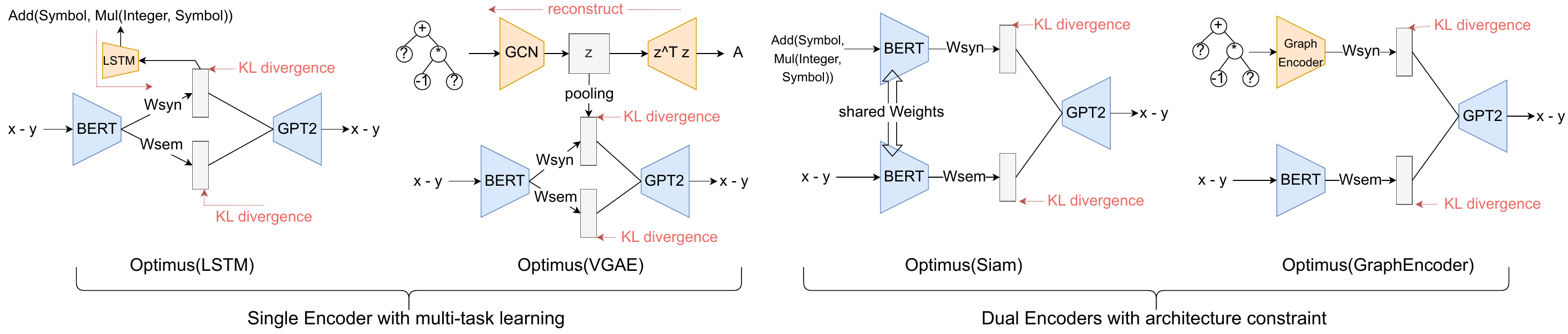}
    \caption{Overview of different methods to explicitly represent and disentangle syntactic information in the latent space of Transformer-based VAEs.}
    \label{fig:sem_syn_baselines}
\end{figure*}

\section{Training setups} \label{sec:enc_baselines}

\paragraph{Encoding Architecture.} Figure \ref{fig:sem_syn_baselines} illustrates the four architectures of encoding baselines for learning syntax representation.
\paragraph{Datasets.} Table \ref{tab:stats_data} displays the statistical information of the datasets used in the experiment. As for the AutoEncoder setup, we use the non-repetitive explanations selected from both WorldTree \cite{jansen2018worldtree} and EntailmentBank \cite{https://doi.org/10.48550/arxiv.2104.08661} corpus as the experimental data. The mathematical expressions are derived from \cite{meadows2023symbolic}.
\begin{table}[ht!]
    \small
    \centering
    \renewcommand\arraystretch{1}
    \begin{tabular}{|c|cc|}
        \hline
        Corpus & Num data. & Avg. length \\ \hline
        WorldTree & 11430 & 8.65 \\
        EntailmentBank & 5134 & 10.35 \\ 
        Math Symbol & 32000 & 6.84 \\
        Math Symbol Inference & 32000 & 51.84 \\ \hline
        
    \end{tabular}
    \caption{Statistics from datasets.} \label{tab:stats_data}
\end{table}
\paragraph{Tokenisation.} As for mathematical expressions, we add specific math tokens, including $\text{frac}, \sin, \cos, \log, e$, into the dictionary of both Bert and GPT2 and consider the remaining tokens as char-level. As for explanatory sentences, we use the default tokenisation in Bert and GPT2.

\paragraph{Syntax Parsing.} As for mathematical expression, we use Expression Trees \footnote{\url{https://docs.sympy.org/latest/tutorials/intro-tutorial/manipulation.html}}, As for explanatory sentences, we use consistency parser\footnote{\url{https://demo.allennlp.org/constituency-parsing}} from AllenNLP library \cite{gardner2018allennlp} to get the flattened syntax tree, and remove all word content from the tree as the input of graph encoder.
\paragraph{Model Implementation.} As for graph encoders, we use \textit{PyTorch Geometric} library \footnote{\url{https://pytorch-geometric.readthedocs.io/en/latest/}}. We deployed two hidden layers for GCN, GraphSAGE, and TransformerCONV. For mathematical expression, we replace the content of variables with random noises following uniform distribution with the range between -1 and 1 during the node embedding stage. The implementation of Optimus is based on their original code \footnote{\url{https://github.com/ChunyuanLI/Optimus}}. The implementation of LSTM-based VAEs is based on the code supplied from \citet{shen2020educating} \footnote{\url{https://github.com/shentianxiao/text-autoencoders}}.

\paragraph{Hyperparameters.} In the experiment, all baselines and our architecture hold the same size of latent representation (768). The training epoch is 100, the learning rate is 5e-5, the batch size is 64.

\section{More Experimental results} \label{sec:exp_res}
\paragraph{Qualitative Evaluation.} Moreover, we randomly sample the points in each k-mean cluster and output the corresponding sentences or syntax parse tree in Table \ref{tab:trav_examples}, \ref{tab:explanation_gcn_sem_trav_examples}, and \ref{tab:explanation_gcn_syn_trav_examples}.
\begin{table}[ht!]
\begin{tcolorbox}[fontupper=\small, fontlower=\small, middle=0.3cm, top=1pt, bottom=1pt, title=Math symbol: Syntax Cluster Traversal]
$C_0$: Pow(cos(Symbol(E)), Symbol(b)) \\
$C_0$: Pow(exp(Symbol(b)), Symbol(A)) \\
$C_0$: Mul(Symbol(F), sin(Symbol(g))) \\
 
$C_4$: exp(Mul(Pow(Symbol(V), Integer(-1)), Symbol(q))) \\
$C_4$: cos(Mul(Pow(Symbol(b), Integer(-1)), Symbol(g))) \\
$C_4$: exp(Mul(Pow(Symbol(T), Integer(-1)), Symbol(a))) \\

$C_8$: sin(Mul(Symbol(A), Symbol(k))) \\
$C_8$: cos(Mul(Symbol(U), Symbol(w))) \\
$C_8$: exp(Mul(Symbol(J), Symbol(l)))
\end{tcolorbox}
\caption{Qualitative evaluation of syntax cluster of Bert-TransCONV encoding.}
\label{tab:trav_examples}
\end{table}
\begin{table}[ht!]
\begin{tcolorbox}[fontupper=\small, fontlower=\small, middle=0.3cm, top=1pt, bottom=1pt, title=Explanations: Semantic Cluster Traversal]
$C_0$: if a pot is exposed to a stove then the pot will become hot \\
$C_0$: if something is used for something else then that something else is the job of that something  \\
$C_0$: if there is a crack in a rock then water can get into the crack  \\
 
$C_8$: decaying plant is a source of nutrients in soil \\
$C_8$: producers are a source of food energy for living things \\
$C_8$: organic matter is a source of nutrients in soil  \\

$C_5$: a magnet is a kind of object \\
$C_5$: a board is a kind of object \\
$C_5$: a wagon is a kind of object
\end{tcolorbox}
\caption{Qualitative evaluation of semantic cluster of Bert-GCN encoding.}
\label{tab:explanation_gcn_sem_trav_examples}
\end{table}
\begin{table*}[ht!]
\begin{tcolorbox}[fontupper=\small, fontlower=\small, middle=0.3cm, top=1pt, bottom=1pt, title=Explanations: Syntax Cluster Traversal]
$C_5$: (S (NP (JJ ) (NN )) (VP (VBZ ) (NP (JJ ) (NN )))) \\
$C_5$: (S (NP (DT ) (NN )) (VP (VBZ ) (NP (DT ) (NN ))))\\
$C_5$: (S (NP (JJ ) (JJ ) (NN )) (VP (VBZ ) (NP (JJ ) (NN ))))  \\
 
$C_6$: (S (NP (NN )) (VP (VBZ ) (PP (IN ) (NP (NP (DT ) (NN )) (SBAR (WHNP (WDT )) (S (VP (VBZ ) (VP (VBN ) (PP (IN ) (NP (NN )))))))))))  \\
$C_6$: (S (NP (NN )) (VP (VBZ ) (NP (NP (DT ) (NN )) (PP (IN ) (SBAR (WHADVP (WRB )) (S (NP (DT ) (NN )) (VP (VBZ ) (VP (VBN ))))))))) \\
$C_6$: (S (NP (NN )) (VP (VBZ ) (NP (NP (DT ) (NN )) (SBAR (WHNP (WDT )) (S (VP (VBZ ) (ADJP (JJ ) (JJS ) (PP (IN ) (NP (DT ) (NNP ))))))))))   \\

$C_9$: (S (NP (NNS )) (VP (VBP ) (NP (NN )) (PP (IN ) (NP (NNS ))))) \\
$C_9$: (S (NP (NNS )) (VP (VBP ) (PP (IN ) (NP (NN )))))  \\
$C_9$: (S (NP (NNS )) (VP (MD ) (VP (VB ) (NP (NN ) (NN )) (PP (IN ) (NP (DT ) (NN )))))) 
\end{tcolorbox}
\caption{Qualitative evaluation of semantic cluster of Bert-GCN encoding.}
\label{tab:explanation_gcn_syn_trav_examples}
\end{table*}

Besides, in Table \ref{tab:rec}, we provide the comparison of reconstructed sentences between normal Optimus and Bert-TransCONV(addition QKV).

\begin{table*}[ht!]
    \small
    \centering
\renewcommand\arraystretch{1.3}
\resizebox{\columnwidth}{!}{
    \begin{tabular}{p{5cm}p{5cm}p{5cm}}  \toprule
        \textbf{Gold explanations} & \textbf{BERT-GPT2} & \textbf{Bert/TransCONV-GPT2} \\ \hline
        lenses are a kind of object & frog is a kind of object & lenses are a kind of object \\
        the chemical symbol for helium is he & a substance has a physical shape & the chemical symbol for helium is He \\
        a rose is a kind of plant & a window pane is a kind of surface & a rose is a kind of flower \\
        a body of water contains water & a flood has a large amount of rainfall & a body of water contains water \\
        growing is a kind of process & population is a kind of process & growing is a kind of process \\
        air is a kind of gas & farming is a kind of human & air is a kind of gas \\
        action means activity & feed means use & activity means action \\
        soda water is a kind of carbonated beverage & condensing is a kind of change in temperature & soda water is a kind of carbonated beverage \\
        plasma is a kind of state of matter & black probability is a kind of event & plasma is a kind of state of matter \\
        earth is a kind of celestial object & sun is a kind of light & earth is a kind of celestial object \\
        a bee is a kind of living thing & a frog is a kind of amphibian & a bee is a kind of living thing
        \\ 
        green is a kind of color & deforestation is a kind of process & green is a kind of color \\ 
        a wooded area is a kind of forest & a coal mine is a kind of natural resource & a wooded area is a kind of forest \\ \toprule
    \end{tabular}
    }
    \caption{Explanation reconstruction (left: original explanations from WorldTree corpus, middle: explanations from Optimus, right: explanations from Bert-TransCONV (addition Q)).} 
    \label{tab:rec}
\end{table*}

\begin{figure}[ht!]
    \centering
    \includegraphics[width=\columnwidth]{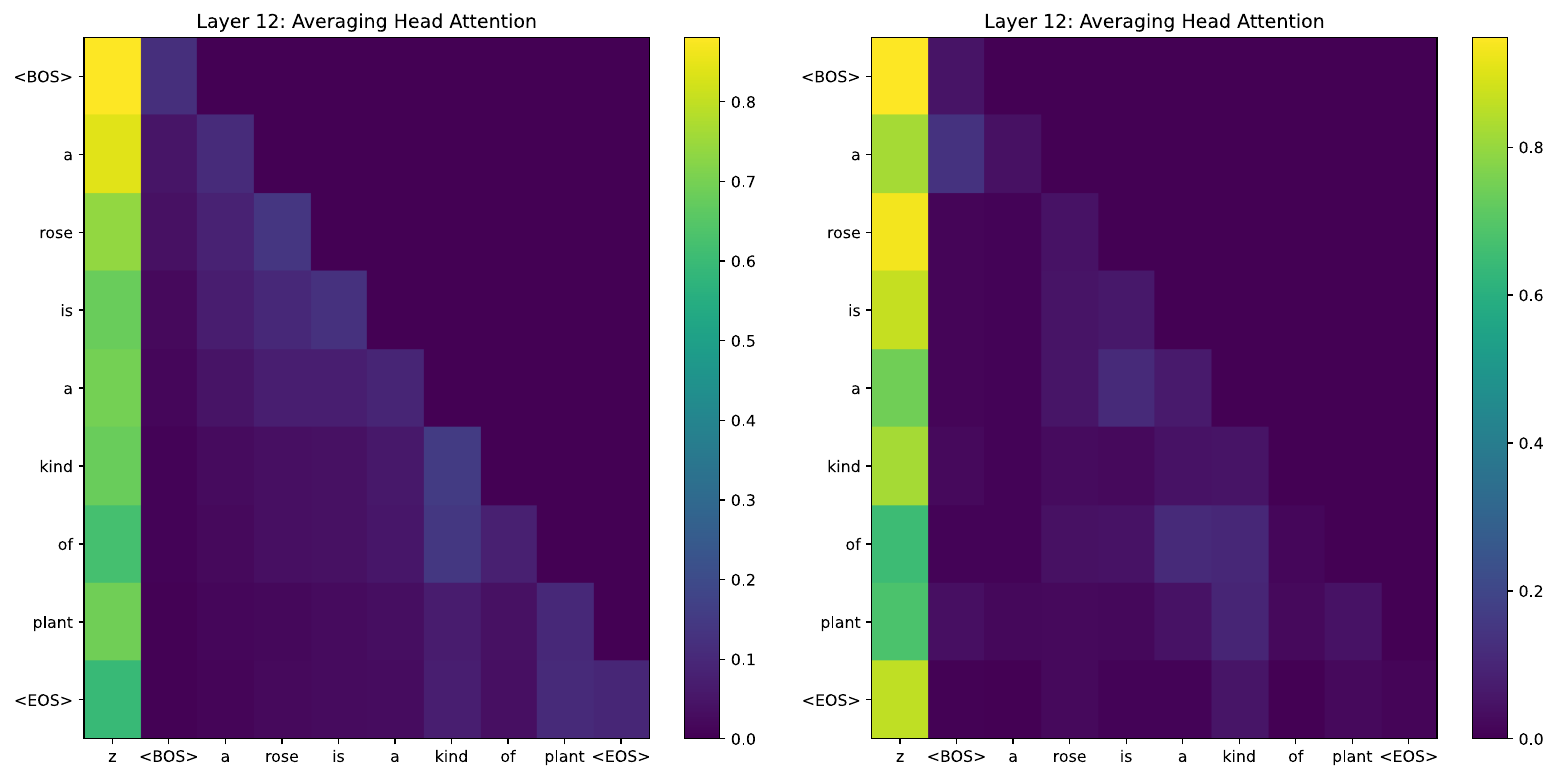}
    \caption{\textit{a rose is a kind of plant}.}
    \label{fig:heatmap_1}
\end{figure}

\begin{figure}[ht!]
    \centering
    \includegraphics[width=\columnwidth]{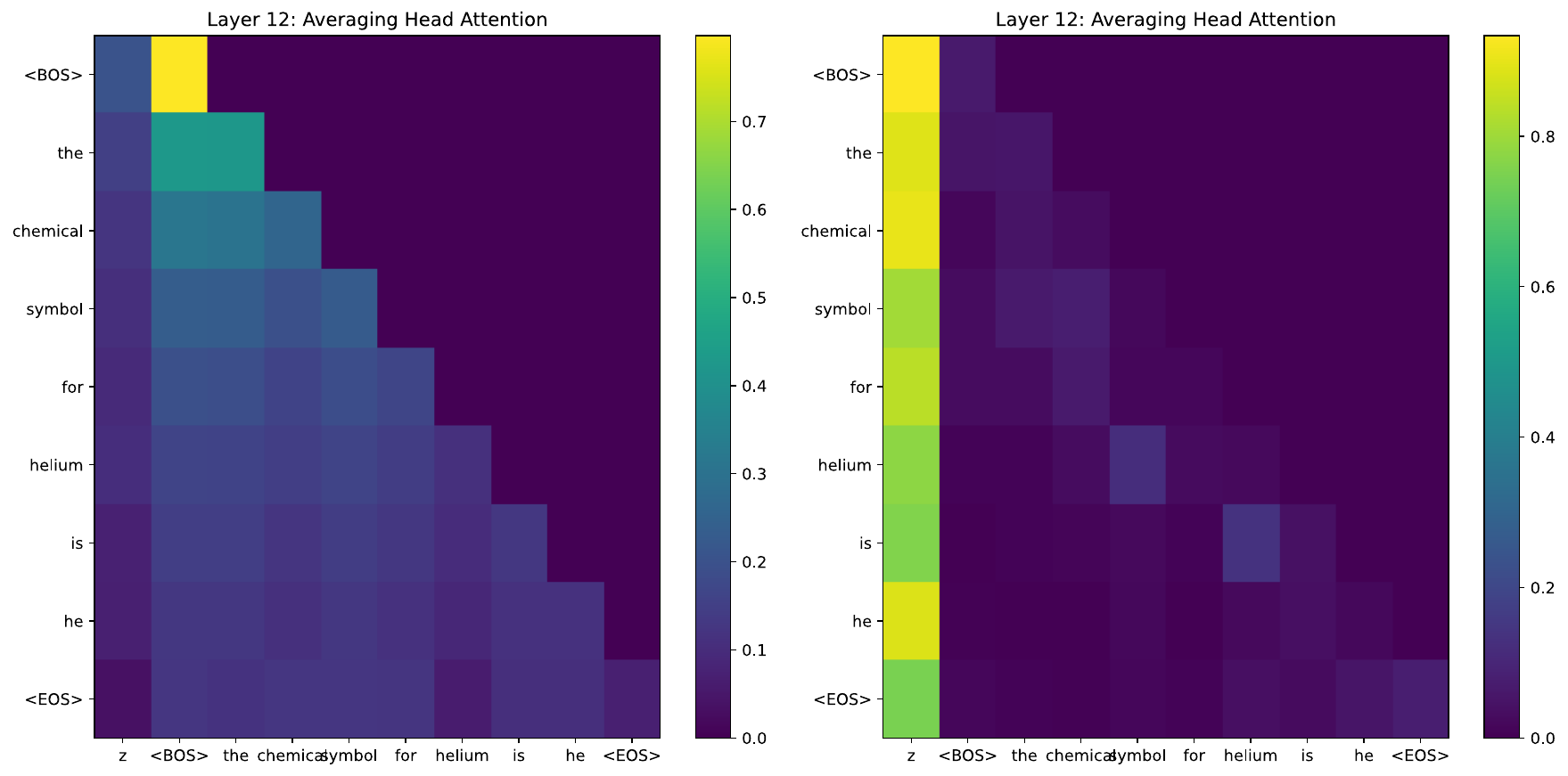}
    \caption{\textit{the chemical symbol for helium is he}.}
    \label{fig:heatmap_2}
\end{figure}

\paragraph{Attention Heatmap.} \label{sec:attn_map} We provide more attention heatmap of different sentences in Figure \ref{fig:heatmap_1} and \ref{fig:heatmap_2}. Similar observation as before, the latent representation can better capture word content information under the graph-language encoding setup.

\paragraph{Traversal.} \label{sec:app_traversal} We provide the traversed sentences of semantic space and syntax space in table \ref{tab:traversal_example} and \ref{tab:traversal_example_1}, respectively. From it, we can observe that the geometrical neighbour sentences traversed via \textit{Ornstein-Uhlenbeck} random walk can hold similar lexical information (“sea/river/ocean”).

More specifically, regarding the traversal of the syntactic space (Table \ref{tab:traversal_example_1}), we can find that the semantics of the generated sentences exhibit higher variability (compared to the variability in syntactic structures when we traverse the semantic space). We conjecture this is mainly because a change in syntactic structure is intrinsically connected with a change in semantics (that is, a perfect separation between the two spaces is extremely hard to achieve). For example, the traversal of the syntactic structure such as the one in Table \ref{tab:traversal_example_1} (e.g., from (S (NP) (VP (ADJP (PP (NP))))) —> (S (NP) (VP (NP (NP) (PP (NP (NP) (PP (NP (ADJP(PP (NP))))))))))) will intrinsically require changes in the semantics of the generated sentences. However, while the intrinsic semantics is expected to change, an alleviation of the information bottleneck is expected to reduce at least the lexical variability of the sentences (that is including entities and relations that are more closely related) derived from our semantic-syntactic separation. In this case, we can observe better results when we compare our approach with Optimus.

\begin{table}[t]
\begin{tcolorbox}[fontupper=\small, fontlower=\small, middle=0.3cm, top=1pt, bottom=1pt, title=Semantic Space Traversal]
Optimus: \\
 0: a desert is a land found in desert environments \\
 1: a forest is a large structure that contains lots of trees \\
 2: a river is a nonliving thing \\
 3: a canyon is a very deep valley \\
 4: a mountain is a large land mass
    \\ \\
Bert-TransCONV: \\
 0: a sea is a source of water for humans \\
 1: a sea is a source of freshwater \\
 2: a river is a source of water \\
 3: an ocean is a source of water for residents
\end{tcolorbox}
\caption{Qualitative evaluation of traversed examples of Optimus (top) and Bert-TransCONV (addition QKV) (bottom).}
\label{tab:traversal_example}
\end{table}
\begin{table}[t]
\begin{tcolorbox}[fontupper=\small, fontlower=\small, middle=0.3cm, top=1pt, bottom=1pt, title=Syntax Space Traversal]
Bert-TransCONV: \\
 0: a river is synonymous with a coastline \\
 1: a hurricane is composed of water vapor and dust \\
 2: a hurricane is the source of most of water vapor in the atmosphere \\
 3:  hurricane is mainly made of water vapor \\
 4: a hurricane is measuring the amount of water in an area
\end{tcolorbox}
\caption{Qualitative evaluation of traversed examples of Bert-TransCONV (addition QKV).}
\label{tab:traversal_example_1}
\end{table}
\chapter{Reasoning Control via Inference Types}
\section{Annotation Details} \label{sec:annotation}
\paragraph{Annotation procedure.} Annotation was performed manually for 5134 entailment triples (two premises, one conclusion) from the EntailmentBank \cite{dalvi2021explaining}, according to Algorithm \ref{alg:annotation}. Graph subset relations and root matching are relaxed for non-argument (:ARG*, op*) edges, meaning relations such as \textit{:manner} or \textit{:time} can be ignored for this purpose. Two independent annotators with post-graduate level backgrounds in Computational Linguistics were used in this process, on a consensus-based annotation scheme where a first annotator defined the transformations and a second annotator verified and refined the annotation scheme, in two iterations. The annotation of the AMR graph is based on an off-the-shelf parser \cite{damonte-17}. The descriptions for each inference type category are as follows:

\textbf{ARG-SUB} (Figure \ref{fig:amr_argsub}): the conclusion is obtained by replacing one argument with another argument.

\textbf{PRED-SUB}: the conclusion is obtained by replacing one verb with another verb.

\textbf{FRAME-SUB}: the conclusion is obtained by replacing a frame of one of the premises with one from the other premise.

\begin{figure}[ht!]
    \centering
    \includegraphics[width=0.5\columnwidth]{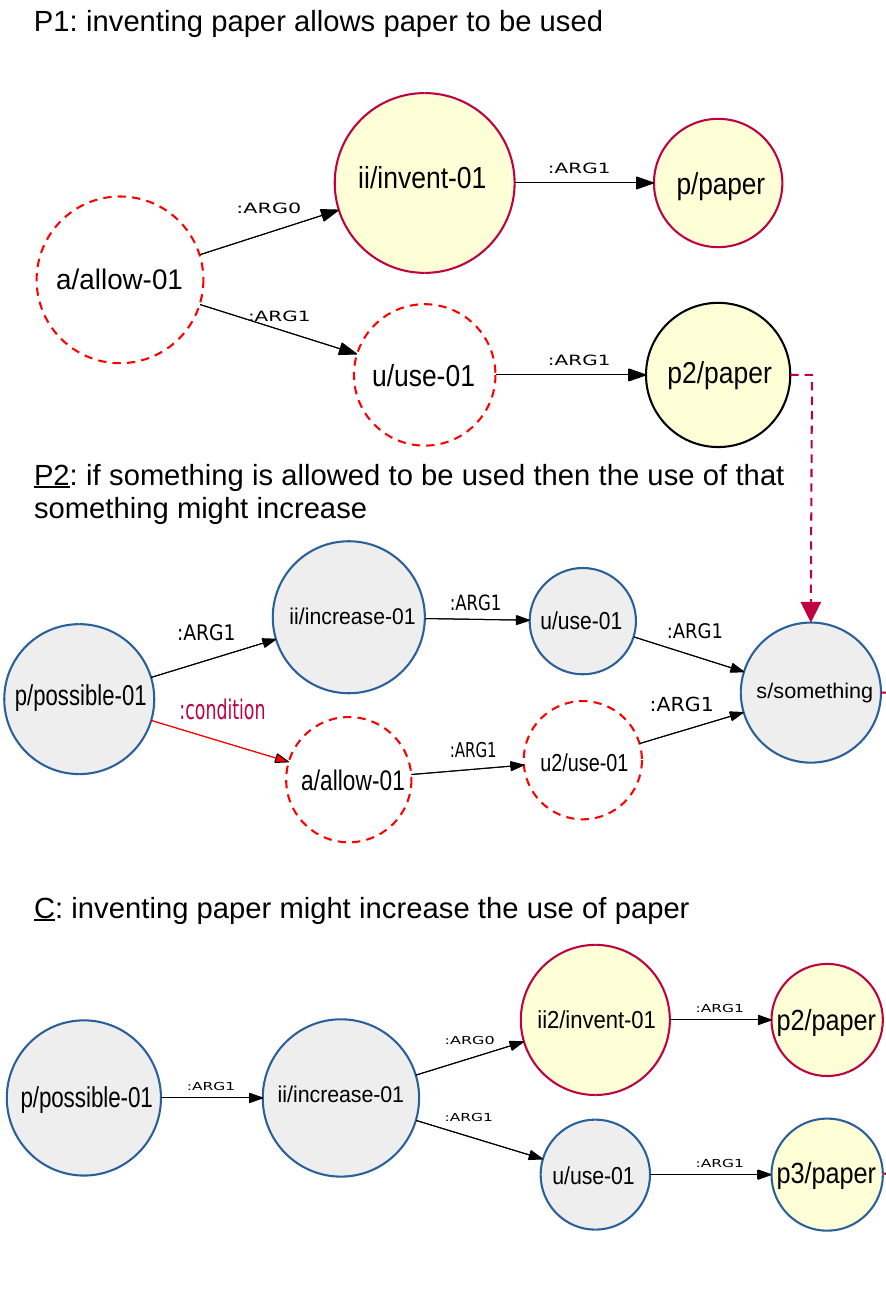}
    \caption{AMR conditional frame insertion (COND-FRAME).}
    \label{fig:amr_condframesub}
\end{figure}

\textbf{COND-FRAM} (Figure \ref{fig:amr_condframesub}): the conclusion is obtained according to the conditional premise with keyword ``if".

\begin{figure}[ht!]
    \centering
    \includegraphics[width=0.5\columnwidth]{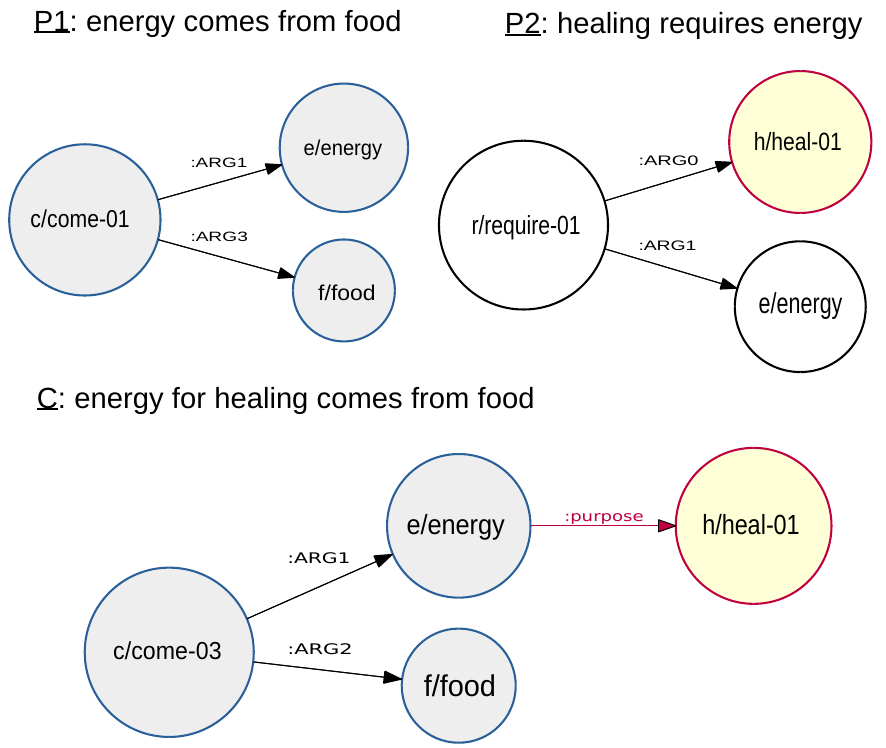}
    \caption{AMR argument insertion (ARG-INS).}
    \label{fig:amr_arginsert}
\end{figure}

\textbf{ARG-INS} (Figure \ref{fig:amr_arginsert}): the conclusion is obtained by connecting an argument from one of the premises to a frame of the other.

\textbf{FRAME-CONJ}: the conclusion is obtained by using connectives to connect two premises.

\begin{figure}[ht!]
    \centering
    \includegraphics[width=0.5\columnwidth]{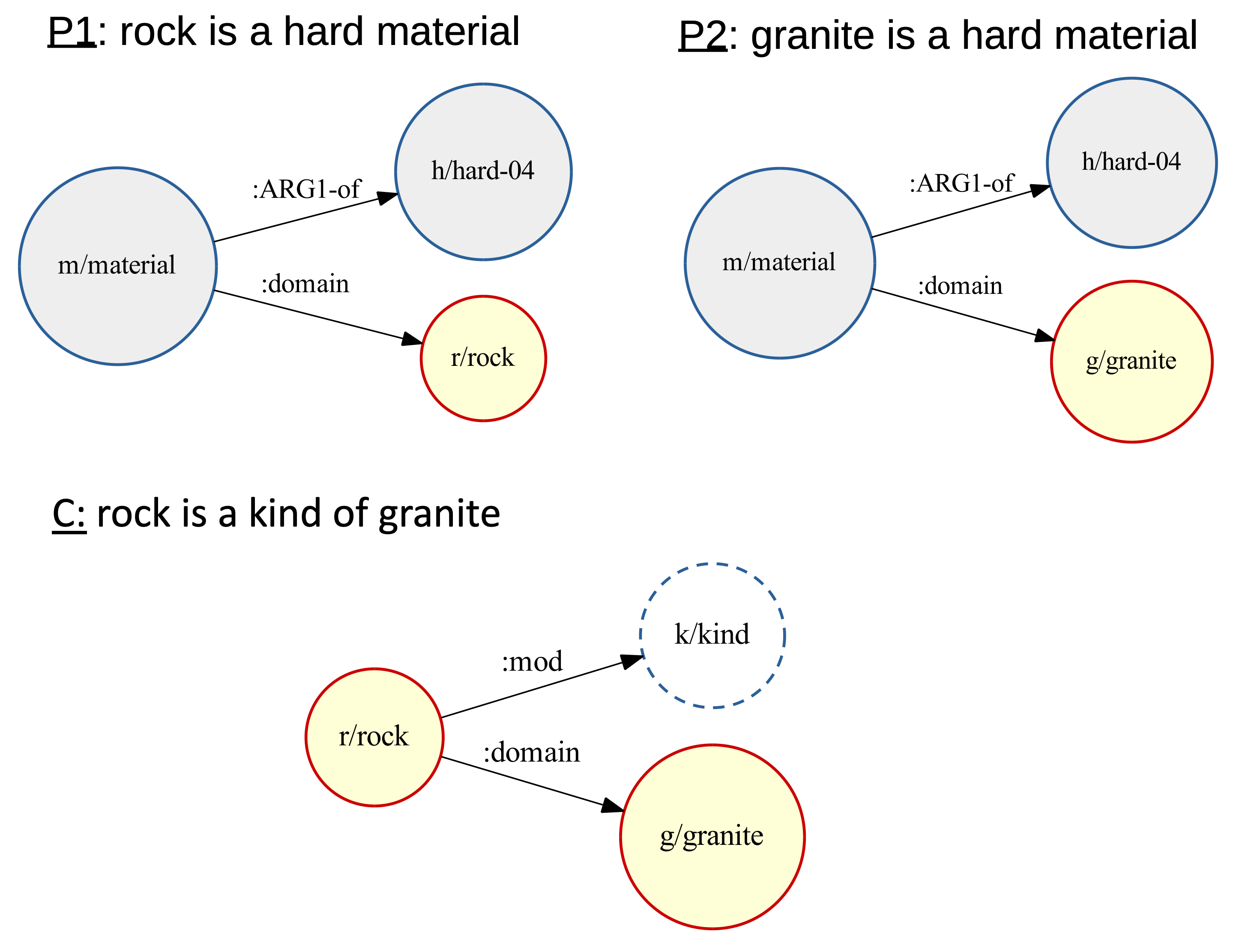}
    \caption{AMR argument generalisation (ARG-GEN).}
    \label{fig:amr_argpredgeneralis}
\end{figure}

\textbf{ARG/PRED-GEN} (Figure \ref{fig:amr_argpredgeneralis}): a new \textit{:domain} relation frame is created in the conclusion if both premise graphs differ by a single predicate/argument term.

\begin{figure}[ht!]
    \centering
    \includegraphics[width=0.5\columnwidth]{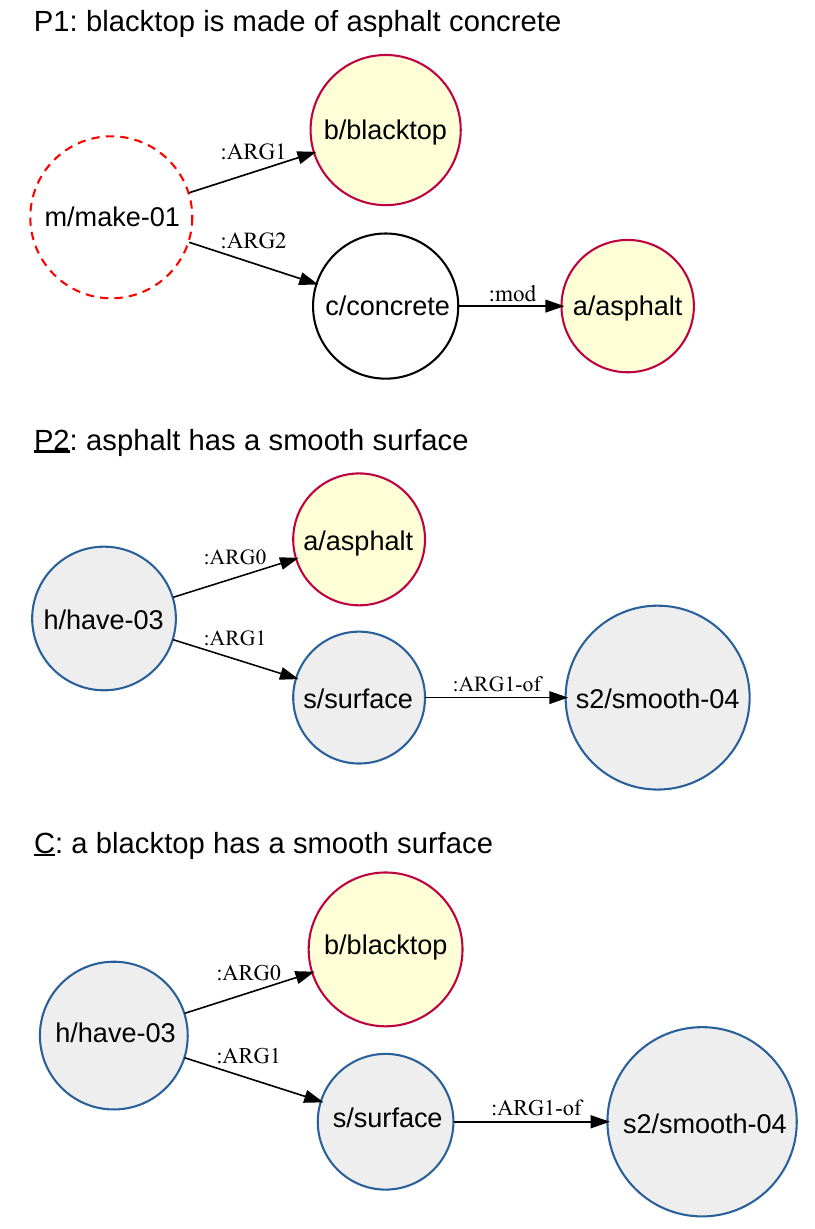}
    \caption{AMR argument substitution (property inheritance) (ARG-SUB-PROP).}
    \label{fig:amr_argsubprop}
\end{figure}

\textbf{ARG-SUB-PROP} (Figure \ref{fig:amr_argsubprop}): one of the premises describes a ``\textit{is made of}'' relationship between the entity in the other premise and its replacement.

\textbf{IFT}: the conclusion should be a conditional sentence.

\textbf{EXAMPLE}: the conclusion should contain the keyword ``example".

\paragraph{Unknown (UNK) category.} In this work, our annotation occupies 84\% based on the EntailmentBank corpus. As for other unknown categories, we do not further specify them, as they either require knowledge outside of the scope of the premises or do not have a consistent symbolic transformation expression. An additional subtype called \textit{premise copy} was included for the cases where the conclusion has the same graph as one of the premises. 

\section{Experimental Details} \label{sec:hyper_param}

\subsection{Dataset}
Table \ref{tab:stats_data} describes the statistical information of the corpus used in the experiment. For experiments: the EntailmentBank dataset is split into train $60\%$, valid $20\%$, and test $20\%$ sets. For the explanation inference retrieval task, we follow the same experimental setup provided online. \footnote{\url{https://github.com/ai-systems/hybrid_autoregressive_inference}}
\begin{table}[ht!]
    \small
    \centering
    \renewcommand\arraystretch{1.1}
      \resizebox{7.6cm}{!}{
    \begin{tabular}{|c|cc|}
        \hline
        Corpus & Num data. & Avg. length \\ \hline
        WorldTree \cite{jansen-etal-2018-worldtree} & 11430 & 8.65 \\
       EntailmentBank \cite{dalvi2021explaining} & 5134 & 10.35 \\ \hline
    \end{tabular}
    }
    \caption{Statistics from explanations datasets. WorldTree is used in the Explanation Inference Retrieval task.} \label{tab:stats_data}
\end{table}

\subsection{T5 Bottleneck Architecture} 
\begin{figure*}[ht!]
\centering
\includegraphics[width=0.9\linewidth]{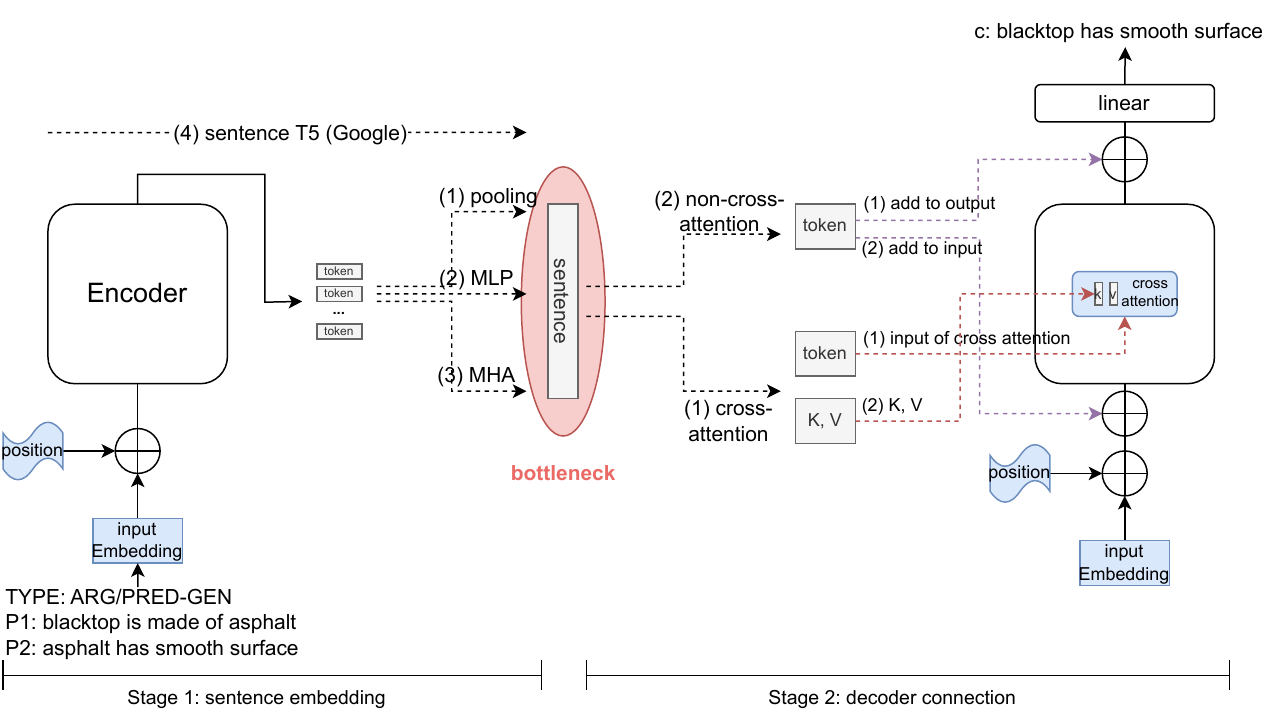}
\caption{The architectural configuration of T5 bottleneck, it consists of two stages: sentence embedding and decoder connection.}
\label{fig:architecture}
\end{figure*}

Figure \ref{fig:architecture} shows the architecture of the T5 bottleneck for learning latent sentence space. It includes two stages: sentence embedding and decoder connection. The sentence embedding aims to transform token embeddings into a sentence (single) embedding. Decoder connection aims to connect the encoder and decoder.

\paragraph{Latent sentence space:} While designing the sentence bottleneck, we compare the four most frequently used mechanisms to transform token embeddings into sentence embeddings: 

(1) Mean pooling: calculating the mean of each dimension on all token embeddings and feeding the resulting vector into a multi-layer perceptron to obtain the sentence embedding. (2) multi-layer perceptron (MLP): applying an MLP to reduce the dimensionality of token embeddings, and the resulting embeddings are concatenated to form a single sentence embedding: $z = \text{concat}\Big[ \text{MLP}_1(x_1); ...; \text{MLP}_T(x_T) \Big]$ where $\text{MLP}_i(x_i)$ represents the $i$-th neural network for input representation of token $x_i$, $z$ is the latent sentence representation, and $T$ is the maximum token length for a sentence. (3) multi-head attention: feeding each token embedding into the multi-head attention and considering the first output embedding as the sentence embedding \cite{montero2021sentence}: $z = \text{MultiHead}\left( XW^q, XW^k, XW^v \right)[0]$ where $X=[x_1, ..., x_T]$ and $W^q$, $W^k$, and $W^v$ are the weights for learning $q$, $k$, $v$ embeddings in self-attention, respectively. (4) Sentence T5: re-loading the pre-trained sentence T5 (S-T5, \citet{https://doi.org/10.48550/arxiv.2108.08877}).

\paragraph{Conditional generation:} Next, we consider four strategies to inject sentence embeddings into the decoder. (1) Cross-attention input embedding (CA Input): reconstructing the token embeddings from a sentence representation and directly feeding them into the cross-attention layers of the decoder: $\hat{Y} = \text{MultiHead}\left(YW^q, \text{MLP}(z)W^k, \text{MLP}(z)W^v\right)$ where $\hat{Y}$ is the reconstruction of decoder input sequence $Y=[y_1, ..., y_K]$. (2) Cross-attention KV embedding (CA KV): instead of reconstructing the token embeddings, it consists of directly learning the Key and Value \cite{hu-etal-2022-fuse,li2020optimus}, which is formalised as $\hat{Y} = \text{MultiHead}\Big(YW^q, \text{MLP}_k(z), \text{MLP}_v(z)\Big)$, where $\text{MLP}_k$ and $\text{MLP}_v$ are neural layers for learning $k$ $v$ embeddings. (3) Non-cross-attention input connection (NCA Input): reconstructing the token embeddings and element-wisely adding them with the input embeddings of the decoder \cite{https://doi.org/10.48550/arxiv.2101.00828}. (4) Non-cross-attention output connection (NCA Output): adding the reconstructed token embeddings to the output embedding of the decoder. The code is provided in the codebase\footnote{\url{https://anonymous.4open.science/r/Inference_type-5E07/}}.

\begin{table}[ht!]
\centering
\resizebox{7.8cm}{!}{
\renewcommand\arraystretch{1}
\begin{tabular}{cccccc} \toprule
\multicolumn{6}{c}{\textit{Train: architecture}} \\
\specialrule{0pt}{2pt}{2pt}
\multicolumn{2}{c}{Decoder Connection} & \shortstack{CA \\ Input} & \shortstack{CA \\ KV} & \shortstack{NCA \\ Input}  & \shortstack{NCA \\ Output}  \\ \hline
\multirow{4}{*}{\shortstack{Sentence \\ Embedding}}  &Pooling & 1.41 & 1.44 & 1.86  & 2.42  \\ 
&MLP & 1.71 & 1.94  & 2.09  & 2.62 \\ 
&MHA & 1.51 & 2.24 & 2.31  & 3.03 \\  
&S-T5 & \textcolor{blue}{\textbf{1.24}} & 1.42 & 1.81  & 2.22 \\ \toprule
\end{tabular}
}
\caption{Comparison of different setups on test loss via cross-entropy (CA: cross-attention, NCA: non-cross-attention), bottom: comparison of different baselines on EntailmentBank testset.} \label{tab:sent_setup}
\end{table}



\subsection{Implementation Details} 

\paragraph{Hyper-parameters.} \textbf{1.} Size of Sentence Representation: in this work, we consider 768 as the size of the sentence embedding. Usually, the performance of the model improves as the size increases. \textbf{2.} Multi-head Attention (MHA): in the experiment, MHA consists of 8 layers, each layer containing 12 heads. The dimensions of Query, Key, and Value are 64 in each head. The dimension of token embedding is 768. Training hyperparameters are: \textbf{3.} For all models, the max epoch: 40, learning rate: 5e-5. During fine-tuning the T5 bottleneck, we first freeze the pre-trained parameters in the first epoch and fine-tune all parameters for the remaining epochs. \textbf{4.} All models are trained on a single A6000 GPU device.

\begin{figure}[t]
    \centering
    \includegraphics[width=0.5\columnwidth]{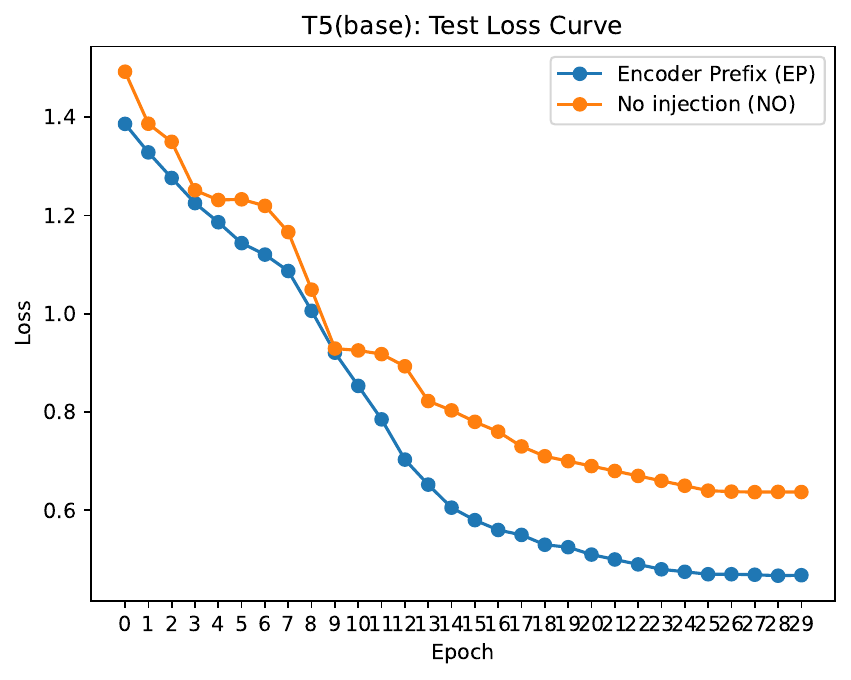}
    \caption{The test loss curve indicates that EP facilitates better convergence, indicating the supervision on inference types aligns the model’s reasoning trajectory with target inference behaviours, improving conclusion prediction accuracy.}
    \label{fig:loss_curve}
\end{figure}

\paragraph{Baselines.} In the experiment, we implement five LSTM-based autoencoders, including denoising AE (\citet{10.1145/1390156.1390294}, DAE), $\beta$-VAE \cite{Higgins2016betaVAELB}, adversarial AE (\citet{makhzani2016adversarial}, AAE), label adversarial AE (\citet{rubenstein2018latent}, LAAE), and denoising adversarial autoencoder (\citet{shen2020educating}, DAAE). Their implementation relies on the open-source codebase available at the URL \footnote{\url{https://github.com/shentianxiao/text-autoencoders}}. As for transformer-based VAEs, we implement Optimus \cite{li2020optimus}\footnote{\url{https://github.com/ChunyuanLI/Optimus}} and Della \cite{hu-etal-2022-fuse}\footnote{\url{https://github.com/OpenVLG/DELLA}}. All baseline models undergo training and evaluation with the hyper-parameters provided by their respective sources. A latent dimension of 768 is specified to ensure a uniform and equitable comparative analysis.

\paragraph{Metrics.} To evaluate the generated conclusions against the reference conclusions, we employ BLEU scores for 1- to 3-gram overlaps and report the average score. Additionally, to assess semantic similarity, we calculate the cosine similarity between the generated and reference conclusions by encoding both using the pretrained Sentence-T5 model\footnote{\url{https://huggingface.co/sentence-transformers/sentence-t5-base}} and computing the cosine similarity of their resulting embeddings. 

\section{Complementary Results} \label{sec:ablation}
\paragraph{Ablation studies.} We remove the inference types from the dataset and evaluate the T5 model performance using the same metrics. In this case, we can compare the model performance trained with or without that inference type. From Table \ref{tab:ablation_study}, we can observe that the baselines (T5 small and base) achieve higher BLEU and BLEURT scores without the data with ARG-INS, COND-FRAME, and UNK inference type, respectively. This result indicates that the T5 cannot generalise well over those inference types. Also, removing the UNK inference type from data can achieve lower loss and PPL, which indicates that it has a negative impact on model training.
\begin{table}[ht!]
\centering
\setlength\tabcolsep{2.5pt}
\resizebox{7.8cm}{!}{
\centering
\begin{tabular}{ccccccc} \toprule
Remove & T5 & BLEU & BLEURT & Cosine & Loss $\downarrow$ & PPL $\downarrow$ \\ \hline
\multirow{2}{*}{\shortstack{FRAME-\\SUB}} & small & 0.50 & 0.19 & 0.95 & 0.95 & 2.58 \\ 
& base & 0.60 & 0.33 & 0.96 & 0.72 & 1.95 \\ \hline

\multirow{2}{*}{ARG-INS} & small & \textcolor{blue}{\textbf{0.54}} & \textcolor{blue}{\textbf{0.27}} & 0.95 & 0.82 & 2.22 \\ 
& base & \textcolor{blue}{\textbf{0.63}} & \textcolor{blue}{\textbf{0.46}} & 0.97 & 0.64 & 1.73 \\ \hline

\multirow{2}{*}{\shortstack{FRAME-\\CONJ}} & small & 0.53 & 0.26 & 0.96 & 0.84 & 2.28 \\ 
& base & 0.60 & 0.35 & 0.96 & 0.65 & 1.76 \\ \hline

\multirow{2}{*}{\shortstack{COND-\\FRAME}} & small & \textcolor{blue}{\textbf{0.55}} & \textcolor{blue}{\textbf{0.25}} & 0.96 & 0.88 & 2.39 \\ 
& base & \textcolor{blue}{\textbf{0.59}} & \textcolor{blue}{\textbf{0.36}} & 0.96 & 0.69 & 1.87 \\ \hline

\multirow{2}{*}{UNK} & small & \textcolor{blue}{\textbf{0.55}} & \textcolor{blue}{\textbf{0.23}} & 0.95 & \underline{0.53} & \underline{1.44} \\ 
& base & \textcolor{blue}{\textbf{0.62}} & \textcolor{blue}{\textbf{0.40}} & 0.96 & \underline{0.58} & \underline{1.57} \\ \hline

No & small & 0.54 & 0.22 & 0.96 & 0.69 & 2.22 \\ 
No & base & 0.57 & 0.33 & 0.96 & 0.61 & 1.65 \\ \toprule

\end{tabular}
}
\caption{Ablation study over inference type (No: no inference types are removed).} \label{tab:ablation_study}
\end{table}

\paragraph{More controllable inference examples.} \label{sec:example_control}
We provide more controlled examples based on the Original T5 in Table \ref{tab:control_generation_comparison} and \ref{tab:more_example_3}. All examples reveal that the inference type can provide quasi-symbolic inference control to language models.
\begin{table}[ht!]
\begin{tcolorbox}[fontupper=\small, fontlower=\small, title=Quasi-symbolic NLI control]
\underline{P1: a \textcolor{orange}{pumpkin} contains \textcolor{blue}{seeds}} \\
\underline{P2: \textcolor{green}{fruit} contains \textcolor{blue}{seeds}}\\

Original T5: \\
ARG-INS: a \textcolor{green}{fruit} in a \textcolor{orange}{pumpkin} contains \textcolor{blue}{seeds} \\
FRAME-CONJ: a \textcolor{orange}{pumpkin} and \textcolor{green}{fruit} both contains \textcolor{blue}{seeds} \\
FRAME-SUB: \textcolor{green}{fruit} is a kind of \textcolor{orange}{pumpkin}
\tcblower
T5 bottleneck: \\
ARG-INS: \textcolor{green}{fruit} is a part of \textcolor{orange}{pumpkin} that contains \textcolor{blue}{seeds} \\
FRAME-CONJ: a \textcolor{green}{fruit} contains \textcolor{blue}{seeds} \\
FRAME-SUB: a \textcolor{orange}{pumpkin} is a kind of plant
\end{tcolorbox}
\caption{Controlled generation. original T5(base) (top) and T5 bottleneck (bottom).}
\label{tab:control_generation_comparison}
\end{table}


\paragraph{Qualitative evaluation for LLM evaluators.} \label{sec:non_consist}
We conduct a qualitative evaluation through manual inspection. However, this assessment is not systematic or rigorously structured as we discussed in the Limitations section. Tables \ref{tab:non_consist} and \ref{tab:ep_no} present examples with discrepancies in scores between ChatGPT4o and GPT4o-mini, as well as a comparison of predictions between encoder prefix injection (EP) and the absence of inference-type injection (NO), respectively.

From both tables, we observe that ChatGPT4o tends to be more accurate than GPT4o-mini and that EP outperforms NO in generating correct predictions.
\begin{table*}[ht!]
\scriptsize
\small
\centering
\resizebox{15.6cm}{!}{
\begin{tabular}{p{4cm}p{4cm}p{4cm}p{1.5cm}p{1.3cm}p{2cm}}
\toprule
\textbf{Premises} & \textbf{Prediction(NO)} & \textbf{Golden} & \textbf{ChatGPT4o} & \textbf{GPT4o-mini} & \textbf{Human Check} \\ \hline
p1: the metal on the roof of a car is in contact with air & the car roof contains water vapor and oxygen & the metal on the roof of a car is in contact with oxygen and water vapor & 0 & 1 & 0 (invalid predicate ``contains'') \\
p2: air contains oxygen and water vapor & \\ \hline

p1: friction occurs when the student is rubbing his hands together & rubbing your hands together causes the temperature of the object to increase & friction causes the temperature of student's hands to increase & 1 & 0 & 1 (replacing ``friction'' with ``rubbing hands together'') \\
p2: friction causes the temperature of an object to increase & \\ \hline

p1: a caterpillar is a kind of insect & metamorphosis is when a caterpillar changes from an immature form to an adult form & an example of metamorphosis is when a caterpillar changes from an immature form to an adult form & 1 & 0 & 1 (replacing ``insect'' with ``metamorphosis'') \\
p2: metamorphosis is when an insect changes from an immature form to an adult form & &&  && \\ \hline
p1: an increase in water has a positive impact on alligators & a flood has a positive impact on alligators & a flood has a positive impact on alligators & 1 & 0 & 1 (exact match) \\
p2: a flood is caused by an increase in water & \\ \hline

p1: predators eat prey & predators catching prey requires catching prey & predators must catch prey to eat prey & 0 & 1 & 0 (fail to do substitution between ``eating'' and ``catching'')\\
p2: eating prey requires catching prey \\ \hline

p1: a leaf uses chlorophyll to produce carbohydrates & a leaf uses chlorophyll to produce sugars & a leaf uses chlorophyll to produce sugar & 0 & 1 & 1 (valid inference)\\
p2: carbohydrates are made of sugars & \\ \hline

p1: salt is a kind of pure substance & salt and pepper are kinds of substances & salt and pepper are two substances & 1 & 0 & 1 (valid conjunction both ``salt'' and ``pepper'') \\
p2: pepper is a kind of substance \\ \hline





p1: different solids will have the same physical properties & one solid will form a mixture & different solids that are combined will become a mixture & 1 & 0 & 0 (incorrect ``one solid'') \\
p2: an mixture is formed by two or more substances combined together physically & \\ \bottomrule
\end{tabular}
}
\caption{Qualitative evaluation for examples with discrepancies in scores between ChatGPT4o and GPT4o-mini (NO: no inference type injection, 0: invalid, 1: valid). We can observe that the ChatGPT4o tends to be more accurate than GPT4o-mini by human check.} \label{tab:non_consist}
\end{table*}

\begin{table*}[ht!]
\scriptsize
\small
\centering
\resizebox{15.6cm}{!}{
\begin{tabular}{p{3.5cm}p{3.5cm}p{3.5cm}p{3.5cm}p{2cm}p{2cm}}
\toprule
\textbf{Premises} & \textbf{Prediction(NO)} & \textbf{Prediction(EP)} & \textbf{Golden} & \textbf{ChatGPT4o} & \textbf{Human Check} \\ \hline
p1: the metal on the roof of a car is in contact with air & the car roof contains water vapor and oxygen & the car roof is in contact with oxygen and water vapor & the metal on the roof of a car is in contact with oxygen and water vapor & NO:0, EP:1 & NO:0, EP:1 \\
p2: air contains oxygen and water vapor & \\ \hline
p1: a beak is used for catching food by some birds & ads are used for eating by birds to catch food & a beak is used for eating by some birds & a beak is used for eating food by some birds & NO:0, EP:1 & NO:0, EP:1\\
p2: eating food requires catching food & \\ \hline
p1: predators must catch prey to eat prey & animals must catch and eat prey & animals must catch prey to eat prey & some animals must catch prey to eat & NO:0, EP:1 & NO:0, EP:1 \\
p2: a predator is a kind of animal & \\ \hline
p1: an adaptation is a kind of change & an adaptation is something a living thing responds to a change in an environment & adaptation is when a living thing responds to a change in an environment & an adaptation is a kind of change in response to a change in an environment & NO:0, EP:1 & NO:0, EP:0 \\
p2: adapting is when a living thing responds to a change in an environment & \\ \hline
p1: a doorbell is a kind of electric device & closing a doorbell causes the doorbell to function & closing an electric circuit causes a doorbell to function & an electric circuit causes a doorbell to function & NO:0, EP:1 & NO:0, EP:1 \\
p2: closing an electric circuit causes an electrical device to function & \\ \hline
p1: green plants are made of plant cells & a producer is made of plant cells & producers are made of plant cells & producers are made of plant cells & NO:1, EP:0 & NO:1, EP:1 \\
p2: green plants are a kind of producer & \\ \hline
p1: the iron nail has rusted &  iron nails rusting is when the iron nails chemically react with water and oxygen to form iron nail & the iron nail has rusted & a chemical reaction has happened on the iron nail & NO:0, EP:1 & NO:0, EP:0 \\
p2: rusting is when iron chemically reacts with water and oxygen & \\ \hline
p1: wood burns & wood burns when introduced to wood & wood chips burn & wood chips burn & NO:0, EP:1 & NO:0, EP:1 \\
p2: wood chips are made of wood & \\ \hline
p1: plant reproduction requires pollinating animals for pollination & plants reproduction requires bees that carry pollen & a bee can help plant reproduction by carrying pollen & a bee can help on pollination in plant reproduction by carry pollen & NO:0, EP:1 & NO:0, EP:1 \\
p2: a bee can help on pollination by carrying pollen & \\ \hline
p1: a leaf uses chlorophyll to produce carbohydrates & a leaf uses chlorophyll to produce sugars & a leaf uses chlorophyll to produce sugar & a leaf uses chlorophyll to produce sugar & NO:1, EP:1 & NO:1, EP:1 \\
p2: carbohydrates are made of sugars \\ \bottomrule
\end{tabular}
}
\caption{Qualitative evaluation for prediction through EP and NO (NO: no inference type, EP: encoder prefix), we can observe that EP outperforms NO in generating correct predictions.} \label{tab:ep_no}
\end{table*}
\begin{table*}[ht!]
\scriptsize
\small
\centering
\resizebox{15.6cm}{!}{
\begin{tabular}{p{4.2cm}p{3.2cm}p{8.2cm}}
\toprule
\textbf{Premises} & \textbf{Inference Type} & \textbf{T5 original}  \\ \hline
P1: a pumpkin contains seeds & ARG-INS & a fruit in a pumpkin contains seeds  \\ 
P2: fruit contains seeds & FRAME-CONJ & a pumpkin and fruit both contain seeds \\ 
 & IFT & if a pumpkin contains fruit then the fruit may contain seeds \\ 
 & EXAMPLE & fruit is an example of pumpkins being sown \\
 & ARG/PRED-GEN & a pumpkin is a kind of fruit \\
 & ARG-SUB & fruit can contain pumpkin seeds \\
 & UNK & a pumpkin can contain seeds \\
 & FRAME-SUB & fruit is a kind of pumpkin \\ \hline
P1: eating something has a negative impact on that something & FRAME-SUB & eating cacti has a negative impact on that cacti \\
P2: some animals eat cacti & PRED-SUB & some animals may have a negative impact on cacti \\
& ARG-INS & some animals have a negative impact on cacti by eating cacti \\
& EXAMPLE & cooking cacti is an example of a negative impact on a cactus \\
& INF & if a cactus has a negative impact on an animal, that cactus could be devoured \\ \hline
P1: seeing requires light & ARG-SUB & reading requires light \\
P2: reading requires seeing & ARG-INS & light is a kind of requirement for reading \\
& INF & if light is moving then reading may be taken \\
& EXAMPLE & a light bulb will be used for reading \\
& UNK & light will help you read \\
\toprule
\end{tabular}
}
\resizebox{15.6cm}{!}{
\begin{tabular}{p{4.2cm}p{3.2cm}p{8.2cm}}
\toprule
\textbf{Premises} & \textbf{Inference Type} & \textbf{T5 bottleneck}  \\ \hline
P1: a pumpkin contains seeds & ARG-INS & fruit is part of a pumpkin that contains seeds  \\ 
P2: fruit contains seeds & FRAME-CONJ & a fruit contains seeds \\ 
 & FRAME-SUB & a pumpkin is a kind of plant \\ \hline
P1: if a pot is exposed to a stove then that pot may become hot & COND-FRAME & the pot may become hot \\ 
P2: the pot is exposed to a stove & ARG/PRED-GEN & the pot may be a source of heat \\ \hline
P1: eating something has a negative impact on that something & FRAME-SUB & eating cacti has a negative impact on that cacti \\
P2: some animals eat cacti & PRED-SUB & animals have a negative impact on cacti \\
& ARG-INS & some animals have a negative impact on cacti by eating cacti \\ \hline

P1: seeing requires light & ARG-SUB & reading requires light \\
P2: reading requires seeing & FRAME-CONJ & reading and feeling can both be used \\
& INF & if something is visible then that something will be seen \\ \toprule
\end{tabular}
}
\caption{controllable NLI via inference type (Top: original T5, bottom: T5 bottleneck).} \label{tab:more_example_3}
\end{table*}

\begin{algorithm*}[ht!]
    \caption{Annotation procedure} \label{alg:annotation}
    \begin{algorithmic}[1]
    \State Find premise $P_x$ most similar to the conclusion $C$, $P_{\bar{x}}$ being the other premise.
    \State $G_{x, \bar{x}, C}~~\gets~~ $AMR~graph~of~$P_x, P_{\bar{x}}, C$,~~respectively. 
    \If{$G_x = G_c$ or $G_{\bar{x}} = G_c$}
        \State $type = PREM\textnormal{-}COPY$ 
    \ElsIf{$P_x$ and $C$ differ by one word $w$} 
        \If{$w$ is a verb}
            \State $type = PRED\textnormal{-}SUB$
        \Else
            \State $type = ARG\textnormal{-}SUB$
        \EndIf
    \Else \\ 
        \State Get AMR graphs $G_1$, $G_2$, $G_c$ for $P_1$, $P_2$ and $C$ respectively. $P_x \rightarrow G_x$.
        \If{$\exists~\textnormal{:ARG*}(x, a)$ $\in C$ and $a \in P_{\bar{x}}$}
            \If{$\exists$ :condition($root(G_x)$, $root(G_{\bar{x}})$)} \\ 
                \State $type = COND\textnormal{-}FRAME$ 
            \ElsIf{$root(a)$ is a noun}
                \If{$root(G_{\bar{x}}) =$ ``make-01'' and $\exists$ :ARG*($root(G_{\bar{x}})$, a)} \\ 
                    \State $type = ARG\textnormal{-}SUB\textnormal{-}PROP$
                \Else
                    \State $type = ARG\textnormal{-}SUB$ 
                \EndIf
            \Else
                \State $type = FRAME\textnormal{-}SUB$
            \EndIf \\
        \ElsIf{$G_x \subset G_c$ and $G_{\bar{x}} \subset G_C$}
            \State $type = FRAME\textnormal{-}CONJ$
        \ElsIf{$\exists x, y$ :domain($root(G_x)$, $x$) and :domain($root(G_{\bar{x}}$, $y$) and :op*(``and'', $x$) $\in G_c$ and :op*(``and'', y) $\in G_c$} 
            \State $type = FRAME\textnormal{-}CONJ$
        \ElsIf{$G_x \subset G_c$}
            \State $d \gets G_c - G_x$
            \If{$root(d)$ is a noun}
                \State $type = ARG\textnormal{-}INS$ 
            \Else
                \State $type = FRAME\textnormal{-}INS$ 
            \EndIf \\
        \ElsIf{$\exists$ :domain($root(G_c)$, $y$) and ($root(G_c) \in G_x$ and $y \in G_{\bar{x}}$) or ($root(G_c) \in G_{\bar{x}}$ and $y \in G_x$)}
            \State $type = ARG/PRED\textnormal{-}GEN$
        \Else
            \State $type = UNK$
        \EndIf
    \EndIf
    \end{algorithmic}
\end{algorithm*}

\begin{table*}[ht!]
\begin{tcolorbox}[fontupper=\small, fontlower=\small, title=Prompts for automatic evaluation]
\textbf{Consistency:} \\
You are a scoring expert in natural language reasoning. Given two premises and a conclusion, your goal is to evaluate whether the conclusion violates the premises. During your inference process, please only consider the information from the premises.

you can directly give your score (0 or 1) based on the following criteria:

0: the conclusion violates the premises.

1: the conclusion doesn't violate the premises. \\

The output format is just the score. You don't need to analyse the reasoning process.

\tcblower
\textbf{Alignment:} \\
You are a scoring expert. Given two premises, a conclusion, and an inference type, your goal is to evaluate whether the (premises, conclusion) pair is aligned with the inference type. \\

The following is the description of 10 inference types:

1. ARG-SUB: the conclusion is obtained by replacing one argument with another argument.

2. PRED-SUB: the conclusion is obtained by replacing one verb with another verb.

3. FRAME-SUB: the conclusion is obtained by replacing a frame of one of the premises with one from the other premise.

4. COND-FRAM: the conclusion is obtained according to the conditional premise with keyword “if”.

5. ARG-INS: the conclusion is obtained by connecting an argument from one of the premises to a frame of the other.

6. FRAME-CONJ: the conclusion is obtained by using connectives to connect two premises.

7. ARG/PRED-GEN: a new “:domain” relation frame is created in the conclusion if both premise graphs differ by a single predicate/argument term.

8. ARG-SUB-PROP: one of the premises describes a “is made of” relationship between the entity in the other premise and its replacement.

9. IFT: the conclusion should be a conditional sentence.

10. EXAMPLE: the conclusion should contain the keyword “example”.
\\ \\
When evaluating, some premises might not be able to deduce more than one conclusions. You can ignore those cases. \\

Finally, you can directly give your score (0 or 1) based on the following criteria:

0: the (premises, conclusion) pair is not aligned with the inference type.

1: the (premises, conclusion) pair is aligned with the inference type. \\

The output format is just the score. You don't need to analyse the reasoning process.




\end{tcolorbox}
\caption{Empirically designed prompt for automatically evaluating the controllability in Section \ref{sec:quasi_sym_inf_eval}.}
\label{tab:prompt}
\end{table*}

\end{document}